\documentclass[10pt,twocolumn,letterpaper]{article}

\usepackage{cvpr}
\usepackage{times}
\usepackage{epsfig}
\usepackage{graphicx}
\usepackage{amsmath}
\usepackage{amssymb}
\usepackage{multirow}
\usepackage{subfig}
\usepackage{threeparttable}
\usepackage{multirow}
\usepackage{makecell}
\usepackage[normalem]{ulem}
\usepackage{caption}
\usepackage{titling}
\usepackage[breaklinks=true,bookmarks=false]{hyperref}
\cvprfinalcopy 

\def\cvprPaperID{3443} 

\date{}

\vspace{-5.5mm}
\begin{document}

\title{A-CNN: Annularly Convolutional Neural Networks on Point Clouds}

\author{Artem Komarichev \qquad Zichun Zhong \qquad Jing Hua\\
Department of Computer Science, Wayne State University\\
{\tt\small \{artem.komarichev,zichunzhong,jinghua\}@wayne.edu} 
}

\maketitle


\begin{abstract}
  Analyzing the geometric and semantic properties of 3D point clouds through the deep networks is still challenging due to the irregularity and sparsity of samplings of their geometric structures. This paper presents a new method to define and compute convolution directly on 3D point clouds by the proposed annular convolution. This new convolution operator can better capture the local neighborhood geometry of each point by specifying the (regular and dilated) ring-shaped structures and directions in the computation. It can adapt to the geometric variability and scalability at the signal processing level. We apply it to the developed hierarchical neural networks for object classification, part segmentation, and semantic segmentation in large-scale scenes. The extensive experiments and comparisons demonstrate that our approach outperforms the state-of-the-art methods on a variety of standard benchmark datasets (e.g., ModelNet10, ModelNet40, ShapeNet-part, S3DIS, and ScanNet).
\end{abstract}
\vspace{-0.5cm}

\section{Introduction} 
Nowadays, the ability to understand and analyze 3D data is becoming increasingly important in computer vision and computer graphics communities. During the past few years, the researchers have applied deep learning methods to analyze 3D objects inspired by the successes of these techniques in 2D images and 1D texts. Traditional low-level handcrafted shape descriptors suffer from not being able to learn the discriminative and sufficient features from 3D shapes~\cite{ahmed2018deep}. Recently, deep learning techniques have been applied to extract hierarchical and effective information from 3D shape features captured by low-level descriptors~\cite{liu2014high,bu2015local}. 3D deep learning methods are widely used in shape classification, segmentation, and recognition, etc. But all these methods are still constrained by the representation power of the shape descriptors.

One of the main challenges to directly apply deep learning methods to 3D data is that 3D objects can be represented in different formats, i.e., regular / structured representation (e.g., multi-view images and volumes), and irregular / unstructured representation (e.g., point clouds and meshes). There are extensive approaches based on regular / structured representation, such as multi-view convolutional neural networks (CNNs)~\cite{su2015multi,qi2016volumetric,huang2018learning} and 3D volumetric / grid CNN methods and its variants~\cite{wu20153d,qi2016volumetric,riegler2017octnet,wang2017cnn,Wang-2018-AOCNN,le2018pointgrid,hua2018pointwise}. These methods can be conveniently developed and implemented in 3D data structure, but they easily suffer from the heavy computation and large memory expense. So it is better to define the deep learning computations based on 3D shapes directly, i.e., irregular / unstructured representation, such as point cloud based methods~\cite{qi2017pointnet,qi2017pointnet++,klokov2017escape,shen2018mining,atzmon2018point,li2018pointcnn,liu2018point2sequence,tatarchenko2018tangent,li2018so,yi2017syncspeccnn,Wang_2018_ECCV,engelmann2017exploring,Ye_2018_ECCV}. 
However, defining the convolution on the irregular / unstructured representation of 3D objects is not an easy task. Very few methods on point clouds have defined an effective and efficient convolution on each point. Meanwhile, several approaches have been proposed to develop convolutional networks on 2D manifolds~\cite{masci2015geodesic,boscaini2016learning,monti2017geometric,Xu_2017_ICCV}. Their representations (e.g., 3D surface meshes) have point positions as well as connectivities, which makes it relatively easier to define the convolution operator on them.
\begin{figure}[t]	
\begin{center}
  \vspace{-0mm}
  \includegraphics[width=0.9\linewidth, trim={0.2cm 0.1cm 0 0.1cm}, clip]{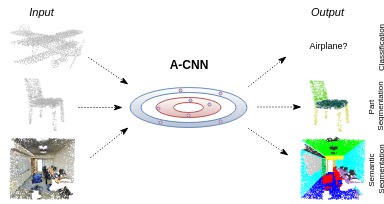}
  \vspace{-5.5mm}
\end{center}
   \caption{The proposed annularly convolutional neural networks (A-CNN) model on point clouds to perform classification, part segmentation, and semantic segmentation tasks.}\vspace{-5mm}
\label{fig:intro}
\end{figure}

In this work, we present a new method to define and compute convolutions directly on 3D point clouds effectively and efficiently by the proposed \emph{annular convolutions}. This new convolution operator can better capture local neighborhood geometry of each point by specifying the (regular and dilated) ring-shaped structures and directions in the computation. It can adapt to the geometric variability and scalability at the signal processing level. Then, we apply it along with the developed hierarchical neural networks to object classification, part segmentation, and semantic segmentation in large-scale scene as shown in Fig.~\ref{fig:intro}. The key \emph{contributions} of our work are as follows:\vspace{-2mm}
\begin{itemize}
  \item We propose a new approach to define convolutions on point cloud. The proposed \textit{annular convolutions} can define arbitrary kernel sizes on each local ring-shaped region, and help to capture better geometric representations of 3D shapes;\vspace{-2mm}
  \item We propose a new multi-level hierarchical method based on \textit{dilated rings}, which leads to better capturing and abstracting shape geometric details. The new dilated strategy on point clouds benefits our proposed closed-loop convolutions and poolings;\vspace{-2mm}
  \item Our proposed network models present new state-of-the-art performance on object classification, part segmentation, and semantic segmentation of large-scale scenes using a variety of standard benchmark datasets.
\end{itemize}

\section{Related Work}
Due to the scope of our work, we focus only on recently related deep learning methods, which are proposed on different 3D shape representations.

\textbf{Volumetric Methods.} One traditional way to analyze a 3D shape is to convert it into the regular volumetric occupancy grid and then apply 3D CNNs~\cite{wu20153d,qi2016volumetric}. The major limitation of these approaches is that 3D convolutions are more expensive in computations than 2D cases. In order to make the computation affordable, the volume grid size is usually in a low resolution. However, lower resolution means loosing some shape geometric information, especially in analyzing large-scale 3D shapes / scenes. To overcome these problems, octree-based methods~\cite{riegler2017octnet,wang2017cnn,Wang-2018-AOCNN} have been proposed to allow applying 3D CNNs on higher / adaptive resolution grids. PointGrid~\cite{le2018pointgrid} is a 3D CNN that incorporates a constant number of points within each grid cell and allows it to learn better local geometric details. Similarly, Hua et al.~\cite{hua2018pointwise} presented a 3D convolution operator based on a uniform grid kernel for semantic segmentation and object recognition on point clouds.

\textbf{Point Cloud based Methods.} PointNet~\cite{qi2017pointnet} is the first attempt of applying deep learning directly on point clouds. PointNet model is invariant to the order of points, but it considers each point independently without including local region information. PointNet++~\cite{qi2017pointnet++} is a hierarchical extension of PointNet model and learns local structures of point clouds at different scales. But~\cite{qi2017pointnet++} still considers every point in its local region independently. 
In our work, we address the aforementioned issues by defining the convolution operator that learns the relationship between neighboring points in a local region, which helps to better capture the local geometric properties of the 3D object.

Klokov et al.~\cite{klokov2017escape} proposed a new deep learning architecture called Kd-networks, which uses kd-tree structure to construct a computational graph on point clouds. KCNet~\cite{shen2018mining} improves PointNet model by considering the local neighborhood information. It defines a set of learnable point-set kernels for local neighboring points and presents a pooling method based on a nearest-neighbor graph. PCNN~\cite{atzmon2018point} is another method to apply convolutional neural networks to point clouds by defining extension and restriction operators, and mapping point cloud functions to volumetric functions. SO-Net~\cite{li2018so} is a permutation invariant network that utilizes spatial distribution of point clouds by building a self-organizing map. There are also some spectral convolution methods on point clouds, such as SyncSpecCNN~\cite{yi2017syncspeccnn} and spectral graph convolution~\cite{Wang_2018_ECCV}.
Point2Sequence~\cite{liu2018point2sequence} learns the correlation of different areas in a local region by using attention mechanism, but it does not propose a convolution on point clouds.
PointCNN~\cite{li2018pointcnn} is a different method that proposes to transform neighboring points to the canonical order and then apply convolution.

Recently, there are several approaches proposed to process and analyze large-scale point clouds from indoor and outdoor environments. Engelmann et al.~\cite{engelmann2017exploring} extended PointNet model to exploit the large-scale spatial context. Ye et al.~\cite{Ye_2018_ECCV} proposed a pointwise pyramid pooling to aggregate features at local neighborhoods as well as two-directional hierarchical recurrent neural networks (RNNs) to learn spatial contexts. However, these methods do not define convolutions on large-scale point clouds to learn geometric features in the local neighborhoods.
TangentConv~\cite{tatarchenko2018tangent} is another method that defines the convolution on point clouds by projecting the neighboring points on tangent planes and applying 2D convolutions on them. 
The orientation of the tangent image is estimated according to the local point / shape curvature, but as we know the curvature computation on the local region of the point clouds is not stable and not robust (see the discussion in Sec.~\ref{sec:annular_conv}), which makes it orientation-dependent. Instead, our method proposes an annular convolution, which is invariant to the orientations of local patches.
Also, ours does not require additional input features while theirs needs such features (e.g., depth, height, etc.).

\textbf{Mesh based Methods.} Besides point cloud based methods, several approaches have been proposed to develop convolutional networks on 3D meshes for shape analysis. Geodesic CNN~\cite{masci2015geodesic} is an extension of the Euclidean CNNs to non-Euclidean domains and is based on a local geodesic system of polar coordinates to extract local patches. Anisotropic CNN~\cite{boscaini2016learning} is another generalization of Euclidean CNNs to non-Euclidean domains, where classical convolutions are replaced by projections over a set of oriented anisotropic diffusion kernels. Mixture Model Networks (MoNet)~\cite{monti2017geometric} generalizes deep learning methods to non-Euclidean domains (graphs and manifolds) by combining previous methods, e.g., classical Euclidean CNN, Geodesic CNN, and Anisotropic CNN. MoNet proposes a new type of kernel in parametric construction. Directionally Convolutional Networks (DCN)~\cite{Xu_2017_ICCV} applies convolution operation on the triangular mesh of 3D shapes to address part segmentation problem by combining local and global features. Lastly, Surface Networks~\cite{Kostrikov-cvpr18} propose upgrades to Graph Neural Networks to leverage extrinsic differential geometry properties of 3D surfaces for increasing their modeling power.

\section{Method}
\label{sec:method}
In this work, we propose a new end-to-end framework named as annularly convolutional neural networks (A-CNN) that leverages the neighborhood information to better capture local geometric features of 3D point clouds. In this section, we introduce main technique components of the A-CNN model on point clouds that include: regular and dilated rings, constraint-based k-nearest neighbors (k-NN) search, ordering neighbors, annular convolution, and pooling on rings.

\subsection{Regular and Dilated Rings on Point Clouds}
\label{sec:rings}
To extract local spatial context of the 3D shape, PointNet++~\cite{qi2017pointnet++} proposes multi-scale architecture. The major limitation of this approach is that multiple scaled regions may have overlaps (i.e., same neighboring points could be duplicately included in different scaled regions), which reduces the performance of the computational architecture. Overlapped points at different scales lead to redundant information at the local region, which limits a network to learn more discriminative features.

In order to address the above issue, our proposed framework is aimed to leverage a neighborhood at different scales more wisely. We propose two ring-based schemes, i.e., \emph{regular rings} and \emph{dilated rings}. Comparing to multi-scale strategy, the ring-based structure does not have overlaps (no duplicated neighboring points) at the query point's neighborhood, so that each ring contains its own unique points, as illustrated in Sec.~\ref{sec:suppl_ball_vs_ring_comparison} of Supplementary Material.

The difference between regular rings and dilated rings is that dilated rings have empty space between rings. The idea of proposed dilated rings is inspired by dilated convolutions on image processing~\cite{YuKoltun2016}, which benefits from aggregating multi-scale contextual information. Although each ring may define the same number of computation / operation parameters (e.g., number of neighboring points), the coverage area of each ring is different (i.e., dilated rings will have larger coverage than the regular rings) as depicted in Fig.~\ref{fig:dilated_rings}. Regular rings can be considered as dilated rings with the dilation factor equal to 0.

The proposed regular rings and dilated rings will contribute to neighboring point search, convolution, and pooling in the follow-up processes. First, for k-NN algorithm, we constrain search areas in the local ring-shaped neighborhood to ensure no overlap. Second, the convolutions defined on rings cover larger areas with the same kernel sizes without increasing the number of convolution parameters. Third, the regular / dilated ring architectures will help to aggregate more discriminative features after applying max-pooling at each ring of the local region. We will discuss them in more detail in the following subsections.

To justify the aforementioned statements, we will compare multi-scale approach with our proposed multi-ring scheme on object classification task in the ablation study (Sec.~\ref{sec:ablation}). The results show that ring-based structure captures better local geometric features than previous multi-scale method, since it achieves higher accuracy.
\begin{figure}[t]
\begin{center}
  \includegraphics[width=0.8\linewidth]{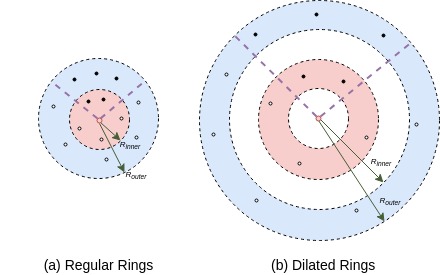}\vspace{-4mm}
\end{center}
\caption{The comparison of the regular and dilated ring-shaped structures (such as with two rings). We can see that comparing two sectors (e.g., black solid points) in the regular and dilated rings, the dilated rings cover larger space by using the same number of neighbors as in regular rings. Moreover, each ring contains unique neighboring points comparing to the other ring.
}\vspace{-5mm}
\centering
\label{fig:dilated_rings}
\end{figure}

\subsection{Constraint-based K-NN Search}
In the original PointNet++ model, the ball query algorithm returns the first $K$ neighbors found inside a search ball specified by a radius $R$ and query point $\mathbf{q}_i$, so that it cannot guarantee that the closest points will always be found. However, our proposed k-NN search algorithm guarantees returning closest points inside the searching area by using the Euclidean metric. Each ring is defined by two parameters: the inner radius $R_{inner}$ and the outer radius $R_{outer}$ (in Fig.~\ref{fig:dilated_rings}); therefore, the \emph{constraint-based k-NN search} ensures that the closest and unique points will be found in each ring.
\vspace{-2mm}
\subsection{Ordering Neighbors}
\label{sec:project_order}

In order to learn relationships between neighboring points in a local regions, we need first to order points in a clockwise / counterclockwise manner and then apply annular convolutions. Our proposed ordering operator consists of two main steps: projection and ordering. The importance of the projection before ordering is that the dot product has its restriction in ordering points. By projecting points on a tangent plane at a query point $\mathbf{q}_i$, we effectively order neighbors in clockwise / counterclockwise direction by taking use of cross product and dot product together. The detailed explanations of normal estimation, orthogonal projection, and ordering are given in the following subsections.
\vspace{-8mm}
\subsubsection{Normal Estimation on Point Clouds}
\label{sec:normal_estimation}
Normal is an important geometric property of a 3D shape. We use it as a tool for projecting and ordering neighboring points at a local domain. The simplest normal estimation method approximates the normal $\mathbf{n}_i$ at the given point $\mathbf{q}_i$ by calculating the normal of the local tangent plane $\mathcal{T}_i$ at that point, which becomes a least-square plane fitting estimation problem~\cite{RusuDoctoralDissertation}.
To calculate normal $\mathbf{n}_i$, one needs to compute eigenvalues and eigenvectors of the covariance matrix $\mathbf{C}$ as:
\vspace{-2mm}
\begin{equation}
\begin{aligned}
  \mathbf{C} &= \frac{1}{K} \sum_{j=1}^{K} (\mathbf{x}_j - \mathbf{q}_i) \cdot (\mathbf{x}_j - \mathbf{q}_i)^T, \\
  \mathbf{C} &\cdot \mathbf{v}_\gamma = \lambda_\gamma \cdot \mathbf{v}_\gamma, \gamma\in\{0,1,2\},
\end{aligned}
\end{equation}
where $K$ is the number of neighboring points $\mathbf{x}_j$s around query point $\mathbf{q}_i$ (e.g., $K$ = 10 in our experiments), $\lambda_\gamma$ and $\mathbf{v}_\gamma$ are the $\gamma {th}$ eigenvalue and eigenvector of the covariance matrix $\mathbf{C}$, respectively. The covariance matrix $\mathbf{C}$ is symmetric and positive semi-definite. The eigenvectors $\mathbf{v}_\gamma$ form an orthogonal frame, in respect to the local tangent plane $\mathcal{T}_i$. The eigenvector $\mathbf{v}_0$ that corresponds to the smallest eigenvalue $\lambda_0$ is the estimated normal $\mathbf{n}_i$.
\vspace{-3mm}
\subsubsection{Orthogonal Projection}
\begin{figure}[t]
\begin{center}
  \includegraphics[width=0.8\linewidth]{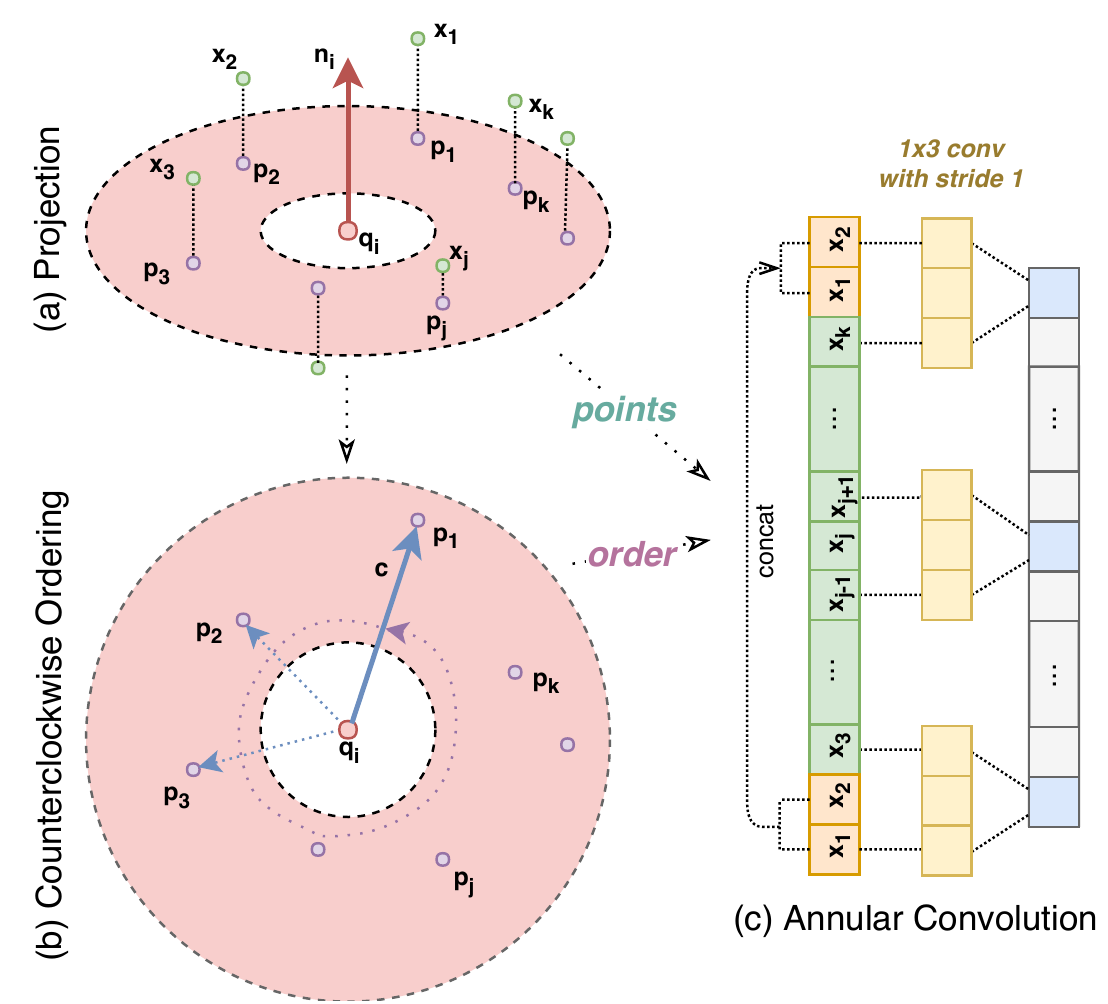}\vspace{-5mm}
\end{center}
\caption{The illustration of the proposed annular convolution on a ring. (a) Projection: $\mathbf{q}_i$ is a query point. After applying the constraint-based k-NN search, neighboring points $\mathbf{X} = \{\mathbf{x}_{j}|j=1, ..., K\}$ are extracted on a ring. Given the normal $\mathbf{n}_i$ at query point $\mathbf{q}_i$, we project the searched points on the tangent plane $\mathcal{T}_i$. (b) Counterclockwise Ordering: After projection, we randomly pick a starting point as our reference direction $\mathbf{c}$ and order points in counterclockwise. It is worth mentioning that we order original points $[\mathbf{x}_1, \mathbf{x}_2, ..., \mathbf{x}_j, ..., \mathbf{x}_K]$ based on their projections. (c) Annular Convolution: Depending on the kernel size, we copy several original points from the beginning position and concatenate them to the end of the ordered points. Finally, we apply annular convolution with the given kernel.}\vspace{-5mm}
\centering
\label{fig:proj_order_conv_horiz}
\end{figure}

\vspace{-1mm}
After extracting neighbors $\mathbf{x}_{j}, j\in\{1, ..., K\}$ for a query point $\mathbf{q}_i$, we calculate projections $\mathbf{p}_{j}$s of these points on a tangent plane $\mathcal{T}_i$ described by a unit normal $\mathbf{n}_i$ (estimated in Sec.~\ref{sec:normal_estimation}) as:
\begin{equation}
\mathbf{p}_{j} = \mathbf{x}_{j} - ((\mathbf{x}_{j} - \mathbf{q}_i)\cdot \mathbf{n}_i)\cdot \mathbf{n}_i, \quad  j\in\{1, ..., K\}.
\end{equation}
Fig.~\ref{fig:proj_order_conv_horiz} (a) illustrates the orthogonal projection of neighboring points on a ring.
\vspace{-3mm}
\subsubsection{Counterclockwise Ordering}\vspace{-1mm}
Firstly, we use the geometric definition of the dot product to compute the angle between two vectors $\mathbf{c}$ (i.e., starts from the query point $\mathbf{q}_i$ and connects with a randomly starting point, such as $\mathbf{p}_1$) and $\mathbf{p}_j - \mathbf{q}_i$ (i.e., starts from the query point $\mathbf{q}_i$ and connects with other neighboring points $\mathbf{p}_j$):
\begin{equation}
cos(\theta_{\mathbf{p}_{j}}) = \frac{\mathbf{c}\cdot (\mathbf{p}_j - \mathbf{q}_i)}{||\mathbf{c}|| ||\mathbf{p}_j - \mathbf{q}_i||}.
\end{equation}

We know that $cos(\theta_{\mathbf{p}_{j}})$ lies in $[-1,1]$, which corresponds to angles between $[0^{\circ}, 180^{\circ}]$. In order to sort the neighboring points around the query point between $[0^{\circ}, 360^{\circ})$, we must to decide which semicircle the considered point $\mathbf{p}_{j}$ belongs to as follows:
\begin{equation}
sign_{\mathbf{p}_j} = (\mathbf{c}\times (\mathbf{p}_j - \mathbf{q}_i))\cdot \mathbf{n}_i,
\end{equation}
where $sign_{\mathbf{p}_j} \geq 0$ is $\theta_{\mathbf{p}_{j}} \in [0^{\circ}, 180^{\circ}]$, and $sign_{\mathbf{p}_j} < 0$ is $\theta_{\mathbf{p}_{j}} \in (180^{\circ}, 360^{\circ})$.

Then, we can recompute the cosine value of the angle as:
\begin{equation}
\angle_{\mathbf{p}_{j}} = \left\{
        \begin{array}{ll}
            - cos(\theta_{\mathbf{p}_{j}}) - 2 & \quad sign_{\mathbf{p}_j} < 0 \\
            cos(\theta_{\mathbf{p}_{j}}) & \quad sign_{\mathbf{p}_j} \geq 0.
        \end{array}
    \right.
\end{equation}
Now the values of the angles lie in $(-3, 1]$, which maps angles between $[0^{\circ}, 360^{\circ})$.

Finally, we sort neighboring points $\mathbf{x}_{j}$ by descending the value of $\angle_{\mathbf{p}_{j}}$ to obtain the counterclockwise order. Fig.~\ref{fig:proj_order_conv_horiz} (b) illustrates the process of ordering in a local neighborhood. The neighboring points can be ordered in the clockwise manner, if we sort neighboring points $\mathbf{x}_{j}$ by ascending the value of $\angle_{\mathbf{p}_{j}}$.

Our experiments show in Sec.~\ref{sec:ablation} that ordering points in the local regions is an important step in our framework and our model achieves better classification accuracy with ordered points than without ordering them.

\subsection{Annular Convolution on Rings}
\label{sec:annular_conv}
Through the previous computation, we have the ordered neighbors represented as an array $[\mathbf{x}_{1}, \mathbf{x}_{2}, ... , \mathbf{x}_{K}]$. In order to develop the \emph{annular convolution}, we need to loop the array of neighbors with respect to the size of the kernel (e.g., $1\times3$, $1\times5$, ...) on each ring. For example, if the convolutional kernel size is $1\times3$, we need to take the first two neighbors and concatenate them with the ending elements in the original array to construct a new circular array $[\mathbf{x}_{1}, \mathbf{x}_{2}, ... , \mathbf{x}_{K}, \mathbf{x}_{1}, \mathbf{x}_{2}]$. Then, we can perform the standard convolutions on this array as shown in Fig.~\ref{fig:proj_order_conv_horiz} (c).

There are some nice properties of the proposed annular convolutions as follows: (1) The annular convolution is invariant to the orientation of the local patch. That is because the neighbors are organized and ordered in a closed loop in each ring by concatenating the beginning with the end of the neighboring points' sequence. Therefore, we can order neighbors based on any random starting position, which does not negatively affect the convolution results. Compared with some previous convolutions defined on 3D shapes~\cite{boscaini2016learning,Xu_2017_ICCV,tatarchenko2018tangent}, they all need to compute the real principal curvature direction as the reference direction to define the local patch operator, which is not robust and cumbersome. In particular, 3D shapes have large areas of flat and spherical regions, where the curvature directions are arbitrary. (2) As we know, in reality, the normal direction flipping issues are widely existing in point clouds, especially the large-scale scene datasets. Under the annular convolution strategy, no matter the neighboring points are ordered in clockwise or counterclockwise manner, the results are the same. (3) Another advantage of annular convolution is that we can define an arbitrary kernel size, instead of just $1\times1$ kernels~\cite{qi2017pointnet,qi2017pointnet++}. Therefore, the annular convolution can provide the ability to learn the relationship between ordered points inside each ring as shown in Fig.~\ref{fig:proj_order_conv_horiz} (c).

Annular convolutions can be applied on both regular and dilated rings. By applying annular convolutions with the same kernel size on different rings, we can cover and convolve larger areas by using the dilated structure, which helps us to learn larger spatial contextual information in the local regions. The importance of annular convolutions is shown in the ablation study in Sec.~\ref{sec:ablation}.
\vspace{-2mm}
\subsection{Pooling on Rings}\vspace{-1mm}
After applying a set of annular convolutions sequentially, the resulting convolved features encode information about its closest neighbors in each ring as well as spatial remoteness from a query point. Then we aggregate the convolved features across all neighbors on each ring separately. We apply the max-pooling strategy in our framework. Our proposed ring-based scheme allows us to aggregate more discriminative features. The extracted max-pooled features contain the encoded information about neighbors and the relationship between them in the local region, unlike the pooling scheme in PointNet++~\cite{qi2017pointnet++}, where each neighbor is considered independently from its neighbors. In our pooling process, the non-overlapped regions (rings) will aggregate different types of features in each ring, which can uniquely describe each local region (ring) around the query point. The multi-scale approach in PointNet++ does not guarantee this and might aggregate the same features at different scales, which is redundant information for a network. The (regular and dilated) ring-based scheme helps to avoid extracting duplicate information but rather promotes extracting multi-level information from different regions (rings). This provides a network with more diverse features to learn from. After aggregating features at different rings, we concatenate and feed them to another abstract layer to further learn hierarchical features.
\vspace{-1mm}
\section{A-CNN Architecture}\vspace{-1mm}
Our proposed A-CNN model follows a design where the hierarchical structure is composed of a set of abstract layers. Each abstract layer consists of several operations performed sequentially and produces a subset of input points with newly learned features. Firstly, we subsample points by using Farthest Point Sampling (FPS) algorithm~\cite{moenning2003fast} to extract centroids randomly distributed on the surface of each object. Secondly, our constraint-based k-NN extracts neighbors of a centroid for each local region (i.e., regular / dilated rings) and then we order neighbors in a counterclockwise manner using projection. Finally, we apply sequentially a set of annular convolutions on the ordered points and max-pool features across neighbors to produce new feature vectors, which uniquely describe each local region.

Given the point clouds of 3D shapes, our proposed end-to-end network is able to classify and segment the objects. 
In the following, we discuss the classification and segmentation network architectures on 3D point clouds.

\begin{figure*}[t]
\begin{center}
  \includegraphics[width=0.9\linewidth]{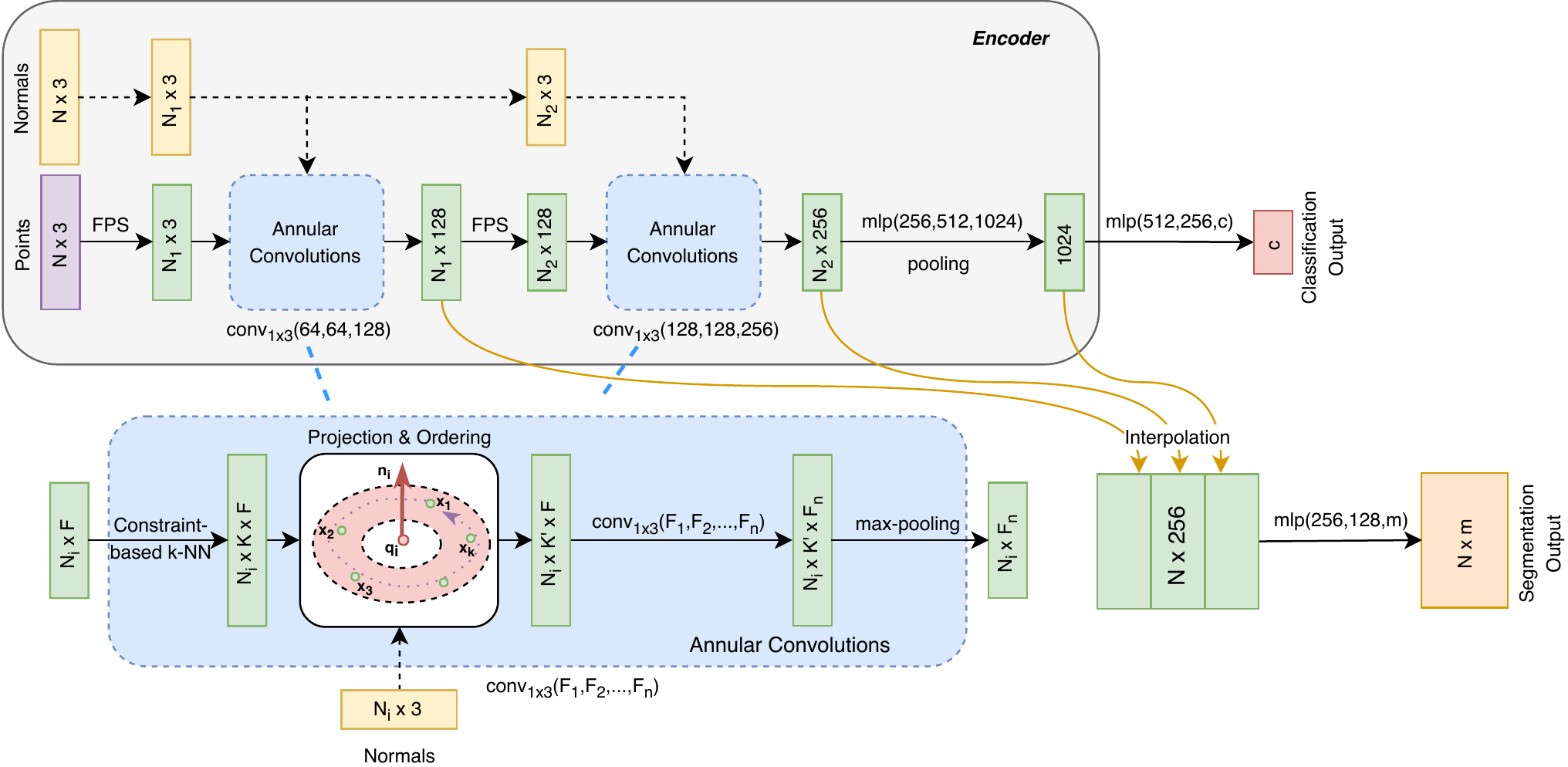}\vspace{-5mm}
\end{center}
\caption{The architecture of A-CNN. Both classification and segmentation networks share encoder part for the feature extraction. Normals are used only to determine the order of neighboring points in the local regions (dashed arrows mean no backpropagation during training) and not used as additional features, unless it is mentioned explicitly in the experiments. $N$, $N_1$, $N_2$ (where $N > N_1 > N_2$) are the numbers of points as input, after  the first and second layer, respectively. $K$ and $K'$ are the unordered and ordered points inside the local rings, respectively. $c$ is the number of classification classes. $m$ is the number of segmentation classes. ``FPS'' stands for Farthest Point Sampling algorithm. ``mlp'' stands for multi-layer perceptron. $conv_{1\times3}(F_1, F_2, ..., F_n)$ stands for annular convolutions with the kernel size $1 \times 3$ applied sequentially with corresponding feature map sizes $F_i, i\in{1, ..., n}$.}\vspace{-3mm}
\centering
\label{fig:arch}
\end{figure*}
\vspace{-1mm}
\subsection{Classification Network}\vspace{-1mm}
The classification network is illustrated at the top of Fig.~\ref{fig:arch}. It consists of two major parts: encoder and classification. The encoder extracts features from each ring independently inside every layer and concatenates them at the end to process further to extract high-level features. The proposed architecture includes both \emph{regular rings} and \emph{dilated rings}. We end up using two rings per layer, because it gives us pretty good experimental results as shown in the Sec.~\ref{sec:experiments}. It can be easily extended to more than two rings per layer, if necessary.

We use regular rings in the first layer and dilated rings in the second layer in the encoder. Annular convolutions with the kernel sizes $1\times3$ and stride 1 are applied in the first two layers, followed by a batch normalization~\cite{ioffe2015batch} (BN) and a rectified linear unit~\cite{nair2010rectified} (ReLU). Different rings of the same query point are processed in parallel. Then, the aggregated features from each ring concatenate together to propagate to the next layer. The last layer in the encoder performs convolutions with kernel sizes $1\times1$ followed by BN and ReLU layers, where only spatial positions of the sampled points are considered. After that aggregated high-level features are fed to the set of fully-connected layers with integrated dropout~\cite{JMLR:v15:srivastava14a} and ReLU layers to calculate probability of each class. The output size of the classification network is equal to the number of classes in the dataset.
\vspace{-1mm}
\subsection{Segmentation Network}\vspace{-1mm}
The segmentation network shares encoder part with the classification network as shown in Figure~\ref{fig:arch}. In order to predict the segmentation label per point, we need to upsample the sampled points in the encoder back to the original point cloud size. As pointed out by~\cite{yu2018pu}, the consecutive feature propagation proposed by~\cite{qi2017pointnet++} is not the most efficient approach. Inspired from~\cite{yu2018pu}, we propagate features from different levels from the encoder directly to the original point cloud size, and concatenate them by allowing the network to learn the most important features from different levels as well as to learn the relationship between them.

The output of each level has different sizes due to the hierarchical feature extractions, so we have to restore hierarchical features from each level back to the original point size by using an interpolation method~\cite{qi2017pointnet++}. The interpolation method is based on the inverse squared Euclidean distance weighted average of the three nearest neighbors as:\vspace{-2mm}
\begin{equation}
f^{(l+1)}(\mathbf{x}) = \sum^3_{j=1} f^{(l)}(\mathbf{x}_j)\frac{w_j(\mathbf{x})}{\sum^3_{j=1} w_j(\mathbf{x})},\vspace{-1.5mm}
\end{equation}
where $w_j(\mathbf{x})=\frac{1}{d(\mathbf{x},\mathbf{x}_j)^2}$ is an inverse squared Euclidean distance weight.

Then, we concatenate upsampled features from different levels and pass them through $1\times1$ convolution to reduce feature space and learn the relationship between features from different levels. Finally, the segmentation class distribution for each point is calculated.
\vspace{-1mm}
\section{Experiments}\vspace{-1mm}
\label{sec:experiments}
We evaluate our A-CNN model on various tasks such as point cloud classification, part segmentation, and large-scale scene segmentation. In the following subsections, we demonstrate more details on each task. It is noted that for the comparison experiments, best results in the tables are shown in bold font.

All models in this paper are trained on a single NVIDIA Titan Xp GPU with 12 GB GDDR5X. The training time of our model is faster than that of PointNet++ model. More details about the network configurations, training settings and timings in our experiments can be found in Sec.~\ref{sec:suppl_training_details} and Tab.~\ref{table:network_configs} of Supplementary Material. The source code of the framework will be made available later.
\vspace{-1mm}
\subsection{Point Cloud Classification}\vspace{-1mm}
We evaluate our classification model on two datasets: \textit{ModelNet10} and \textit{ModelNet40}~\cite{wu20153d}. ModelNet is a large-scale 3D CAD model dataset. \textit{ModelNet10} is a subset of ModelNet dataset that consists of 10 different classes with 3991 training and 908 testing objects. \textit{ModelNet40} includes 40 different classes with 9843 objects for training and 2468 objects for testing. Point clouds with 10,000 points and normals are sampled from meshes, normalized into a unit sphere, and provided by~\cite{qi2017pointnet++}.

For experiments on \textit{ModelNet10} and \textit{ModelNet40}, we sample 1024 points with normals, where normals are only used to order points in the local region. For data augmentation, we randomly scale object sizes, shift object positions, and perturb point locations. For better generalization, we apply point shuffling in order to generate different centroids for the same object at different epochs.

In Tab.~\ref{table:modelnet40}, we compare our method with several state-of-the-art methods in the shape classification results on both \textit{ModelNet10} and \textit{ModelNet40} datasets. Our model achieves better accuracy among the point cloud based methods (with 1024 points), such as PointNet~\cite{qi2017pointnet}, PointNet++~\cite{qi2017pointnet++} (5K points + normals), Kd-Net (depth 15)~\cite{klokov2017escape}, Pointwise CNN~\cite{hua2018pointwise}, KCNet~\cite{shen2018mining}, PointGrid~\cite{le2018pointgrid}, PCNN~\cite{atzmon2018point}, and PointCNN~\cite{li2018pointcnn}. Our model is slightly better than Point2Sequence~\cite{liu2018point2sequence} on \emph{ModelNet10} and shows comparable performance on \emph{ModelNet40}.

Meanwhile, our model performs better than other volumetric approaches, such as O-CNN~\cite{wang2017cnn} and AO-CNN~\cite{Wang-2018-AOCNN}; while we are a little worse than SO-Net~\cite{li2018so}, which uses denser input points, i.e., 5000 points with normals as the input (1024 points in our A-CNN); MVCNN-MultiRes~\cite{qi2016volumetric}, which uses multi-view 3D volumes to represent an object (i.e., 20 views of $30 \times 30 \times 30$ volume); the VRN Ensemble~\cite{brock2016generative}, which involves an ensemble of six models.

We also provide some feature visualization results in Sec.~\ref{sec:suppl_feature_visualization} of Supplementary Material, including global feature (e.g., t-SNE clustering) visualization and local feature (e.g., the magnitude of the gradient per point) visualization.
\vspace{-2mm}
\begin{table}[t]
\centering
\caption{Classification results on \emph{ModelNet10} and \emph{ModelNet40} datasets. AAC is accuracy average class, OA is overall accuracy.}\vspace{-2mm}
\scalebox{0.75}{
\begin{tabular}{l|c|c|c|l}
\hline
 & \multicolumn{2}{c}{\textit{ModelNet10}} & \multicolumn{2}{c}{\textit{ModelNet40}} \\ \hline
 & AAC & OA & AAC & OA \\ \hline
 \multicolumn{5}{c}{\textit{different methods with additional input or more points}} \\ \hline
 AO-CNN~\cite{Wang-2018-AOCNN} & - & - & - & 90.5  \\
 O-CNN~\cite{wang2017cnn} & - & - & - & 90.6  \\
 PointNet++~\cite{qi2017pointnet++} & - & - & - & 91.9 \\
 SO-Net~\cite{li2018so} & 95.5 & 95.7 & 90.8 & 93.4 \\
 MVCNN-MultiRes~\cite{qi2016volumetric} & - & - & 91.4 & 93.8  \\
 VRN Ensemble~\cite{brock2016generative} & - & 97.1 & - & 95.5 \\ \hline
 \multicolumn{5}{c}{\textit{point cloud based methods with 1024 points}} \\ \hline
 PointNet~\cite{qi2017pointnet} & - & - & 86.2 & 89.2   \\
 Kd-Net (depth 15)~\cite{klokov2017escape} & 93.5 & 94.0 & 88.5 & 91.8  \\
 Pointwise CNN~\cite{hua2018pointwise} & - & - & 81.4 & 86.1 \\
 KCNet~\cite{shen2018mining} & - & 94.4 & - & 91.0  \\
 PointGrid~\cite{le2018pointgrid} & - & - & 88.9 & 92.0  \\
 PCNN~\cite{atzmon2018point} & - & 94.9 & - & 92.3  \\
 PointCNN~\cite{li2018pointcnn} & - & - & 88.1 & 92.2 \\
 Point2Sequence~\cite{liu2018point2sequence} & 95.1 & 95.3 & \textbf{90.4} & \textbf{92.6}  \\ \hline
 A-CNN (our) & \textbf{95.3} & \textbf{95.5} & 90.3 & \textbf{92.6}  \\ \hline
\end{tabular}} \vspace{-2mm}
\label{table:modelnet40}
\end{table}
\vspace{-2mm}
\subsection{Point Cloud Segmentation}\vspace{-1mm}

We evaluate our segmentation model on \textit{ShapeNet-part}~\cite{yi2016scalable} dataset. The dataset contains 16,881 shapes from 16 different categories with 50 label parts in total. The main challenge of this dataset is that all categories are highly imbalanced. There are 2048 points sampled for each shape from the dataset, where most shapes contain less than six parts. We follow the same training and testing splits provided in~\cite{qi2017pointnet,yi2016scalable}. For data augmentation, we perturb point locations with the point shuffling for better generalization.

We evaluate our segmentation model with two different inputs. One of the models is trained without feeding normals as additional features and the other model is trained with normals as additional features. The quantitative results are provided in Tab.~\ref{table:s3dis_eval}, where mean IoU (Intersection-over-Union) is reported. The qualitative results are visualized in Fig.~\ref{fig:shapenetpart_eval}.
Our approach with point locations only as input outperforms PointNet~\cite{qi2017pointnet}, Kd-Net~\cite{klokov2017escape}, KCNet~\cite{shen2018mining}, and PCNN~\cite{atzmon2018point}; and shows slightly worse performance comparing to PointGrid~\cite{le2018pointgrid} (volumetric method) and PointCNN~\cite{li2018pointcnn}.
Meanwhile, our model achieves the best performance with the input of point locations and normals, compared with PointNet++~\cite{qi2017pointnet++}, SyncSpecCNN~\cite{yi2017syncspeccnn}, SO-Net~\cite{li2018so}, SGPN~\cite{wang2018sgpn}, O-CNN~\cite{wang2017cnn}, RSNet~\cite{huang2018recurrent}, and Point2Sequence~\cite{liu2018point2sequence}. 
The more detailed quantitative results (e.g., per-category IoUs) and more visualization results are provided in Sec.~\ref{sec:suppl_more_experimental_results} of Supplementary Material.

\setcounter{figure}{-7}
\begin{figure}[t]
\centering
\vspace{-4mm}
\subfloat{\includegraphics[width=0.06\linewidth,trim={0 0 0 0}, clip]{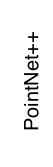} \phantomcaption}
\subfloat{\includegraphics[width=0.18\linewidth,trim={3.5cm 4cm 3.5cm 4cm}, clip]{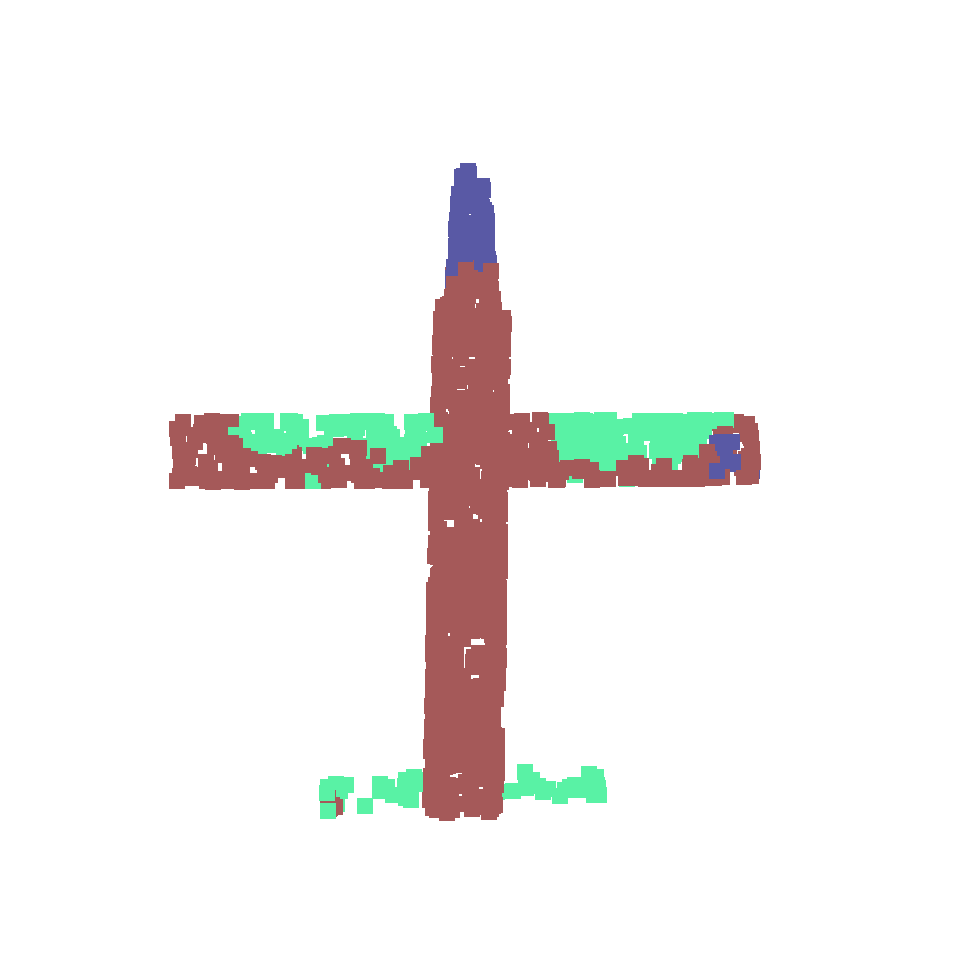} \phantomcaption} \hfill
\subfloat{\includegraphics[width=0.18\linewidth,trim={3.5cm 5cm 3.5cm 4cm}, clip]{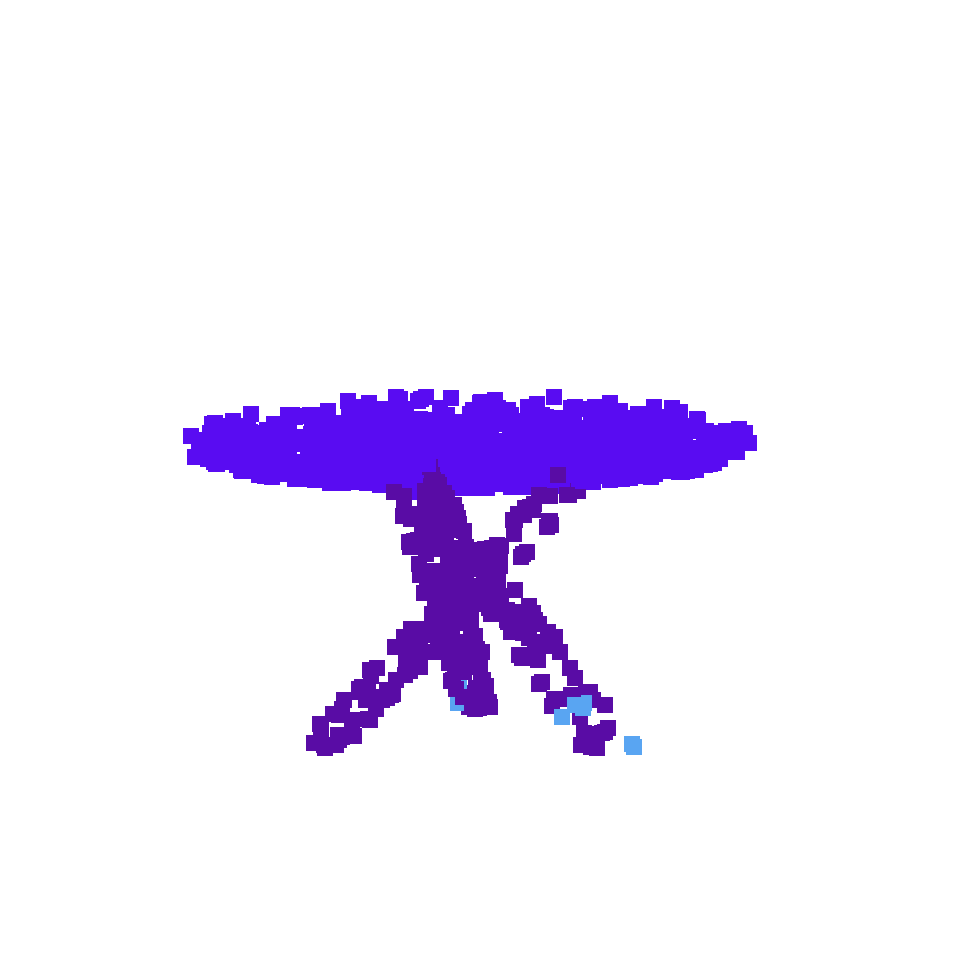} \phantomcaption} \hfill
\subfloat{\includegraphics[width=0.22\linewidth,trim={3.5cm 6cm 3.5cm 4cm}, clip]{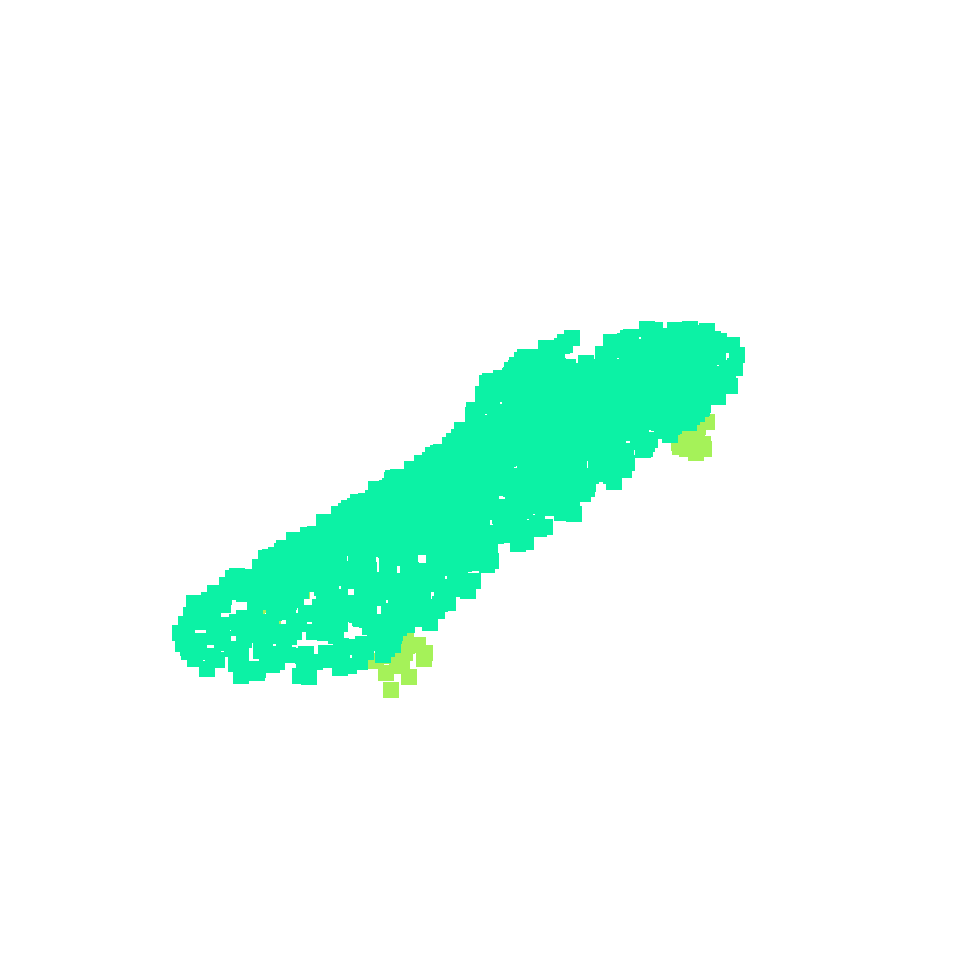} \phantomcaption} \hfill
\subfloat{\includegraphics[width=0.18\linewidth,trim={3.5cm 6cm 5.5cm 4cm}, clip]{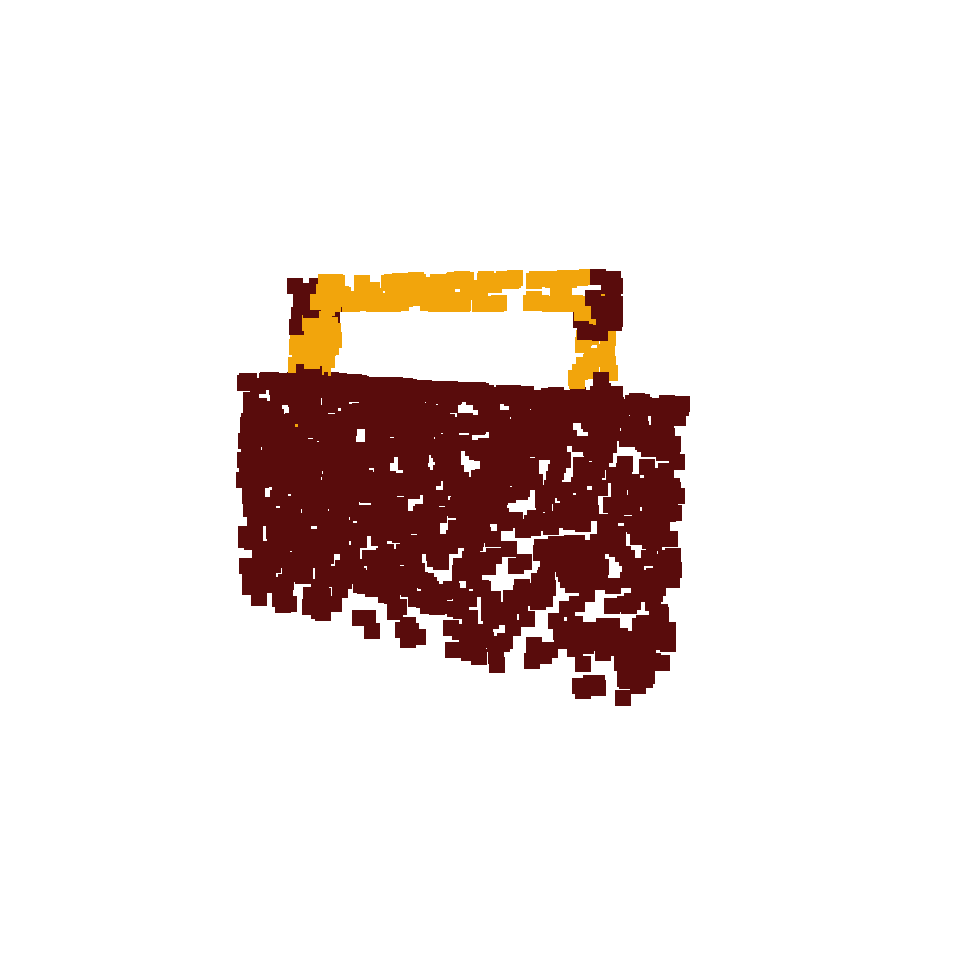} \phantomcaption}
\vspace{-4mm}
\subfloat{\includegraphics[width=0.06\linewidth,trim={0 0 0 0}, clip]{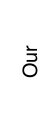} \phantomcaption}
\subfloat{\includegraphics[width=0.18\linewidth,trim={3.5cm 4cm 3.5cm 4cm}, clip]{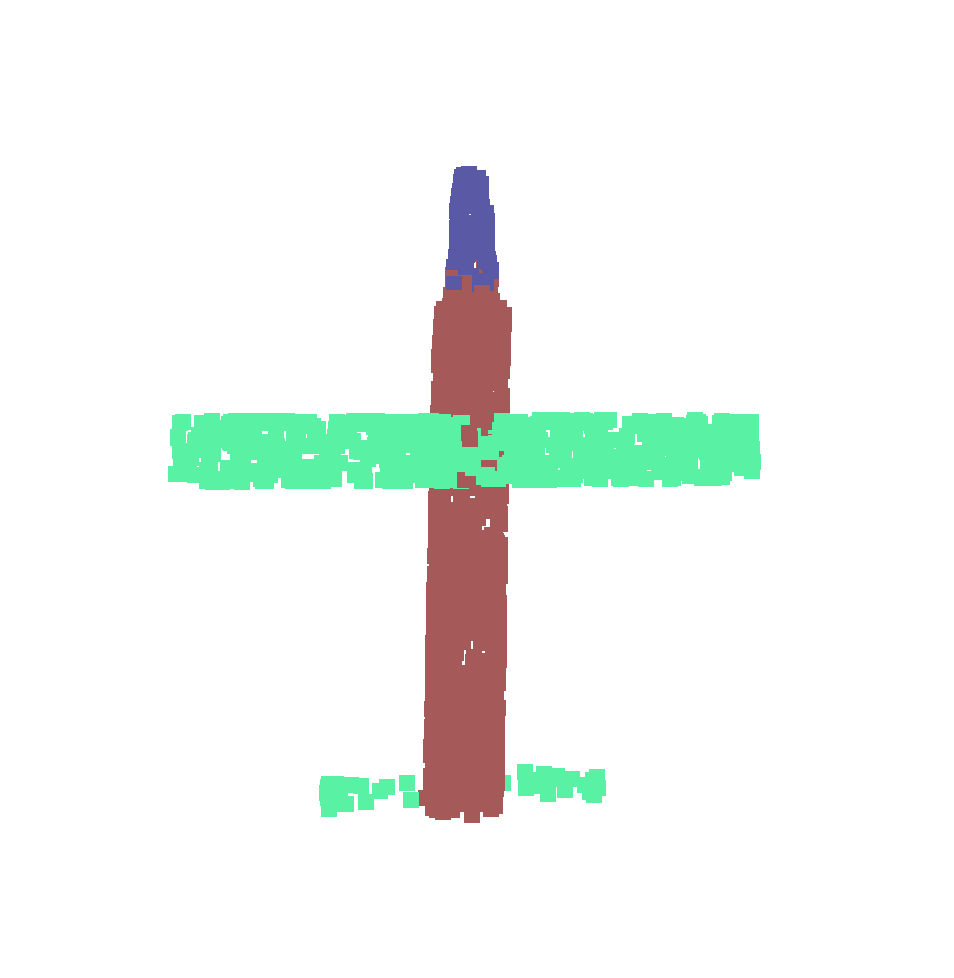} \phantomcaption} \hfill
\subfloat{\includegraphics[width=0.18\linewidth,trim={3.5cm 5cm 3.5cm 4cm}, clip]{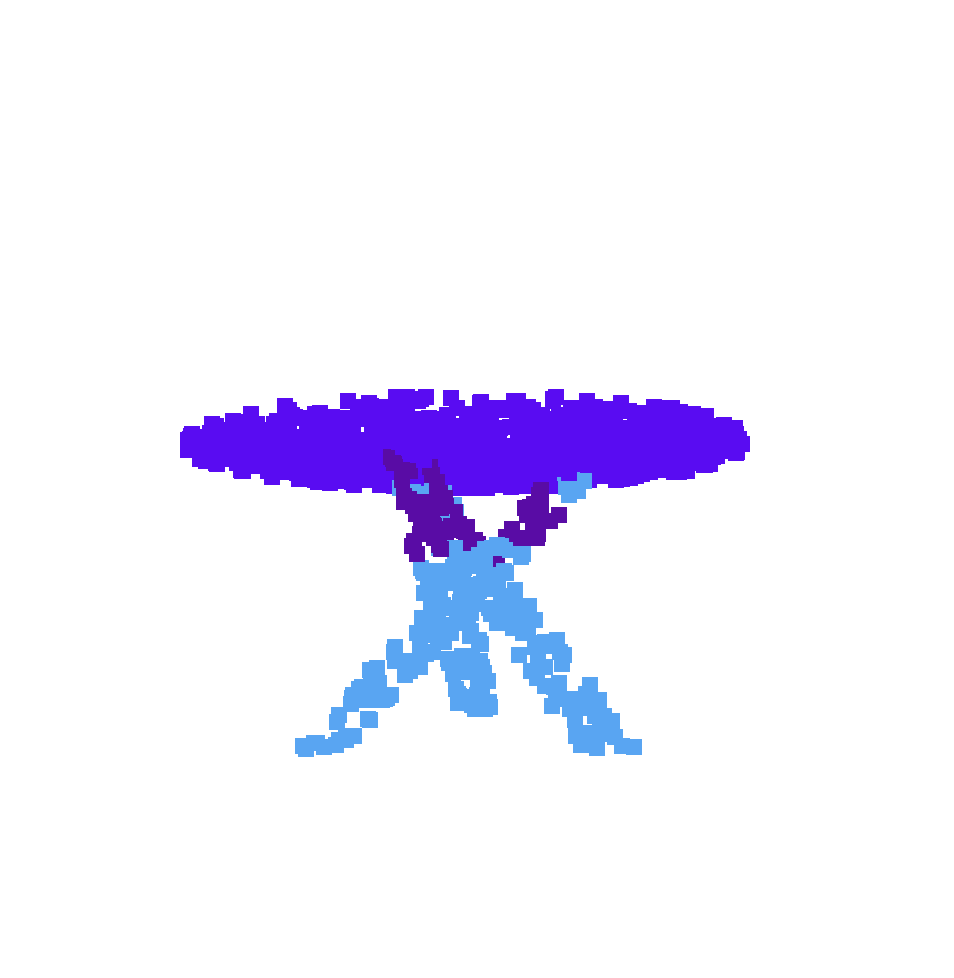} \phantomcaption} \hfill
\subfloat{\includegraphics[width=0.22\linewidth,trim={3.5cm 6cm 3.5cm 4cm}, clip]{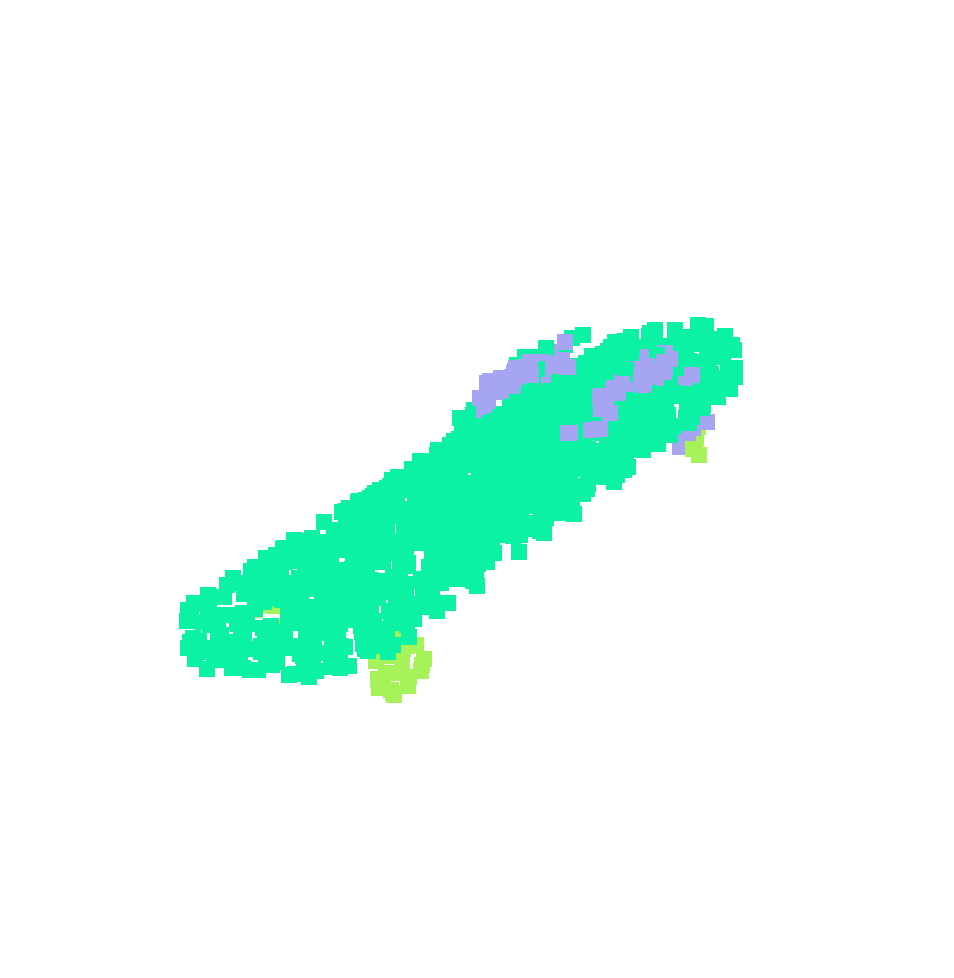} \phantomcaption} \hfill
\subfloat{\includegraphics[width=0.18\linewidth,trim={3.5cm 6cm 5.5cm 4cm}, clip]{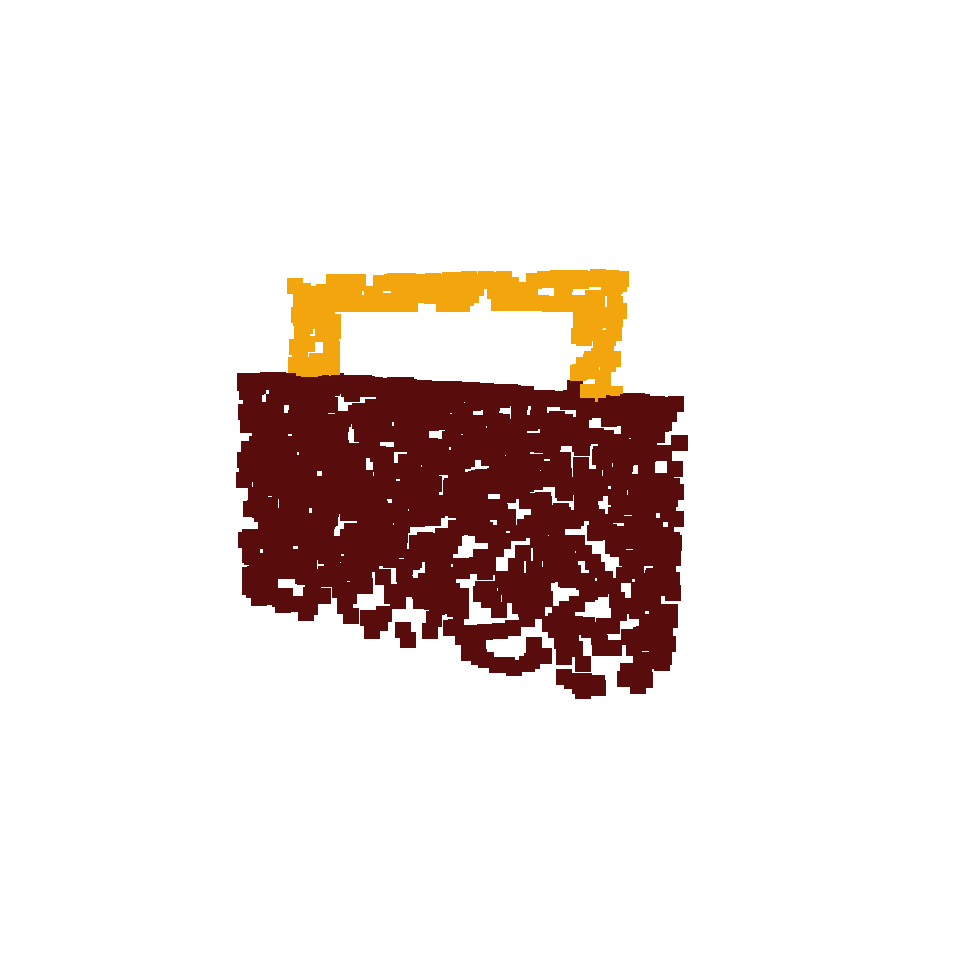} \phantomcaption}
\vspace{-4mm}
\subfloat{\includegraphics[width=0.06\linewidth,trim={0 0 0 0}, clip]{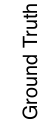} \phantomcaption}
\subfloat[\textbf{Rocket}]{\includegraphics[width=0.18\linewidth,trim={3.5cm 4cm 3.5cm 4cm}, clip]{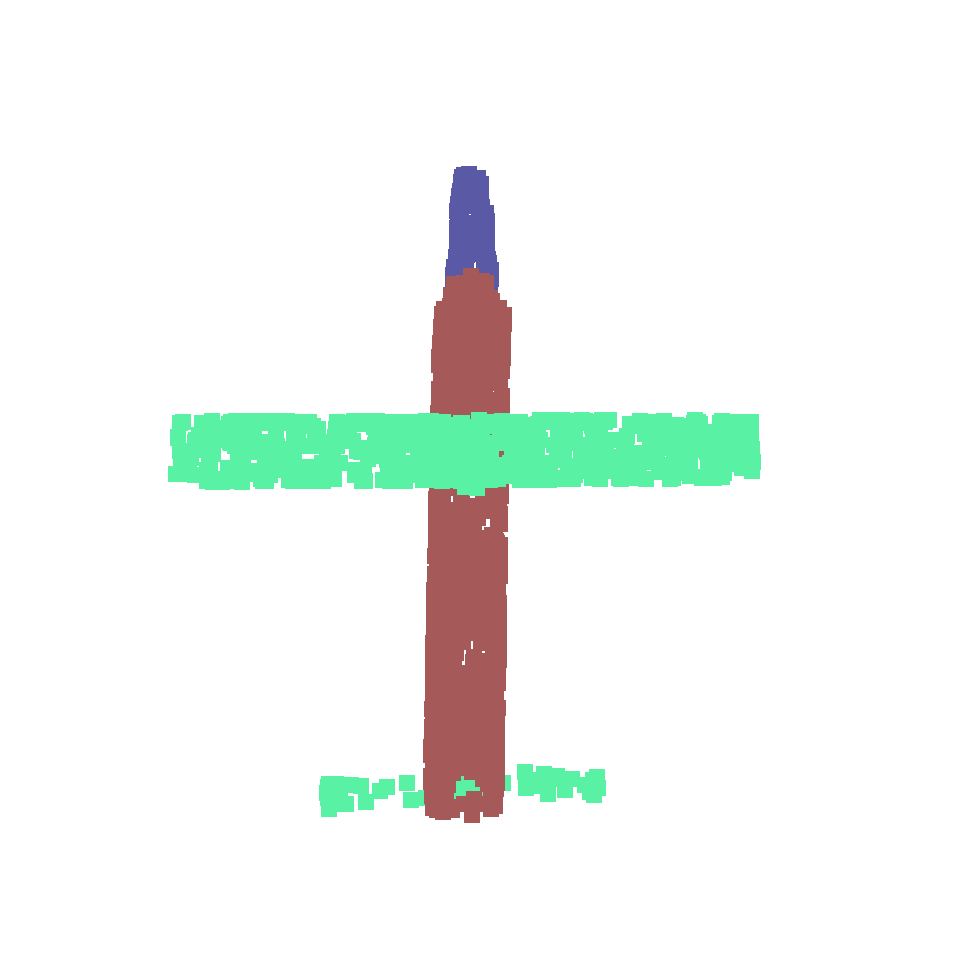}} \hfill
\subfloat[\textbf{Table}]{\includegraphics[width=0.18\linewidth,trim={3.5cm 5cm 3.5cm 4cm}, clip]{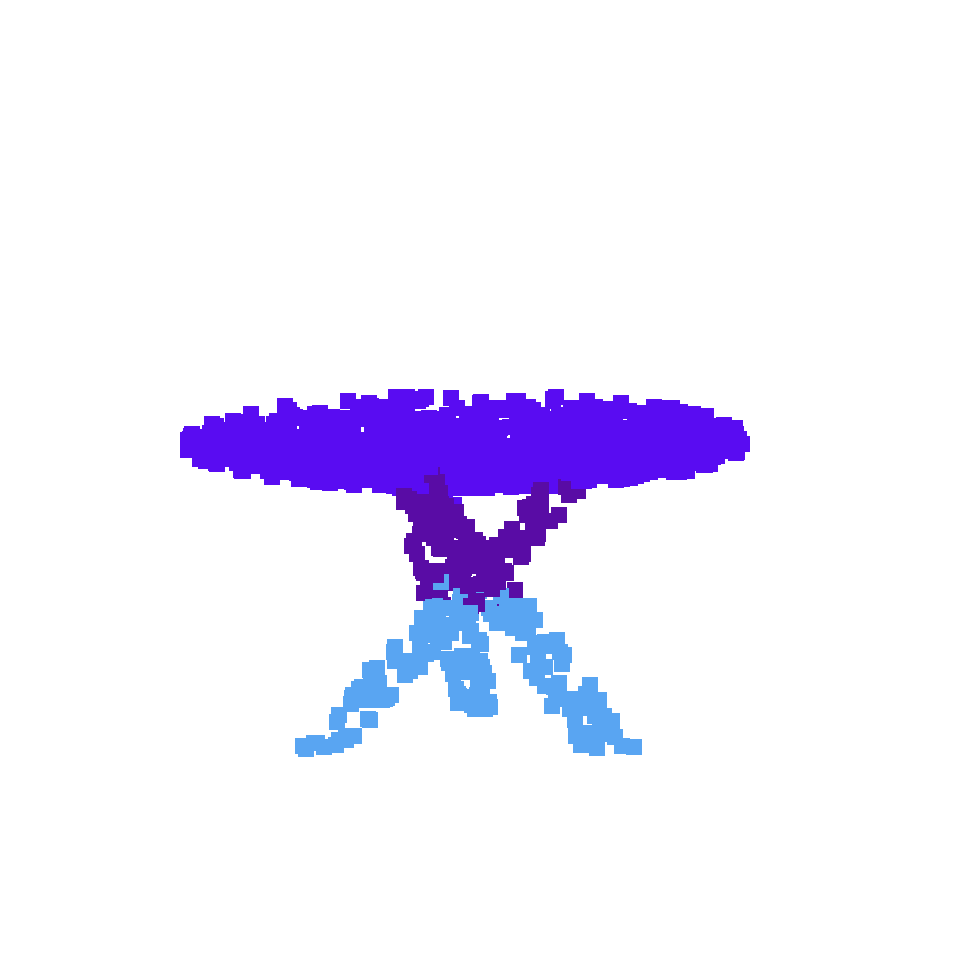}} \hfill
\subfloat[\textbf{Skateboard}]{\includegraphics[width=0.22\linewidth,trim={3.5cm 6cm 3.5cm 4cm}, clip]{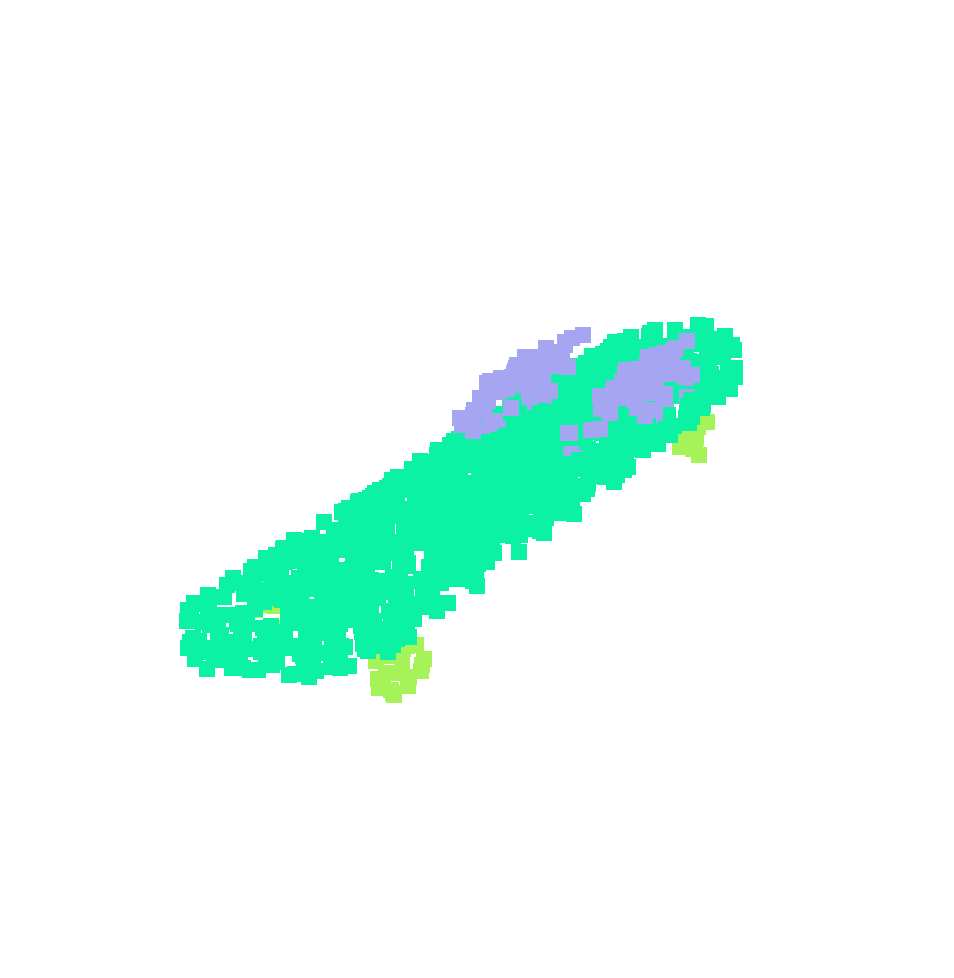}} \hfill
\subfloat[\textbf{Bag}]{\includegraphics[width=0.18\linewidth,trim={3.5cm 6cm 5.5cm 4cm}, clip]{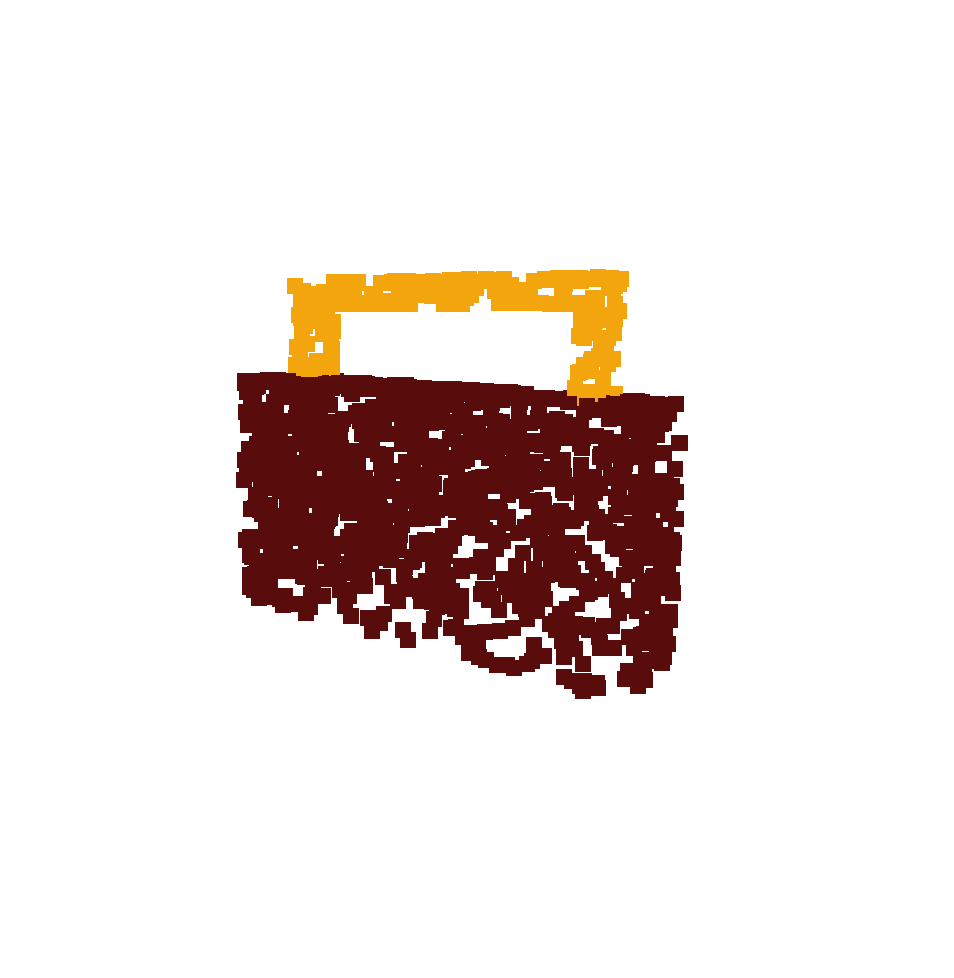}}
\vspace{-3mm}
\caption{Qualitative results on \emph{ShapeNet-part} dataset. We compare our results with PointNet++~\cite{qi2017pointnet++} and ground truth.
}\vspace{-3mm}
\label{fig:shapenetpart_eval}
\end{figure}
\vspace{-1mm}
\subsection{Semantic Segmentation in Scenes}\vspace{-1mm}
\setcounter{figure}{+4}
\begin{figure*}[t]
\centering
\subfloat{\includegraphics[width=0.85\linewidth, clip]{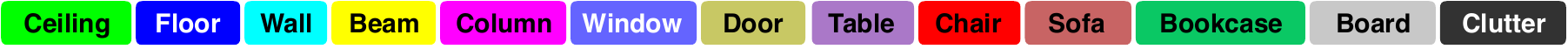} \phantomcaption }
\vspace{-3.5mm}
\subfloat[\textbf{Input}]{\includegraphics[width=0.22\linewidth,trim={0 0 0 0}, clip]{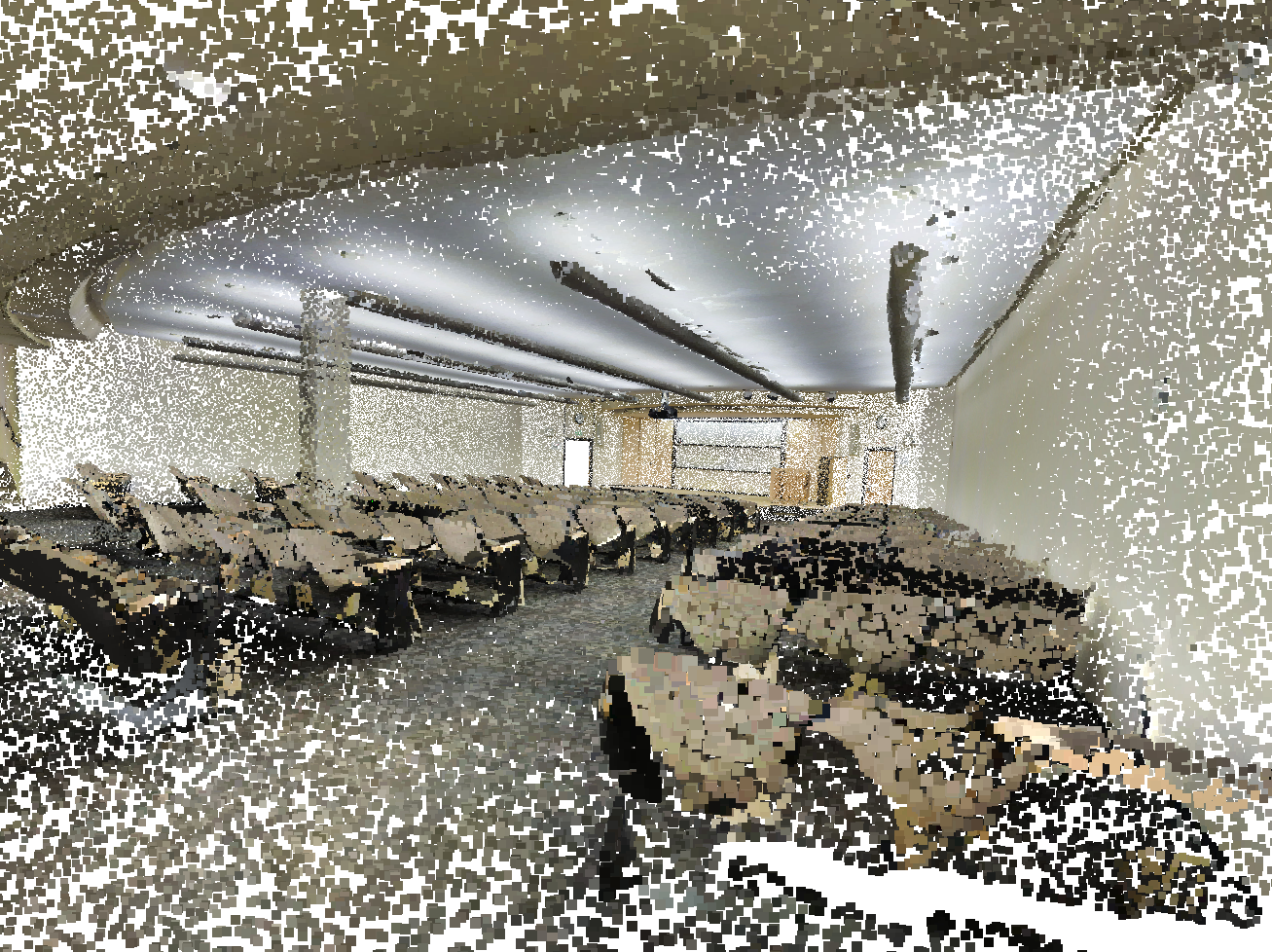} }
\subfloat[\textbf{PointNet~\cite{qi2017pointnet}}] {\includegraphics[width=0.22\linewidth,trim={0 0 0 0}, clip]{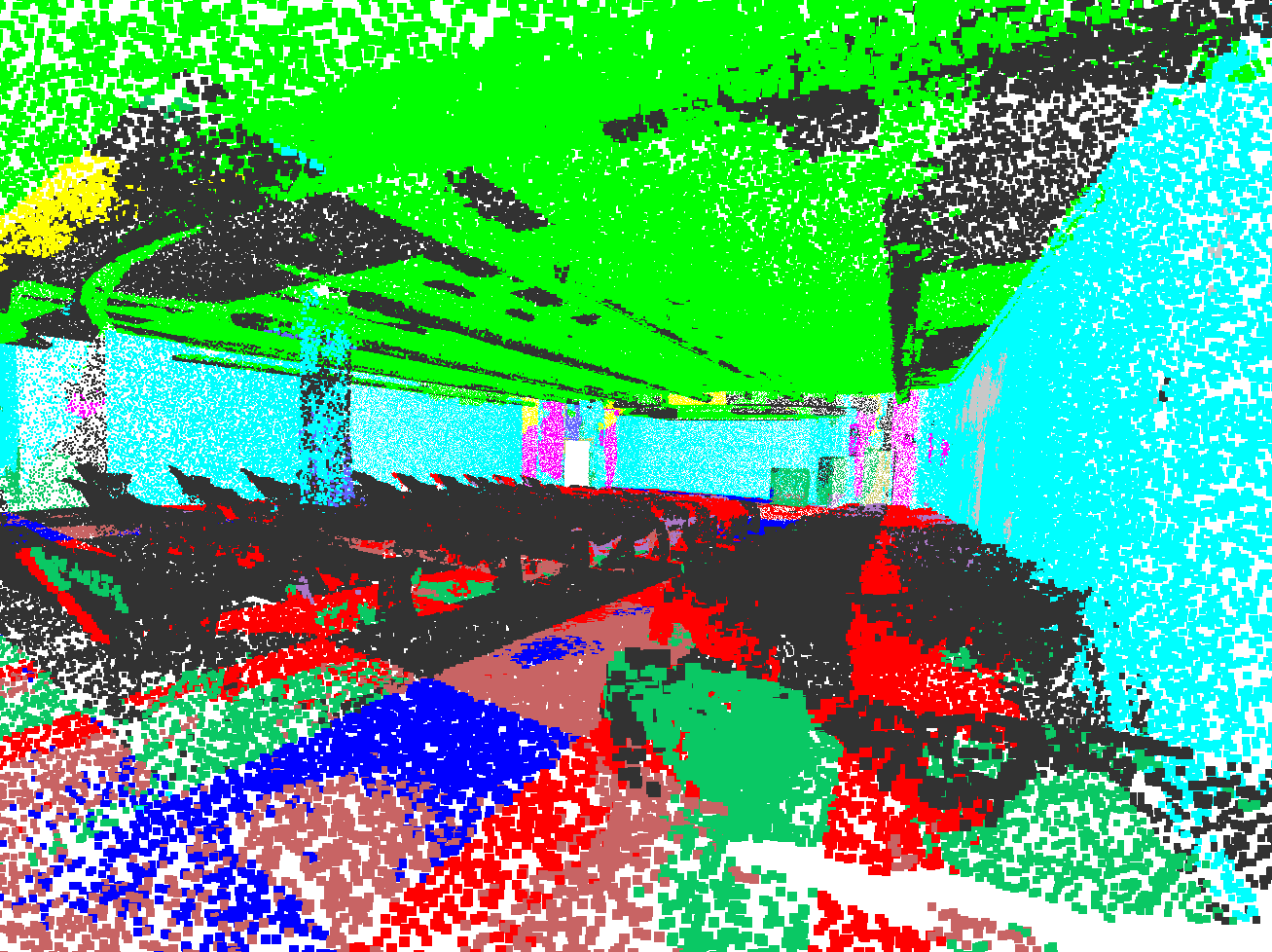} }
\subfloat[\textbf{Our}]{\includegraphics[width=0.22\linewidth,trim={0 0 0 0}, clip]{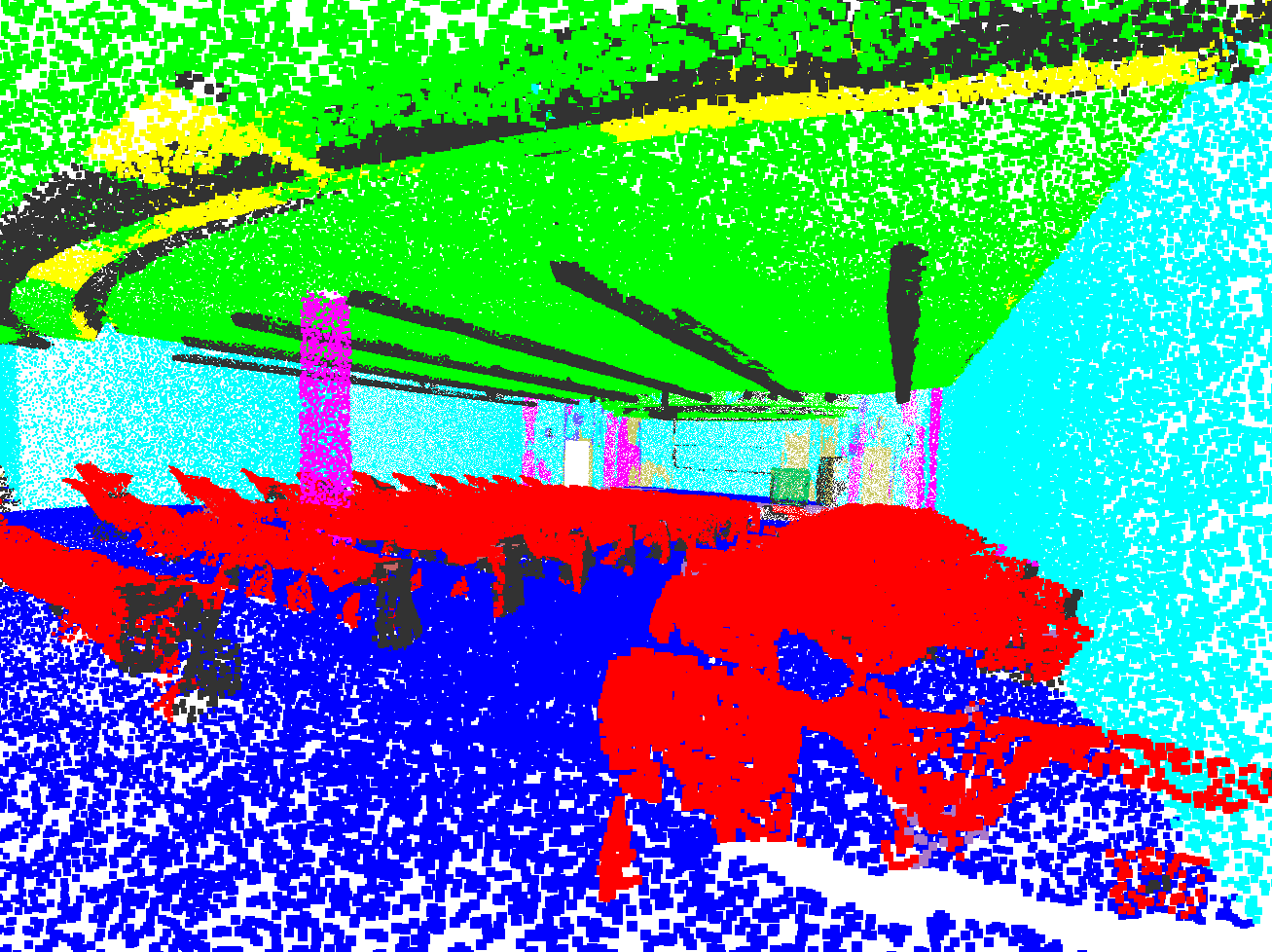} }
\subfloat[\textbf{Ground Truth}]{\includegraphics[width=0.22\linewidth,trim={0 0 0 0}, clip]{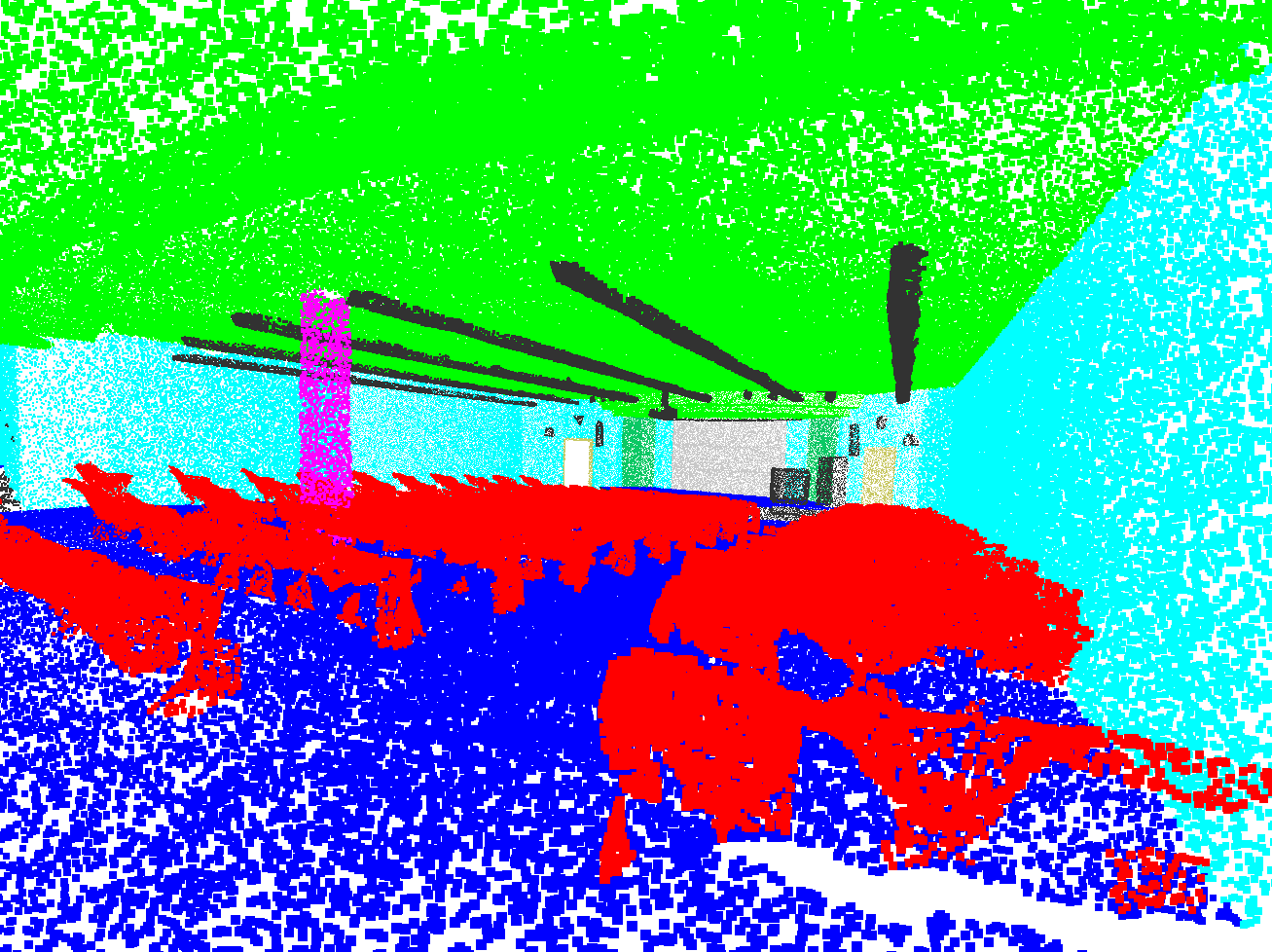} }
\vspace{-3mm}
\caption{Qualitative results on \emph{S3DIS} dataset. We compare our results with PointNet~\cite{qi2017pointnet} and ground truth. 
The auditorium is a challenging room type and appears only in Area 2. Our model produces much better segmentation result, compared with the result of PointNet.}\vspace{-3mm}
\label{fig:s3dis_eval}
\end{figure*}

We also evaluate our segmentation model on two large-scale indoor datasets \emph{Stanford 3D Large-Scale Indoor Spaces (S3DIS)}~\cite{armeni20163d} and \emph{ScanNet}~\cite{dai2017scannet}. \emph{S3DIS} contains 6 large-scale indoor areas with 271 rooms sampled from 3 different buildings, where each point has the semantic label that belongs to one of the 13 categories. 
\emph{ScanNet} includes 1513 scanned indoor point clouds, where each voxel has been labeled with one of the 21 categories.


We employ the same training and testing strategies as PointNet~\cite{qi2017pointnet} on \emph{S3DIS}, where we use 6-fold cross validation over all six areas. The evaluation results are reported in Tab.~\ref{table:s3dis_eval}, and qualitative results are visualized in Fig.~\ref{fig:s3dis_eval}. Our model demonstrates better segmentation results compared with PointNet~\cite{qi2017pointnet}, MS+CU (2)~\cite{engelmann2017exploring}, G+RCU~\cite{engelmann2017exploring}, 3P-RNN~\cite{Ye_2018_ECCV}, SPGraph~\cite{Landrieu2017LargescalePC}, and TangentConv~\cite{tatarchenko2018tangent}. However, our model performs slightly worse than PointCNN~\cite{li2018pointcnn} due to their non-overlapping block sampling strategy with paddings which we do not use. 
Meanwhile, our approach shows the best segmentation results on \emph{ScanNet}~\cite{dai2017scannet} and achieves the state-of-the-art performance, compared with PointNet~\cite{qi2017pointnet}, PointNet++~\cite{qi2017pointnet++}, TangentConv~\cite{tatarchenko2018tangent}, and PointCNN~\cite{li2018pointcnn} according to Tab.~\ref{table:s3dis_eval}.

More qualitative visualization results and data preparation details on both datasets are provided in Sec.~\ref{sec:suppl_data_preparation_details} and Sec.~\ref{sec:suppl_more_experimental_results}, respectively, of Supplementary Material and Video.\vspace{-2mm}

\begin{table}[h]
\centering
\small
\caption{Segmentation results on \emph{ShapeNet-part}, \emph{S3DIS}, and \emph{ScanNet}. ``mean'' is mean IoU (\%), OA is overall accuracy. }\vspace{-2mm}
\scalebox{0.75}{%
\begin{tabular}{@{}l|c|c|c|c}
\hline
			& \multicolumn{2}{c|}{\emph{ShapeNet-part}} & \emph{S3DIS} & \emph{ScanNet} \\ \hline
			& \multicolumn{1}{c|}{\begin{tabular}[c]{@{}c@{}}\emph{without} \emph{normals}\end{tabular}} & \begin{tabular}[c]{@{}c@{}}\emph{with} \emph{normals}\end{tabular} & \multicolumn{1}{c|}{\multirow{2}{*}{OA}} & \multicolumn{1}{c}{\multirow{2}{*}{OA}} \\ \cline{2-3}
             & mean & mean & \multicolumn{1}{c|}{} & \multicolumn{1}{c}{} \\ \hline
PointNet~\cite{qi2017pointnet} & 83.7 & - & 78.5 &  73.9  \\
PointNet++~\cite{qi2017pointnet++} & - & 85.1 & - & 84.5 \\
SyncSpecCNN~\cite{yi2017syncspeccnn} & - & 84.7 & - & - \\
O-CNN~\cite{wang2017cnn} & - & 85.9 & - & - \\
Kd-Net~\cite{klokov2017escape} & 82.3 & - & - & - \\
KCNet~\cite{shen2018mining}    & 84.7 & - & - & -\\
SO-Net~\cite{li2018so}     & - & 84.9 & - & - \\
SGPN~\cite{wang2018sgpn}  &  - & 85.8 & - & - \\
MS+CU (2)~\cite{engelmann2017exploring} & - & - & 79.2 & -  \\
G+RCU~\cite{engelmann2017exploring}   & - & - & 81.1  & -  \\
RSNet~\cite{huang2018recurrent} & - & 84.9 & - & - \\
3P-RNN~\cite{Ye_2018_ECCV}  & - & - & 86.9  & -  \\
SPGraph~\cite{Landrieu2017LargescalePC} & - & - & 85.5 & -  \\
TangentConv~\cite{tatarchenko2018tangent} & - & - & \textbf{*} & 80.9 \\
PCNN~\cite{atzmon2018point}       & 85.1 &  - & - & - \\
Point2Sequence~\cite{liu2018point2sequence} & - & 85.2 & - & - \\
PointGrid~\cite{le2018pointgrid} & \textbf{86.4} & - & - & - \\
PointCNN~\cite{li2018pointcnn}     & 86.1 & - & \textbf{88.1} & 85.1 \\ \hline
A-CNN (our)  & 85.9 & \textbf{86.1} & 87.3 & \textbf{85.4}  \\ \hline
\end{tabular}
} 
\label{table:s3dis_eval}
 \begin{tablenotes}
      \centering
      \footnotesize 
      \item Note: \textbf{*} TangentConv~\cite{tatarchenko2018tangent} OA on \emph{S3DIS} Area 5 is 82.5$\%$ (as reported in their paper), which is worse compared with our OA of 85.5$\%$.\vspace{-3mm}
    \end{tablenotes}
\end{table}

\vspace{-1mm}
\subsection{Ablation Study} \vspace{-1mm}
\label{sec:ablation}

The goal of our ablation study is to show the importance of the proposed technique components (in Sec.~\ref{sec:method}) in our A-CNN model. We evaluate three proposed components, such as rings without overlaps (Sec.~\ref{sec:rings}), ordering (Sec.~\ref{sec:project_order}), and annular convolution (Sec.~\ref{sec:annular_conv}) on the classification task of \emph{ModelNet40} dataset as shown in Tab.~\ref{table:ablation}. In the first experiment, we replace our proposed constraint-based k-NN on ring regions with ball query in~\cite{qi2017pointnet++}, but keep ordering and annular convolutions on. In the second and third experiments, we turn off either annular convolutions or ordering, respectively; and keep the rest two components on. Our experiments show that the proposed ring-shaped scheme contributes the most to our model. It is because multi-level rings positively affect annular convolutions. Finally, A-CNN model with all three components (i.e., rings without overlaps, ordering, and annular convolutions) achieves the best results. We also discover that reducing overlap / redundancy in multi-scale scheme can improve existing methods. We evaluate the original PointNet++~\cite{qi2017pointnet++} with and without overlap as shown in Sec.~\ref{sec:suppl_ball_vs_ring_comparison} of Supplementary Material.\vspace{-2mm}
\begin{table}[h]
\centering
\caption{Ablation experiments on \emph{ModelNet40} dataset. AAC is accuracy average class, OA is overall accuracy.}\vspace{-2mm}
\scalebox{0.75}{
\begin{tabular}{l|l|l} 
\hline
 & AAC & OA \\ \hline
 A-CNN (without rings / with overlap) & 89.2 & 91.7 \\
 A-CNN (without annular conv.) & 89.2 & 91.8 \\
 A-CNN (without ordering) & 89.6 & 92.0 \\ \hline
 A-CNN (with all components) & \textbf{90.3} & \textbf{92.6} \\ \hline
\end{tabular}}\vspace{-2mm}
\label{table:ablation}
\end{table}
\vspace{-3mm}
\section{Conclusion}\vspace{-1mm}
In this work, we propose a new A-CNN framework on point clouds, which can better capture local geometric information of 3D shapes. Through extensive experiments on several benchmark datasets, our method has achieved the state-of-the-art performance on point cloud classification, part segmentation, and large-scale semantic segmentation tasks. Since our work does not solely focus on large-scale scene datasets, we will explore some new deep learning architectures to improve the current results. We will also investigate to apply the proposed framework on large-scale outdoor datasets in our future work.

\vspace{1mm}
\noindent
\textbf{Acknowledgment}. We would like to thank the reviewers for their valuable comments. This work was partially supported by the NSF IIS-1816511, CNS-1647200, OAC-1657364, OAC-1845962, Wayne State University Subaward 4207299A of CNS-1821962, NIH 1R56AG060822-01A1, and ZJNSF LZ16F020002.
{\small
\bibliographystyle{ieee}
\bibliography{egpaper_final}
}

\clearpage
\pagebreak
\appendix

\title{Supplementary Material:\\A-CNN: Annularly Convolutional Neural Networks on Point Clouds\vspace{-13mm}}
\author{}

\makeatletter
\newcommand{\settitle}{\@maketitle}
\makeatother

\maketitle

\section{Ball Query \emph{vs} Ring-based Scheme}
\label{sec:suppl_ball_vs_ring_comparison}
The comparison of multi-scale method proposed in~\cite{qi2017pointnet++} and our ring-based scheme is depicted in Fig.~\ref{fig:dilated_rings}. It is noted that comparing to multi-scale regions, the ring-based structure does not have overlaps (no neighboring point duplication) at the query point's neighborhood. It means that each ring contains its own unique points.
\vspace{-2.5mm}
\begin{figure}[h]
\begin{center}
\includegraphics[width=0.9\linewidth]{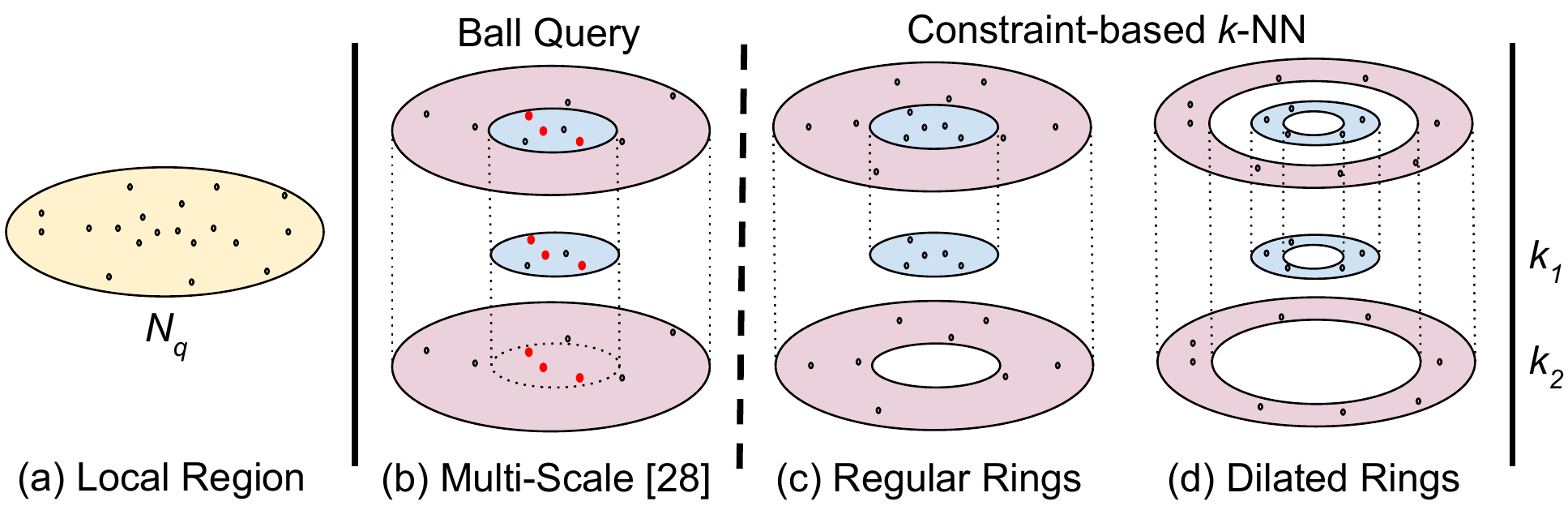}\vspace{-3.5mm}
\end{center}
\caption{A schematic comparison for searching neighbors in a local region with $N_q$ points between multi-scale approach from~\cite{qi2017pointnet++} and our proposed approaches with regular and dilated rings. The number of neighboring points per region (e.g., $k_1$ and $k_2$) is the same between different methods. Regions in multi-scale architecture have neighboring overlaps (red points belong to different regions near the same query point $\mathbf{q}$), while regular and dilated rings have the unique neighbors.
\vspace{-3.5mm}
}
\centering
\label{fig:dilated_rings}
\end{figure}

\begin{table}[h]
\centering
\caption{Experiments on redundancy on \emph{ModelNet40} dataset. AAC is accuracy average class, OA is overall accuracy.}\vspace{-2mm}
\scalebox{0.75}{
\begin{tabular}{l|l|l}
\hline
 & AAC & OA \\ \hline
 PointNet++ (multi-scale / with overlap) & 86.5 & 90.2 \\ 
 PointNet++ (multi-ring / without overlap) & 87.3 & 90.6 \\ \hline
 A-CNN (with all components) & \textbf{90.3} & \textbf{92.6} \\ \hline
\end{tabular}}\vspace{-2mm}
\label{table:ablation}
\end{table}

We have discovered that reducing redundancy can improve the existing multi-scale approach in~\cite{qi2017pointnet++}. We test redundancy issue on original PointNet++ model~\cite{qi2017pointnet++} with and without overlap / redundancy. We compare the original PointNet++ multi-scale model with ball queries (with redundant points) against PointNet++ with our proposed regular rings (without redundant points). Our experiments show that the proposed multi-ring (i.e., without redundant points) outperforms the multi-scale scheme (i.e., with redundant points) on \emph{ModelNet40} according to Tab.~\ref{table:ablation}.

\vspace{-1mm}
\section{Training Details}
\label{sec:suppl_training_details} 
\vspace{-1.5mm}
We use \textit{A-CNN-3L} network configuration in Tab.~\ref{table:network_configs} for all experiments on point cloud classification tasks and \textit{A-CNN-4L} network configuration in Tab.~\ref{table:network_configs} for both part segmentation and semantic segmentation tasks. We use regular rings in $L_1$ and dilated rings in $L_2$ in our \textit{A-CNN-3L} architecture. Similarly, we use regular rings in $L_1$ and dilated rings in $L_2$ and $L_3$ in our \textit{A-CNN-4L} architecture.

We use Adam optimization method with learning rate 0.001 and decay rate 0.7 in classification and decay 0.5 in segmentation tasks. We have trained our classification model for 250 epochs, our part segmentation model for 200 epochs, and our large-scale semantic segmentation models for 50 epochs on each area of \emph{S3DIS} and for 200 epochs on \emph{ScanNet}. The training time of our model is faster than that of PointNet++ model, since we use ring-based neighboring search, which is more efficient and effective than ball query in PointNet++ model. For instance, the training time on the segmentation model for 200 epochs is about 19 hours on a single NVIDIA Titan Xp GPU with 12 GB GDDR5X, and PointNet++ model needs about 32 hours for the same task. The size of our trained model is 22.3 MB and the size of PointNet++ model is 22.1 MB.
\begin{table*}[t]
\centering
\small 
\caption
{Network configurations.}\vspace{-2mm}
\scalebox{0.82}{
 \begin{tabular}{ll|llll}
 \hline
                   &  & $L_1$ & $L_2$ & $L_3$ & $L_4$ \\ \hline
 \multirow{4}{*}{\makecell{\emph{A-CNN-3L} \\ (classification)}} & \textit{C} & 512 & 128 & 1 & - \\
                   & \textit{rings}  & [[0.0, 0.1], [0.1, 0.2]]  & $[[0.1, 0.2], [0.3, 0.4]]$ & - & - \\
                   & \textit{k} & [16, 48] & [16, 48] & 128 & - \\
                   & \textit{F} & [[32,32,64], [64,64,128]] & [[64,64,128], [128,128,256]] & [256,512,1024] & - \\ \hline
 \multirow{4}{*}{\makecell{\emph{A-CNN-4L} \\ (segmentation)}} & \textit{C} & 512 & 128 & 32 & 1 \\
                   & \textit{rings}  & [[0.0, 0.1], [0.1, 0.2]]  & [[0.1, 0.2], [0.3, 0.4]] & [[0.2, 0.4], [0.6, 0.8]] & - \\
                   & \textit{k} & [16, 48] & [16, 48] & [16, 48] & 32 \\
                   & \textit{F} & [[32,32,64], [64,64,128]] & [[64,64,128], [128,128,256]] & [[128,128,256], [256,256,512]] & [512,768,1024] \\ \hline
 \end{tabular}
    }
 \begin{tablenotes}
   \footnotesize 
   \item Note: Both of the models represent encoder part. \emph{A-CNN-3L} model consists of three layers. A-CNN-4L model consists of four layers. For each layer, $C$ is the number of centroids, $rings$ is the inner and outer radiuses of a ring: [$R_{inner}, R_{outer}$], $k$ is number of neighbors, $F$ is feature map size. For example, our \emph{A-CNN-4L} model at the first layer $L_1$ has 512 centroids; two regular rings where first ring constrained by radiuses of 0.0 and 0.1 and the second ring has radiuses of 0.1 and 0.2; k-NN search returns $16$ points in the first ring, and $48$ points in the second ring; the feature map size in the first ring is equal to $[32,32,64]$ and in the second ring is $[64,64,128]$. Convolutional kernel size across different rings and layers is the same and equal to $1\times3$. Also, we have to double the number of centroids in each layer in model \emph{A-CNN-4L} on \emph{ScanNet} as the number of points in each block is twice more than that in \emph{S3DIS}.\vspace{-2mm}
 \end{tablenotes}
 \label{table:network_configs}
\end{table*}

\vspace{-1mm}
\section{Feature Visualization}
\label{sec:suppl_feature_visualization}
\textbf{Local Feature Visualization. }Fig.~\ref{fig:feature_vis_mn10} and Fig.~\ref{fig:feature_vis} visualize the magnitude of the gradient per point in the classification task on \textit{ModelNet10} and \textit{ModelNet40} datasets. Blue color represents low magnitude of the gradients and red color represents high magnitude of the gradients. The points with higher magnitudes get greater updates during training and the learning contribution of them is higher. Therefore, this feature visualization could be thought as the object saliency. For example, in \textit{ModelNet40} dataset our model considers wings and tails as important regions to classify an object as an airplane; bottle neck is important for a bottle; the flowers and leaves are important for a plant; tube or middle part (usually narrow parts) is important for a lamp; legs are important to classify an object as a stool.
\setcounter{figure}{-9}
\begin{figure}[h]
\centering
\vspace{-3.5mm}
\subfloat{\includegraphics[width=0.9\linewidth, clip]{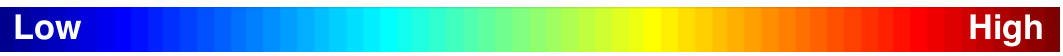}}
\vspace{-3.5mm}
\subfloat{\includegraphics[width=0.09\linewidth,trim={0 2mm 0 0}, clip]{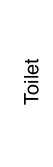} \phantomcaption}
\subfloat{\includegraphics[width=0.27\linewidth,trim={4cm 5cm 4cm 4.5cm}, clip]{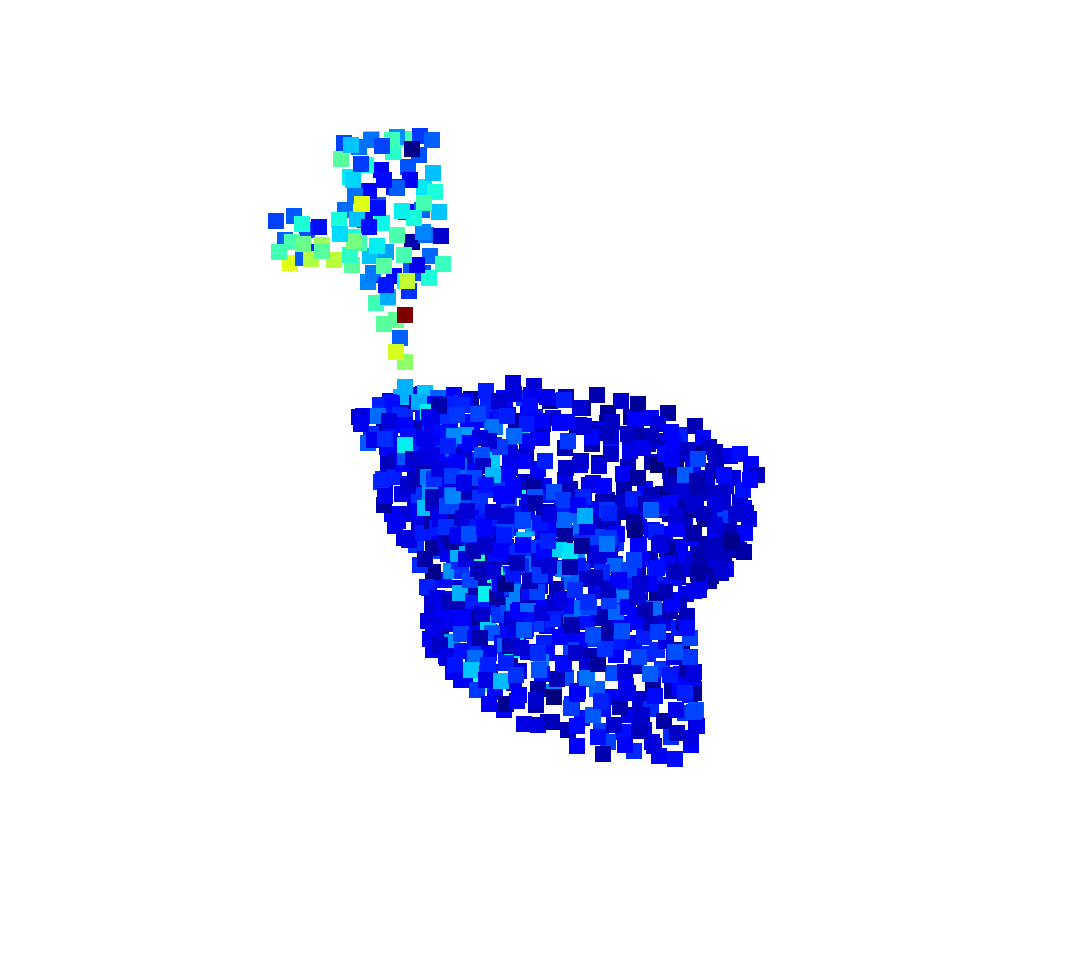} \phantomcaption}
\subfloat{\includegraphics[width=0.27\linewidth,trim={4cm 5.5cm 4cm 4.5cm}, clip]{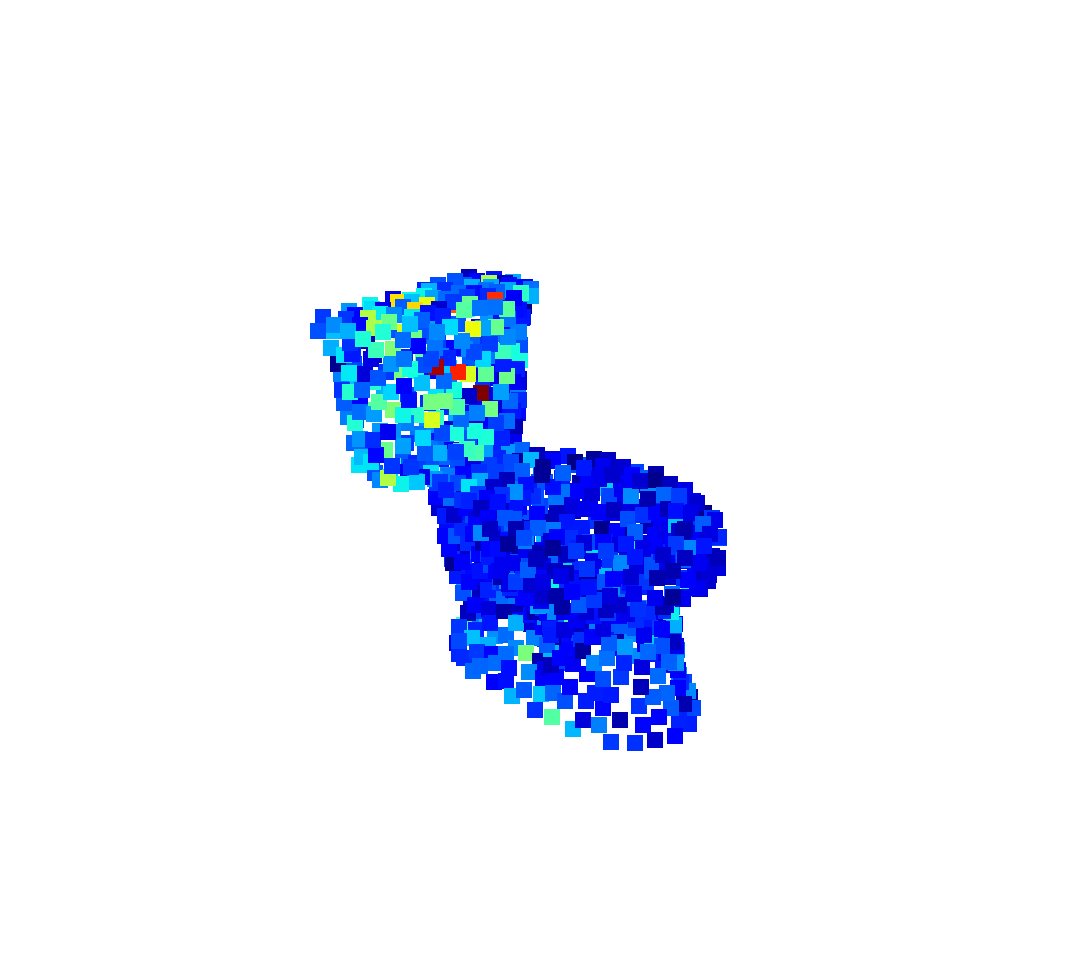} \phantomcaption}
\subfloat{\includegraphics[width=0.27\linewidth,trim={4cm 5cm 4cm 4.5cm}, clip]{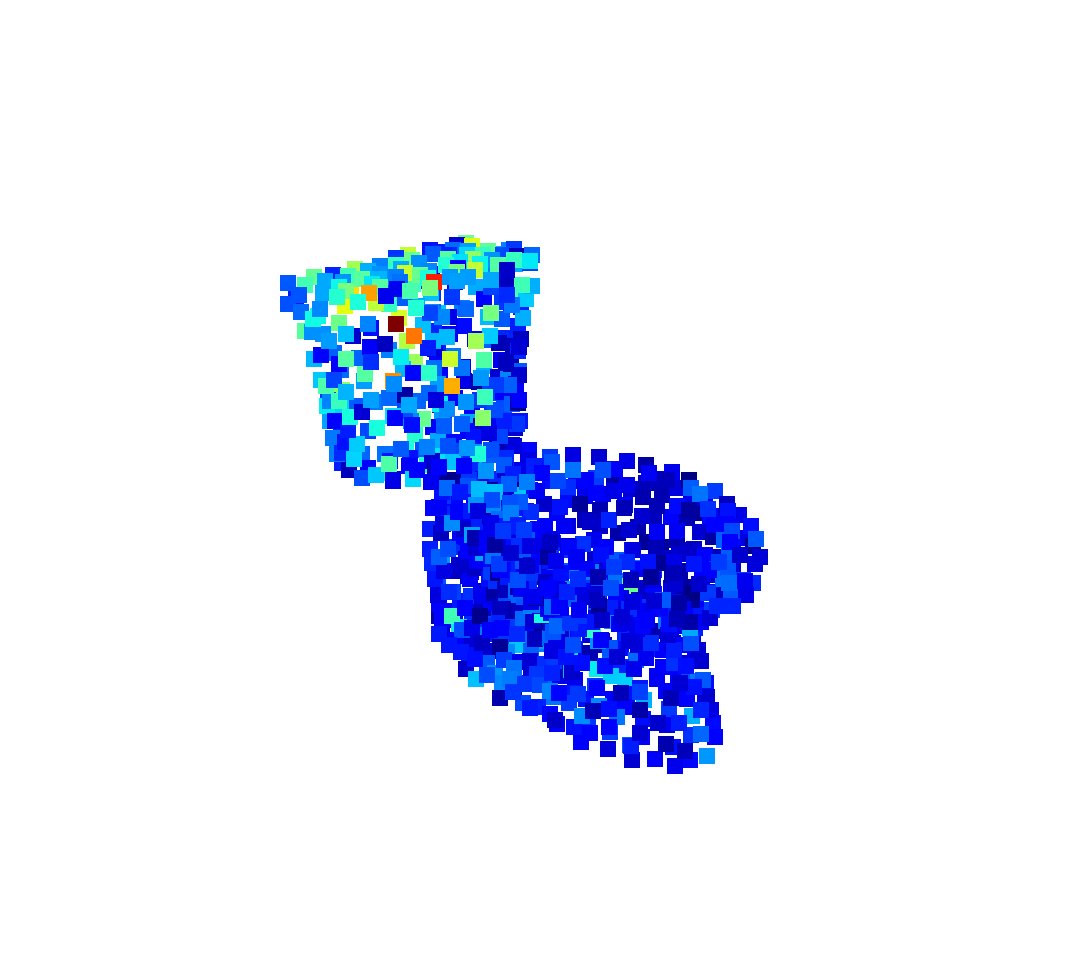} \phantomcaption}
\vspace{-3.5mm}
\subfloat{\includegraphics[width=0.09\linewidth,trim={0 3mm 0 0}, clip]{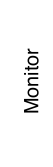} \phantomcaption}
\subfloat{\includegraphics[width=0.27\linewidth,trim={3cm 4cm 3cm 5.5cm}, clip]{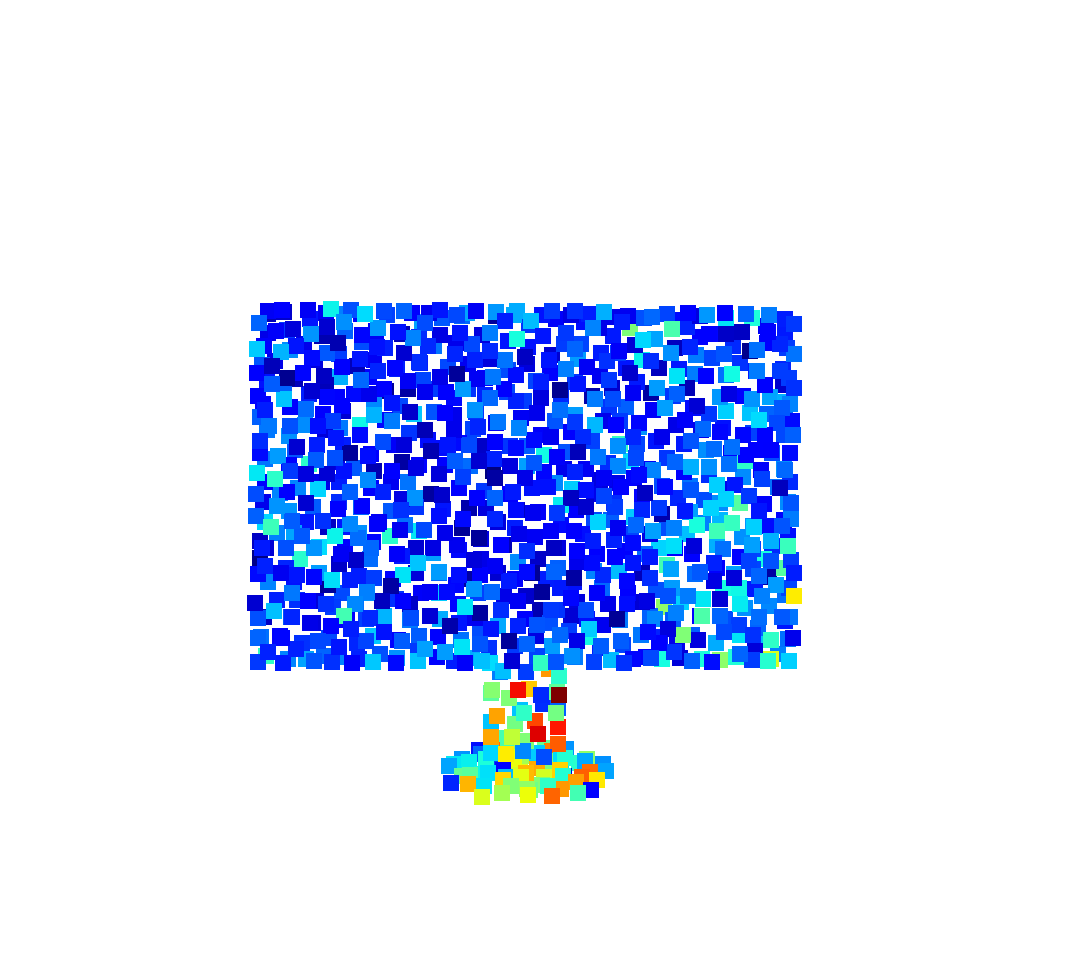} \phantomcaption}
\subfloat{\includegraphics[width=0.27\linewidth,trim={3cm 4.5cm 3cm 5.5cm}, clip]{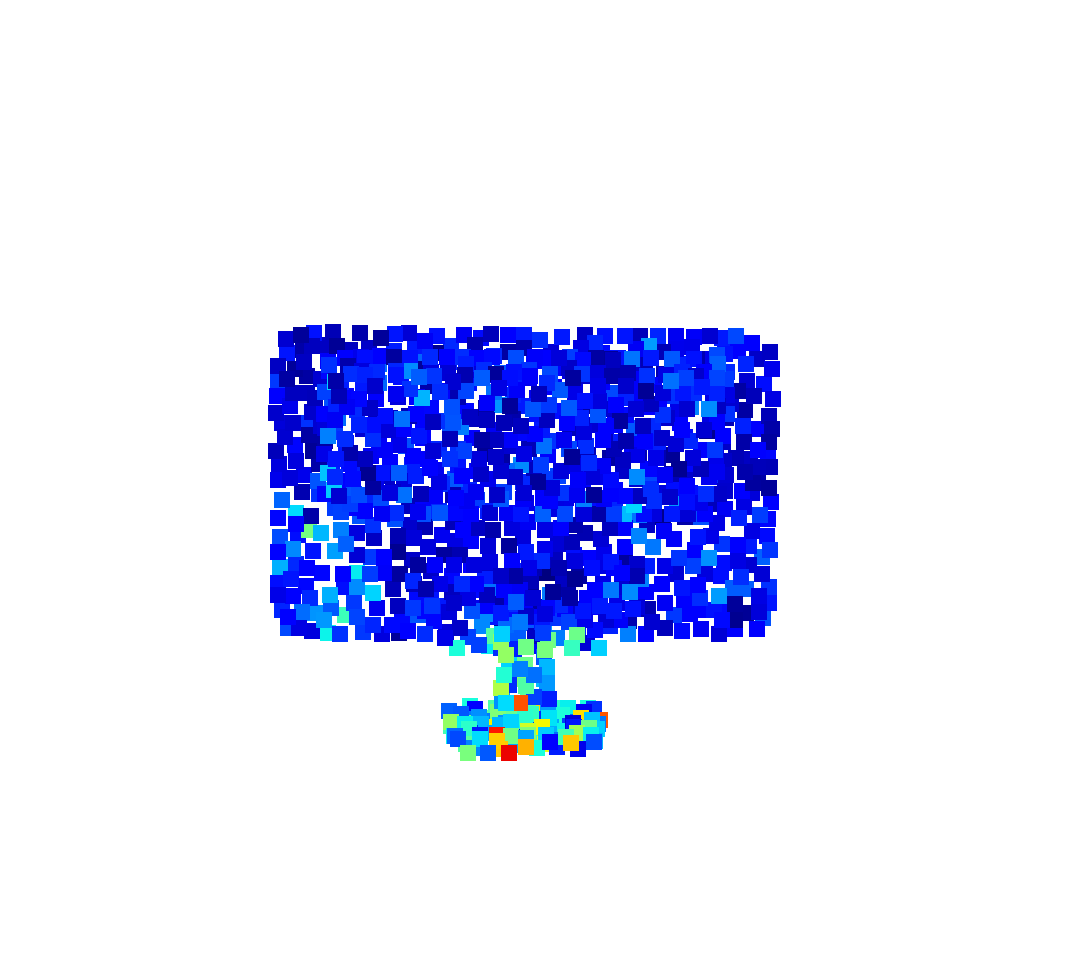} \phantomcaption}
\subfloat{\includegraphics[width=0.27\linewidth,trim={3cm 4cm 3cm 5.5cm}, clip]{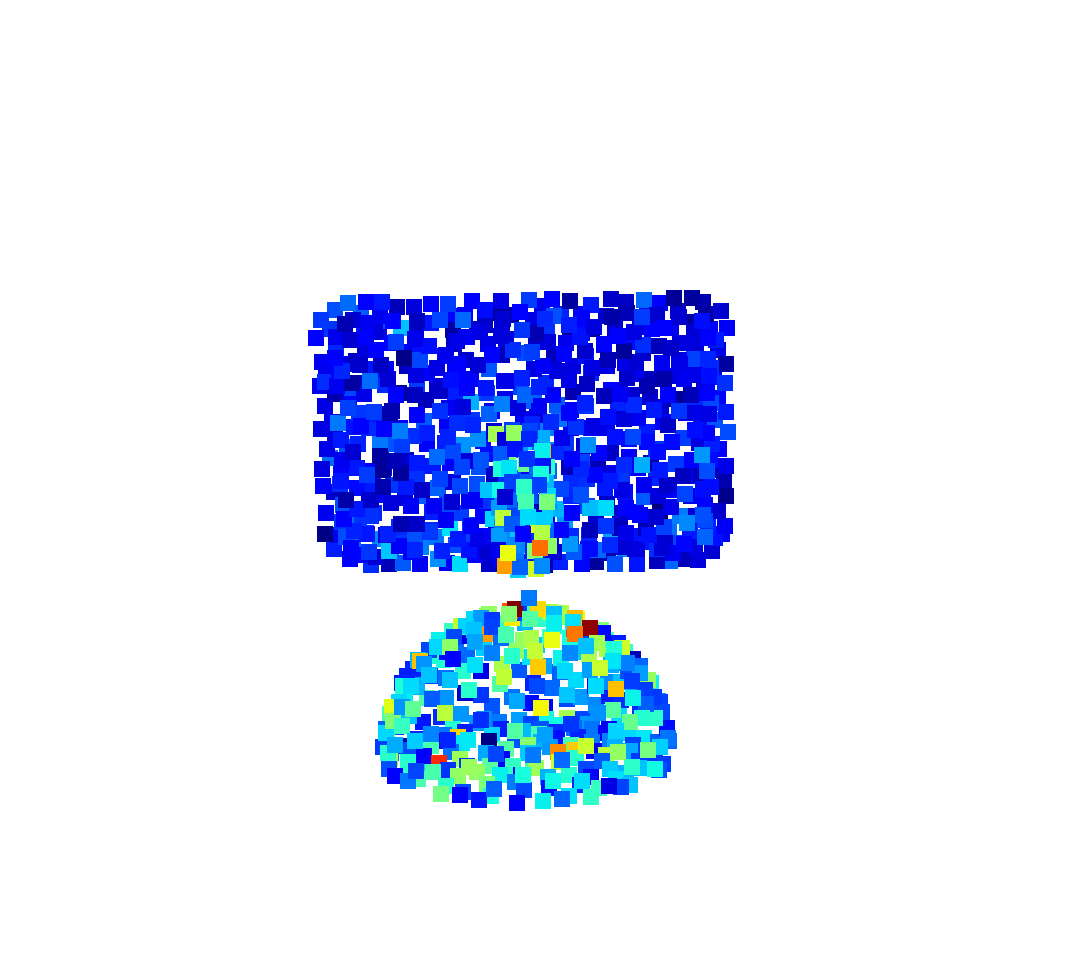} \phantomcaption}
\vspace{-3.5mm}
\subfloat{\includegraphics[width=0.09\linewidth,trim={0 0 0 0}, clip]{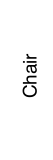} \phantomcaption}
\subfloat{\includegraphics[width=0.27\linewidth,trim={4cm 3cm 4cm 3cm}, clip]{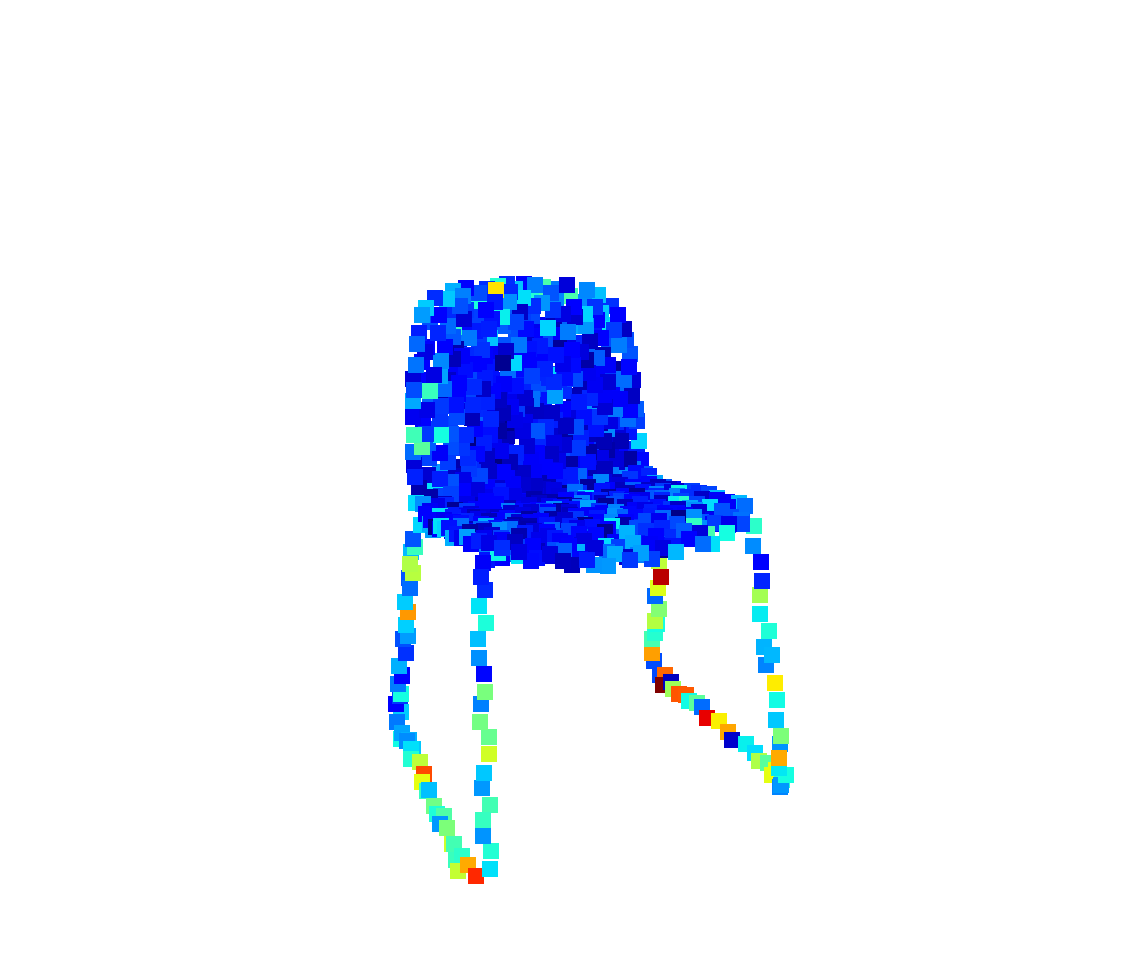} \phantomcaption}
\subfloat{\includegraphics[width=0.27\linewidth,trim={4cm 3cm 4cm 3cm}, clip]{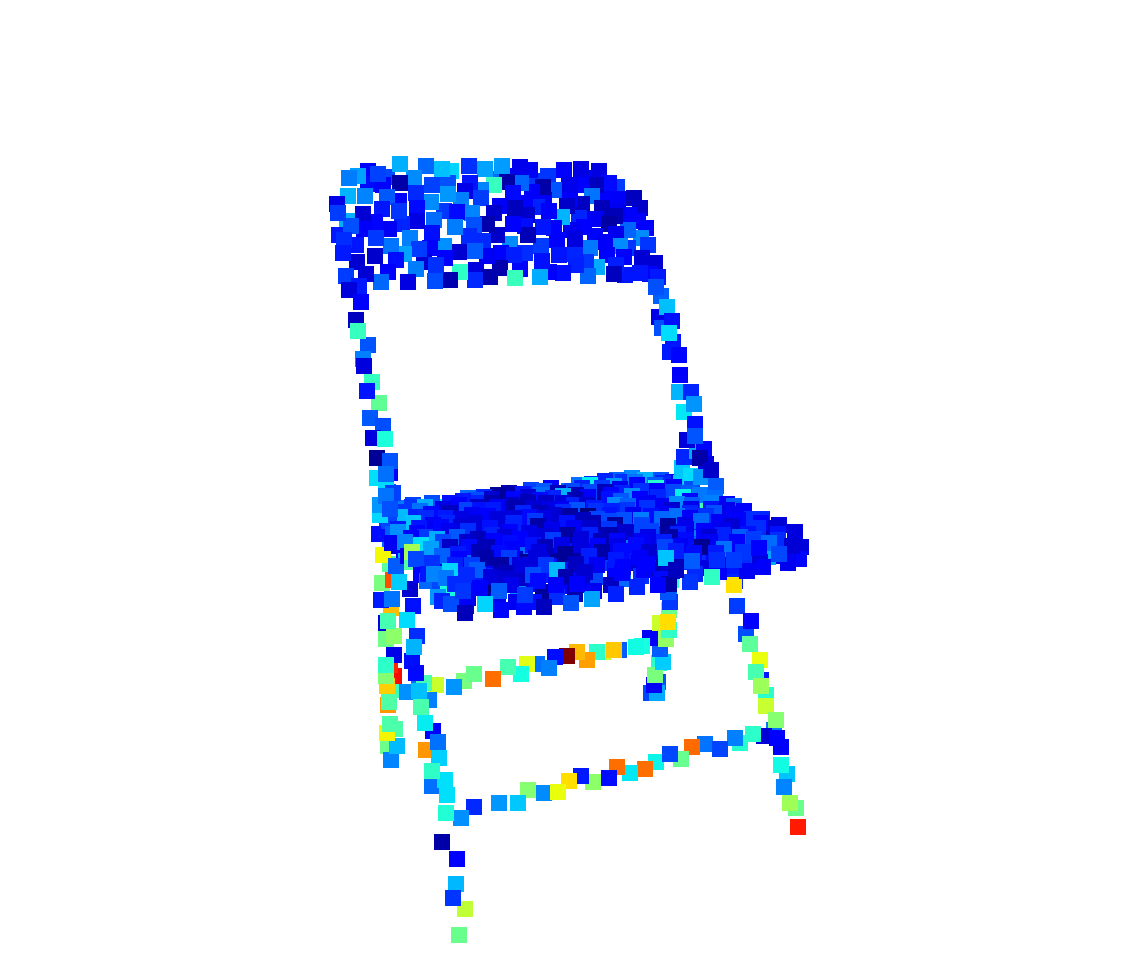} \phantomcaption}
\subfloat{\includegraphics[width=0.27\linewidth,trim={4cm 3cm 4cm 3cm}, clip]{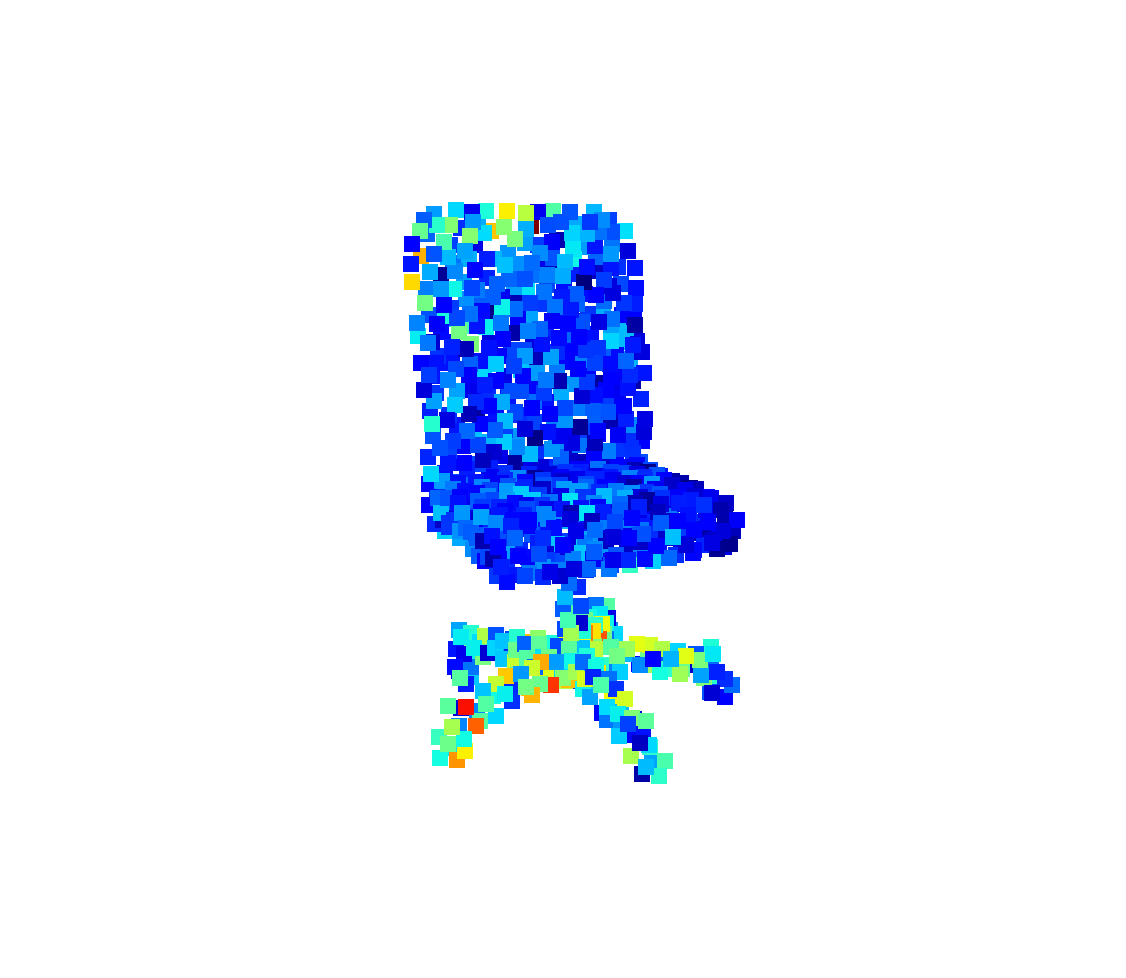} \phantomcaption}
\vspace{-3.5mm}
\subfloat{\includegraphics[width=0.09\linewidth,trim={0 0 0 0}, clip]{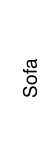} \phantomcaption}
\subfloat{\includegraphics[width=0.27\linewidth,trim={4cm 3cm 4cm 6cm}, clip]{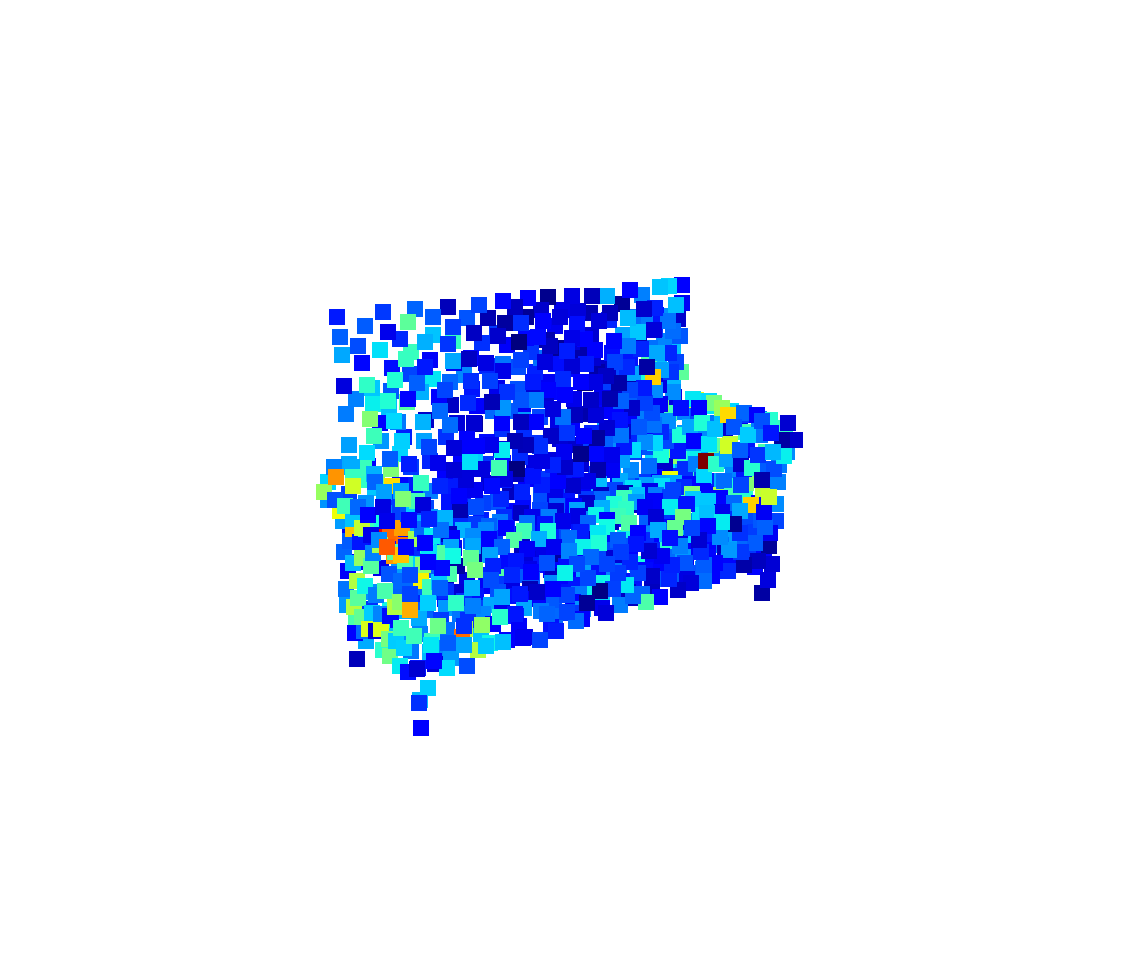} \phantomcaption}
\subfloat{\includegraphics[width=0.27\linewidth,trim={4cm 4cm 4cm 6cm}, clip]{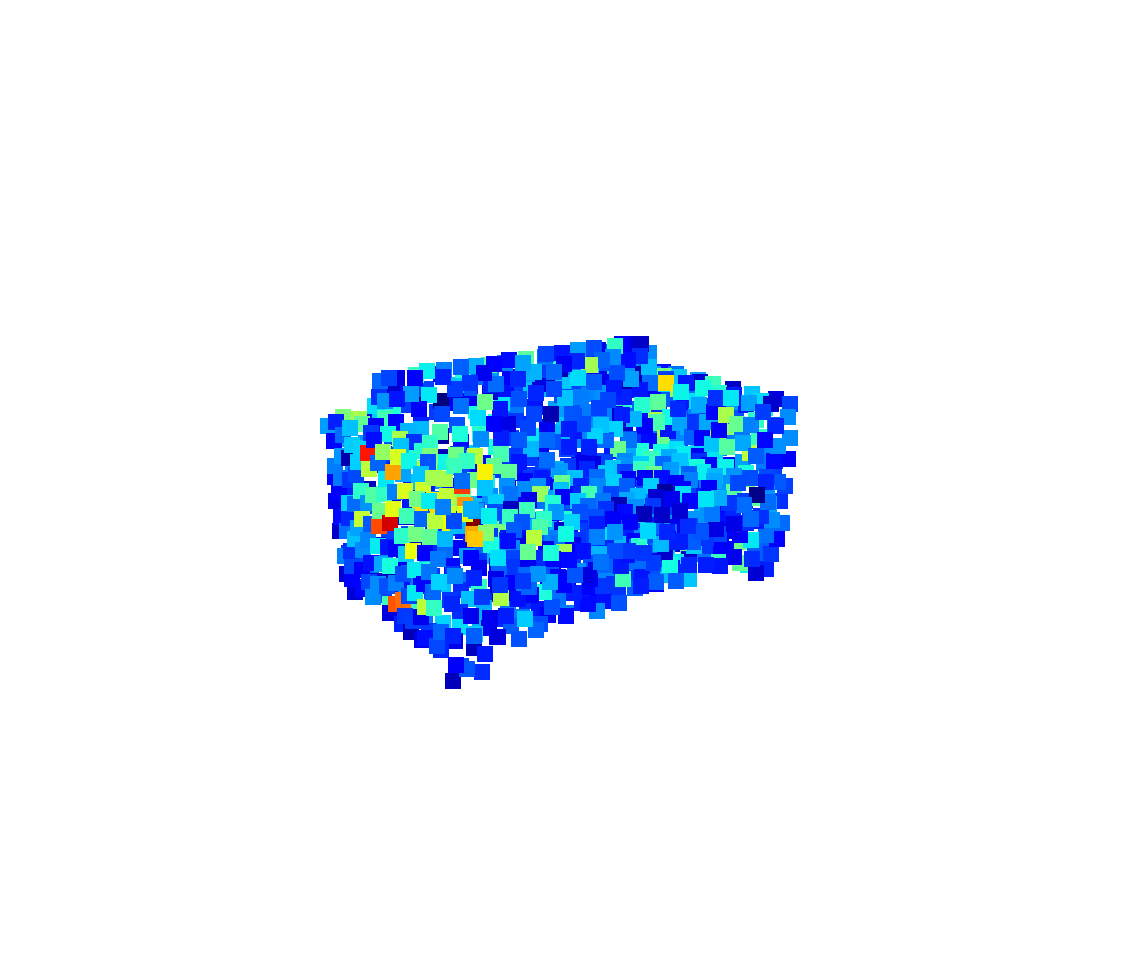} \phantomcaption}
\subfloat{\includegraphics[width=0.27\linewidth,trim={4cm 5cm 4cm 5.5cm}, clip]{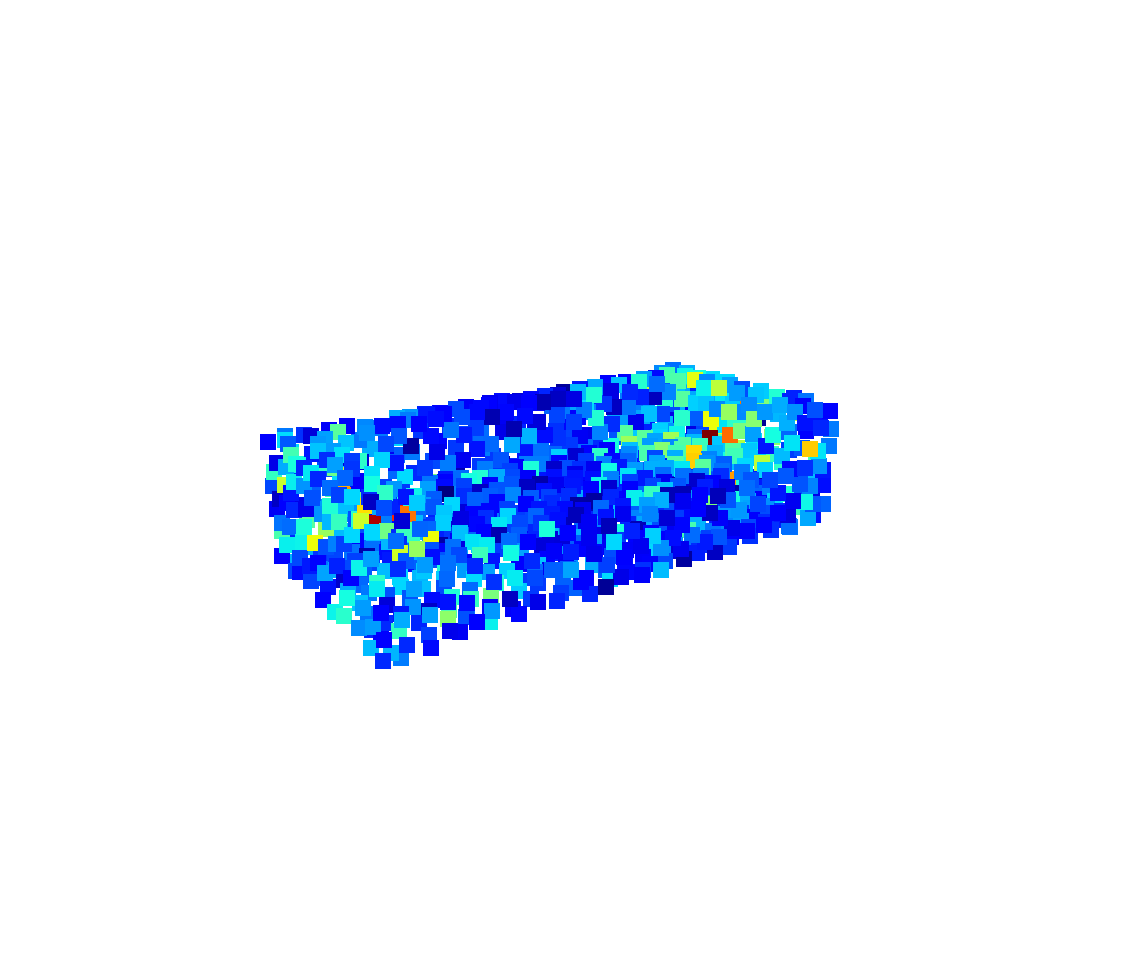} \phantomcaption}
\vspace{-3.5mm}
\caption{The magnitude of the gradient per point in the classification task on \emph{ModelNet10} dataset.}
\vspace{-5.5mm}
\label{fig:feature_vis_mn10}
\end{figure}

\setcounter{figure}{-12}
\begin{figure}[h]
\centering
\vspace{-3.5mm}
\subfloat{\includegraphics[width=0.9\linewidth, clip]{pics/top_jet.pdf}}
\vspace{-3.5mm}
\subfloat{\includegraphics[width=0.09\linewidth,trim={0 0 0 0}, clip]{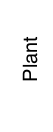} \phantomcaption}
\subfloat{\includegraphics[width=0.27\linewidth,trim={4cm 4.5cm 4cm 4.5cm}, clip]{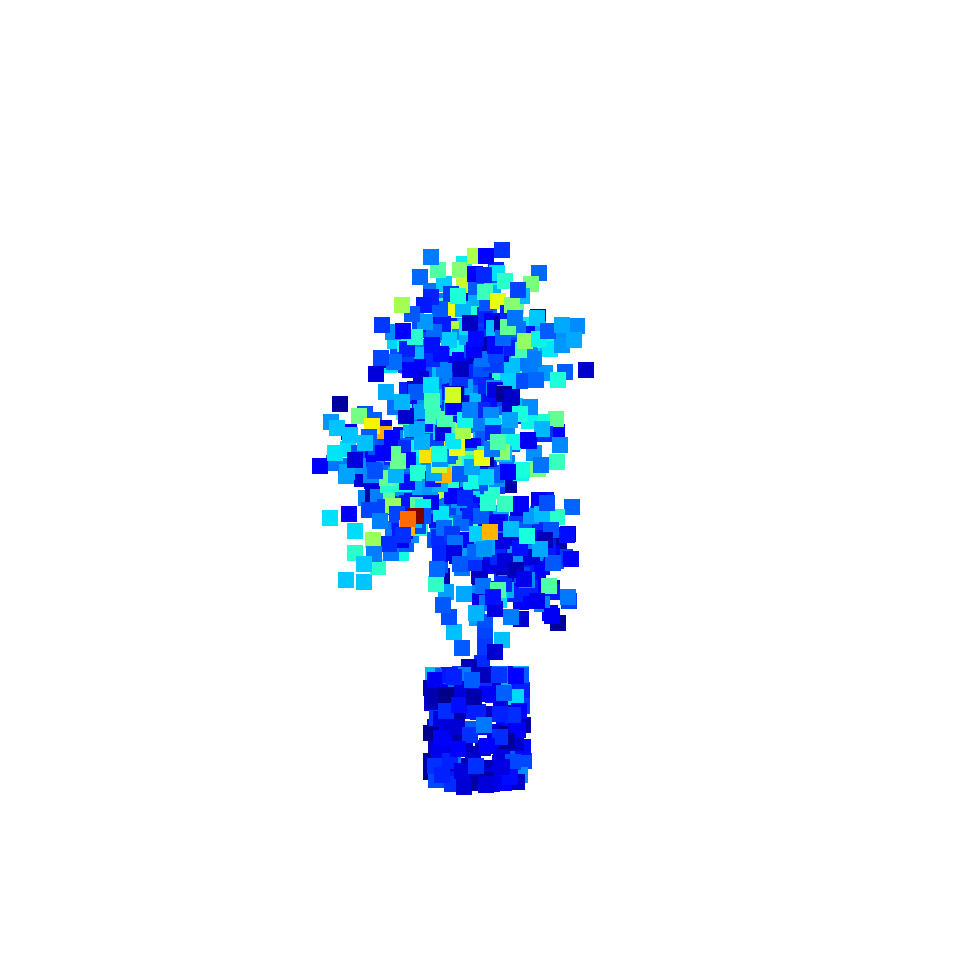} \phantomcaption}
\subfloat{\includegraphics[width=0.27\linewidth,trim={4cm 6cm 4cm 4.5cm}, clip]{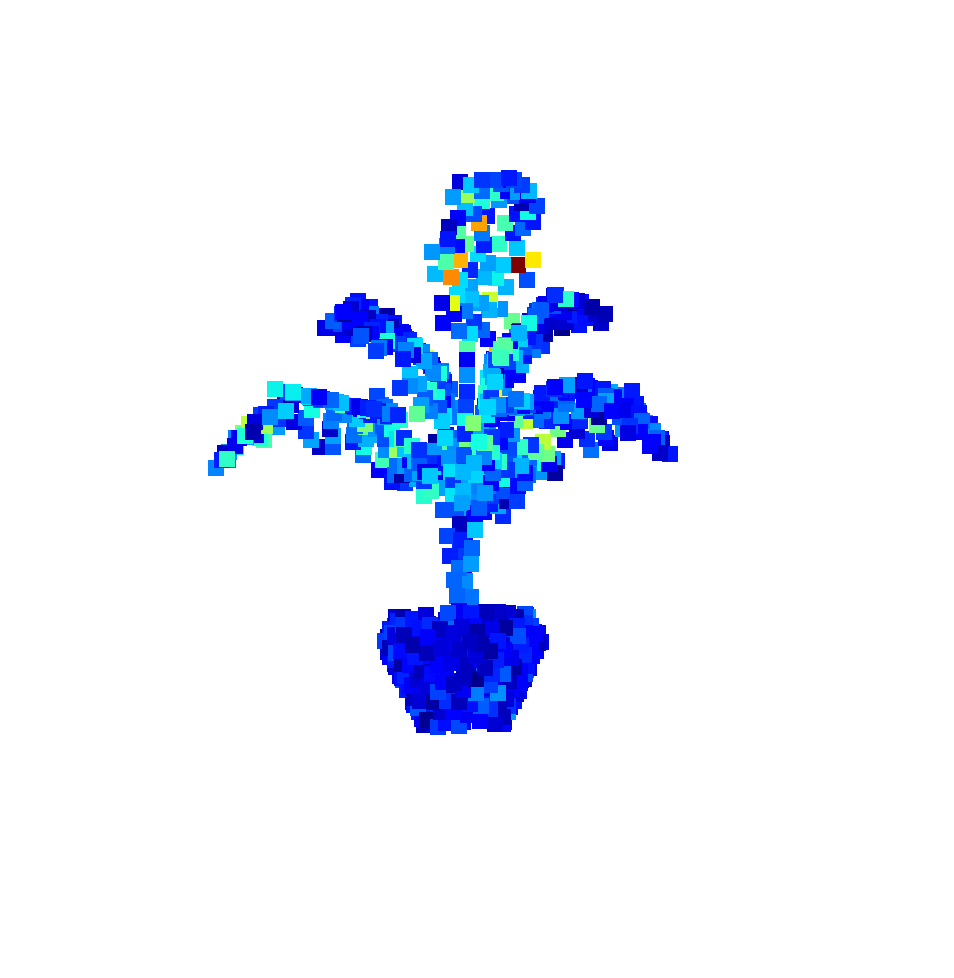} \phantomcaption}
\subfloat{\includegraphics[width=0.27\linewidth,trim={4cm 5cm 4cm 4.5cm}, clip]{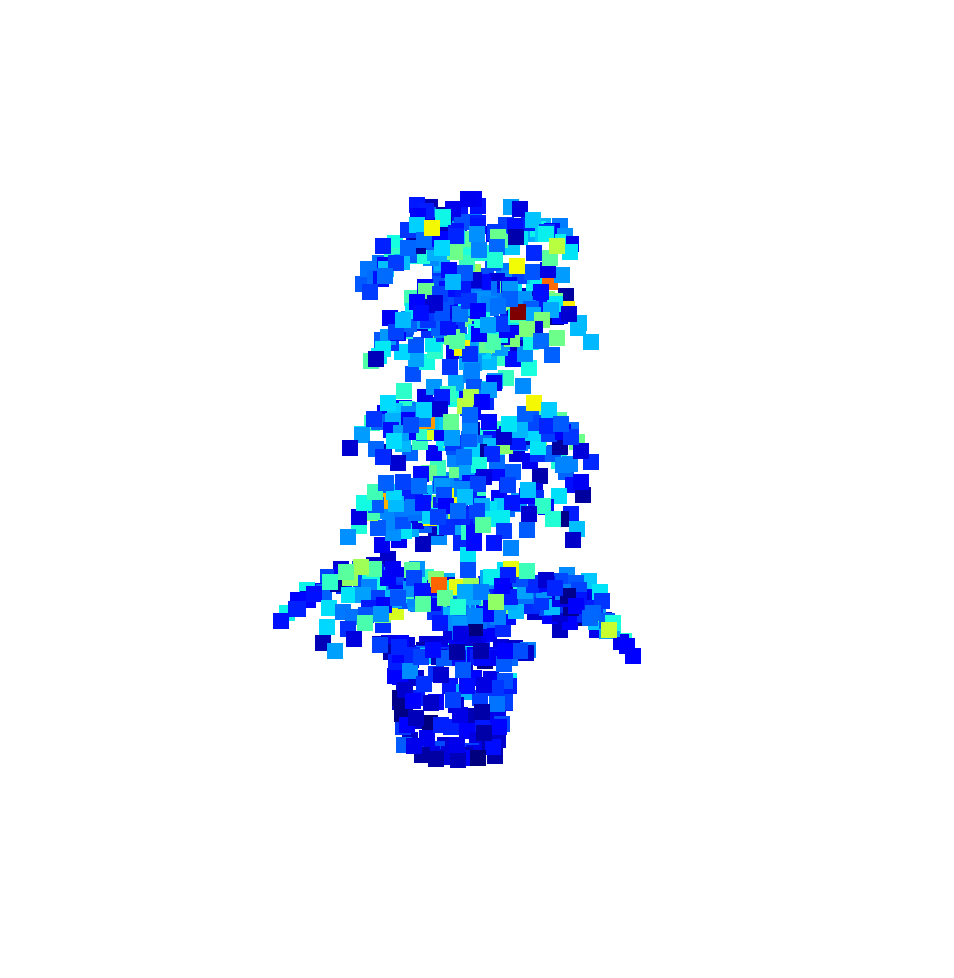} \phantomcaption}
\vspace{-3.5mm}
\subfloat{\includegraphics[width=0.09\linewidth,trim={0 0 0 0}, clip]{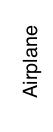} \phantomcaption}
\subfloat{\includegraphics[width=0.27\linewidth,trim={3cm 6.5cm 3cm 5.5cm}, clip]{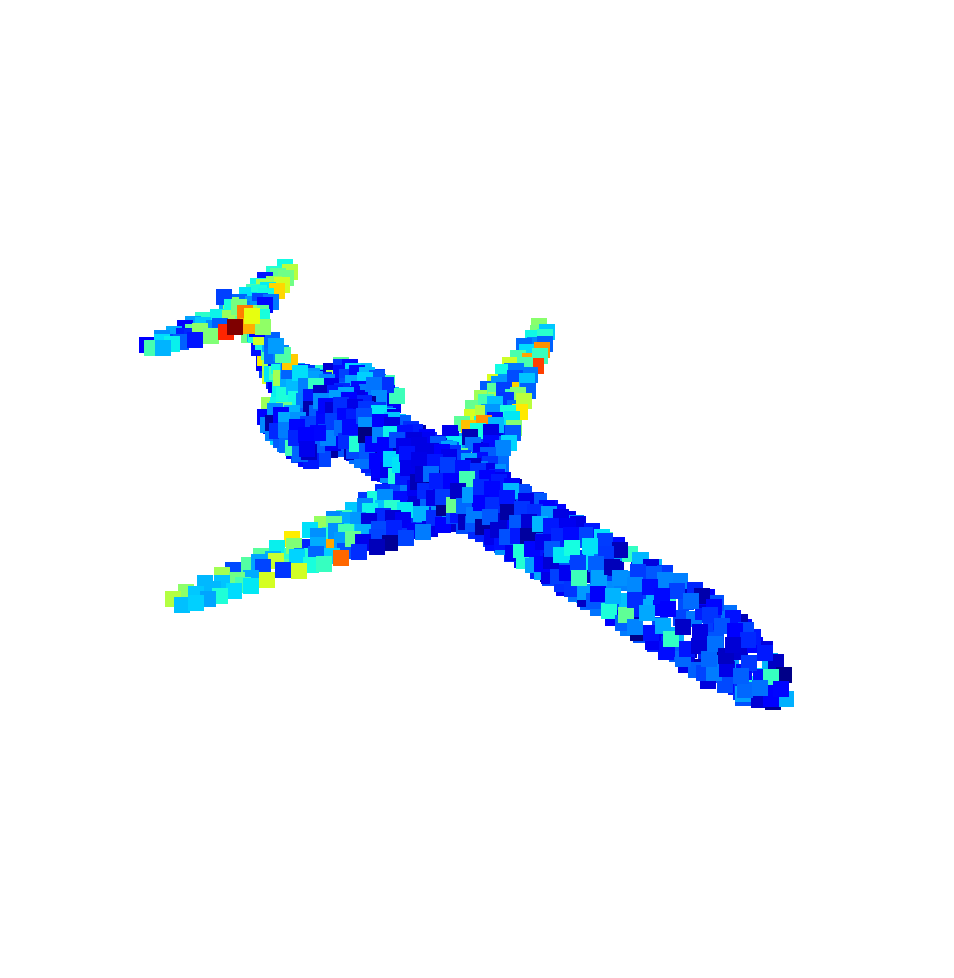} \phantomcaption}
\subfloat{\includegraphics[width=0.27\linewidth,trim={3cm 6.5cm 3cm 5.5cm}, clip]{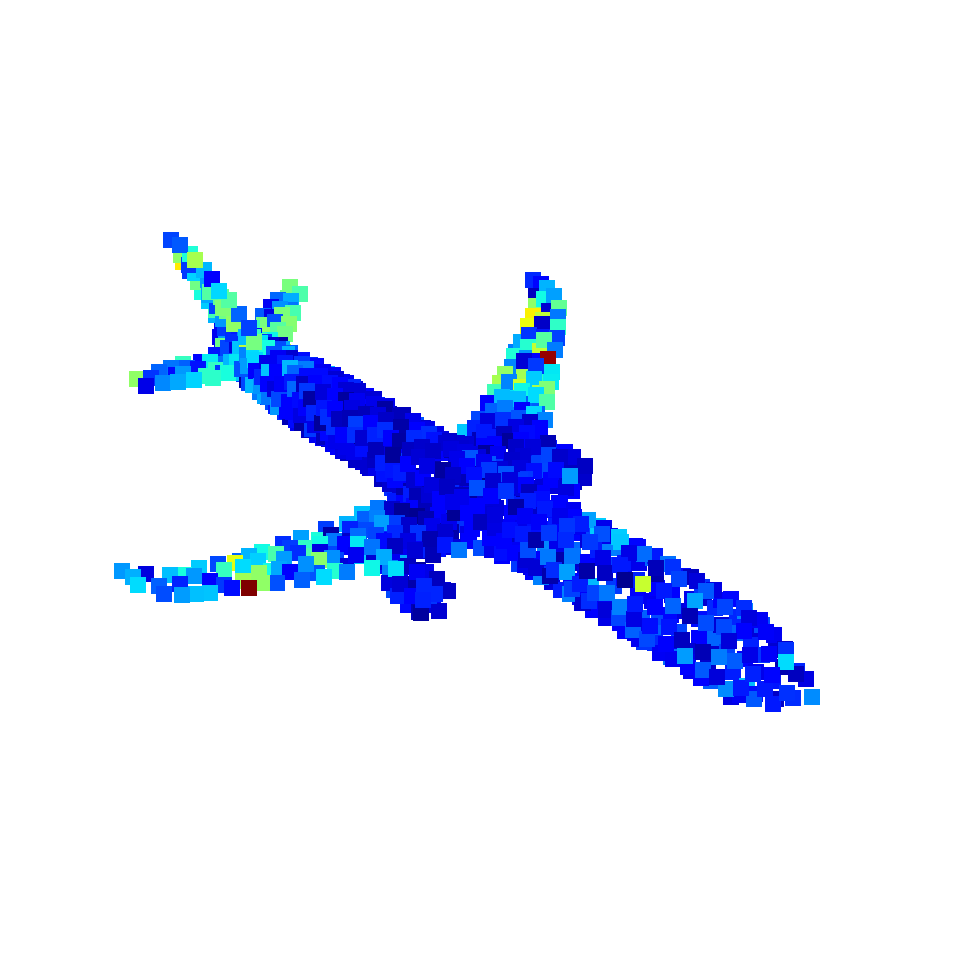} \phantomcaption}
\subfloat{\includegraphics[width=0.27\linewidth,trim={2cm 6.5cm 3cm 5.5cm}, clip]{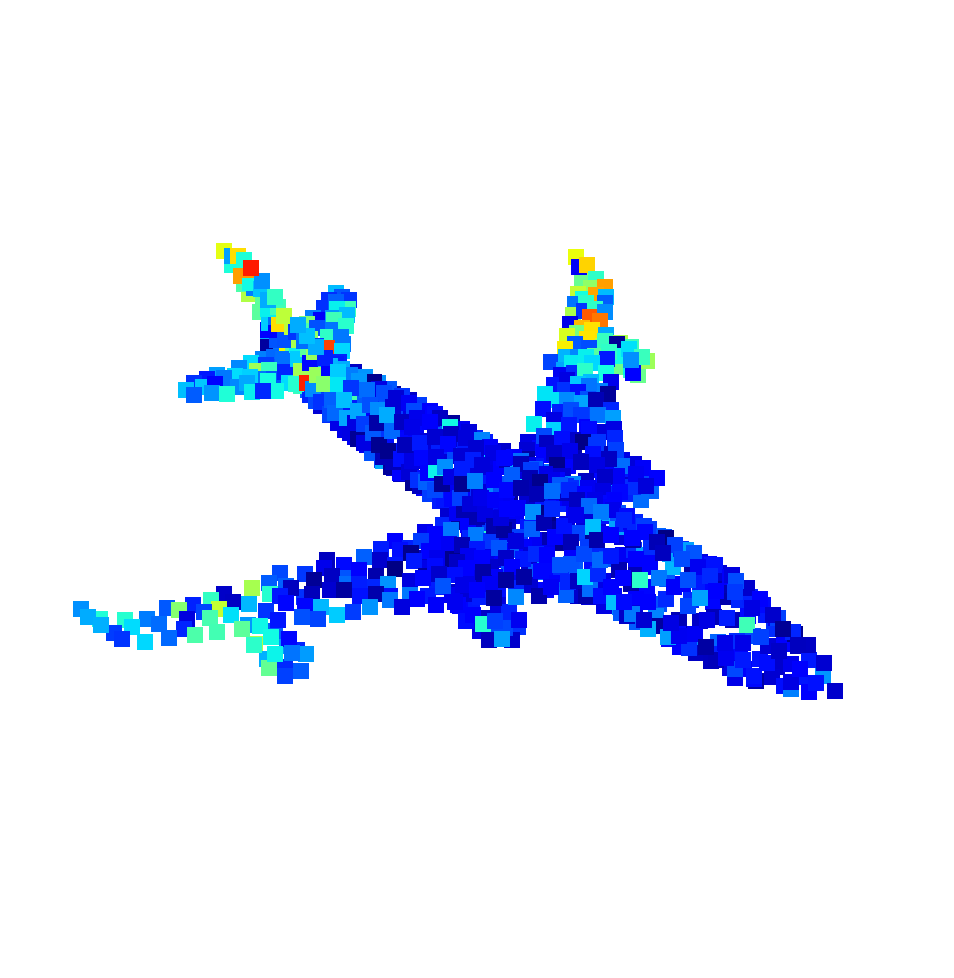} \phantomcaption}
\vspace{-3.5mm}
\subfloat{\includegraphics[width=0.09\linewidth,trim={0 0 0 0}, clip]{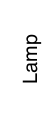} \phantomcaption}
\subfloat{\includegraphics[width=0.27\linewidth,trim={4cm 3cm 4cm 5cm}, clip]{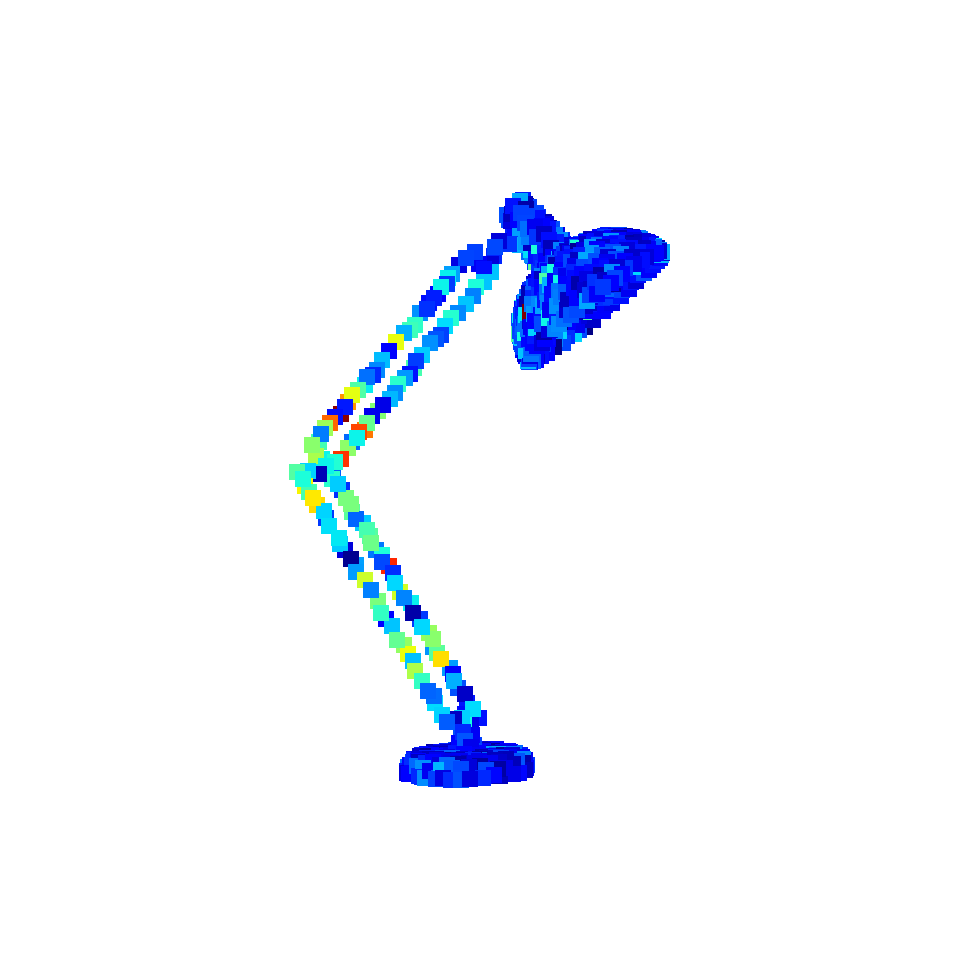} \phantomcaption}
\subfloat{\includegraphics[width=0.27\linewidth,trim={4cm 4.3cm 4cm 3cm}, clip]{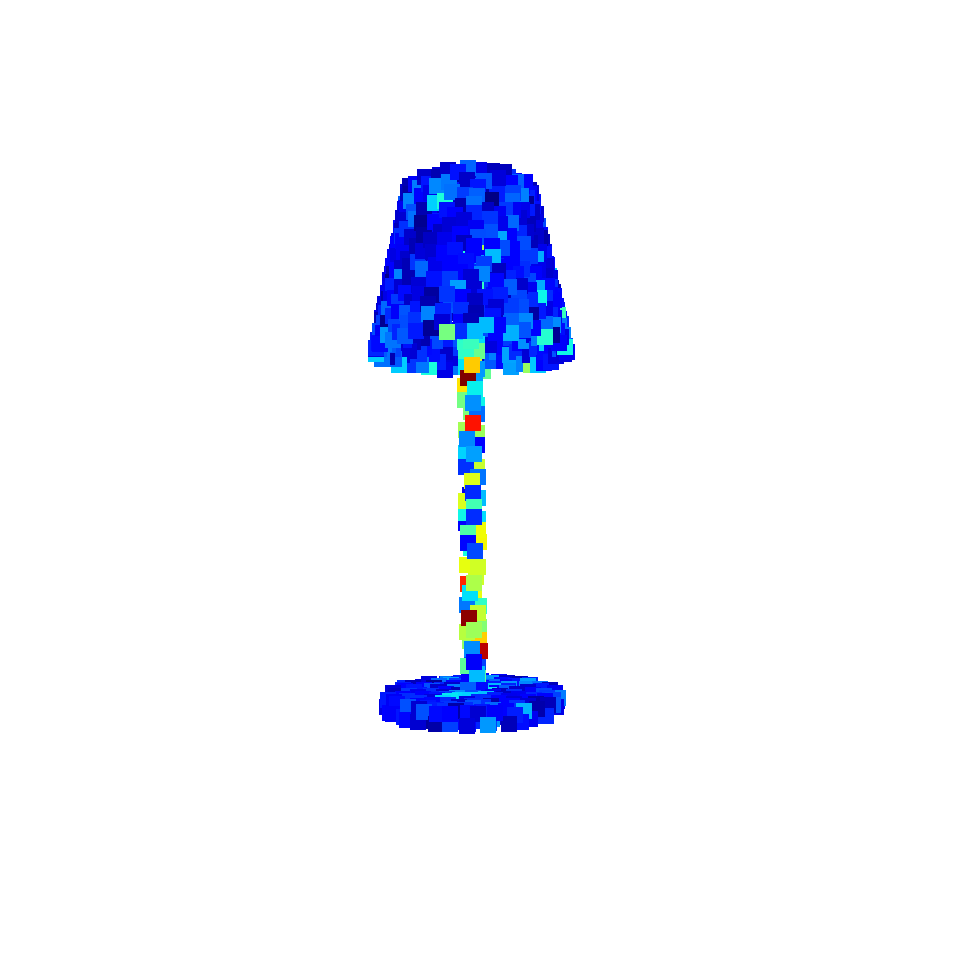} \phantomcaption}
\subfloat{\includegraphics[width=0.27\linewidth,trim={4cm 3cm 4cm 4cm}, clip]{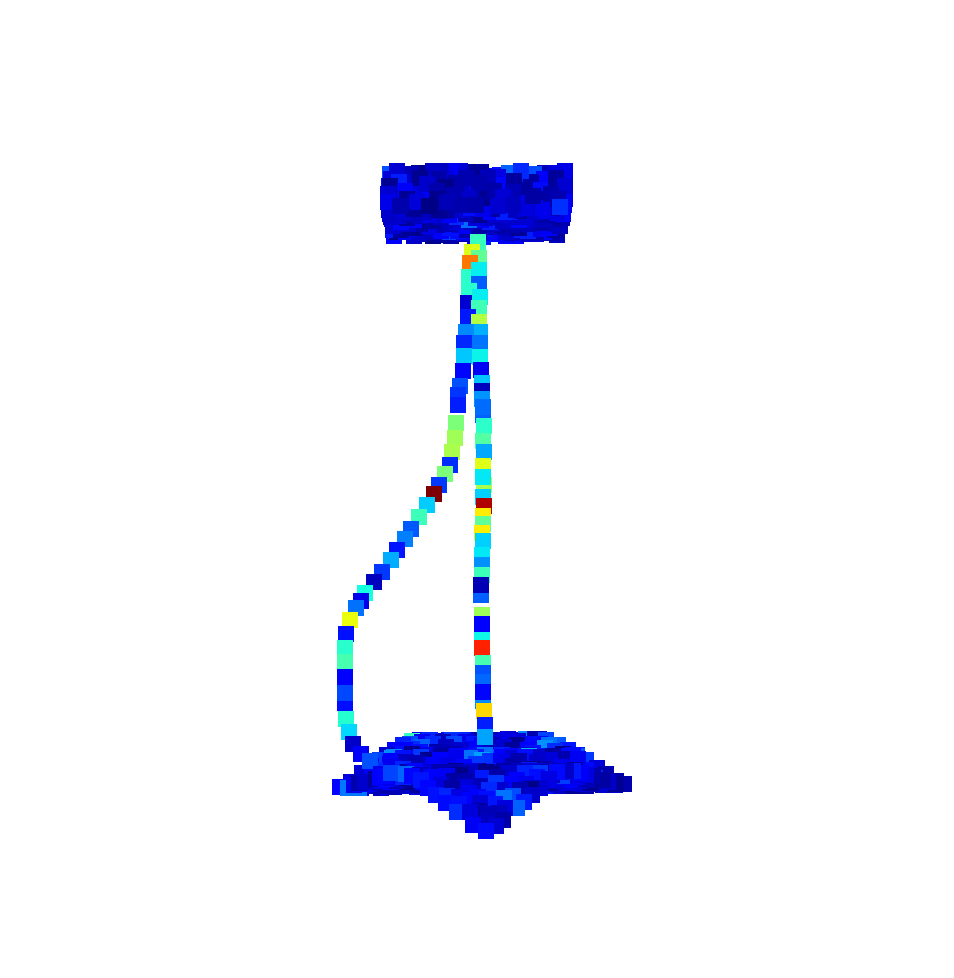} \phantomcaption}
\vspace{-3.5mm}
\subfloat{\includegraphics[width=0.09\linewidth,trim={0 0 0 0}, clip]{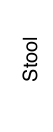} \phantomcaption}
\subfloat{\includegraphics[width=0.27\linewidth,trim={4cm 3cm 4cm 6cm}, clip]{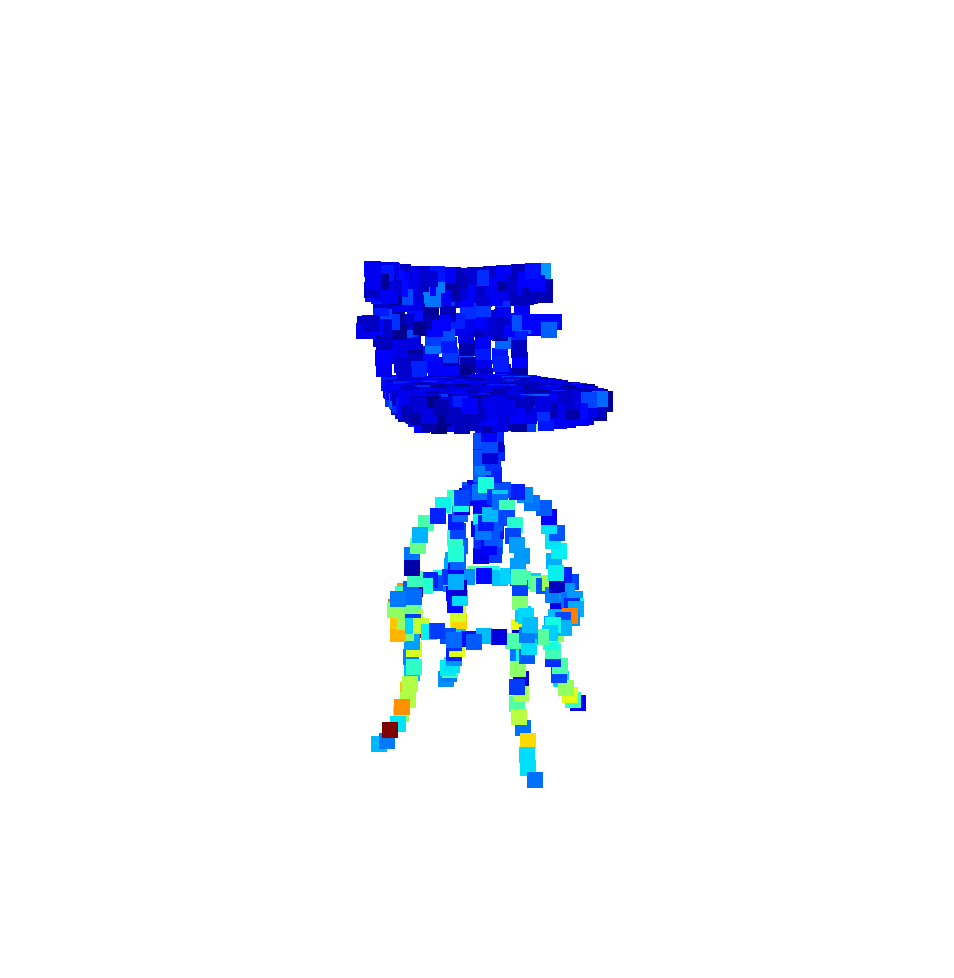} \phantomcaption}
\subfloat{\includegraphics[width=0.27\linewidth,trim={4cm 4cm 4cm 6cm}, clip]{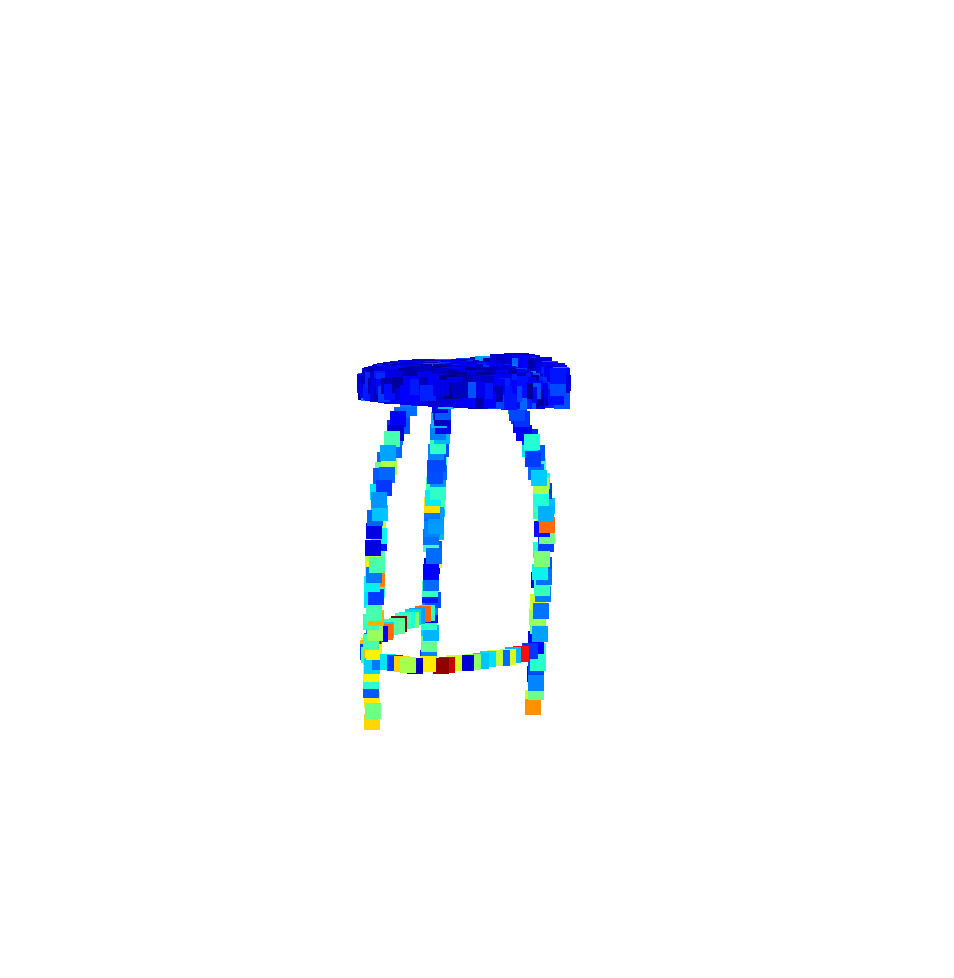} \phantomcaption}
\subfloat{\includegraphics[width=0.27\linewidth,trim={4cm 5cm 4cm 5.5cm}, clip]{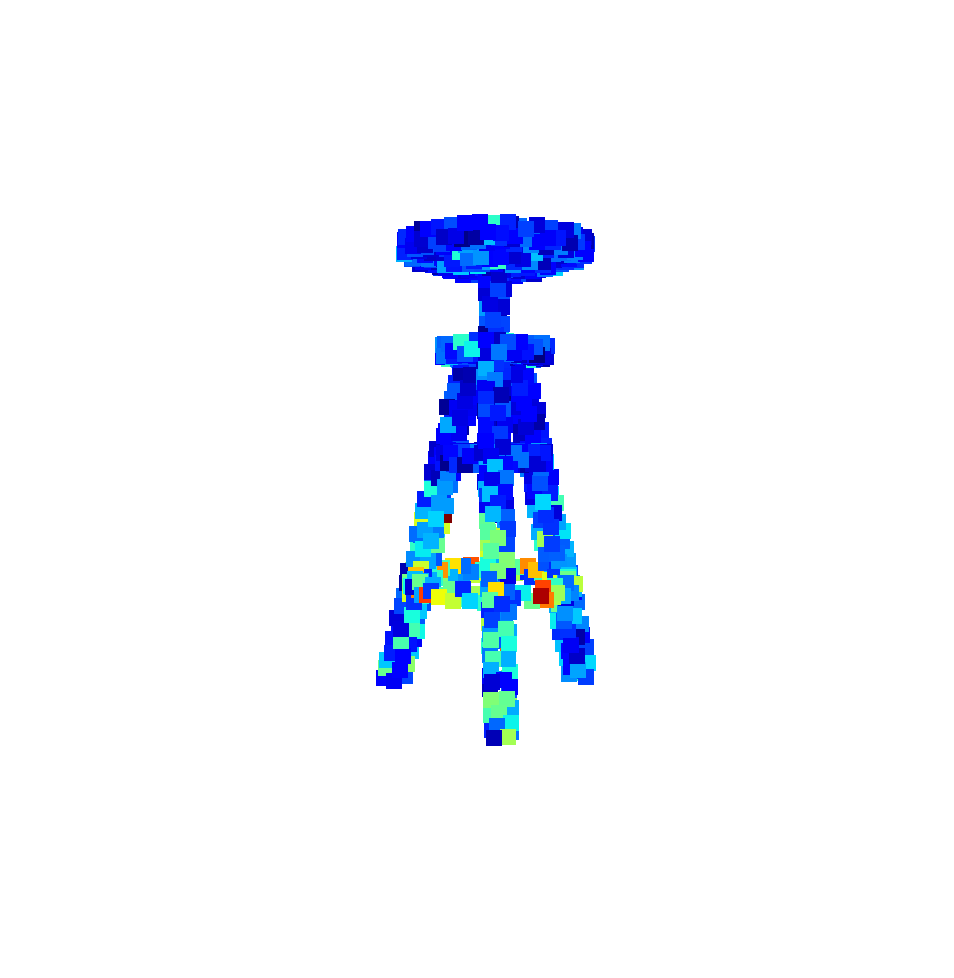} \phantomcaption}
\vspace{-3.5mm}
\subfloat{\includegraphics[width=0.09\linewidth,trim={0 0 0 0}, clip]{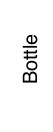} \phantomcaption}
\subfloat{\includegraphics[width=0.27\linewidth,trim={4cm 5cm 4cm 5cm}, clip]{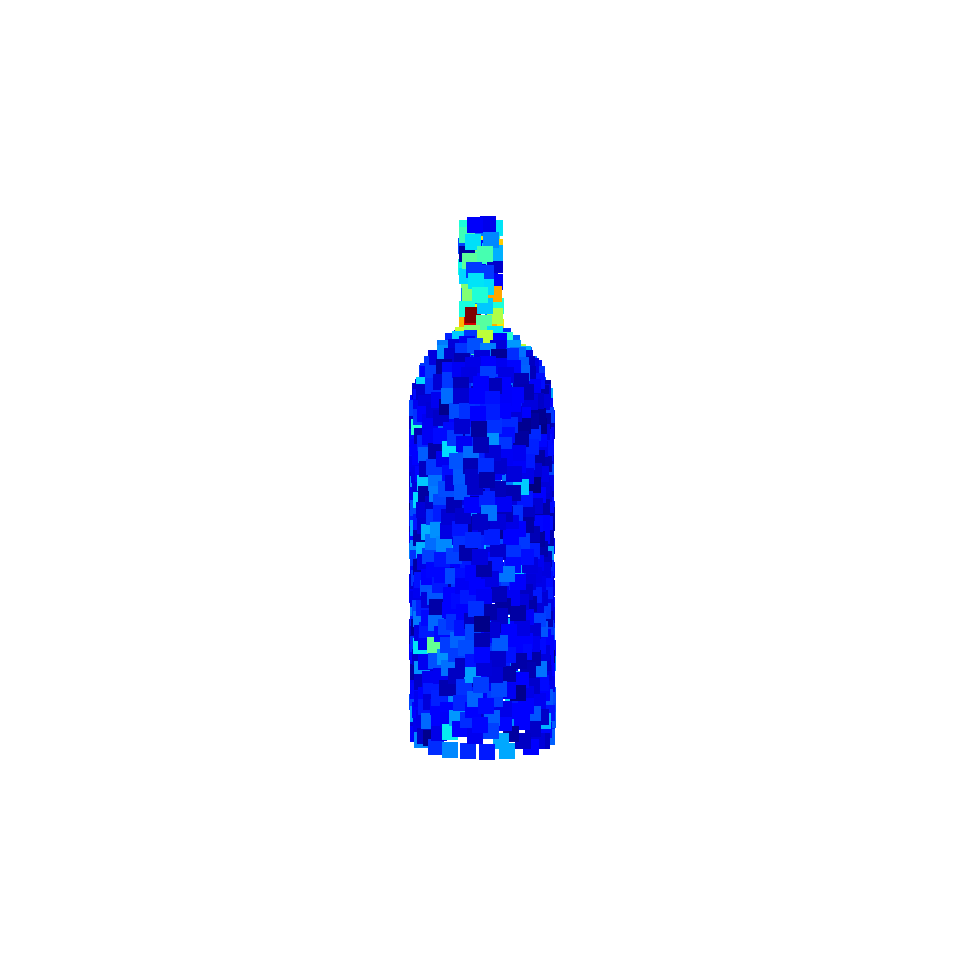} \phantomcaption}
\subfloat{\includegraphics[width=0.27\linewidth,trim={4cm 5cm 4cm 5cm}, clip]{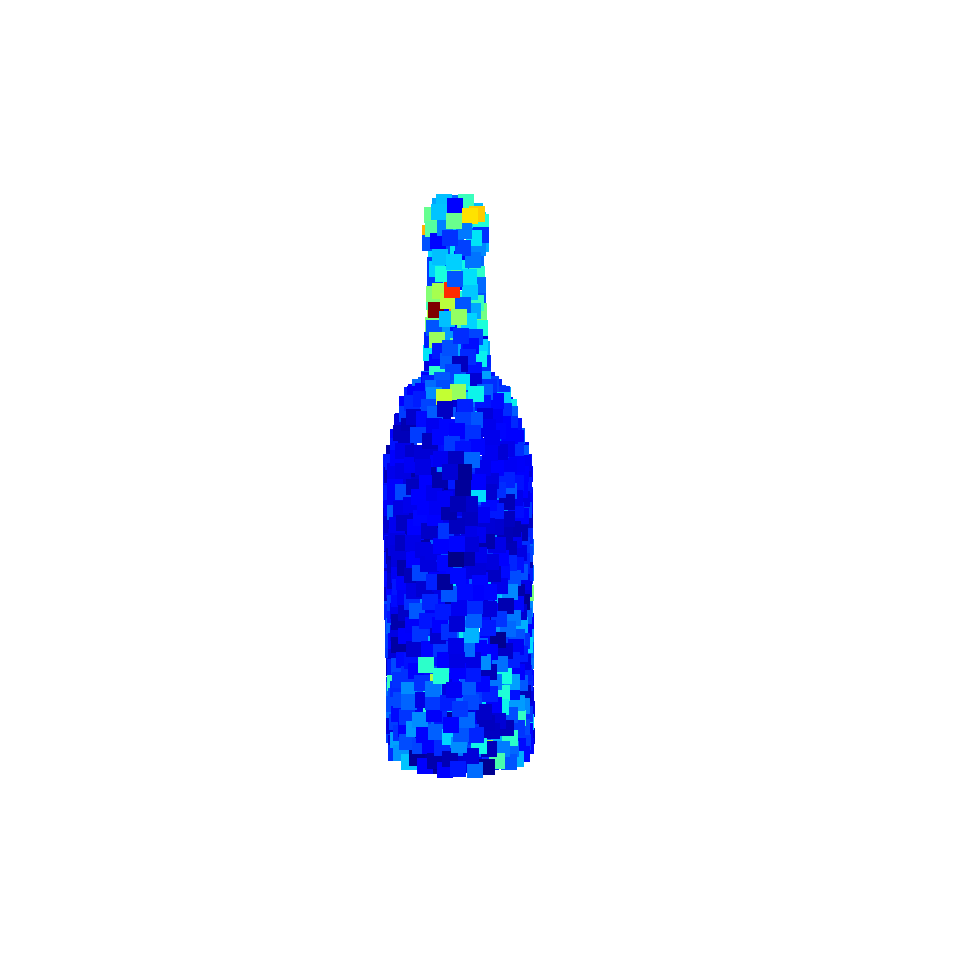} \phantomcaption}
\subfloat{\includegraphics[width=0.27\linewidth,trim={4cm 5cm 4cm 5cm}, clip]{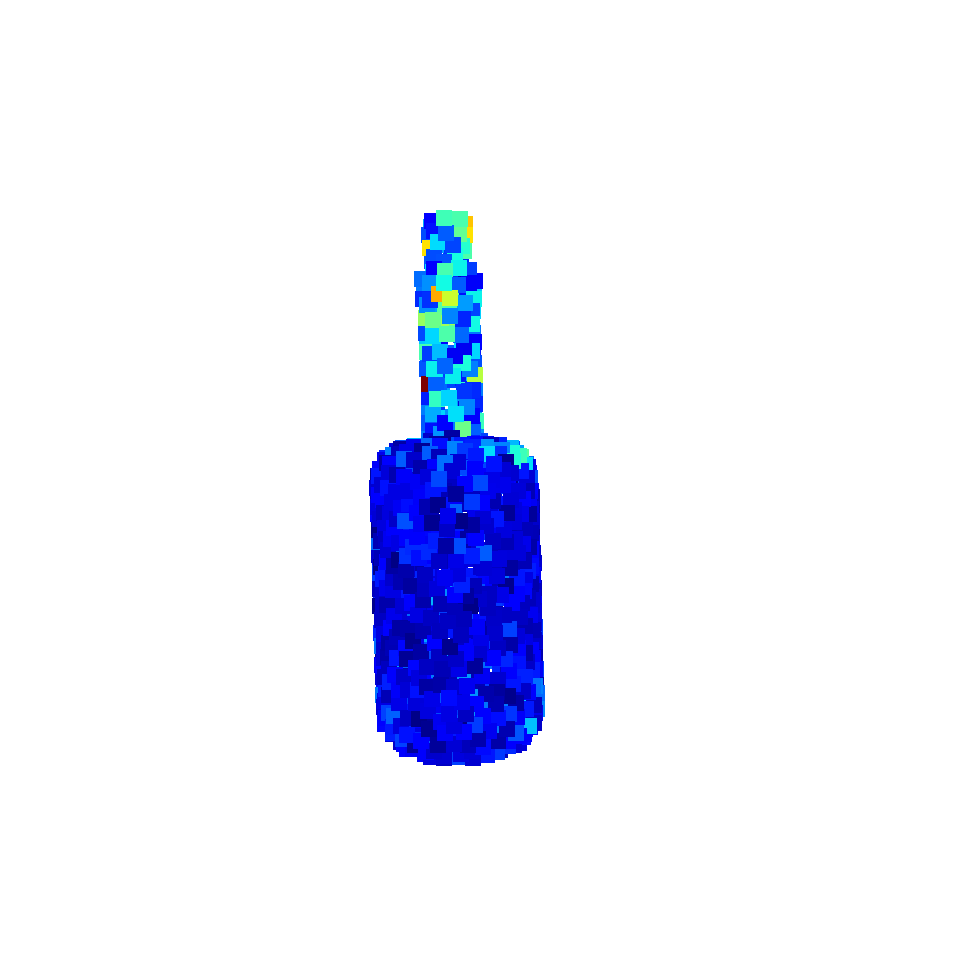} \phantomcaption}
\vspace{-3.5mm}
\caption{The magnitude of the gradient per point in the classification task on \emph{ModelNet40} dataset.}
\vspace{-5.5mm}
\label{fig:feature_vis}
\end{figure}

\begin{figure}[h]
\begin{center}
  \vspace{-5.5mm}
\includegraphics[width=1.0\linewidth, trim={0 0.5cm 0 1.5cm}, clip]{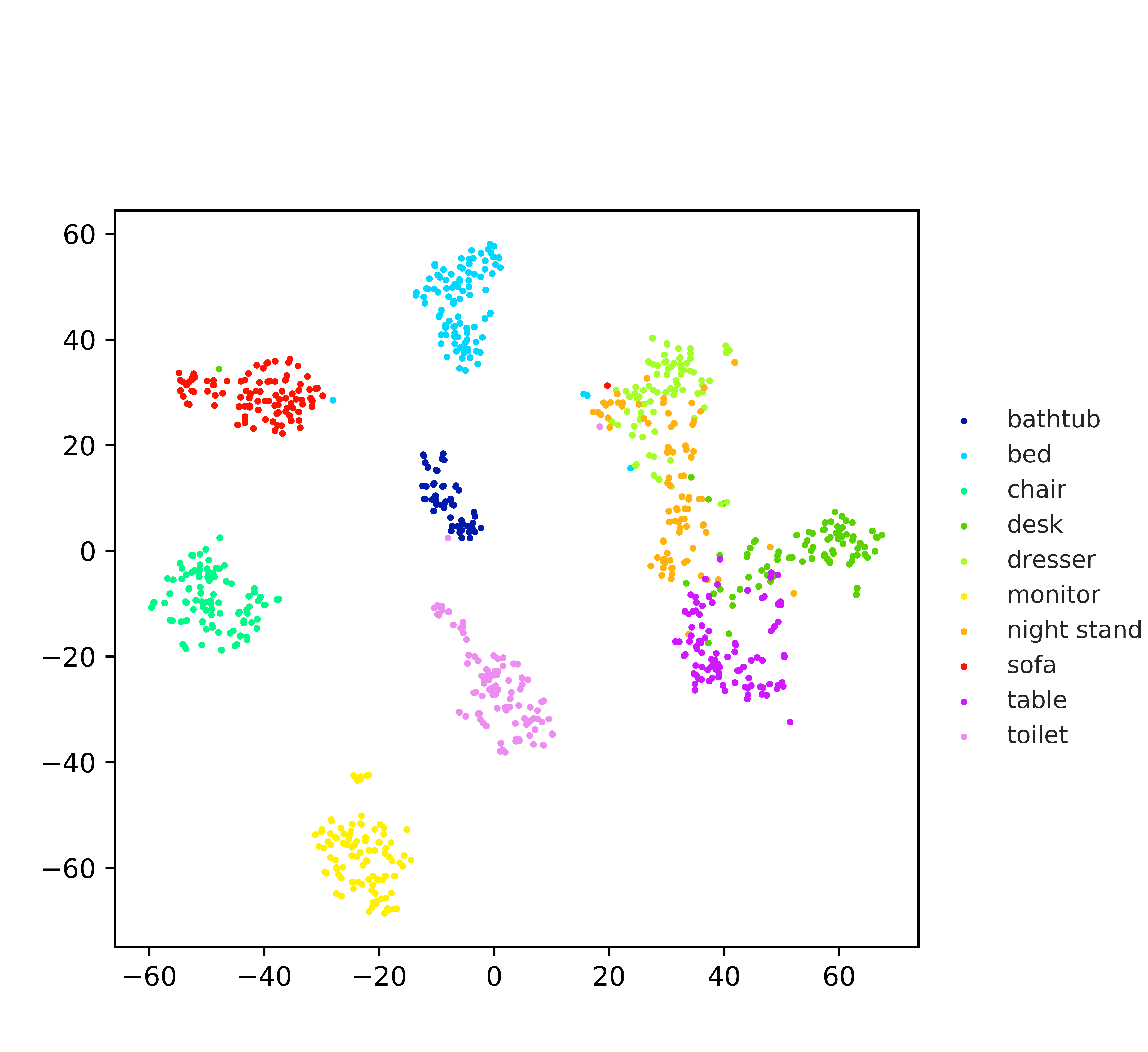}
\end{center}
\caption{The t-SNE clustering visualization of the learned global shape features from the proposed A-CNN model for the shapes in \emph{ModelNet10} test split.}
\centering
\vspace{-5.5mm}
\label{fig:t-sne10}
\end{figure}

\begin{figure}[h]
\begin{center}
\includegraphics[width=0.9\linewidth, trim={0.5cm 0.5cm 1.7cm 2.5cm}, clip]{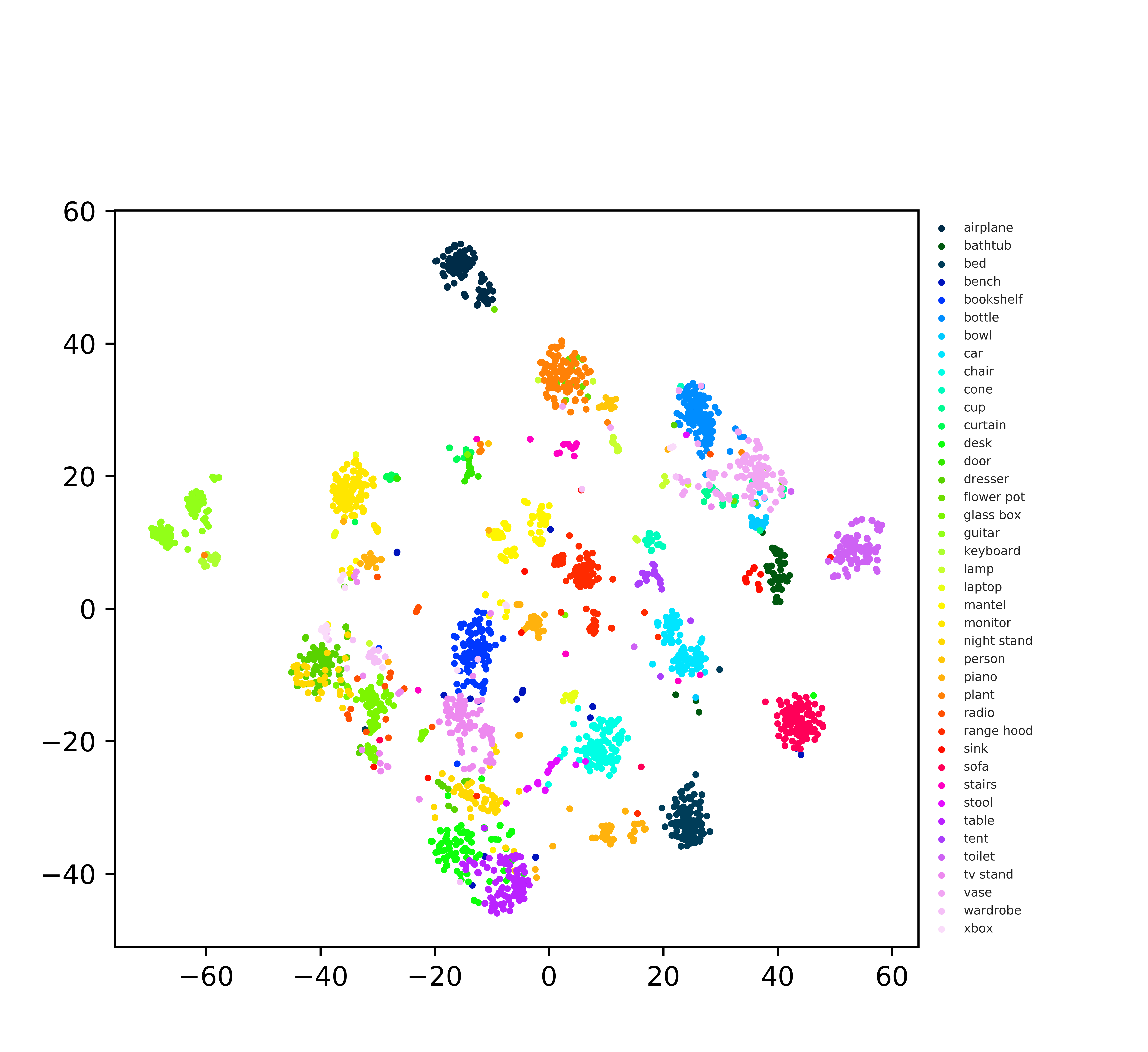}\vspace{-5.5mm}
\end{center}
\caption{The t-SNE clustering visualization of the learned global shape features from the proposed A-CNN model for the shapes in \emph{ModelNet40} test split.}
\centering
\vspace{-5.5mm}
\label{fig:t-sne}
\end{figure}

\textbf{Global Feature Visualization. }Fig.~\ref{fig:t-sne10} and Fig.~\ref{fig:t-sne} shows the t-SNE clustering visualization~\cite{maaten2008visualizing} of the learned global shape features from the proposed A-CNN model for the shape classification tasks in \emph{ModelNet10} and \emph{ModelNet40} test splits. We reduce 1024-dim feature vectors to 2-dim features. We can see that similar shapes are well clustered together according to their semantic categories. For example, in \textit{ModelNet10} dataset the clusters of desk, dresser, night stand, and table classes are closer and even intersect with each other, because the objects from these classes look similar. The perplexity parameters for \emph{ModelNet10} and \emph{ModelNet40} datasets are set as 15 and 50, respectively, to reduce spare space between clusters.

\section{Data Preparation Details}
\label{sec:suppl_data_preparation_details}
\begin{table*}[t]
\centering
\small 
\caption{Segmentation results on \emph{ShapeNet-part} dataset (input is \emph{XYZ} only). Per-category and mean IoUs ($\%$) are reported.}\vspace{-2mm}
\scalebox{0.75}{%
\begin{tabular}{@{}p{3.0cm}|c|p{0.5cm}p{0.5cm}p{0.5cm}p{0.5cm}p{0.5cm}p{0.5cm}p{0.5cm}p{0.5cm}p{0.5cm}p{0.5cm}p{0.5cm}p{0.5cm}p{0.5cm}p{0.5cm}p{0.5cm}p{0.5cm}}
\hline
             & mean & areo & bag & cap & car & chair & ear phone & guitar & knife & lamp & laptop & motor & mug & pistol & rocket & skate board & table  \\ \hline
\# shapes  &       & 2690 & 76   & 55   & 898  & 3758 & 69   & 787  & 392  & 1547 & 451  & 202  & 184  & 283  & 66   & 152  & 5271   \\ \hline
 PointNet~\cite{qi2017pointnet}   &  83.7 & 83.4 & 78.7 & 82.5 & 74.9 & 89.6 & 73.0 & 91.5 & 85.9 & 80.8 & 95.3 & 65.2 & 93.0 & 81.2 & 57.9 & 72.8 & 80.6  \\
 Kd-Net~\cite{klokov2017escape} &  82.3 & 80.1 & 74.6 & 74.3 & 70.3 & 88.6 & 73.5 & 90.2 & 87.2 & 81.0 & 94.9 & 57.4 & 86.7 & 78.1 & 51.8 & 69.9 & 80.3  \\
 KCNet~\cite{shen2018mining}      &  84.7 & 82.8 & 81.5 & \textbf{86.4} & 77.6 & 90.3 & 76.8 & 91.0 & 87.2 & 84.5 & 95.5 & 69.2 & 94.4 & 81.6 & 60.1 & 75.2 & 81.3 \\
 PCNN~\cite{atzmon2018point}       &  85.1 & 82.4 & 80.1 & 85.5 & 79.5 & 90.8 & 73.2 & 91.3 & 86.0 & 85.0 & 95.7 & 73.2 & 94.8 & 83.3 & 51.0 & 75.0 & 81.8  \\
 PointGrid~\cite{le2018pointgrid} & \textbf{86.4} & \textbf{85.7} & 82.5 & 81.8 & 77.9 & \textbf{92.1} & \textbf{82.4} & \textbf{92.7} & 85.8 & 84.2 & 95.3 & 65.2 & 93.4 & 81.7 & 56.9 & 73.5 & \textbf{84.6} \\
 PointCNN~\cite{li2018pointcnn} &  86.1 & 84.1 & 86.5 & 86.0 & \textbf{80.8} & 90.6 & 79.7 & 92.3 & \textbf{88.4} & \textbf{85.3} & \textbf{96.1} & \textbf{77.2} & \textbf{95.2} & \textbf{84.2} & \textbf{64.2} & \textbf{80.0} & 83.0 \\ \hline
 A-CNN (our) &  85.9 & 83.9 & \textbf{86.7} & 83.5 & 79.5 & 91.3 & 77.0 & 91.5 & 86.0 & 85.0 & 95.5 & 72.6 & 94.9 & 83.8 & 57.8 & 76.6 & 83.0 \\ \hline
\end{tabular}
}
\label{table:suppl_shapenetpart_results}
\end{table*}
\begin{table*}[t]
\centering
\small 
\caption
{Segmentation results on \emph{ShapeNet-part} dataset (input is \emph{XYZ + normals}). Per-category and mean IoUs ($\%$) are reported.}\vspace{-2mm}
\scalebox{0.75}{%
\begin{tabular}{@{}p{3.0cm}|c|p{0.5cm}p{0.5cm}p{0.5cm}p{0.5cm}p{0.5cm}p{0.5cm}p{0.5cm}p{0.5cm}p{0.5cm}p{0.5cm}p{0.5cm}p{0.5cm}p{0.5cm}p{0.5cm}p{0.5cm}p{0.5cm}}
\hline
             & mean & areo & bag & cap & car & chair & ear phone & guitar & knife & lamp & laptop & motor & mug & pistol & rocket & skate board & table  \\ \hline
\# shapes  &       & 2690 & 76   & 55   & 898  & 3758 & 69   & 787  & 392  & 1547 & 451  & 202  & 184  & 283  & 66   & 152  & 5271  \\ \hline
PointNet++~\cite{qi2017pointnet++} &  85.1 & 82.4 & 79.0 & 87.7 & 77.3 & 90.8 & 71.8 & 91.0 & 85.9 & 83.7 & 95.3 & 71.6 & 94.1 & 81.3 & 58.7 & 76.4 & 82.6 \\
SyncSpecCNN~\cite{yi2017syncspeccnn} & 84.7 & 81.6 & 81.7 & 81.9 & 75.2 & 90.2 & 74.9 & \textbf{93.0} & 86.1 & 84.7 & 95.6 & 66.7 & 92.7 & 81.6 & 60.6 & 82.9 & 82.1  \\
SO-Net~\cite{li2018so}     &  84.9 & 82.8 & 77.8 & \textbf{88.0} & 77.3 & 90.6 & 73.5 & 90.7 & 83.9 & 82.8 & 94.8 & 69.1 & 94.2 & 80.9 & 53.1 & 72.9 & 83.0  \\
SGPN~\cite{wang2018sgpn}  &  85.8 & 80.4 & 78.6 & 78.8 &  71.5 & 88.6 & 78.0 & 90.9 & 83.0 & 78.8 & \textbf{95.8} & \textbf{77.8} & 93.8 & \textbf{87.4} & 60.1 & \textbf{92.3} & \textbf{89.4} \\
RSNet~\cite{huang2018recurrent} & 84.9 & 82.7 & 86.4 & 84.1 & 78.2 & 90.4 & 69.3 & 91.4 & 87.0 & 83.5 & 95.4 & 66.0 & 92.6 & 81.8 & 56.1 & 75.8 & 82.2 \\
O-CNN (+ CRF)~\cite{wang2017cnn} &  85.9 & \textbf{85.5} & \textbf{87.1} & 84.7 & 77.0 & 91.1 & \textbf{85.1} & 91.9 & \textbf{87.4} & 83.3 & 95.4 & 56.9 & \textbf{96.2} & 81.6 & 53.5 & 74.1 & 84.4 \\
Point2Sequence~\cite{liu2018point2sequence} & 85.2 & 82.6 & 81.8 & 87.5 & 77.3 & 90.8 & 77.1 & 91.1 & 86.9 & 83.9 & 95.7 & 70.8 & 94.6 & 79.3 & 58.1 & 75.2 & 82.8 \\ \hline
A-CNN (our) &  \textbf{86.1} & 84.2 & 84.0 & \textbf{88.0} & \textbf{79.6} & \textbf{91.3} & 75.2 & 91.6 & 87.1 & \textbf{85.5} & 95.4 & 75.3 & 94.9 & 82.5 & \textbf{67.8} & 77.5 & 83.3  \\ \hline
\end{tabular}
}
 \begin{tablenotes}
      \centering
      \footnotesize 
      \item Note: ``CRF'' stands for conditional random field method for final result refinement in O-CNN method.\vspace{0mm}
    \end{tablenotes}
\label{table:suppl_shapenetpart_results2}
\end{table*}

\textbf{\emph{S3DIS} data preparation.} To prepare training and testing datasets, we divide every room into blocks with a size of 1 $m$ $\times$ 1 $m$ $\times$ 2 $m$ and with a stride of 0.5 $m$. We has sampled 4096 points from each block. The height of each block is scaled to 2 $m$ to ensure that our constraint-based k-NN search works optimally with the provided radiuses. In total, the prepared dataset contains 23,585 blocks across all six areas. Each point is represented as a 6D vector ($XYZ$: normalized global point coordinates and centered at origin, $RGB$: colors). We do not use the relative position of the block in the room scaled between 0 and 1 as used in~\cite{qi2017pointnet}, because our model already achieves better results without using this additional information. We calculate point normals for each room by using the Point Cloud Library (PCL) library~\cite{rusu20113d}. The calculated normals are only used to order points in the local region. For data augmentation, we use the same data augmentation strategy as used in the point cloud segmentation on \emph{ShapeNet-part} dataset which is point perturbation with point shuffling.

\textbf{\emph{ScanNet} data preparation}. ScanNet divides original 1513 scanned scenes in 1201 and 312 for training and testing, respectively. We sample blocks from the scenes following the same procedure as in~\cite{qi2017pointnet++}, where every block has a size of 1.5 $m$ $\times$ 1.5 $m$ with 8192 points. We estimate point normals using the PCL library~\cite{rusu20113d}. Each point is represented as a 6D vector ($XYZ$: coordinates of the block centered at origin, $N_xN_yN_z$: normals) without $RGB$ information. For data augmentation, we use the point perturbation with point shuffling.

\section{More Experimental Results}
\label{sec:suppl_more_experimental_results} 
\textbf{Point Cloud Segmentation. }Tab.~\ref{table:suppl_shapenetpart_results} and Tab.~\ref{table:suppl_shapenetpart_results2} show the quantitative results of part segmentation on \emph{ShapeNet-part} dataset with two different inputs. Tab.~\ref{table:suppl_shapenetpart_results} reports results when the input is point position only. Tab.~\ref{table:suppl_shapenetpart_results2} reports results when the input is point position with its normals.

\begin{table*}[t]
\centering
\small
\caption{Segmentation results on \emph{S3DIS} dataset. ``acc'' is overall accuracy and ``mean'' is average IoU over 13 classes.}\vspace{-2mm}
\scalebox{0.82}{%
\begin{tabular}{@{}l|cc|ccccccccccccc}
\hline
             & acc & mean & ceiling & floor & wall & beam & column & window & door & table & chair & sofa & bookcase & board & clutter \\ \hline
PointNet~\cite{qi2017pointnet}  & 78.5 &  47.6 & 88.0 & 88.7 & 69.3 & 42.4 & 23.1 & 47.5 & 51.6 & 54.1 & 42.0 &  9.6 & 38.2 & 29.4 & 35.2 \\
MS+CU (2)~\cite{engelmann2017exploring} & 79.2 &  47.8 & 88.6 & 95.8 & 67.3 & 36.9 & 24.9 & 48.6 & 52.3 & 51.9 & 45.1 & 10.6 & 36.8 & 24.7 & 37.5 \\
G+RCU~\cite{engelmann2017exploring}   & 81.1  &  49.7 & 90.3 & 92.1 & 67.9 & 44.7 & 24.2 & 52.3 & 51.2 & 58.1 & 47.4 &  6.9 & 39.0 & 30.0 & 41.9 \\
RSNet~\cite{huang2018recurrent} &- & 56.5 & 92.5 & 92.8 & 78.6 & 32.8 & 34.4 & 51.6 & 68.1 & 59.7 & 60.1 & 16.4 & 50.2 & 44.9 & 52.0 \\
3P-RNN~\cite{Ye_2018_ECCV}  & 86.9  & 56.3 & 92.9 & 93.8 & 73.1 & 42.5 & 25.9 & 47.6 & 59.2 & 60.4 & 66.7 & 24.8 & 57.0 & 36.7 & 51.6 \\
SPGraph~\cite{Landrieu2017LargescalePC} & 85.5 & 62.1 & 89.9 & 95.1 & 76.4 & 62.8 & 47.1 & 55.3 & \textbf{68.4} & \textbf{73.5} & 69.2 & \textbf{63.2} & 45.9 & 8.7 & 52.9 \\ 
PointCNN~\cite{li2018pointcnn}     & \textbf{88.1} & \textbf{65.4} & \textbf{94.8} & \textbf{97.3} & 75.8 & \textbf{63.3} & \textbf{51.7} & \textbf{58.4} & 57.2 & 71.6 & 69.1 & 39.1 & \textbf{61.2} & \textbf{52.2} & \textbf{58.6}  \\ \hline
A-CNN (our)  & 87.3 & 62.9 & 92.4 & 96.4 & \textbf{79.2} & 59.5 & 34.2 & 56.3 & 65.0 & 66.5 & \textbf{78.0} & 28.5 & 56.9 & 48.0 & 56.8  \\ \hline
\end{tabular}
}
\label{table:suppl_s3dis_eval}
\end{table*}
For \emph{ShapeNet-part} dataset, we visualize more results (besides the segmentation results shown in the paper) in Fig.~\ref{fig:suppl_shapenet_part_eval}. We compare our results with PointNet++~\cite{qi2017pointnet++}, and our A-CNN model can produce better segmentation results than PointNet++ model.

\textbf{Semantic Segmentation in Scenes. }
For \emph{S3DIS} dataset, we pick rooms from all six areas: area 1 (\textit{row 1}), area 2 (\textit{row 2}), area 3 (\textit{row 3}), area 4 (\textit{row 4}), area 5 (\textit{row 5}), and area 6 (\textit{row 6}); and compare them with PointNet~\cite{qi2017pointnet} results and ground truth. The results are shown in Fig.~\ref{fig:suppl_s3dis_eval}. The detailed quantitative evaluation results for each shape class are reported in Tab.~\ref{table:suppl_s3dis_eval}. Our model demonstrates good semantic segmentation results and achieves the state-of-the-art performance on segmenting \textit{walls} and \textit{chairs}. Meanwhile, our model performs slightly worse than PointCNN~\cite{li2018pointcnn} on other categories due to their non-overlapping block sampling strategy with paddings which we do not use. Supplementary Video is included for dynamically visualizing each area in detail.

For \emph{ScanNet} dataset, we pick six challenging scenes and visualize the results of our A-CNN model, PointNet++~\cite{qi2017pointnet++}, and ground truth side by side. The visualization results are provided in Fig.~\ref{fig:suppl_scannet_eval}. Our approach outperforms PointNet++~\cite{qi2017pointnet++} and other baseline methods, such as PointNet~\cite{qi2017pointnet}, TangentConv~\cite{tatarchenko2018tangent}, and PointCNN~\cite{li2018pointcnn} according to Tab.~\ref{table:s3dis_eval} in the main paper.

\setcounter{figure}{-32}
\begin{figure*}[t]
\centering \vspace{-3.5mm}
\subfloat{\includegraphics[width=0.05\linewidth,trim={0 0 0 0}, clip]{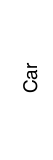} \phantomcaption} \hfill
\subfloat{\includegraphics[width=0.16\linewidth,trim={0 0 0 0}, clip]{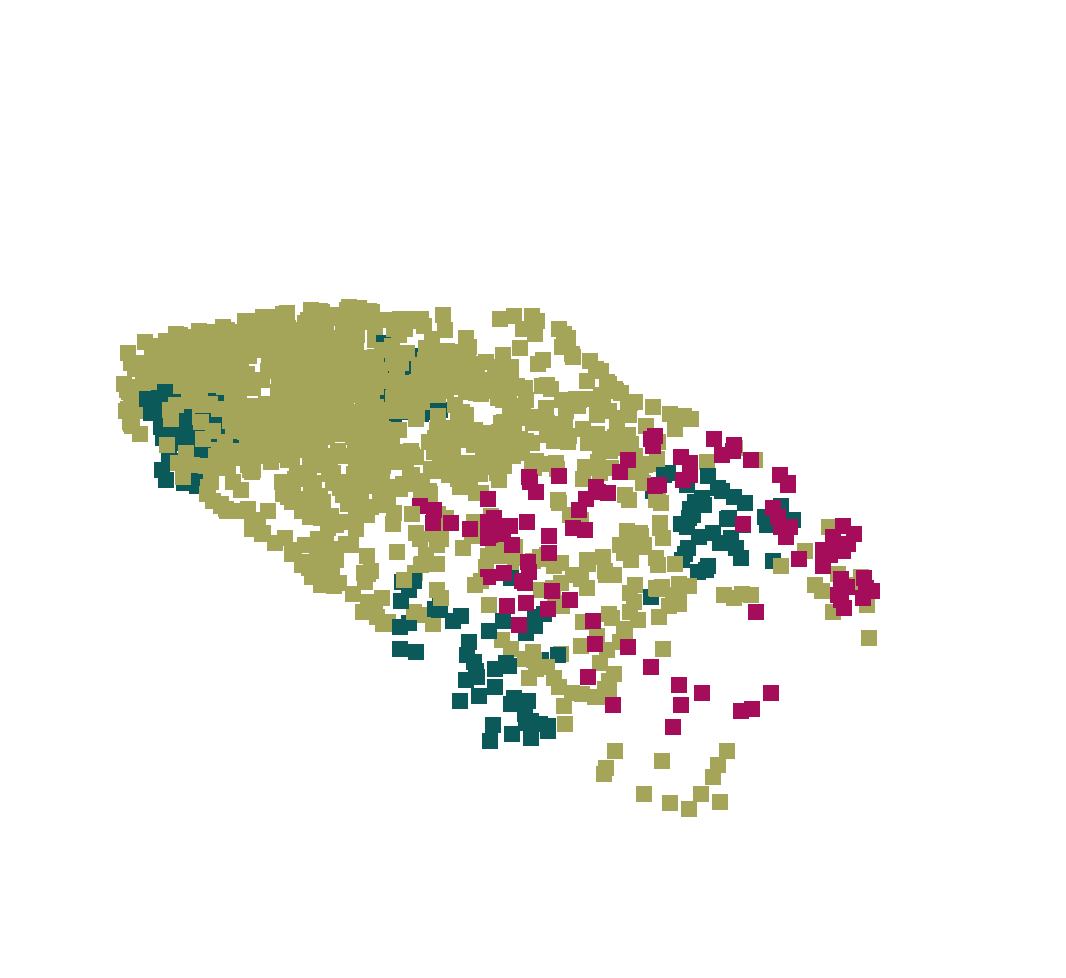} \phantomcaption} \hfill
\subfloat{\includegraphics[width=0.16\linewidth,trim={0 0 0 0}, clip]{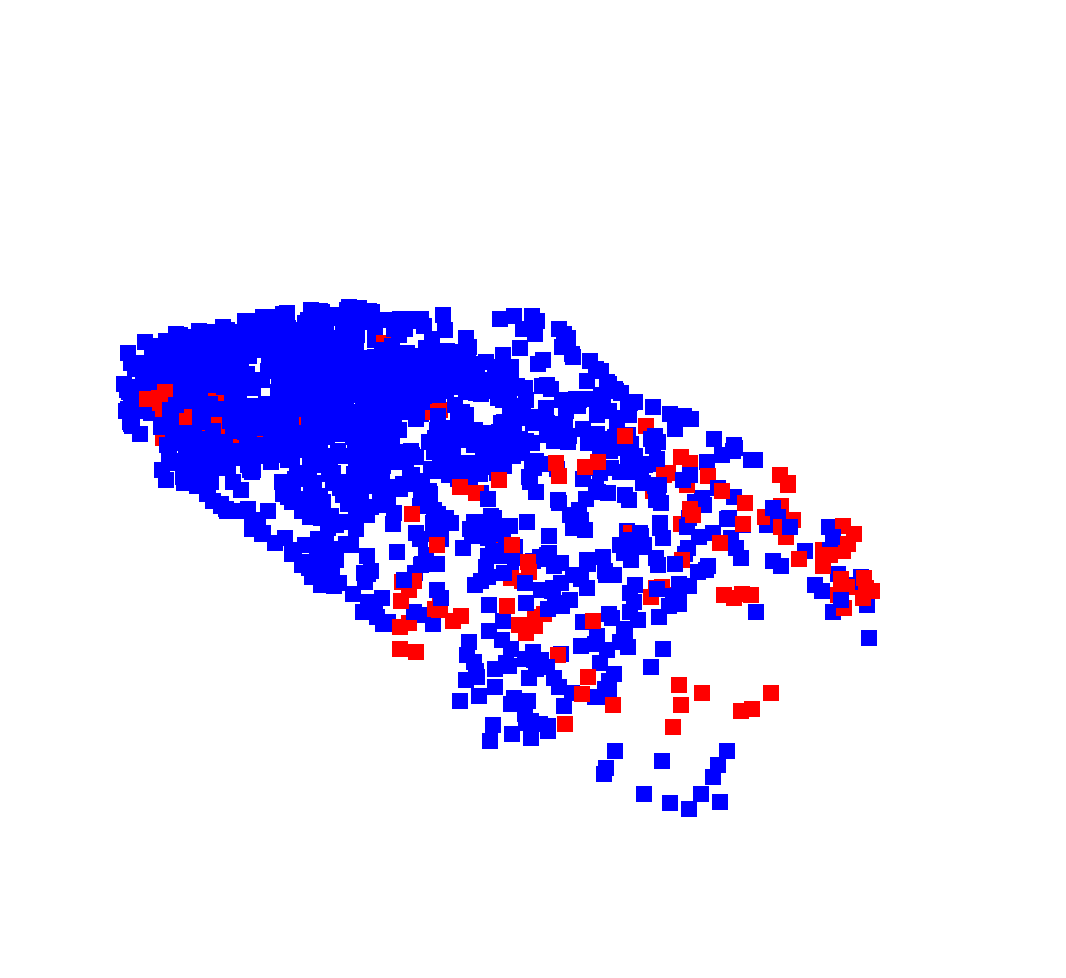} \phantomcaption} \hfill
\subfloat{\includegraphics[width=0.16\linewidth,trim={0 0 0 0}, clip]{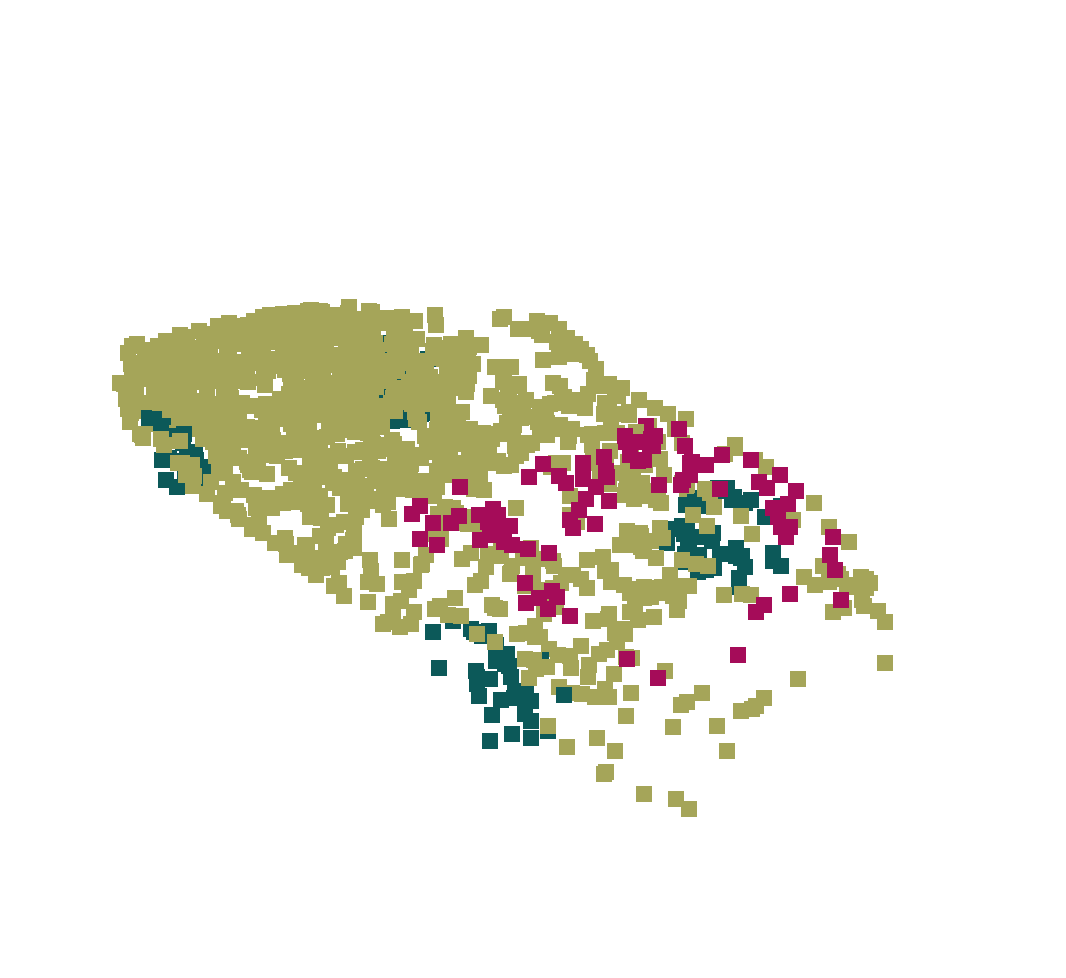} \phantomcaption} \hfill
\subfloat{\includegraphics[width=0.16\linewidth,trim={0 0 0 0}, clip]{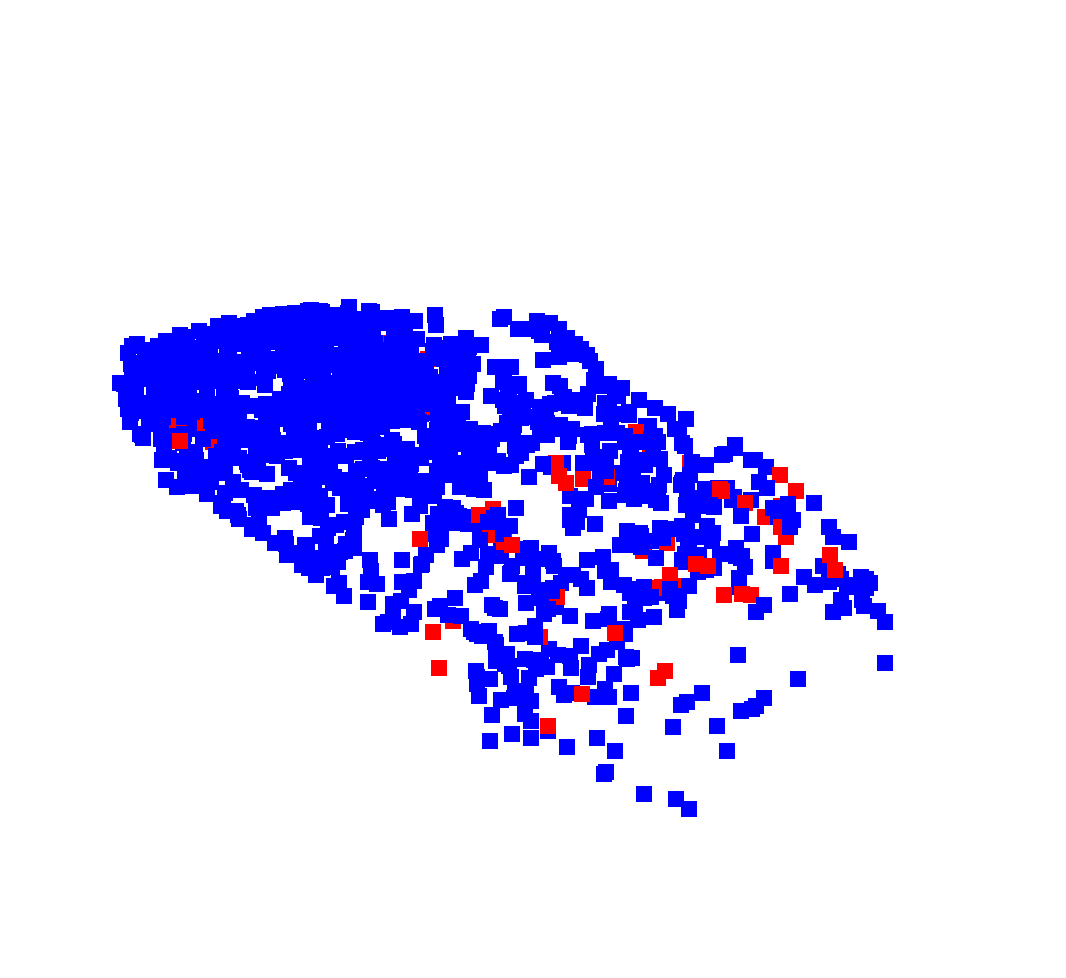} \phantomcaption} \hfill
\subfloat{\includegraphics[width=0.16\linewidth,trim={0 0 0 0}, clip]{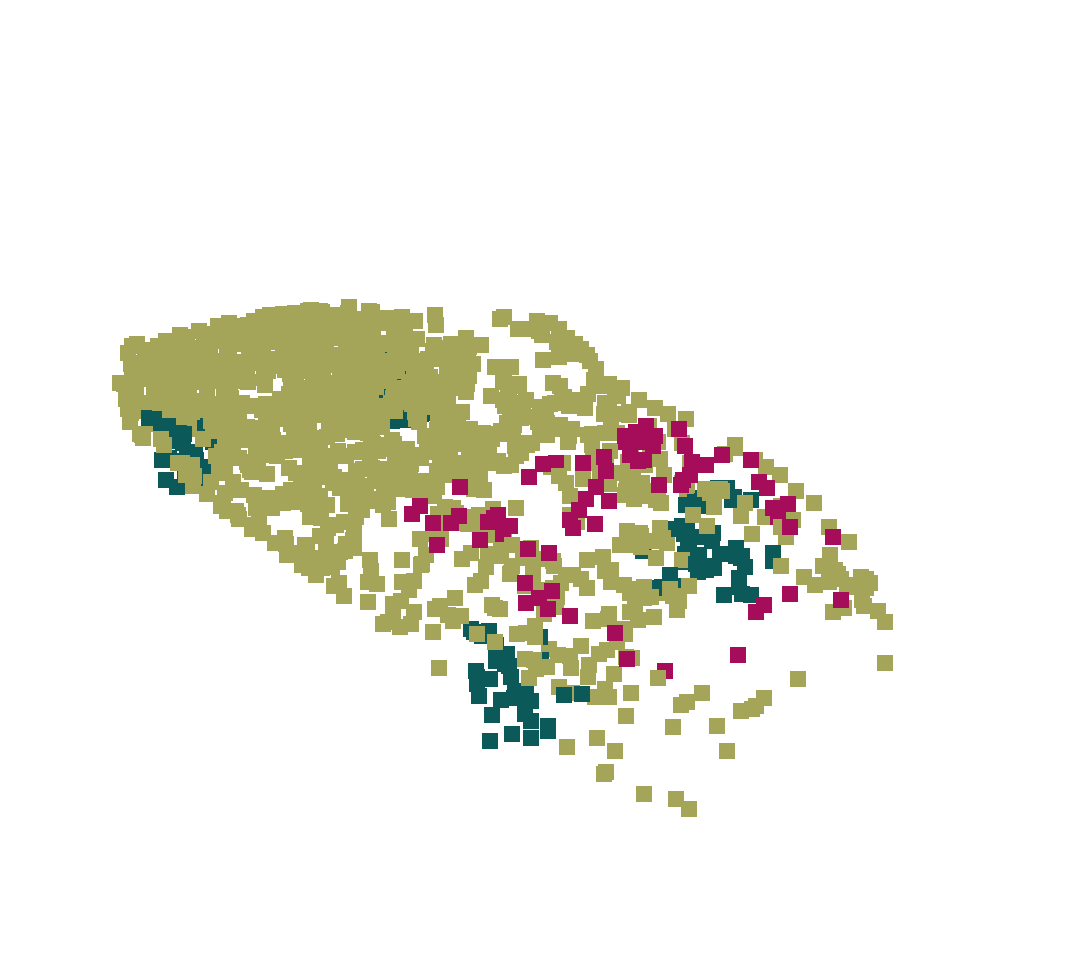} \phantomcaption}\vspace{-3.5mm}

\subfloat{\includegraphics[width=0.05\linewidth,trim={0 0 0 0}, clip]{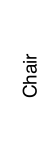} \phantomcaption} \hfill
\subfloat{\includegraphics[width=0.16\linewidth,trim={0 0 0 0}, clip]{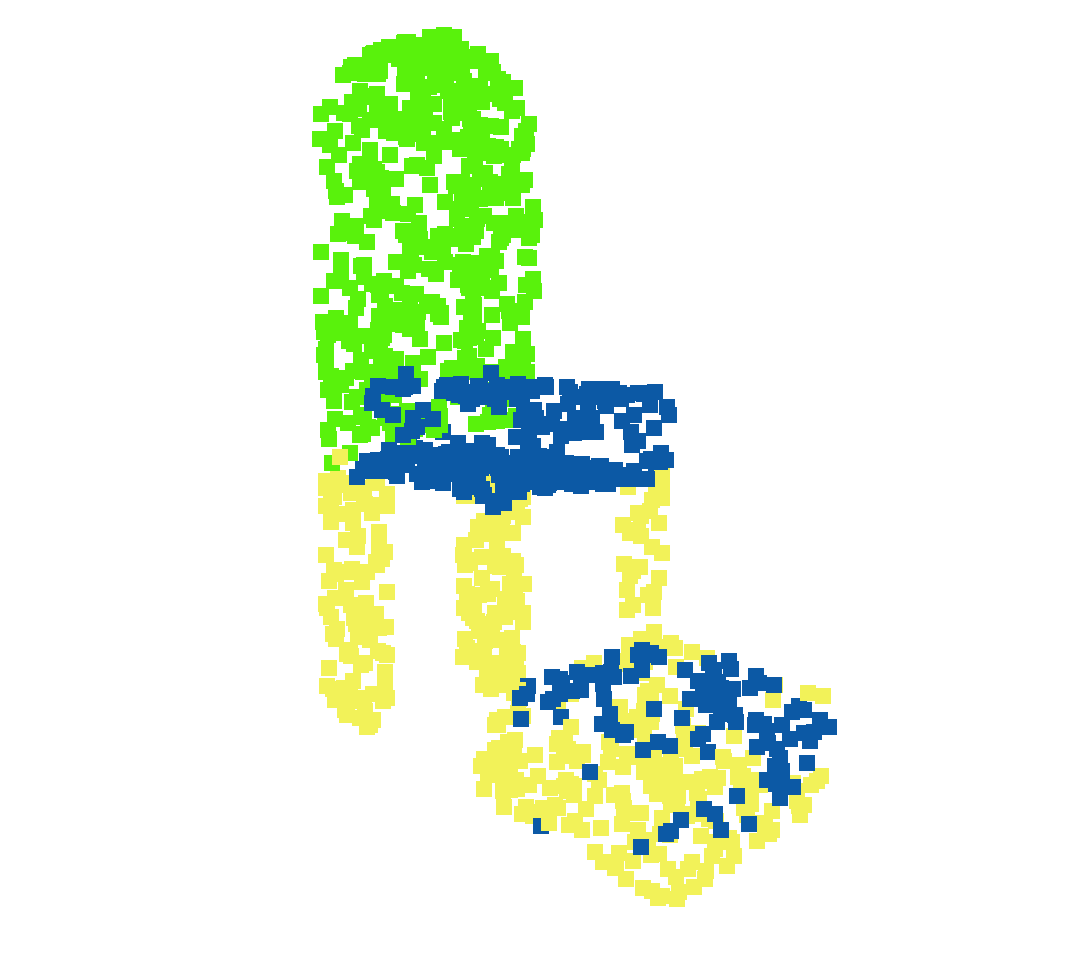} \phantomcaption} \hfill
\subfloat{\includegraphics[width=0.16\linewidth,trim={0 0 0 0}, clip]{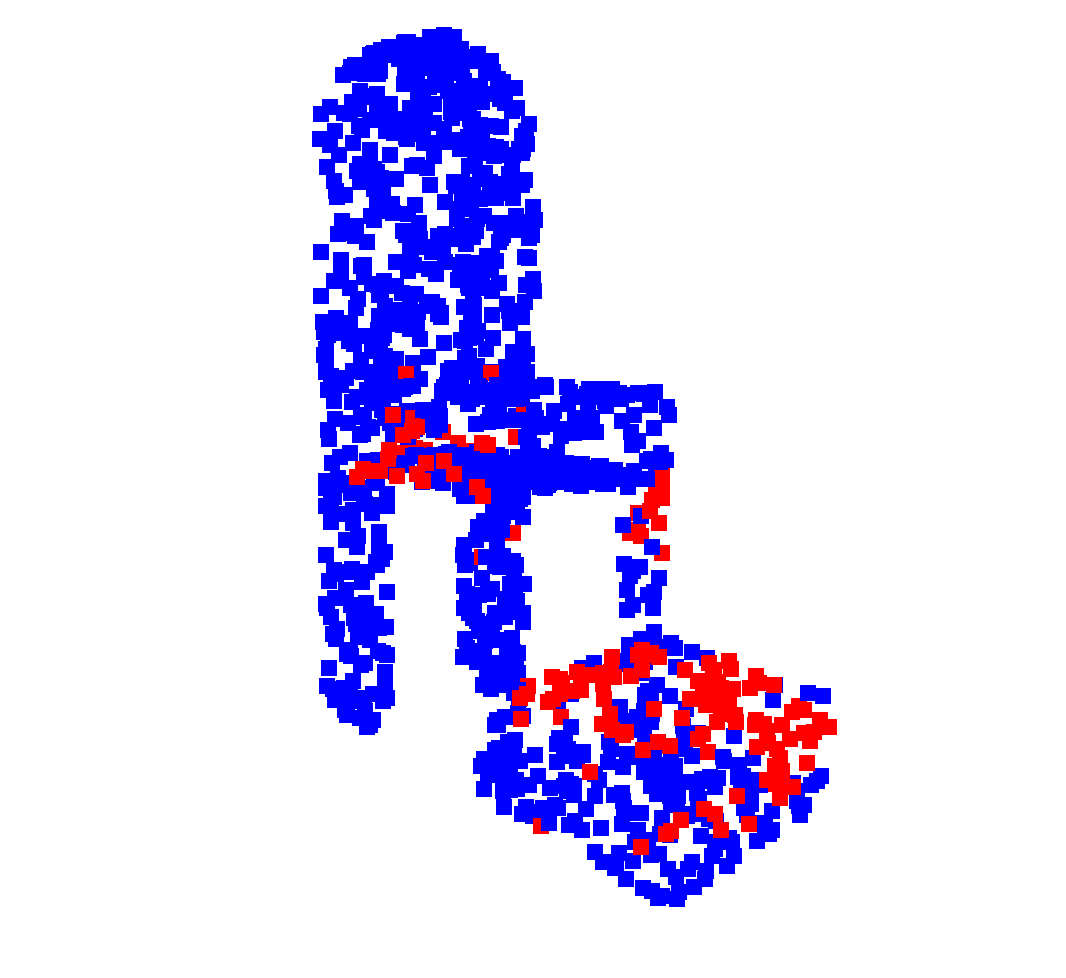} \phantomcaption} \hfill
\subfloat{\includegraphics[width=0.16\linewidth,trim={0 0 0 0}, clip]{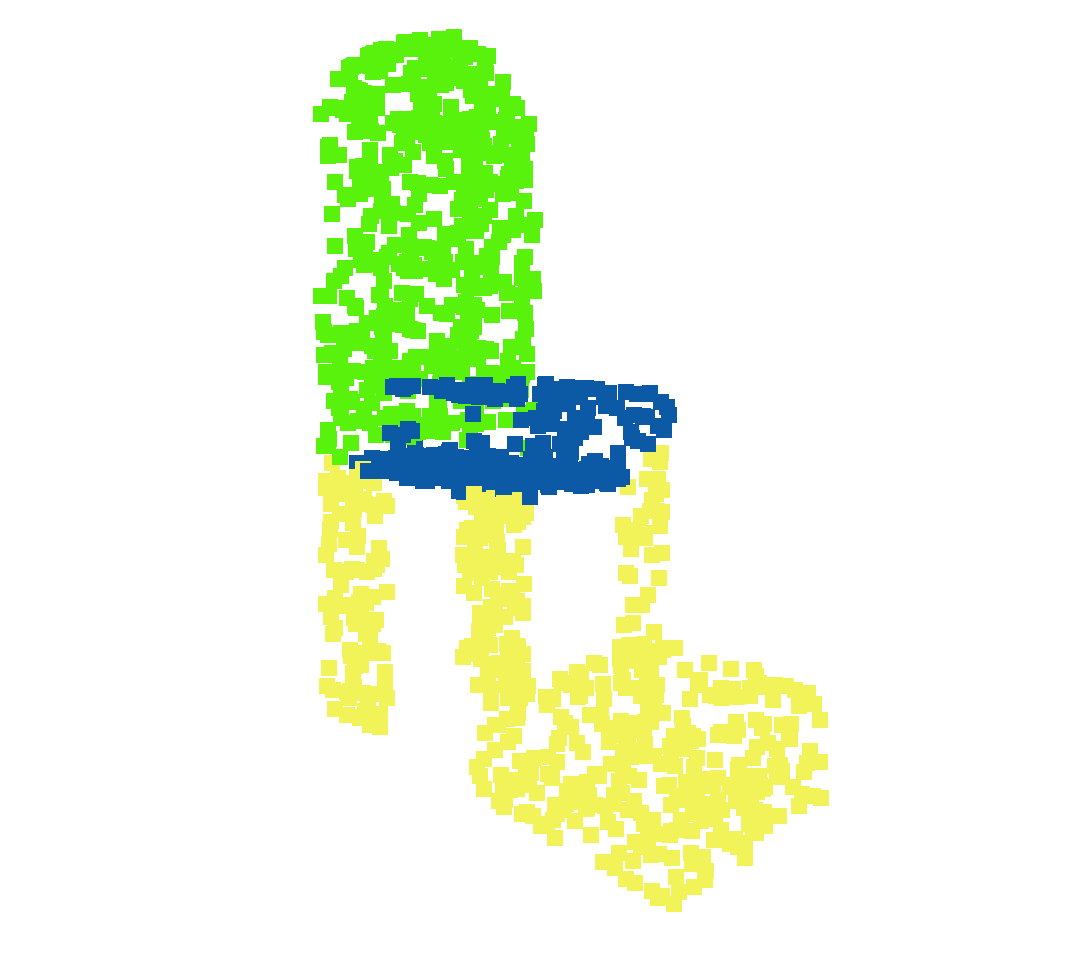} \phantomcaption} \hfill
\subfloat{\includegraphics[width=0.16\linewidth,trim={0 0 0 0}, clip]{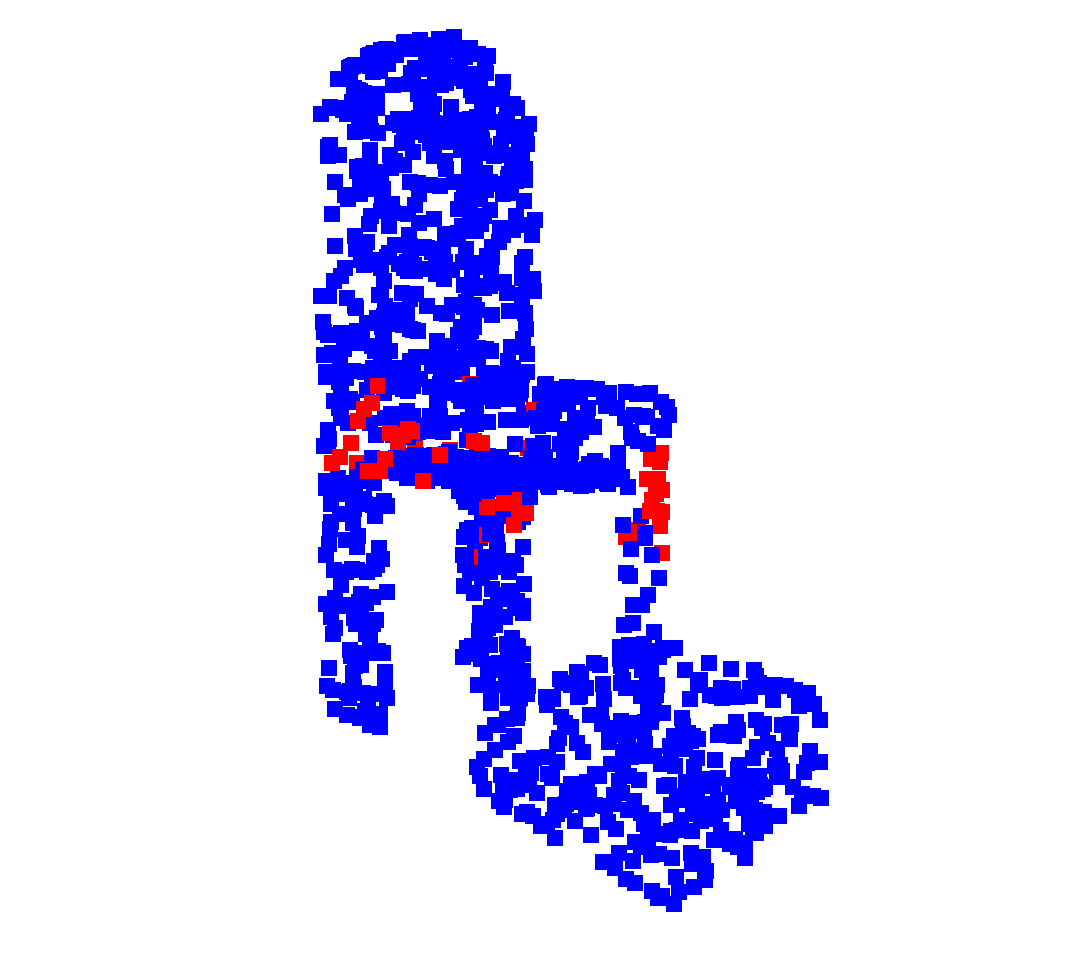} \phantomcaption} \hfill
\subfloat{\includegraphics[width=0.16\linewidth,trim={0 0 0 0}, clip]{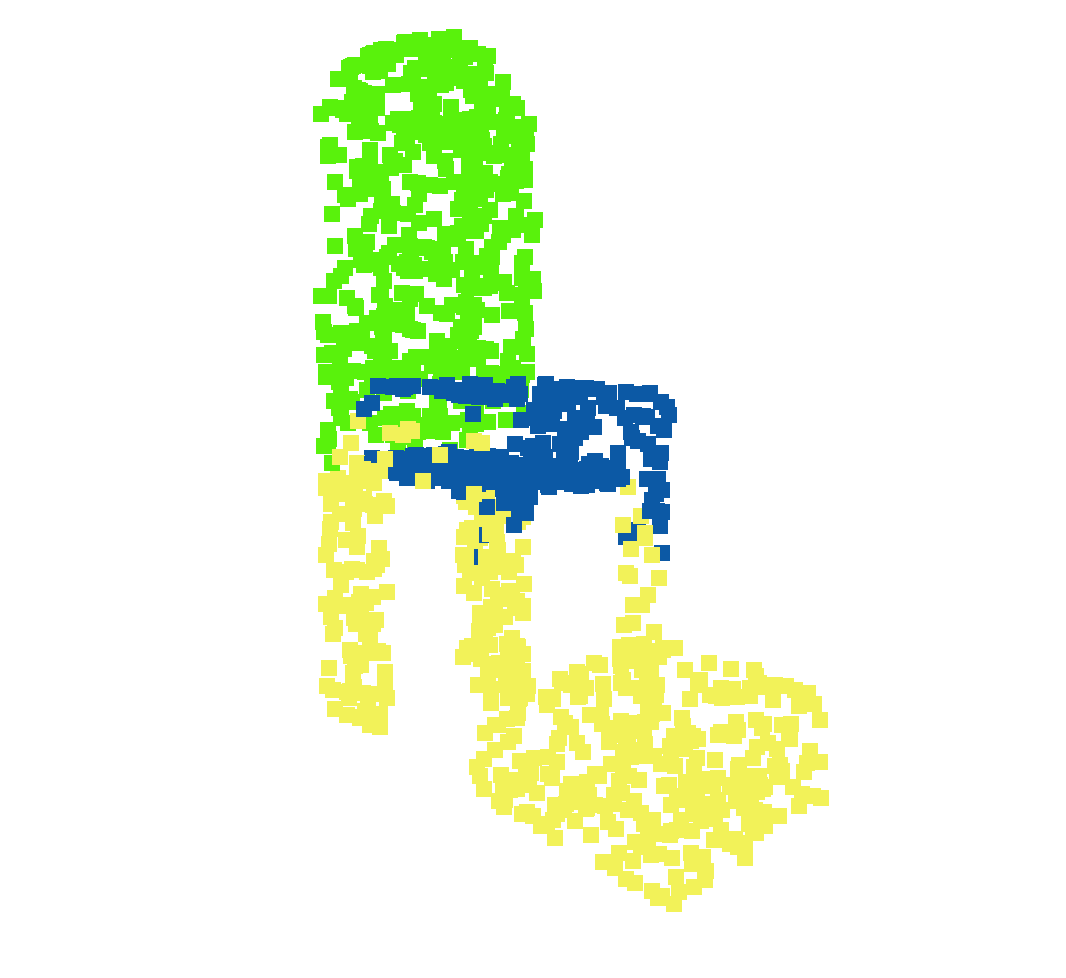} \phantomcaption}\vspace{-3.5mm}

\subfloat{\includegraphics[width=0.05\linewidth,trim={0 0 0 0}, clip]{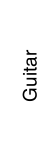} \phantomcaption} \hfill
\subfloat{\includegraphics[width=0.16\linewidth,trim={0 0 0 0}, clip]{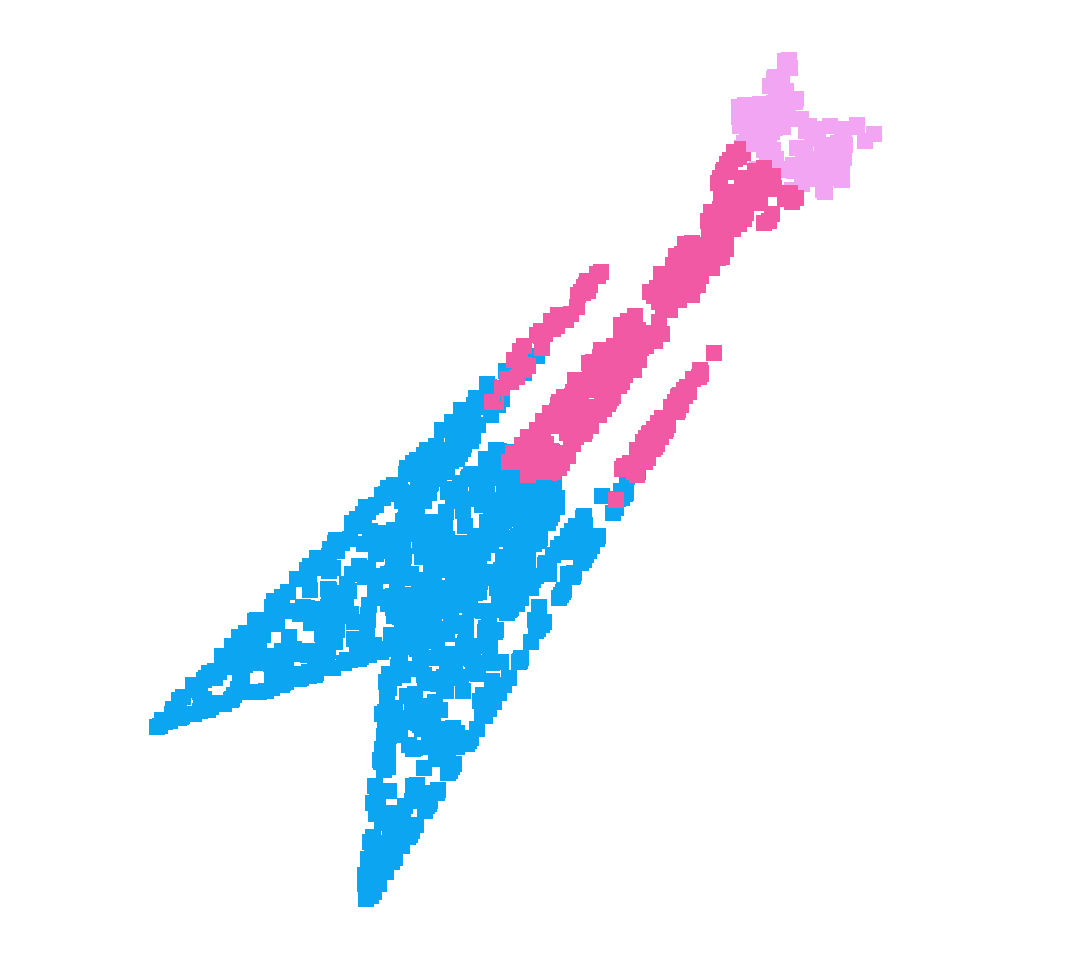} \phantomcaption} \hfill
\subfloat{\includegraphics[width=0.16\linewidth,trim={0 0 0 0}, clip]{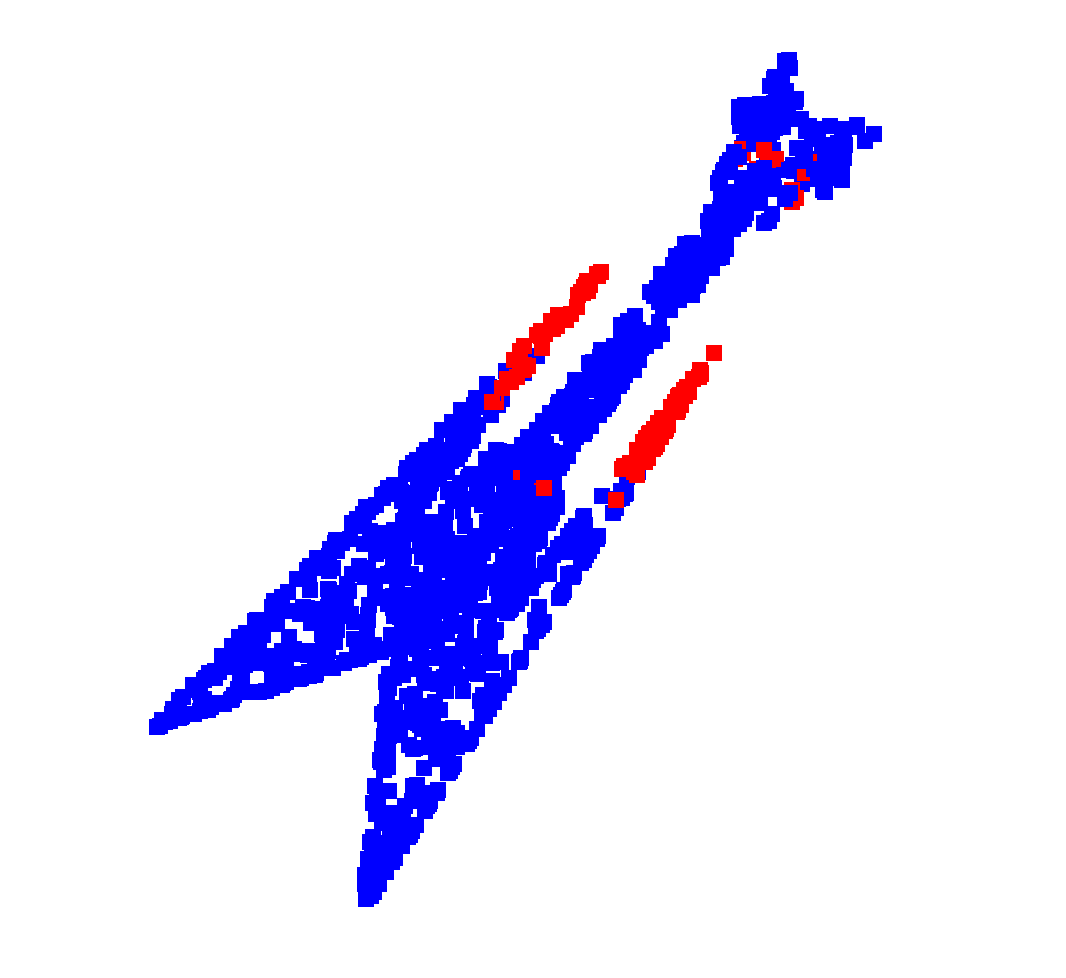} \phantomcaption} \hfill
\subfloat{\includegraphics[width=0.16\linewidth,trim={0 0 0 0}, clip]{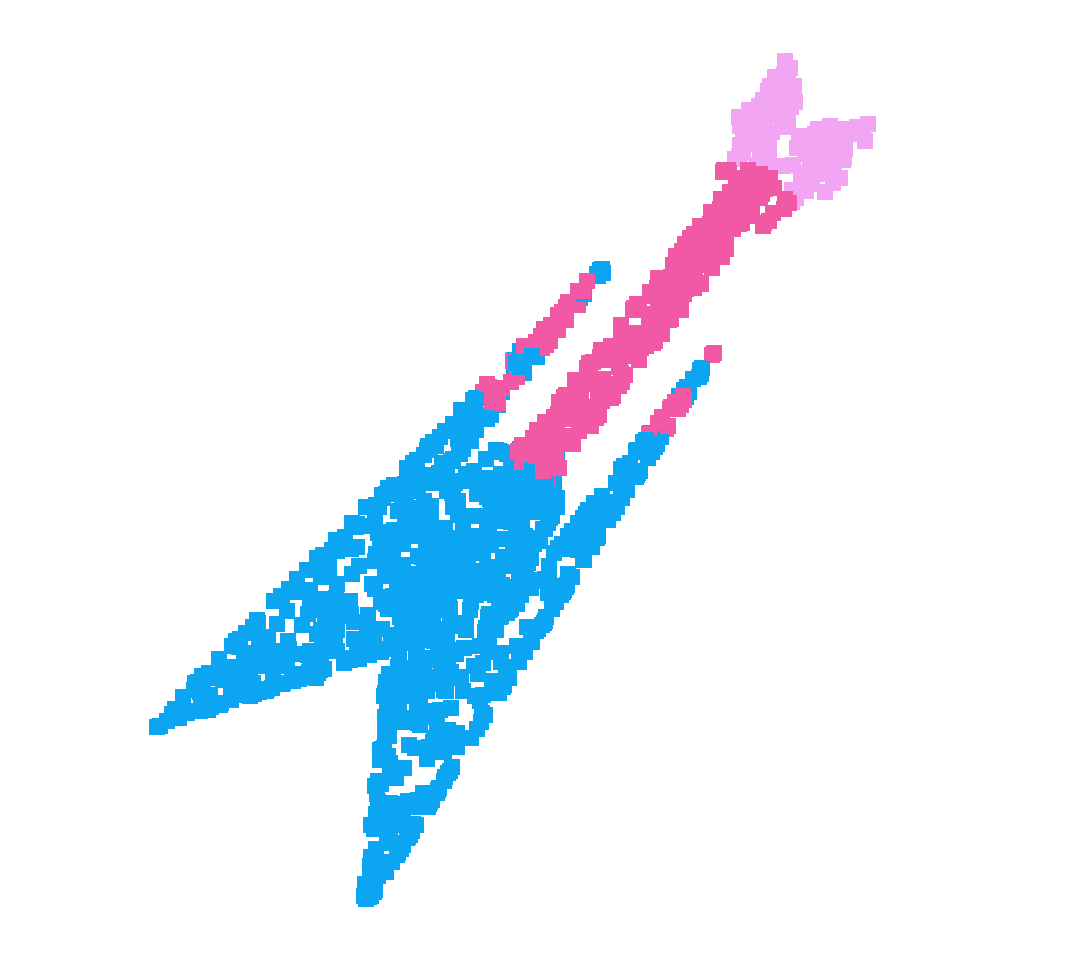} \phantomcaption} \hfill
\subfloat{\includegraphics[width=0.16\linewidth,trim={0 0 0 0}, clip]{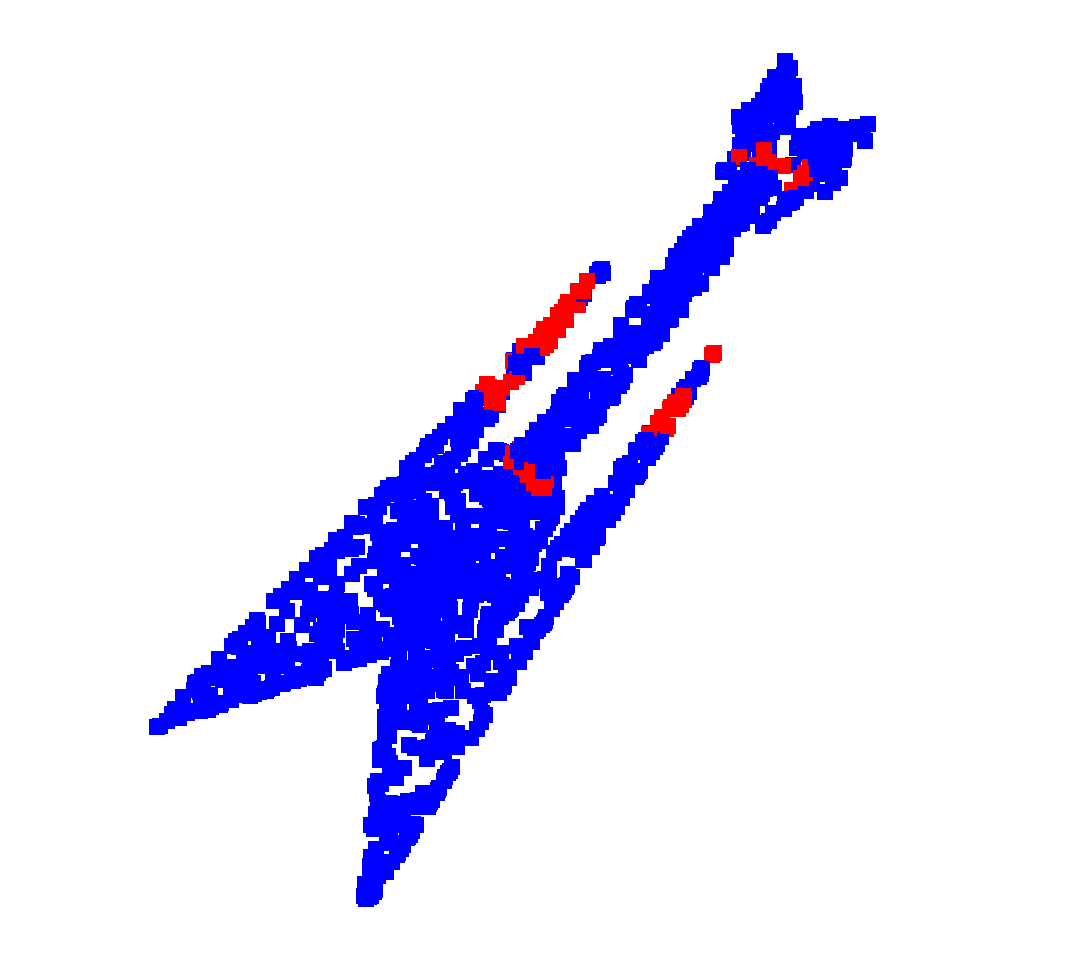} \phantomcaption} \hfill
\subfloat{\includegraphics[width=0.16\linewidth,trim={0 0 0 0}, clip]{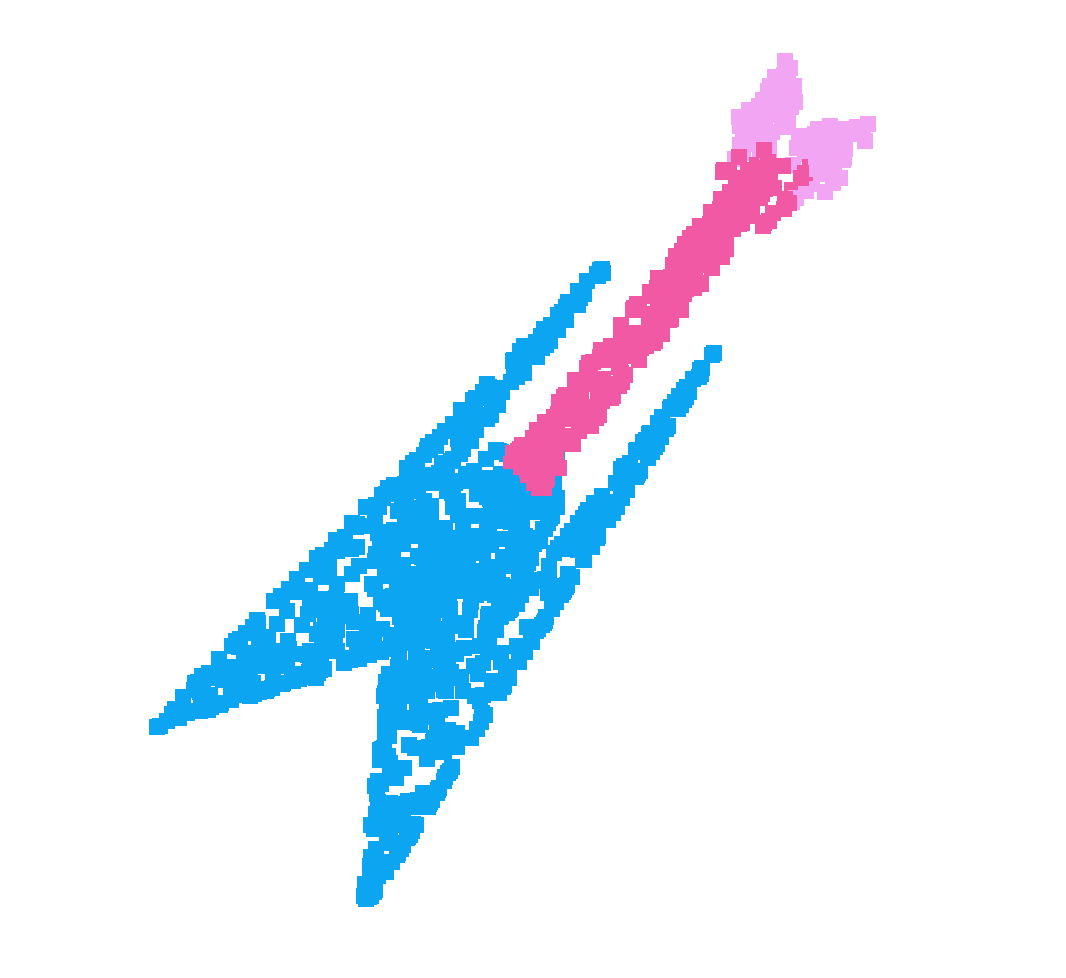} \phantomcaption}\vspace{-3.5mm}

\subfloat{\includegraphics[width=0.05\linewidth,trim={0 0 0 0}, clip]{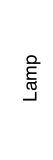} \phantomcaption} \hfill
\subfloat{\includegraphics[width=0.16\linewidth,trim={0 0 0 0}, clip]{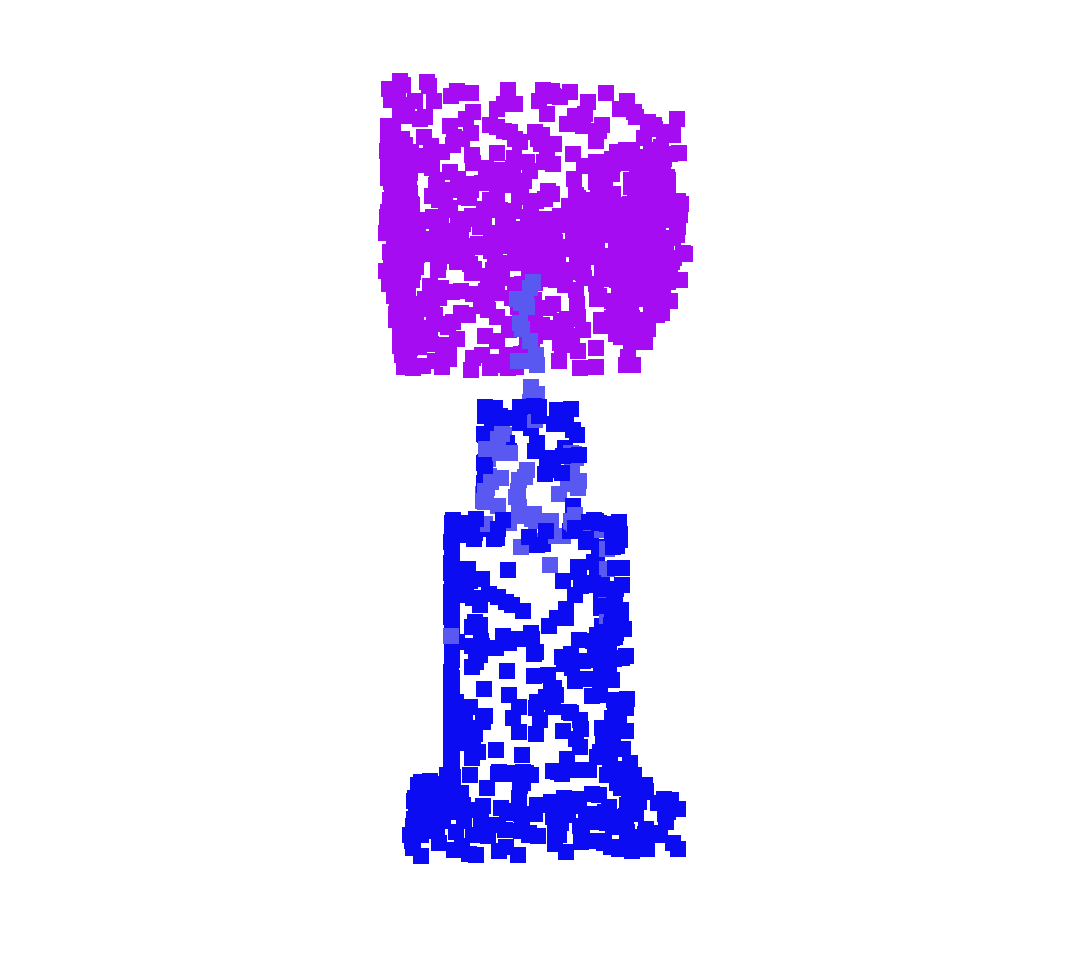} \phantomcaption} \hfill
\subfloat{\includegraphics[width=0.16\linewidth,trim={0 0 0 0}, clip]{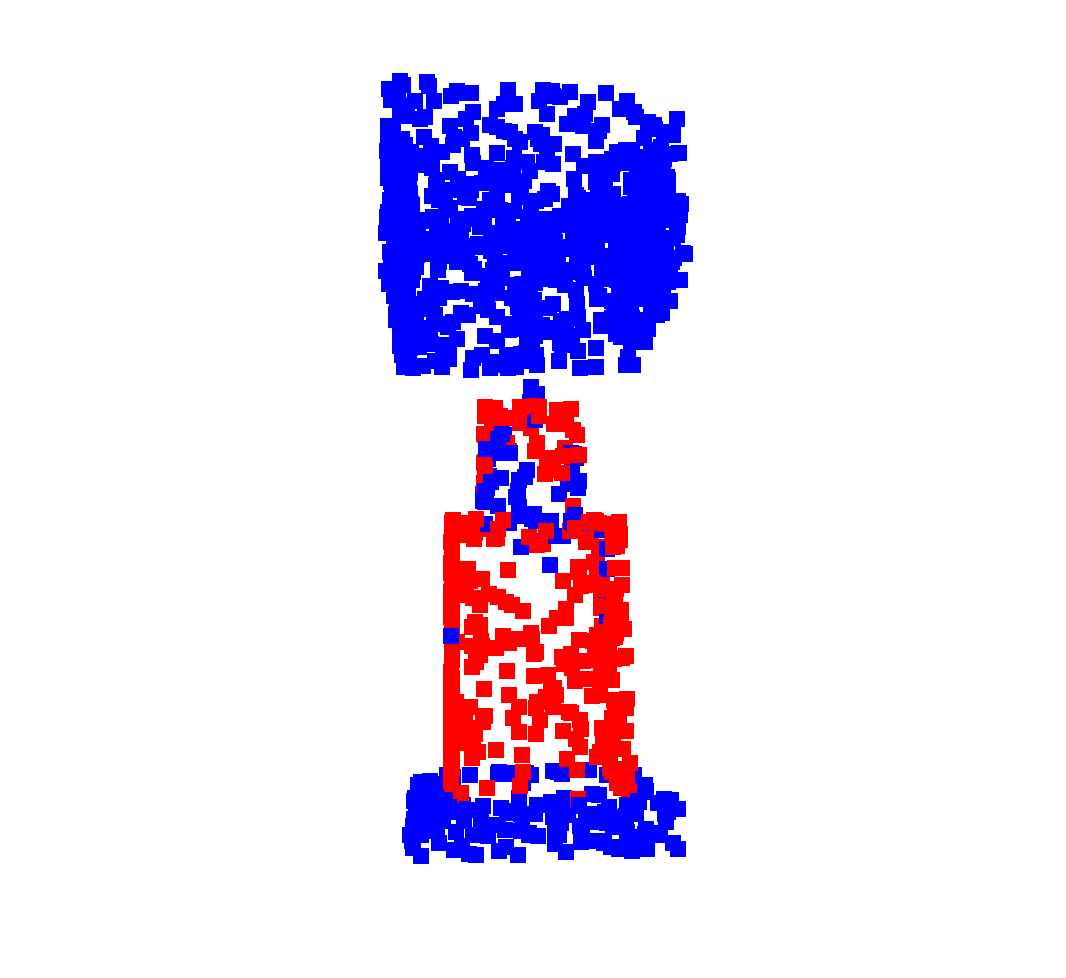} \phantomcaption} \hfill
\subfloat{\includegraphics[width=0.16\linewidth,trim={0 0 0 0}, clip]{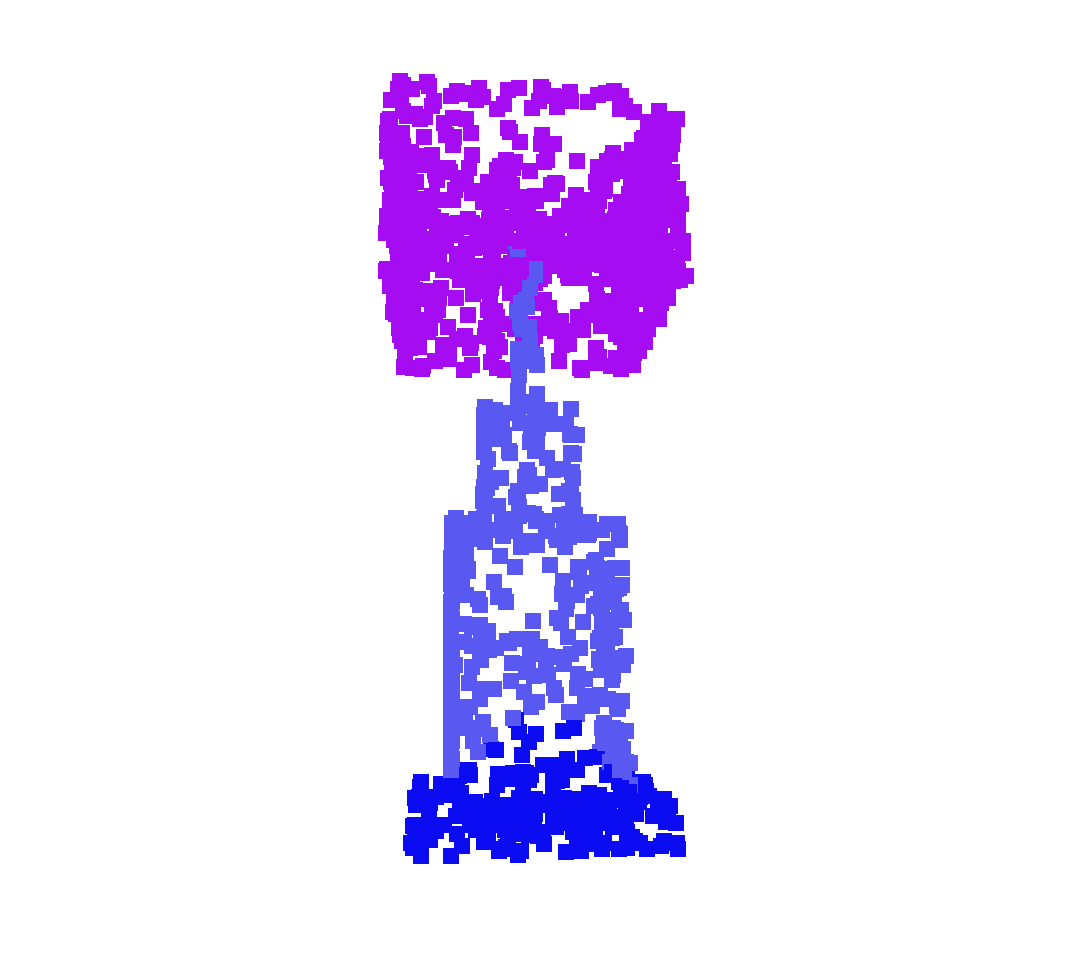} \phantomcaption} \hfill
\subfloat{\includegraphics[width=0.16\linewidth,trim={0 0 0 0}, clip]{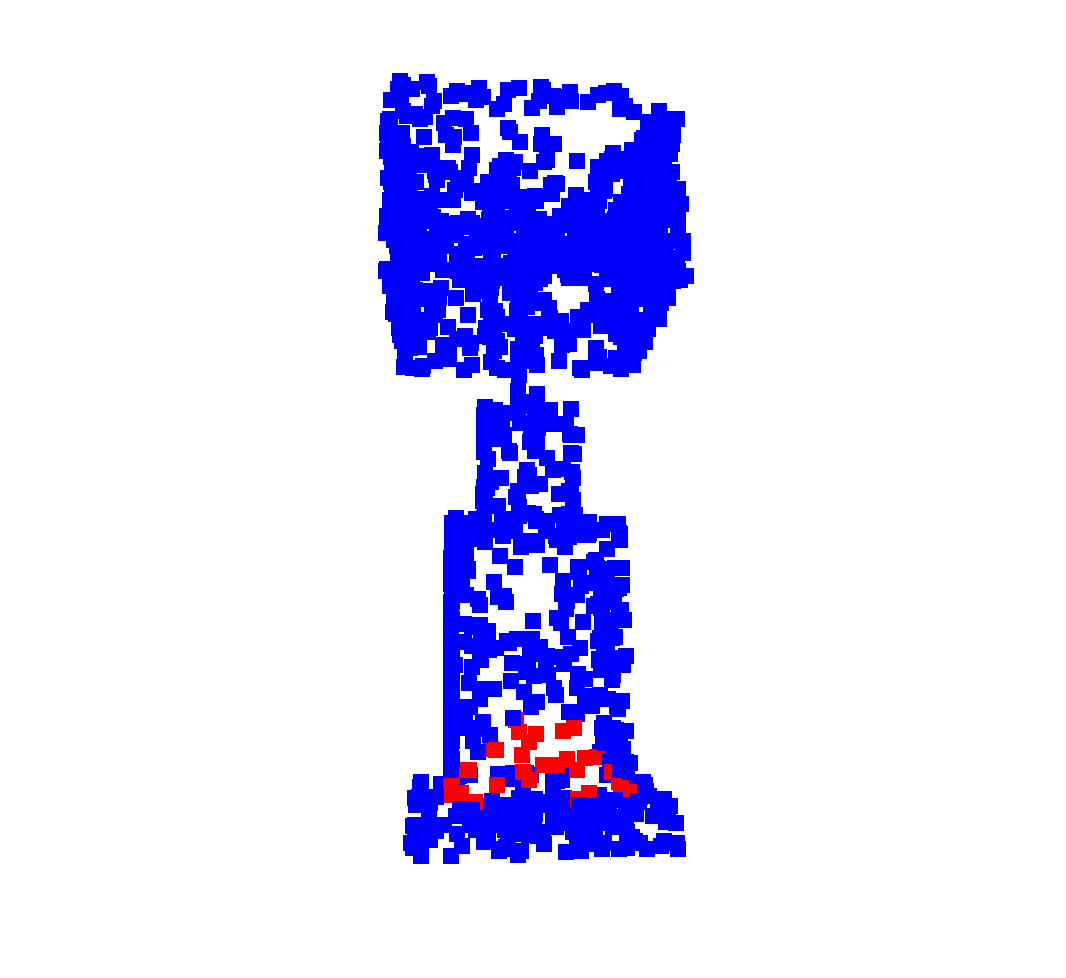} \phantomcaption} \hfill
\subfloat{\includegraphics[width=0.16\linewidth,trim={0 0 0 0}, clip]{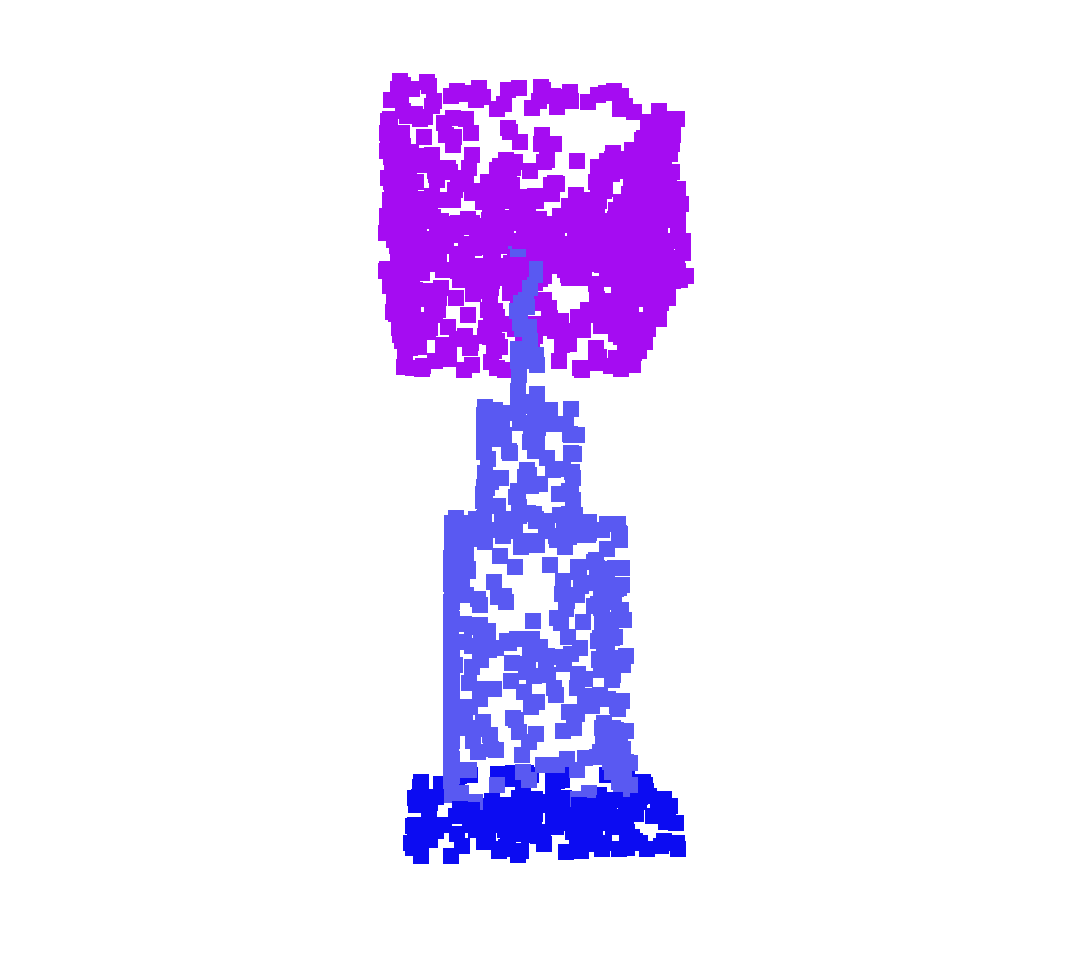} \phantomcaption}\vspace{-3.5mm}

\subfloat{\includegraphics[width=0.05\linewidth,trim={0 0 0 0}, clip]{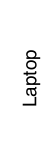} \phantomcaption} \hfill
\subfloat{\includegraphics[width=0.16\linewidth,trim={0 0 0 0}, clip]{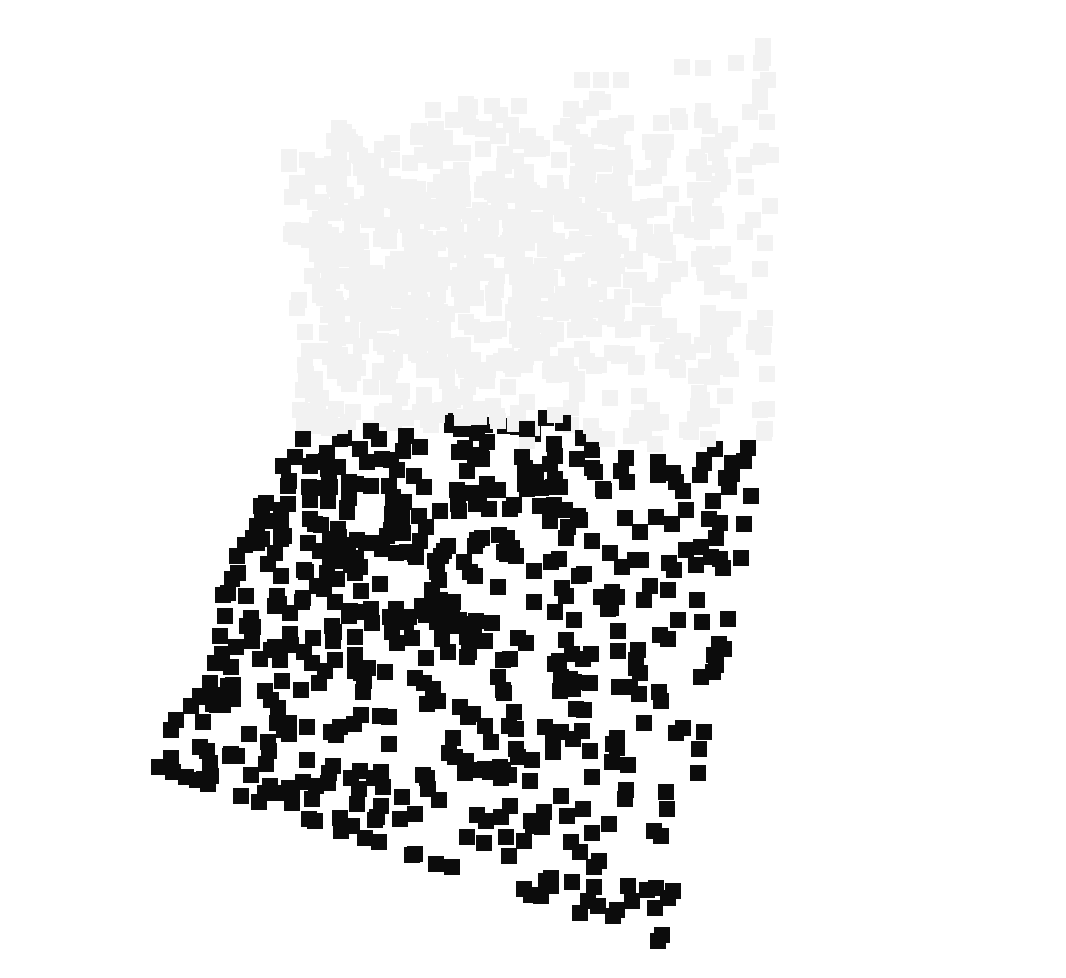} \phantomcaption} \hfill
\subfloat{\includegraphics[width=0.16\linewidth,trim={0 0 0 0}, clip]{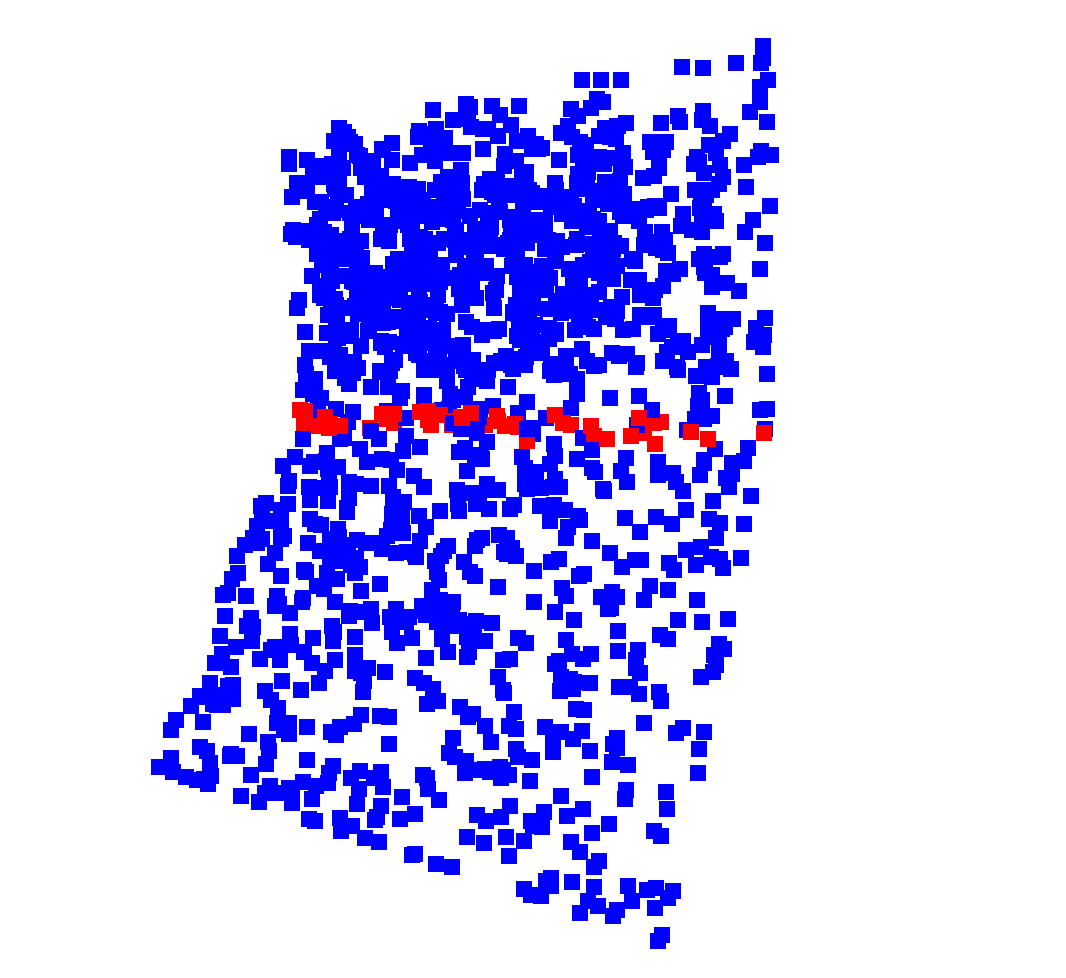} \phantomcaption} \hfill
\subfloat{\includegraphics[width=0.16\linewidth,trim={0 0 0 0}, clip]{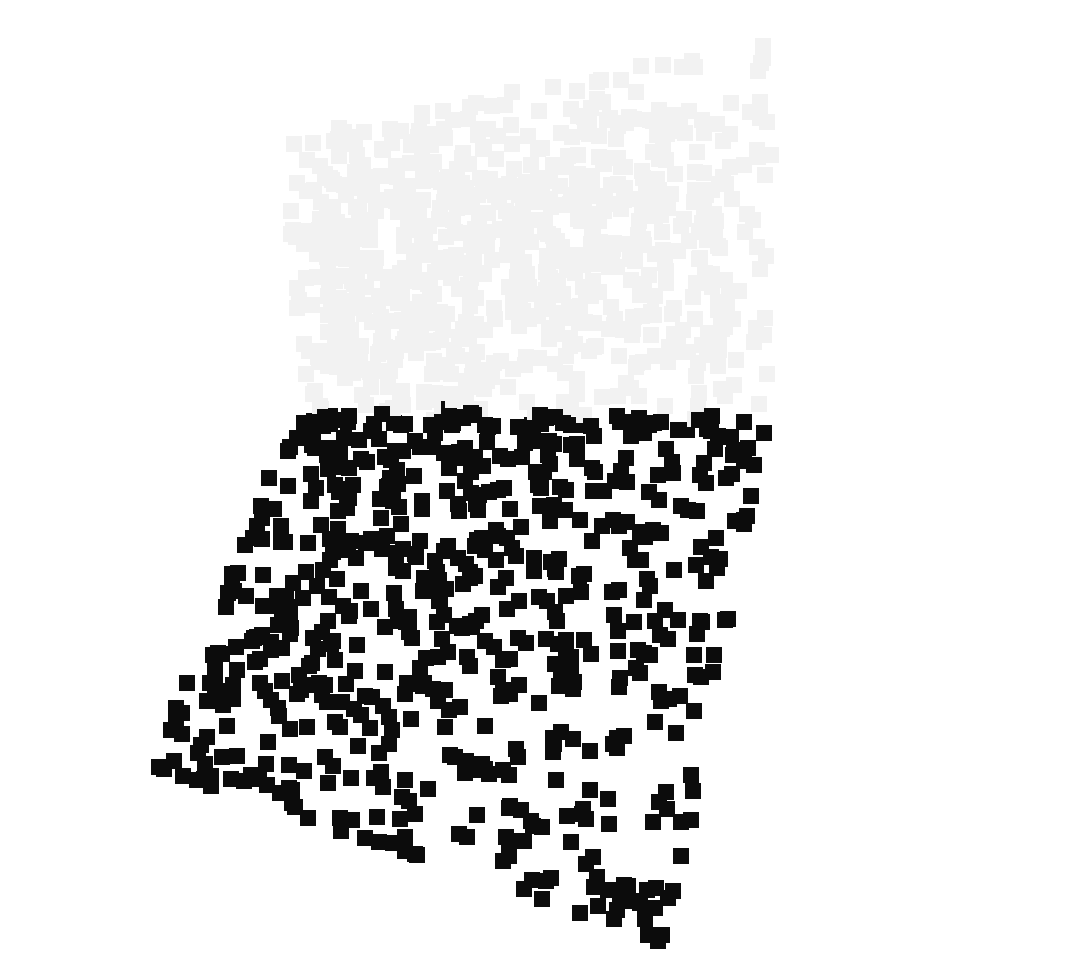} \phantomcaption} \hfill
\subfloat{\includegraphics[width=0.16\linewidth,trim={0 0 0 0}, clip]{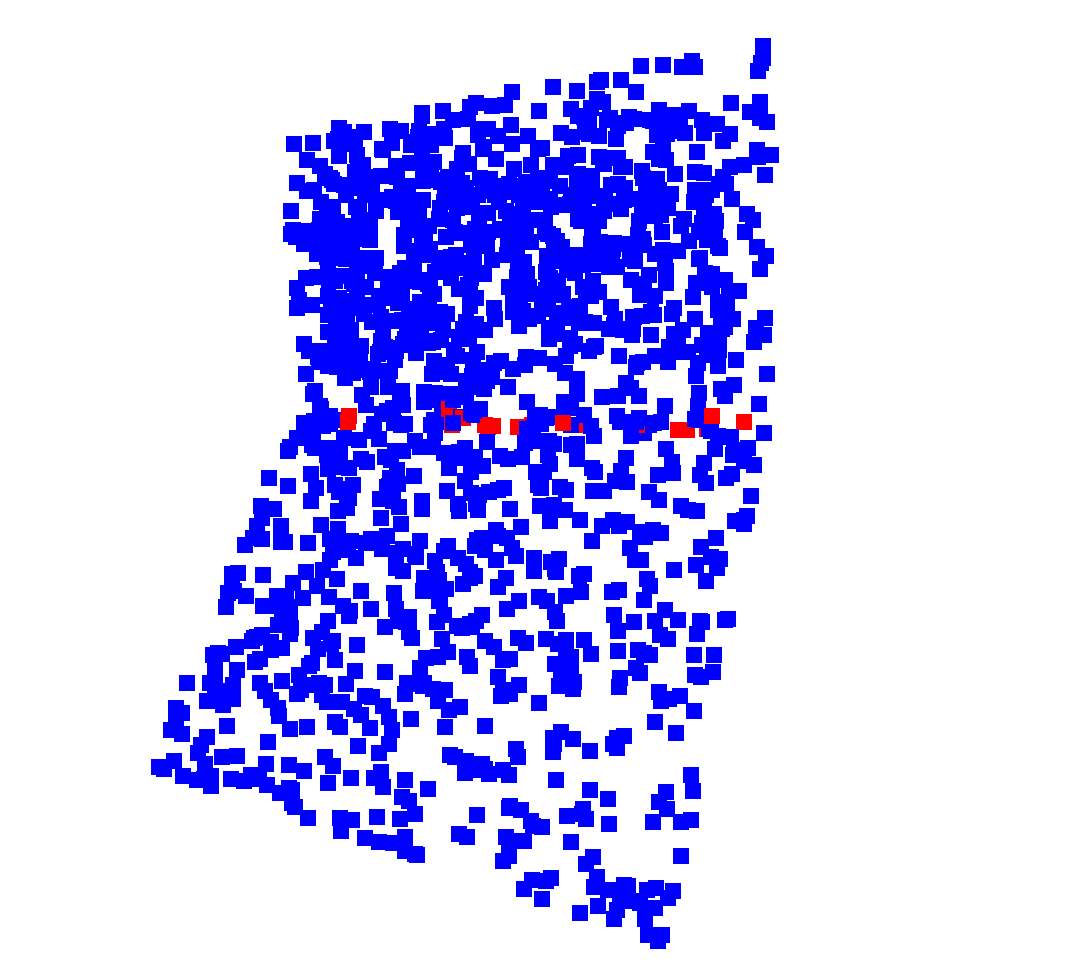} \phantomcaption} \hfill
\subfloat{\includegraphics[width=0.16\linewidth,trim={0 0 0 0}, clip]{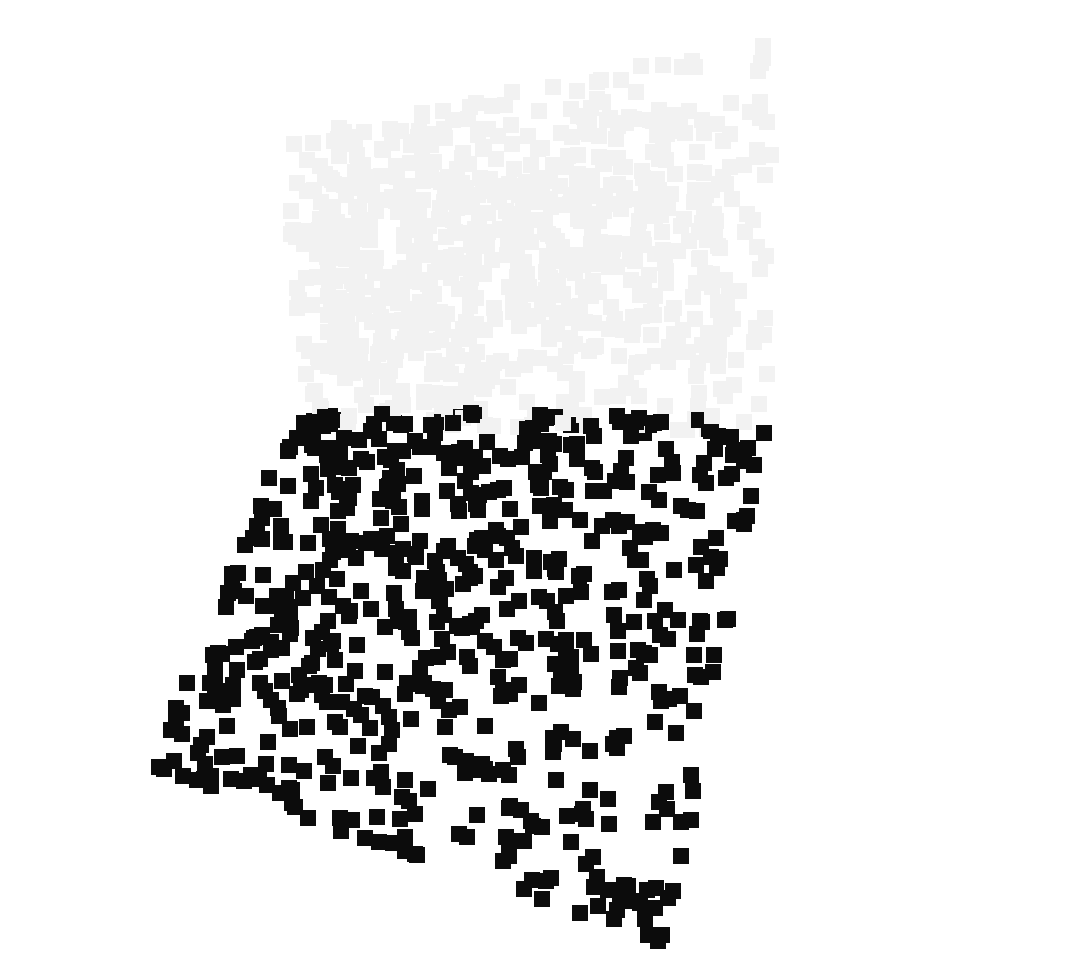} \phantomcaption}\vspace{-3.5mm}

\subfloat{\includegraphics[width=0.05\linewidth,trim={0 0 0 0}, clip]{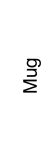} \phantomcaption} \hfill
\subfloat{\includegraphics[width=0.16\linewidth,trim={0 0 0 0}, clip]{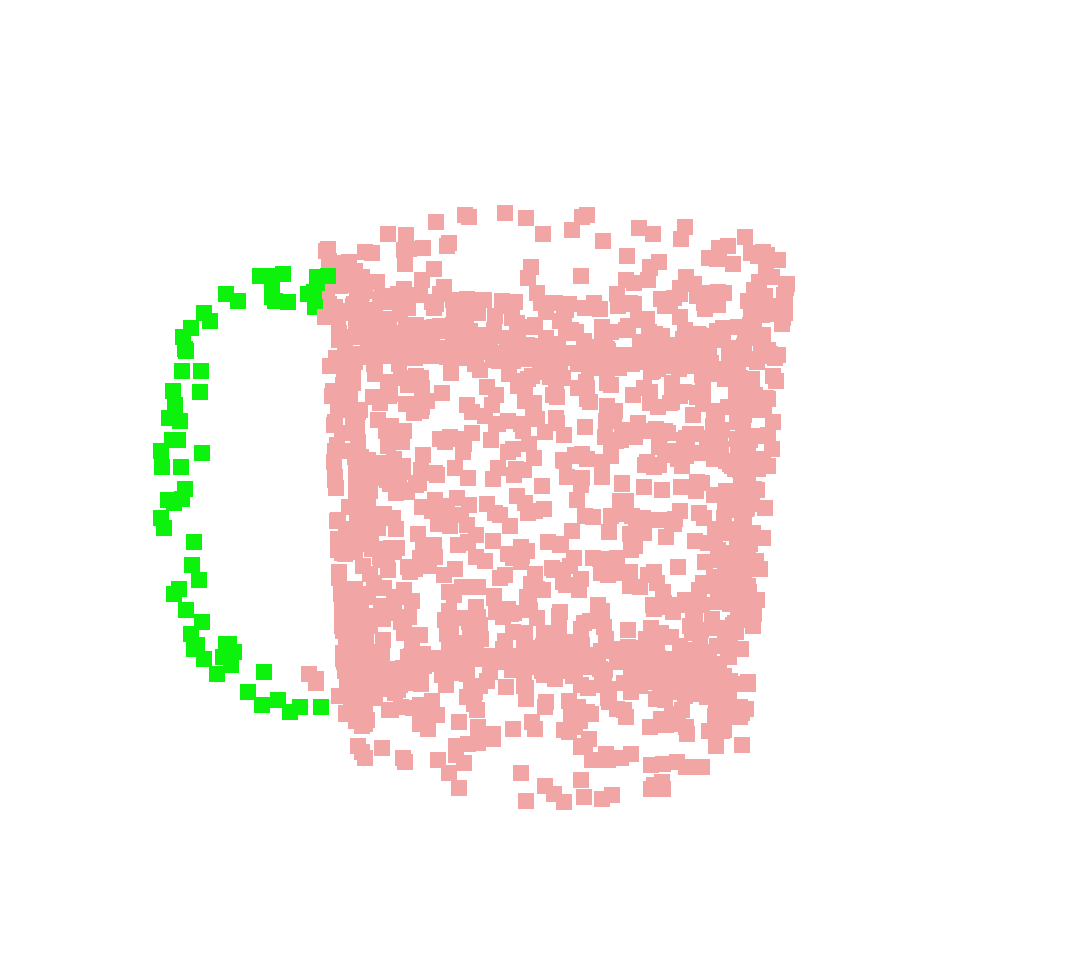} \phantomcaption} \hfill
\subfloat{\includegraphics[width=0.16\linewidth,trim={0 0 0 0}, clip]{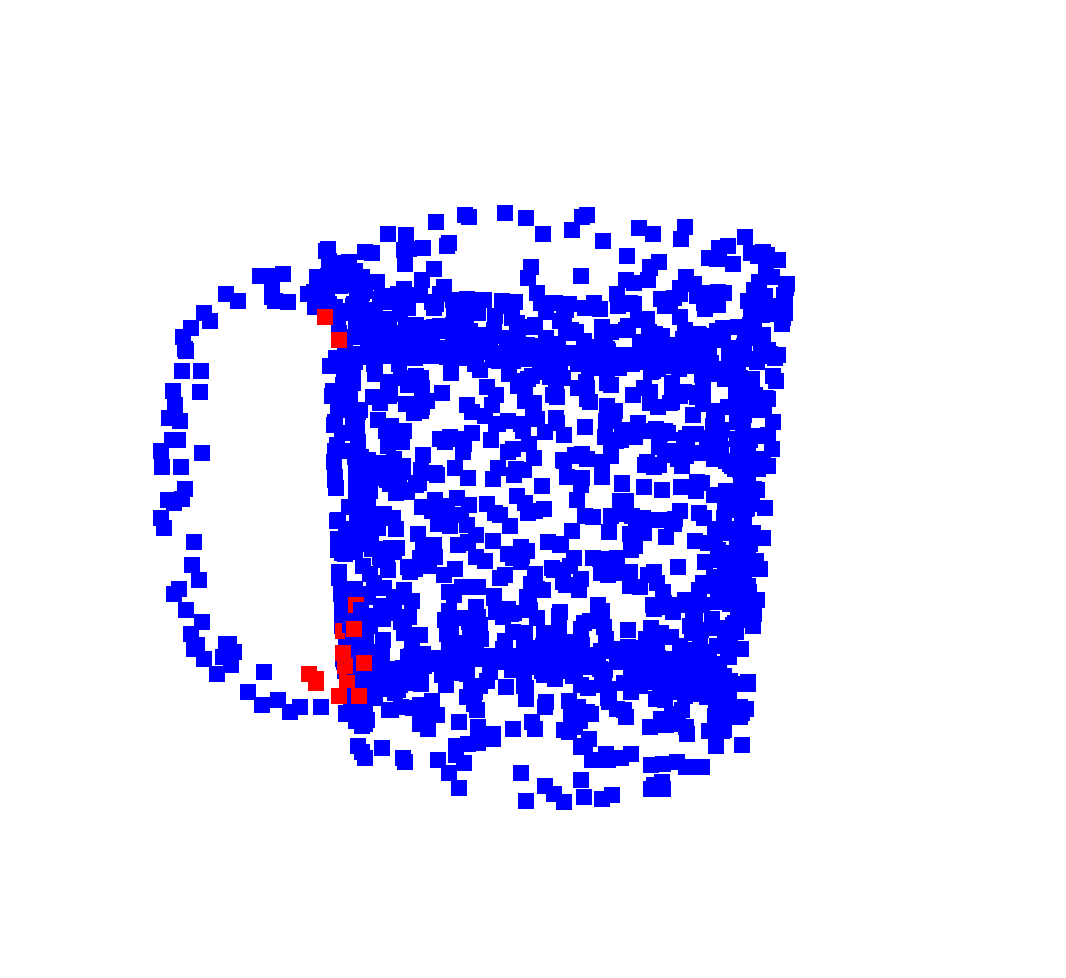} \phantomcaption} \hfill
\subfloat{\includegraphics[width=0.16\linewidth,trim={0 0 0 0}, clip]{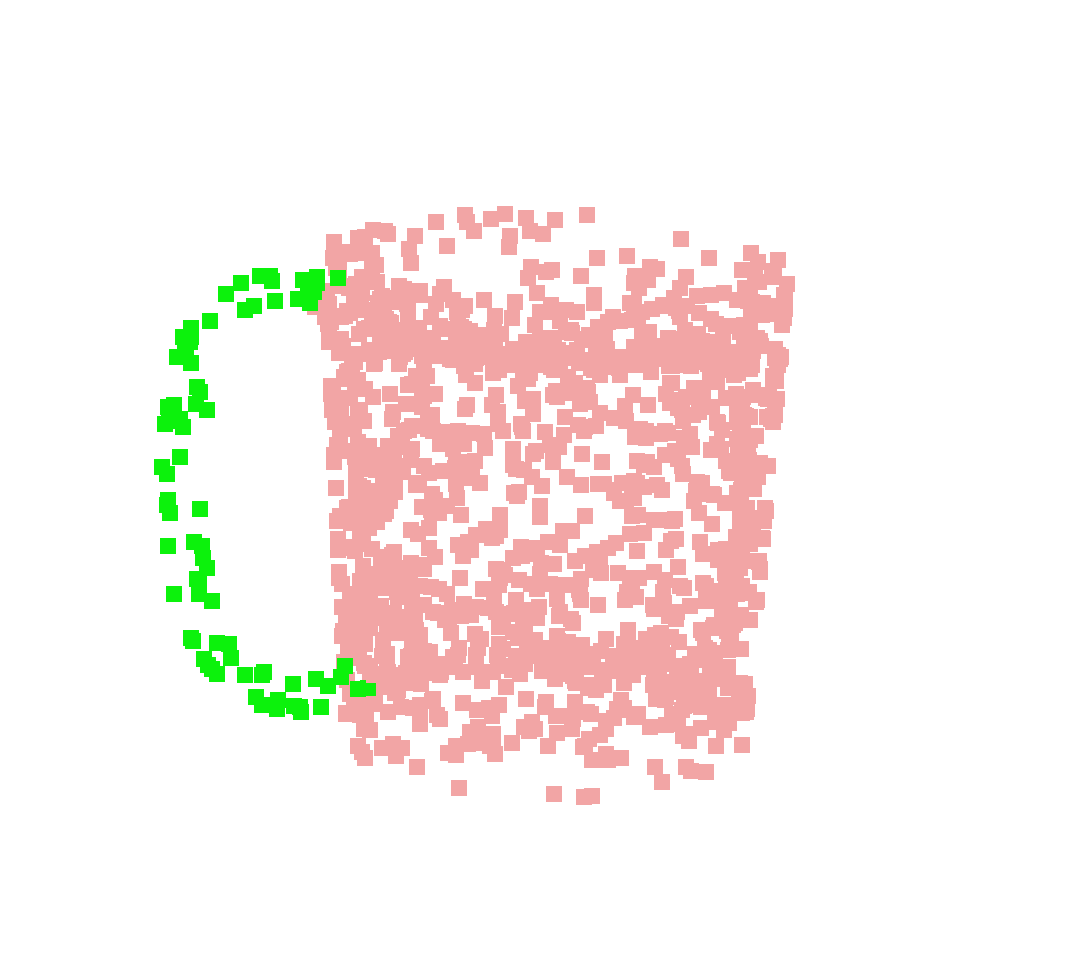} \phantomcaption} \hfill
\subfloat{\includegraphics[width=0.16\linewidth,trim={0 0 0 0}, clip]{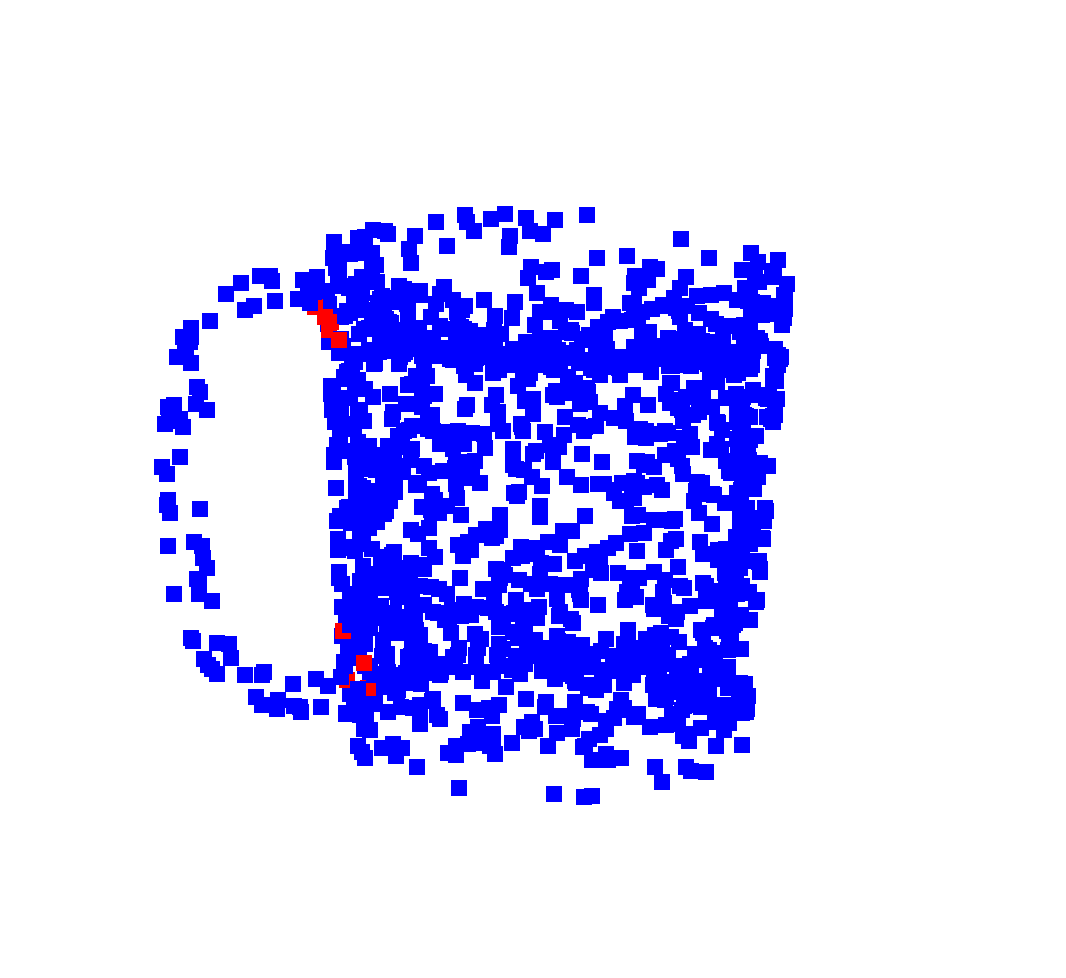} \phantomcaption} \hfill
\subfloat{\includegraphics[width=0.16\linewidth,trim={0 0 0 0}, clip]{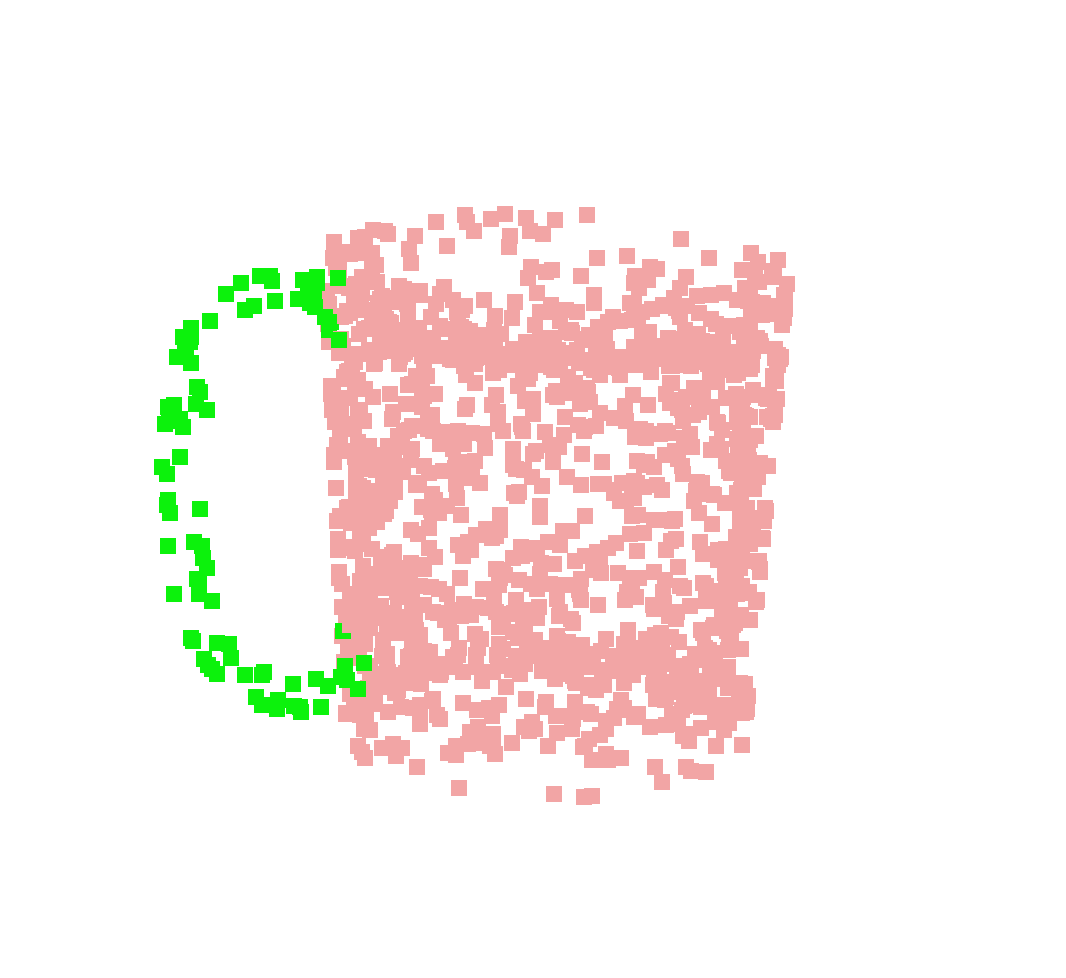} \phantomcaption}\vspace{-3.5mm}

\subfloat{\includegraphics[width=0.05\linewidth,trim={0 0 0 0}, clip]{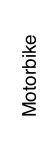} \phantomcaption} \hfill
\subfloat{\includegraphics[width=0.16\linewidth,trim={0 0 0 0}, clip]{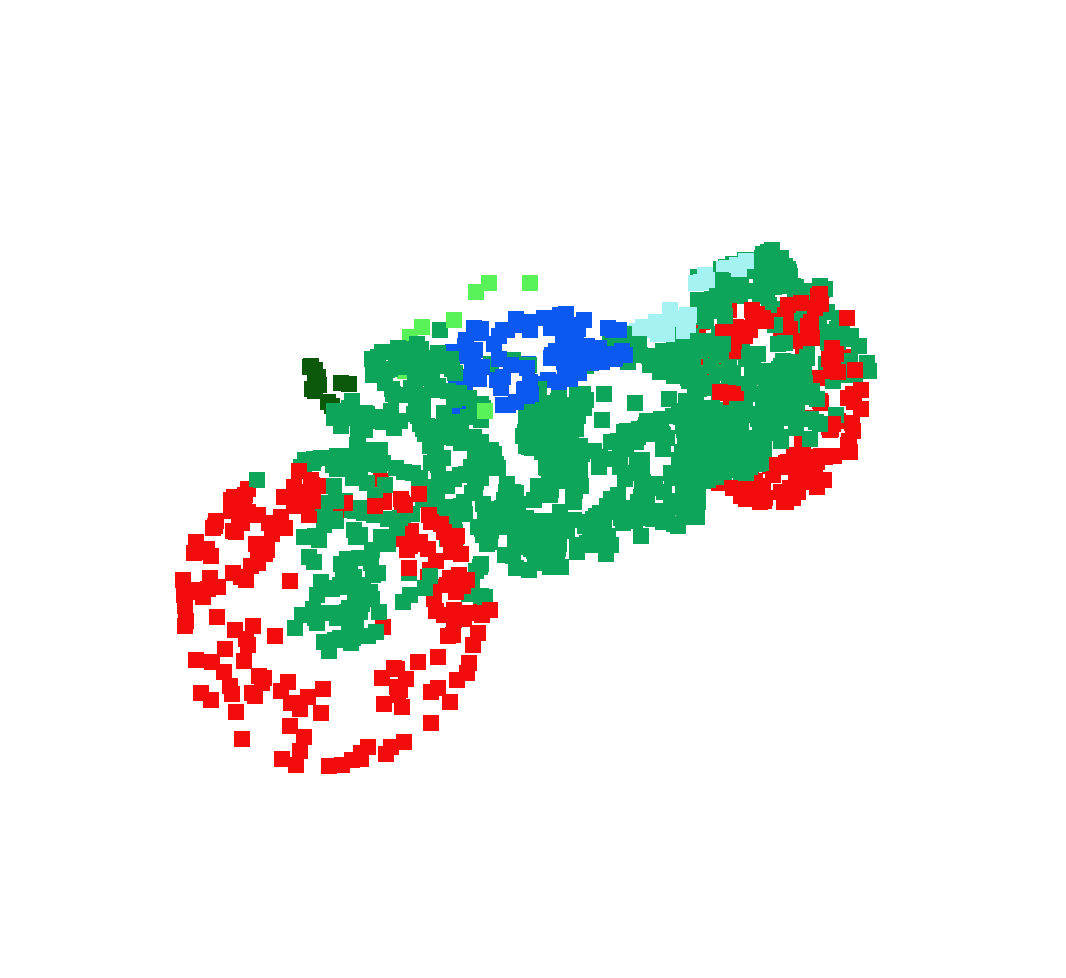} \phantomcaption} \hfill
\subfloat{\includegraphics[width=0.16\linewidth,trim={0 0 0 0}, clip]{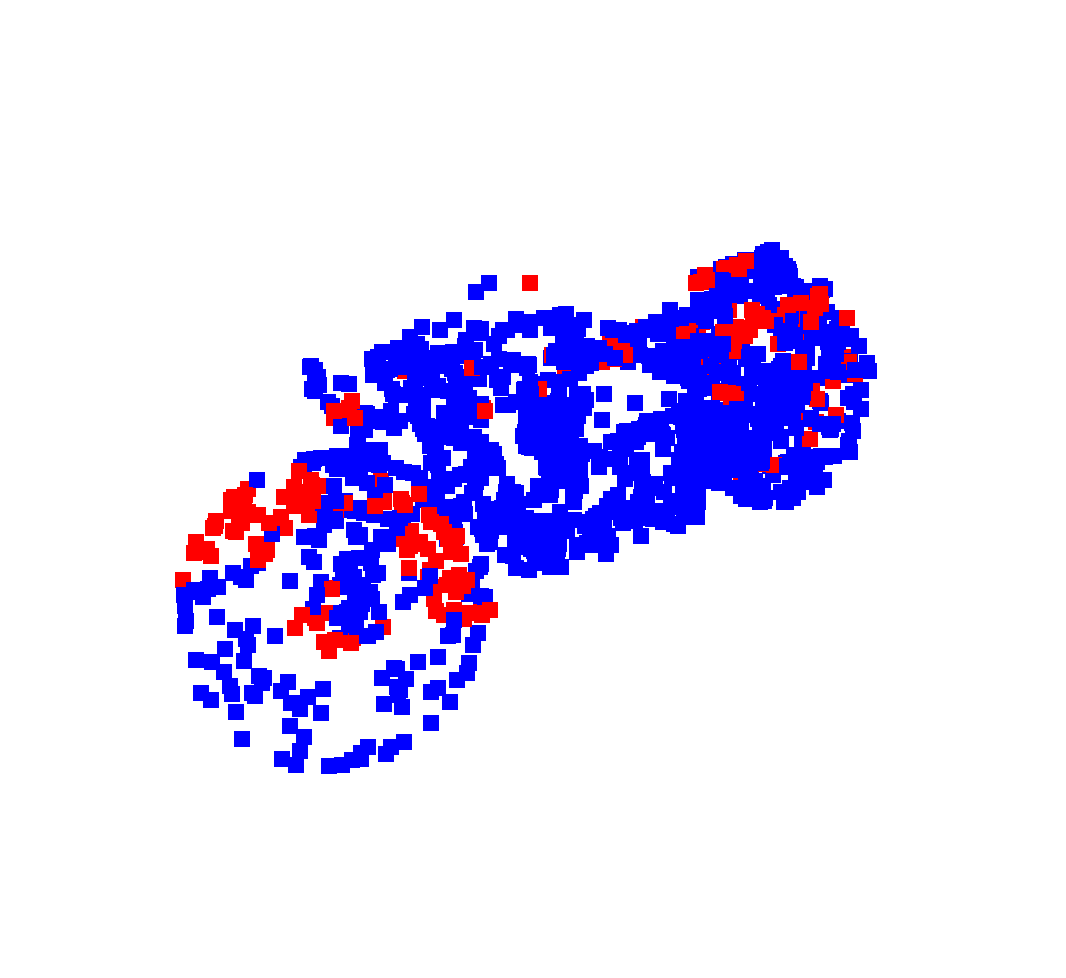} \phantomcaption} \hfill
\subfloat{\includegraphics[width=0.16\linewidth,trim={0 0 0 0}, clip]{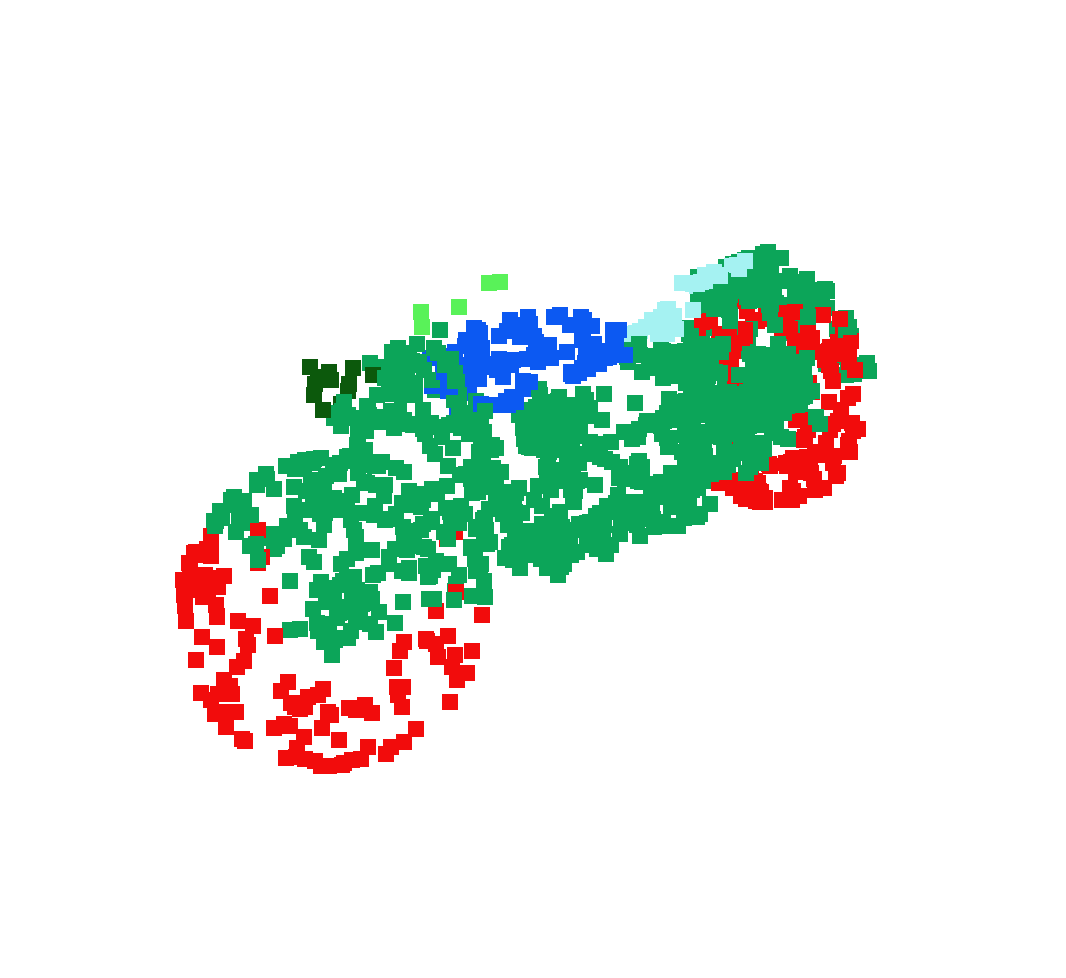} \phantomcaption} \hfill
\subfloat{\includegraphics[width=0.16\linewidth,trim={0 0 0 0}, clip]{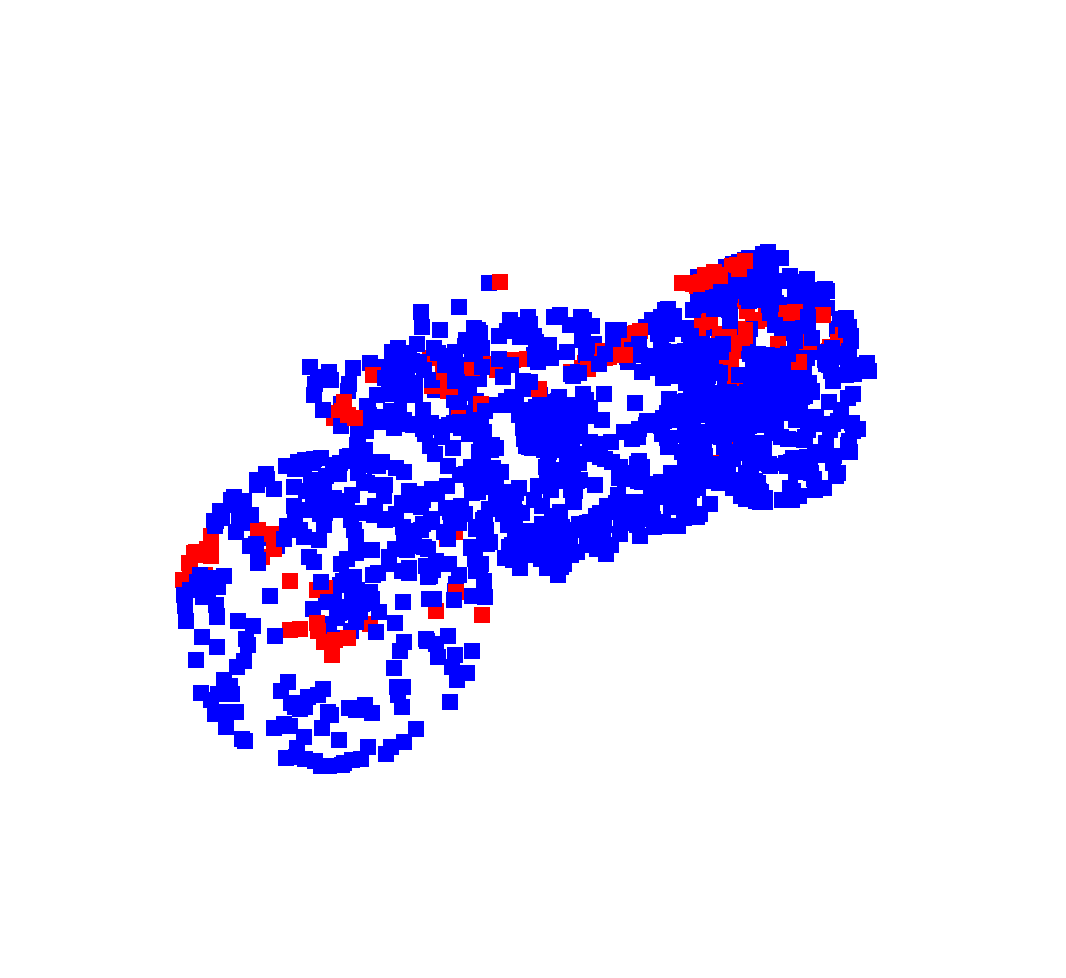} \phantomcaption} \hfill
\subfloat{\includegraphics[width=0.16\linewidth,trim={0 0 0 0}, clip]{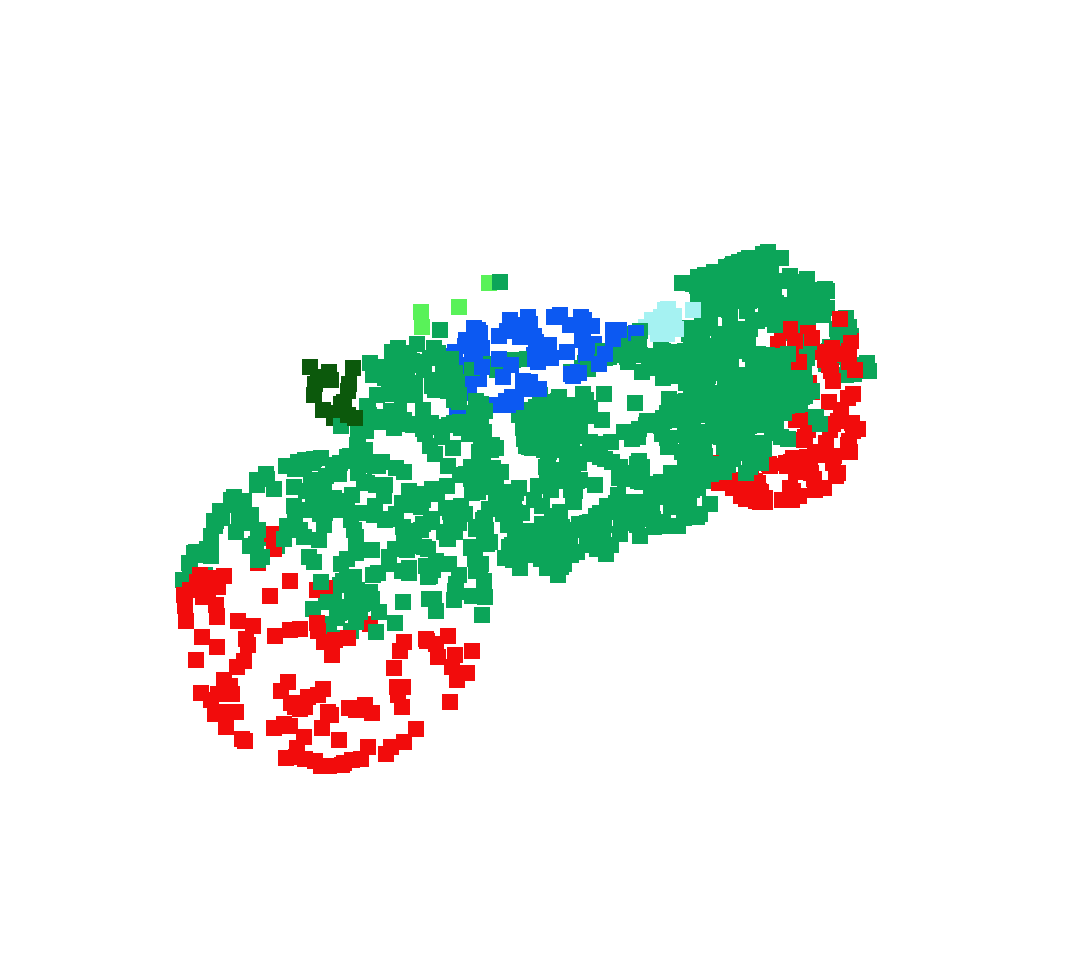} \phantomcaption}\vspace{-3.5mm}

\subfloat{\includegraphics[width=0.05\linewidth,trim={0 0 0 0}, clip]{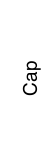} \phantomcaption} \hfill
\subfloat[\textbf{PointNet++~\cite{qi2017pointnet++}}]{\includegraphics[width=0.16\linewidth, trim={0 0 0 0}]{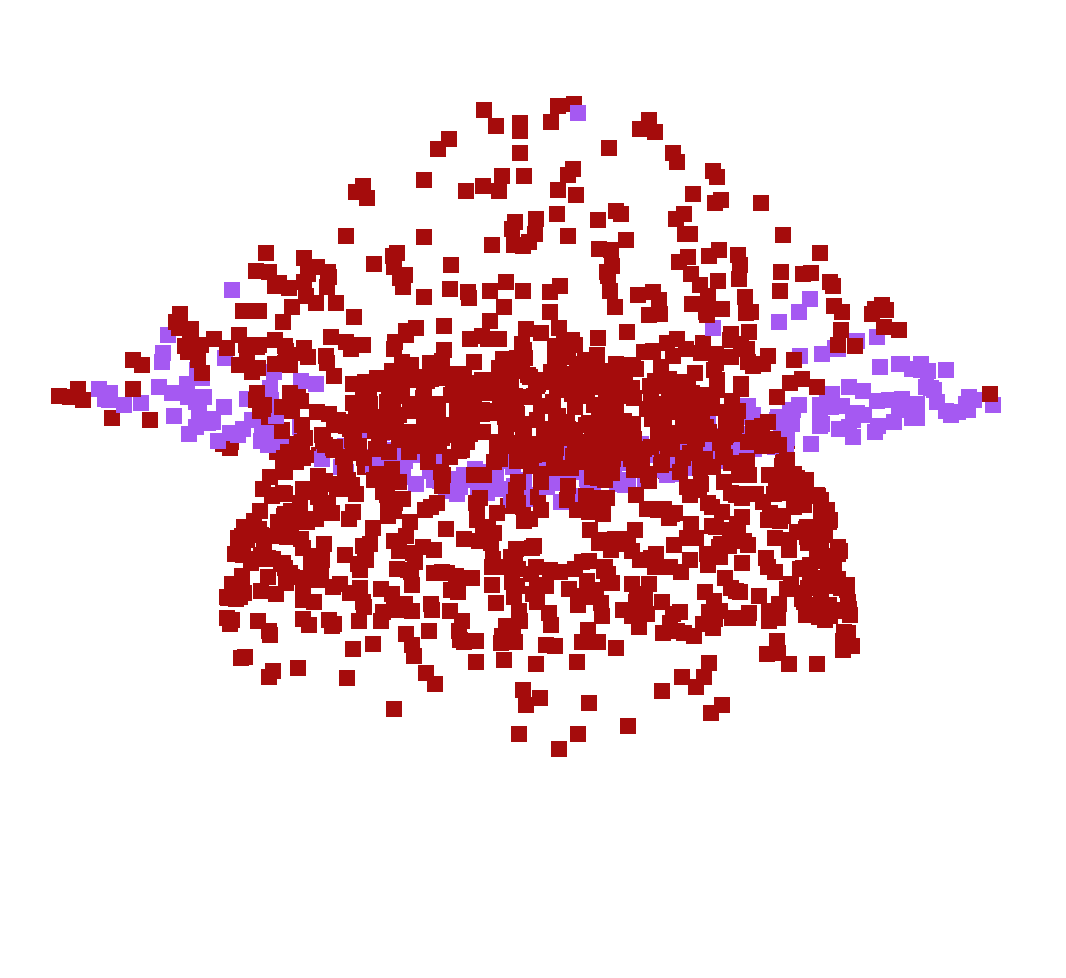}} \hfill
\subfloat[\textbf{PointNet++ (diff)}]{\includegraphics[width=0.16\linewidth, trim={0 0 0 0}]{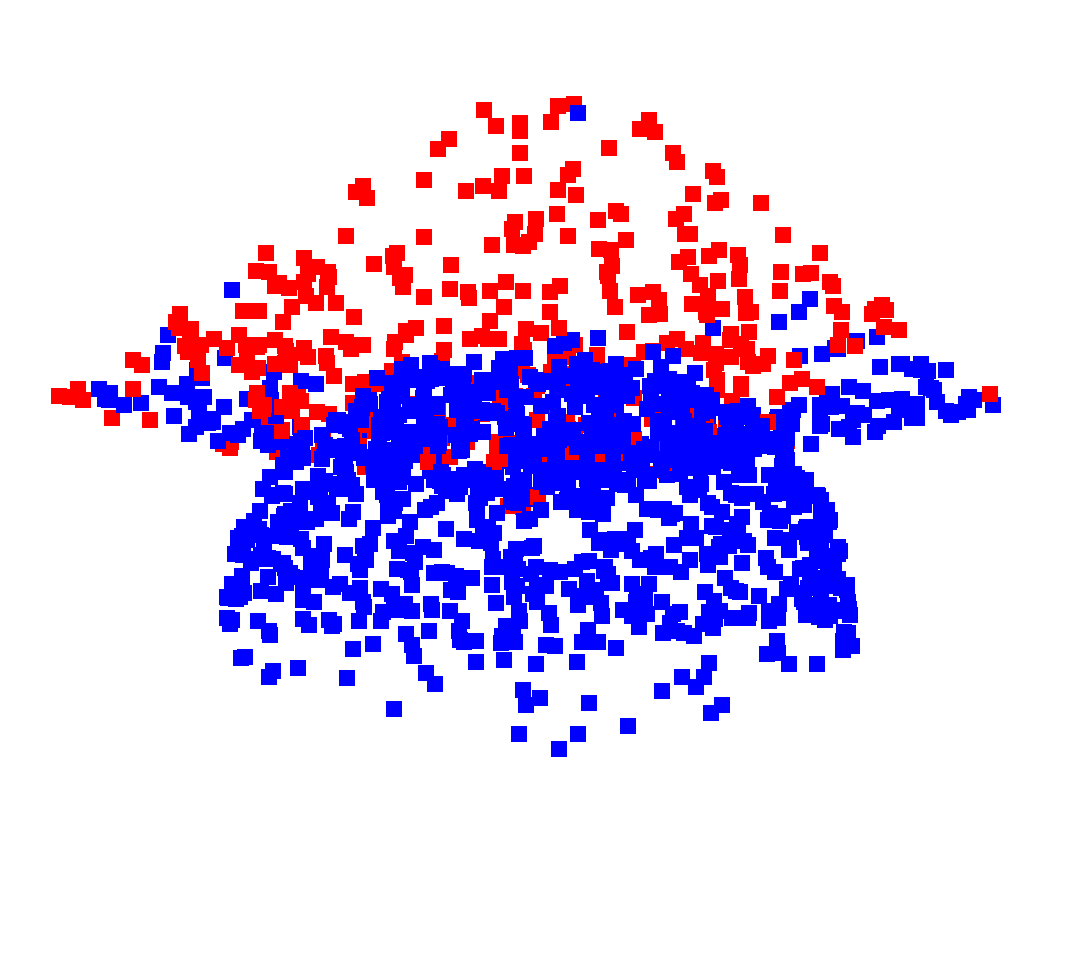}} \hfill
\subfloat[\textbf{Our}]{\includegraphics[width=0.16\linewidth, trim={0 0 0 0}]{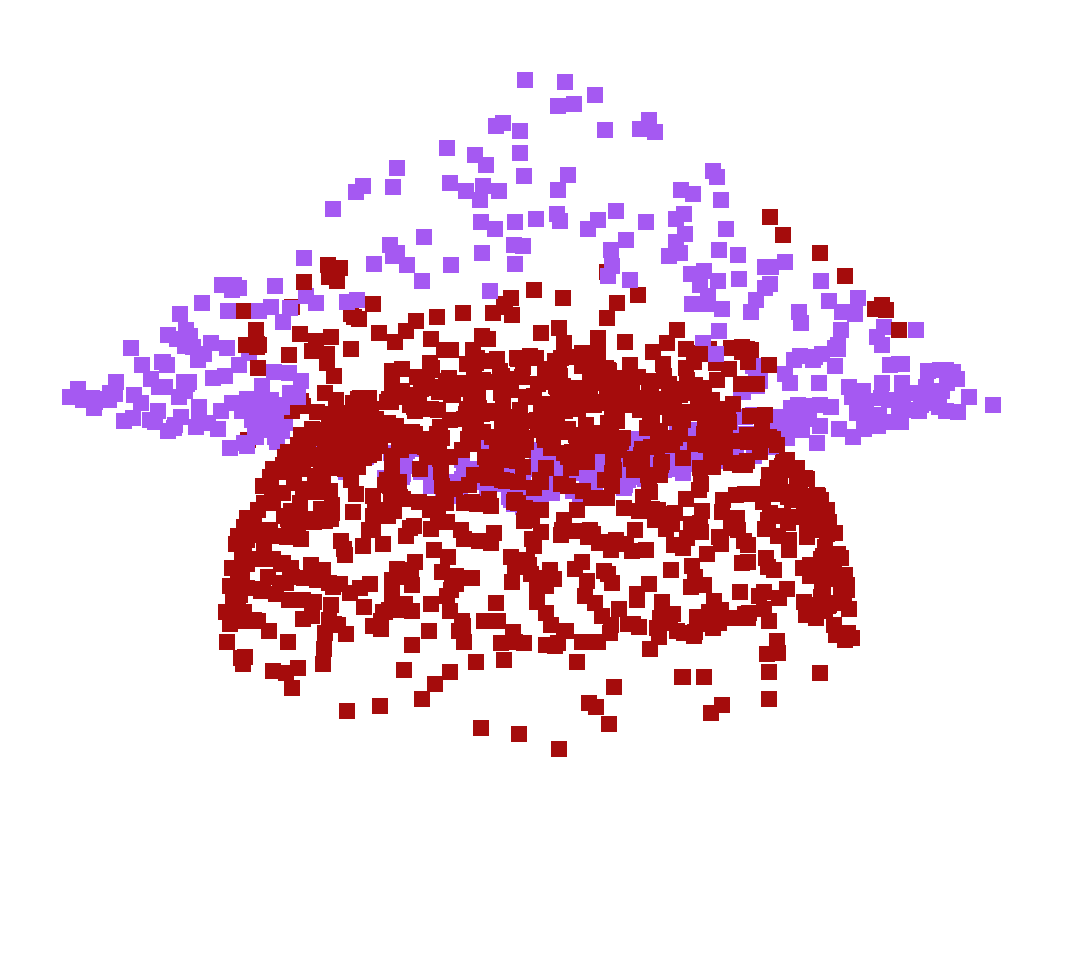}} \hfill
\subfloat[\textbf{Our (diff)}]{\includegraphics[width=0.16\linewidth, trim={0 0 0 0}]{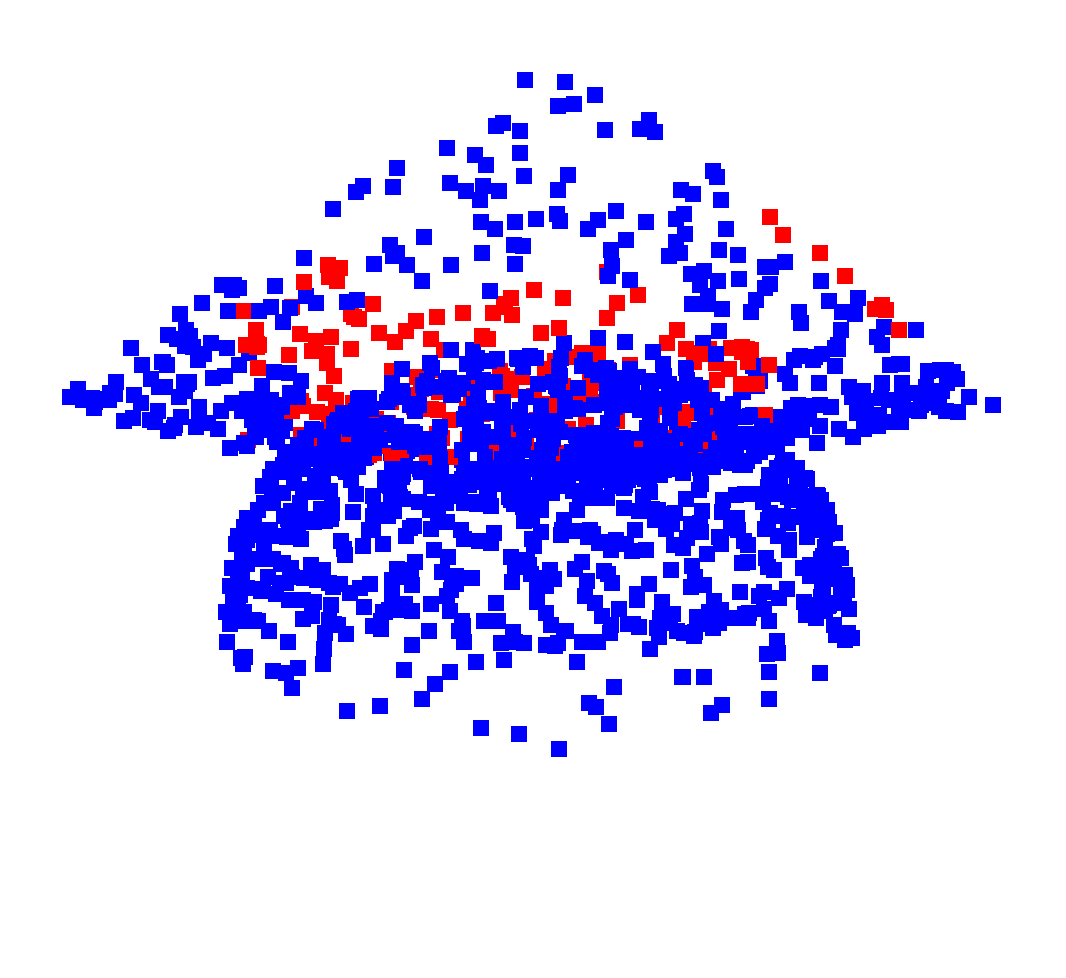}} \hfill
\subfloat[\textbf{Ground Truth}]{\includegraphics[width=0.16\linewidth, trim={0 0 0 0}]{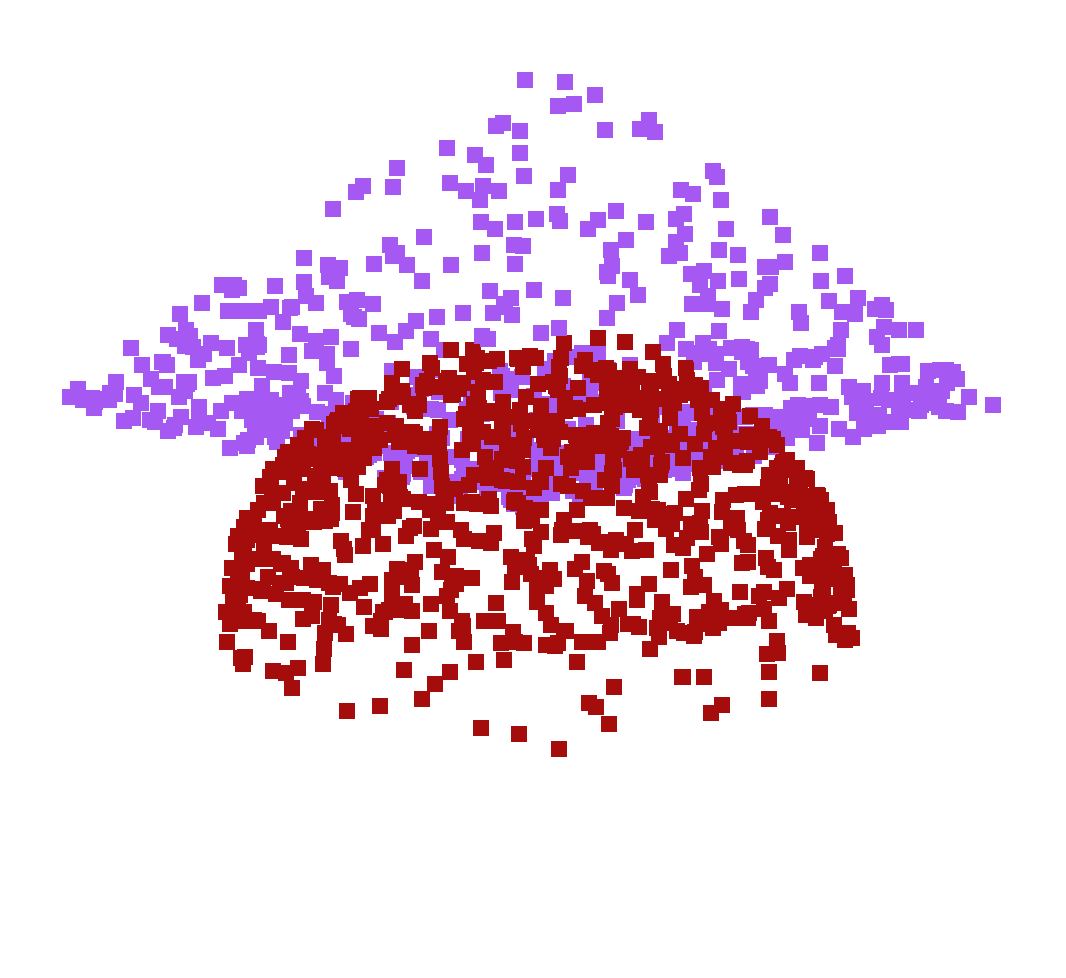}} \hfill
\caption{More segmentation results on \emph{ShapeNet-part} dataset. Second and fourth columns show the differences between ground truth and prediction (red points are mislabeled points) of PointNet++ and our method.}
\vspace{-5.5mm}
\label{fig:suppl_shapenet_part_eval}
\end{figure*}

\setcounter{figure}{-8}
\begin{figure*}[t]
\centering
\subfloat{\includegraphics[width=0.9\linewidth, clip]{pics/s3dis_results/s3dis_label2color.pdf}}
\vspace{-3.5mm}
\subfloat{\includegraphics[width=0.225\linewidth,trim={0 0 0 0}, clip]{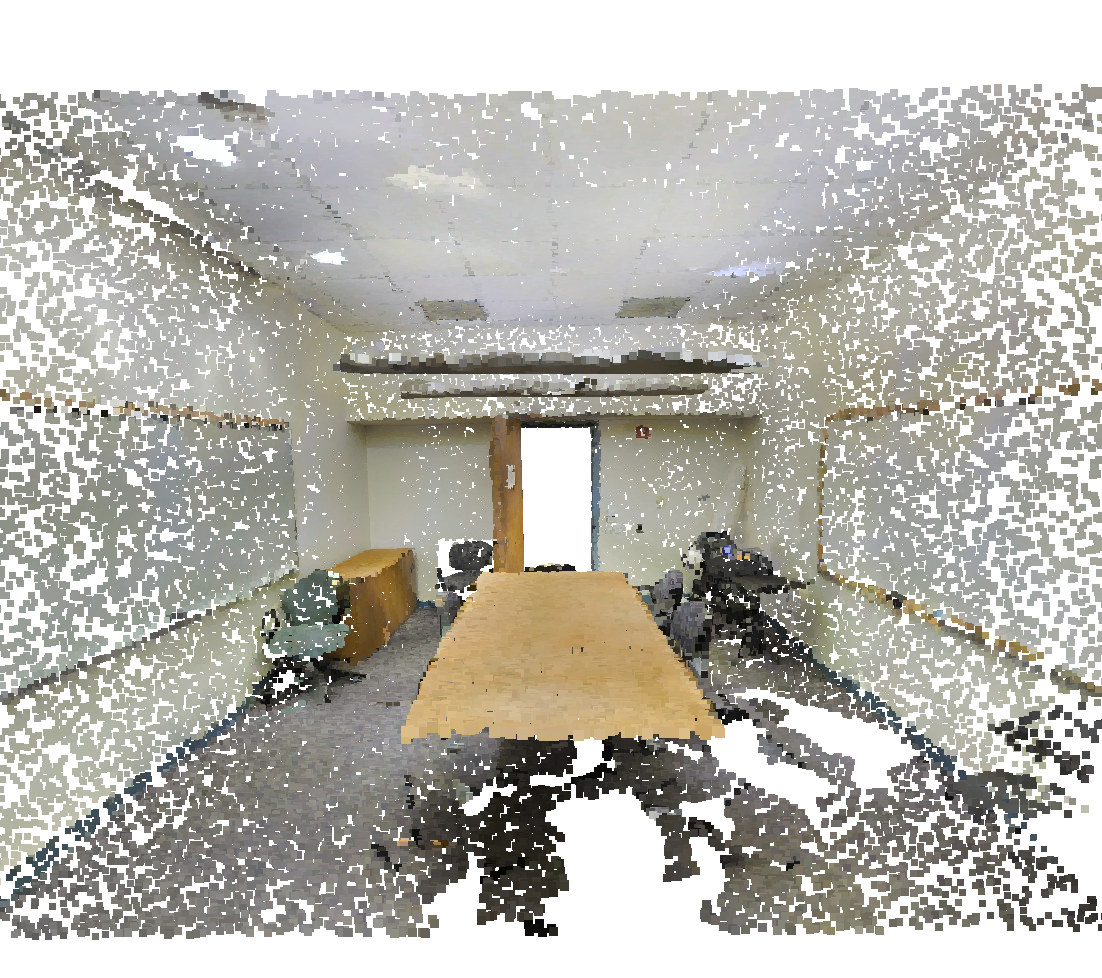} \phantomcaption}
\subfloat{\includegraphics[width=0.225\linewidth,trim={0 0 0 0}, clip]{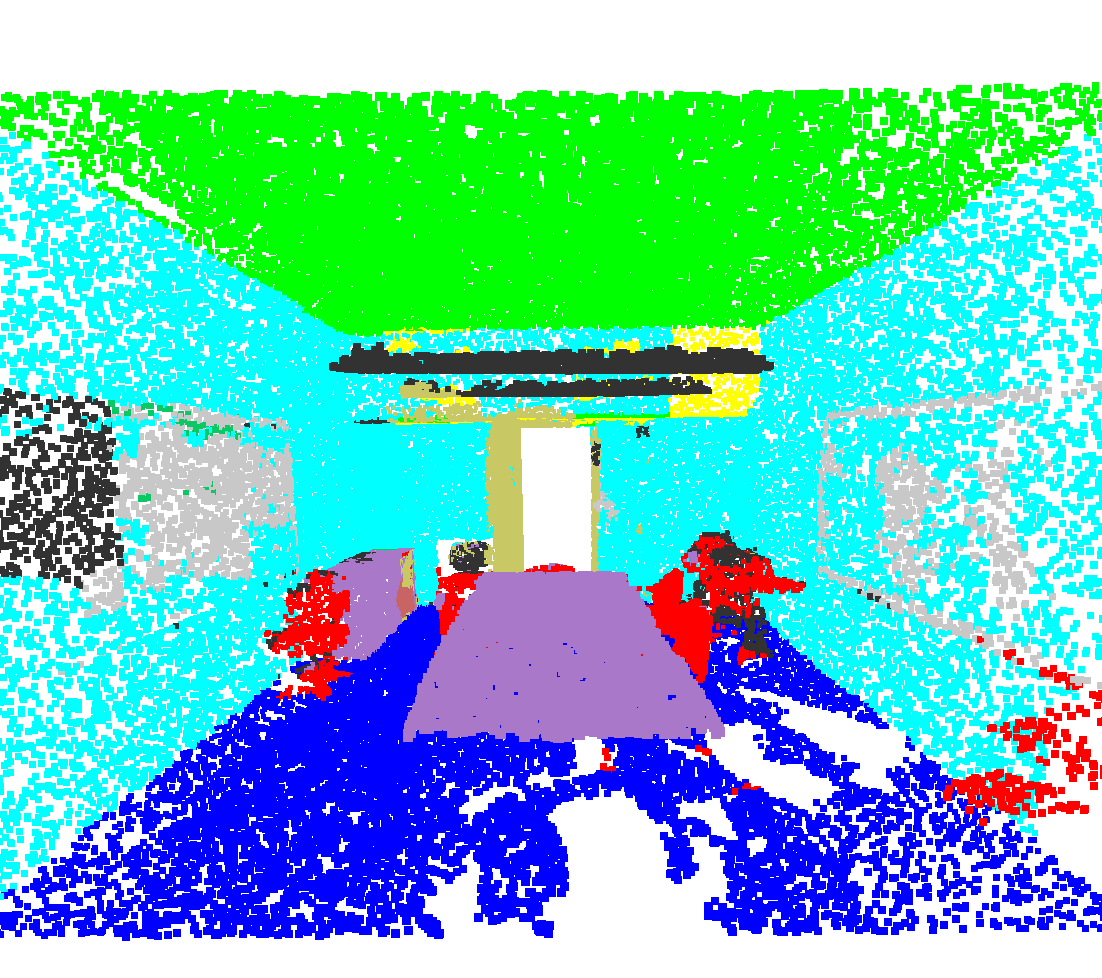} \phantomcaption}
\subfloat{\includegraphics[width=0.225\linewidth,trim={0 0 0 0}, clip]{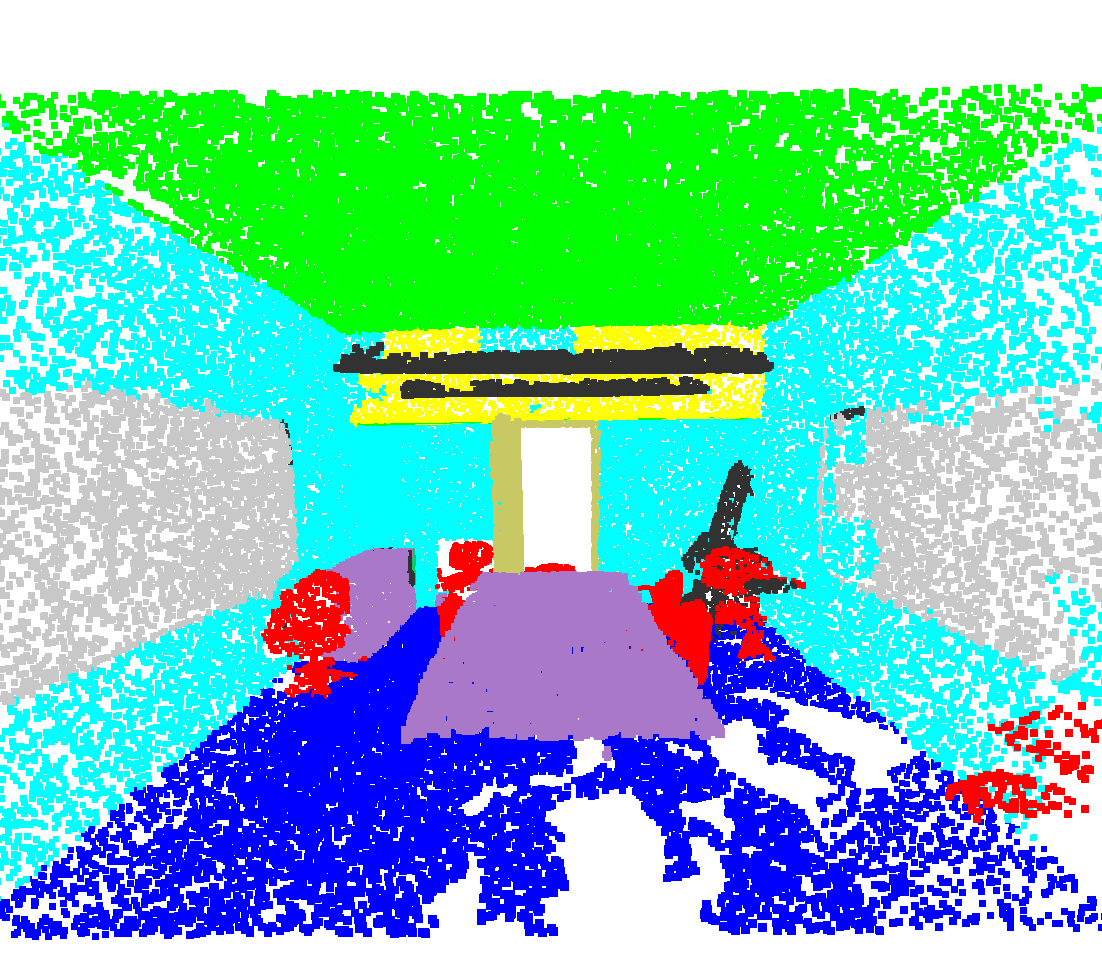} \phantomcaption}
\subfloat{\includegraphics[width=0.225\linewidth,trim={0 0 0 0}, clip]{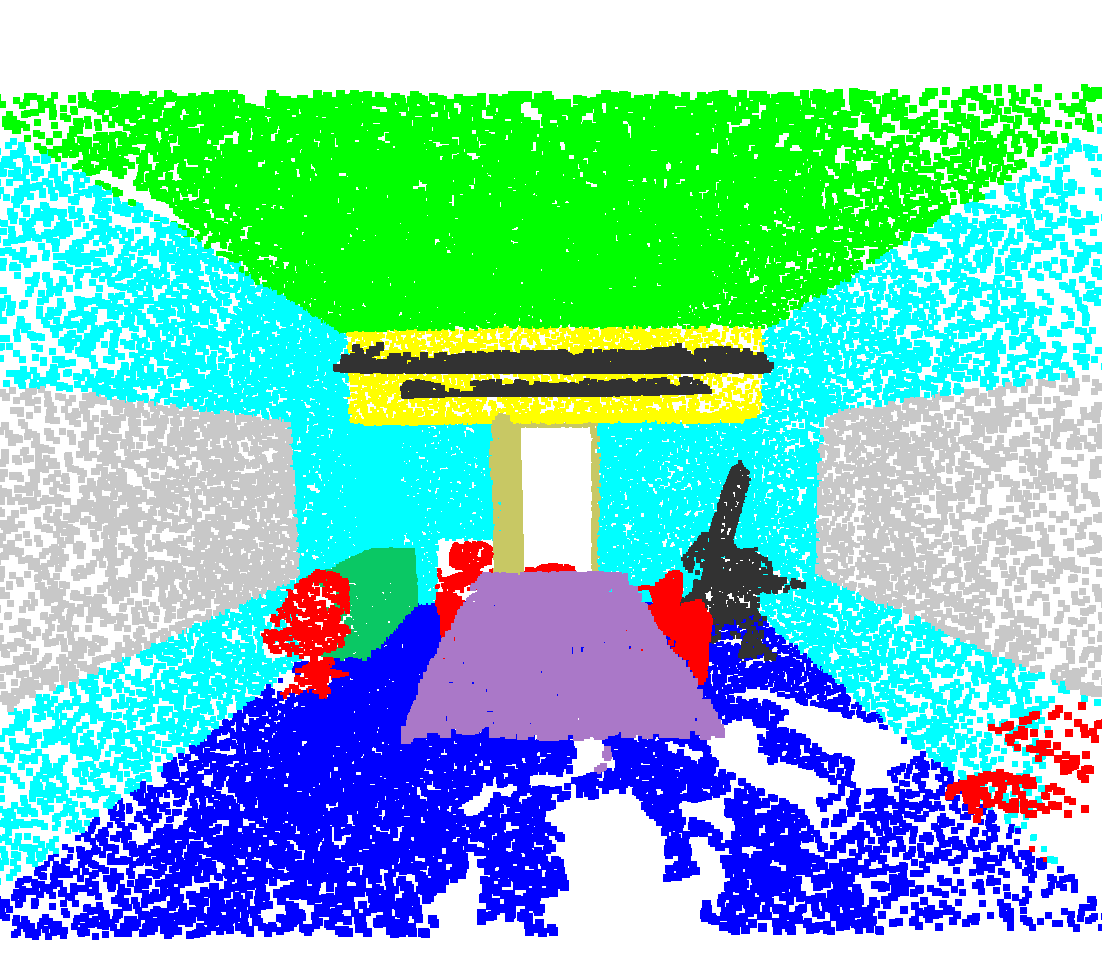} \phantomcaption}
\vspace{-3.5mm}
\subfloat{\includegraphics[width=0.225\linewidth,trim={0 0 0 0}, clip]{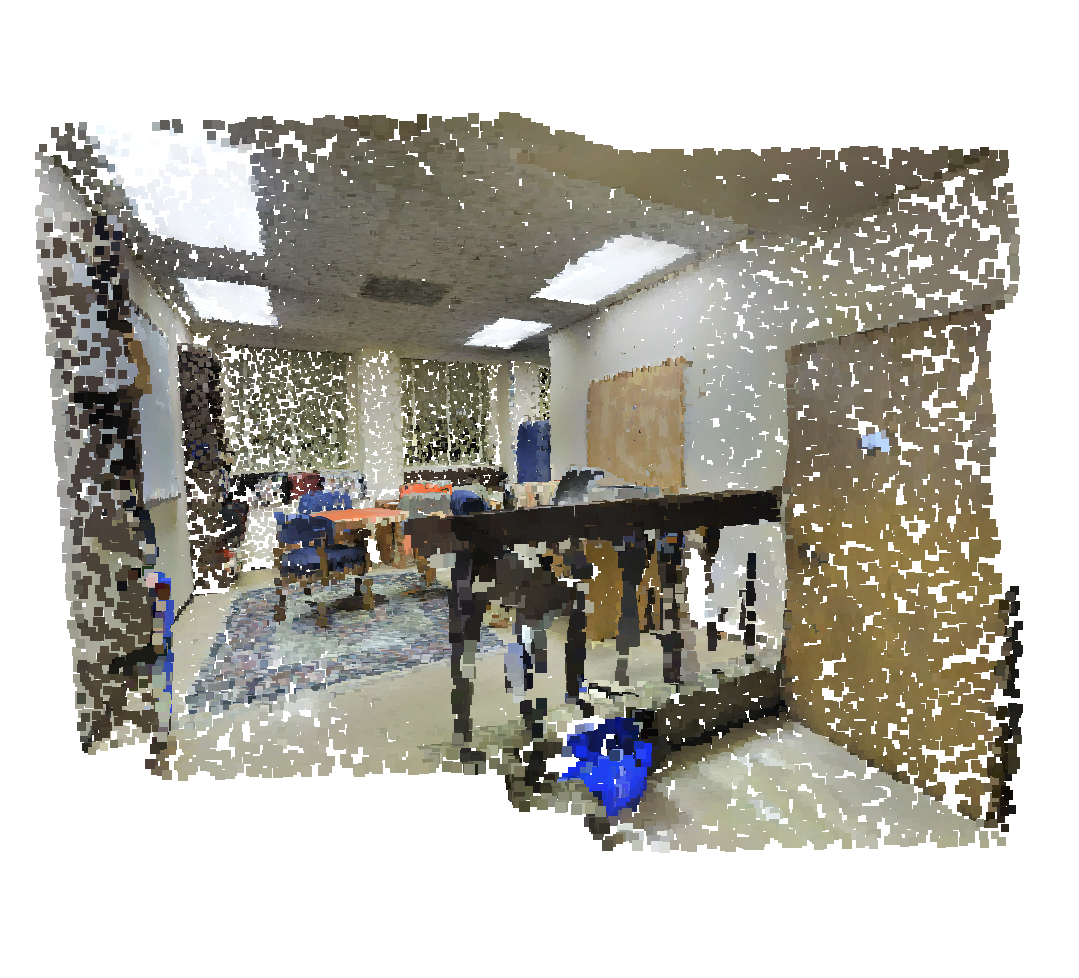} \phantomcaption}
\subfloat{\includegraphics[width=0.225\linewidth,trim={0 0 0 0}, clip]{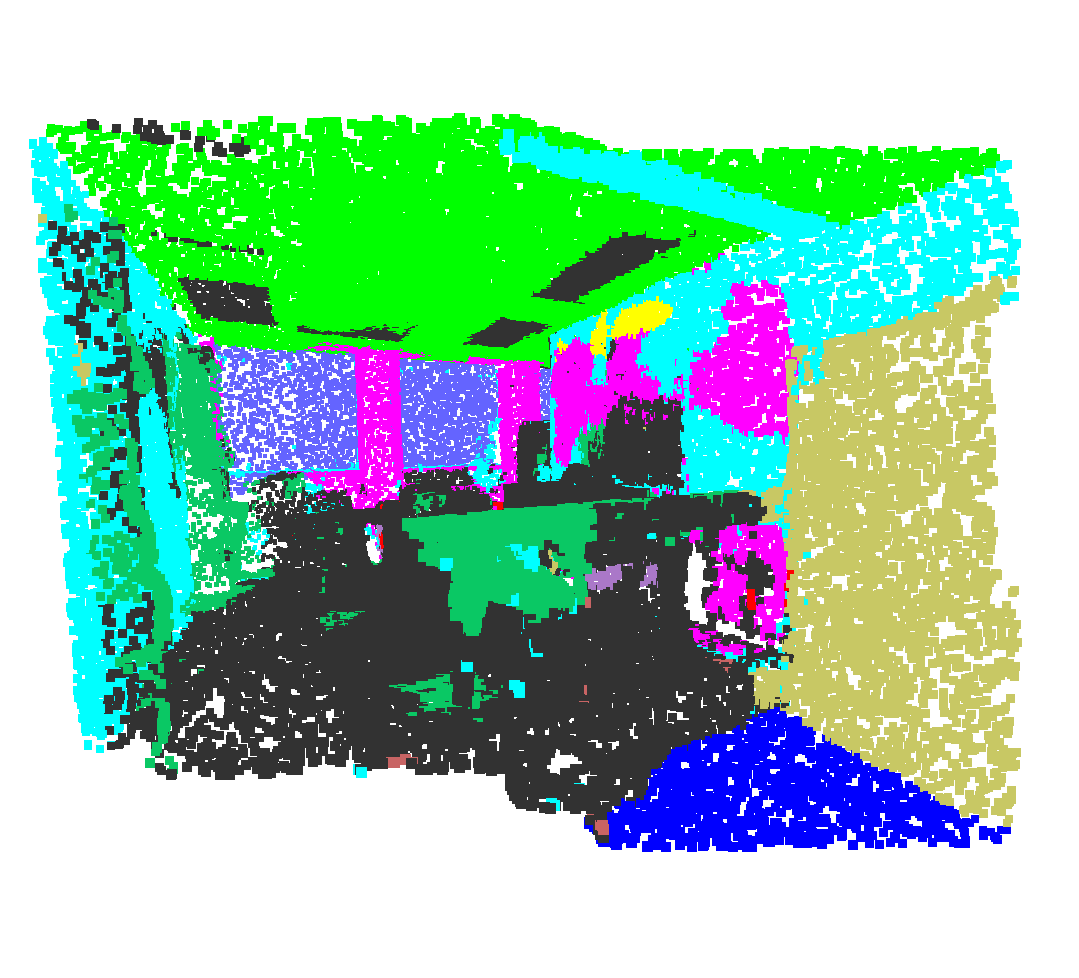} \phantomcaption}
\subfloat{\includegraphics[width=0.225\linewidth,trim={0 0 0 0}, clip]{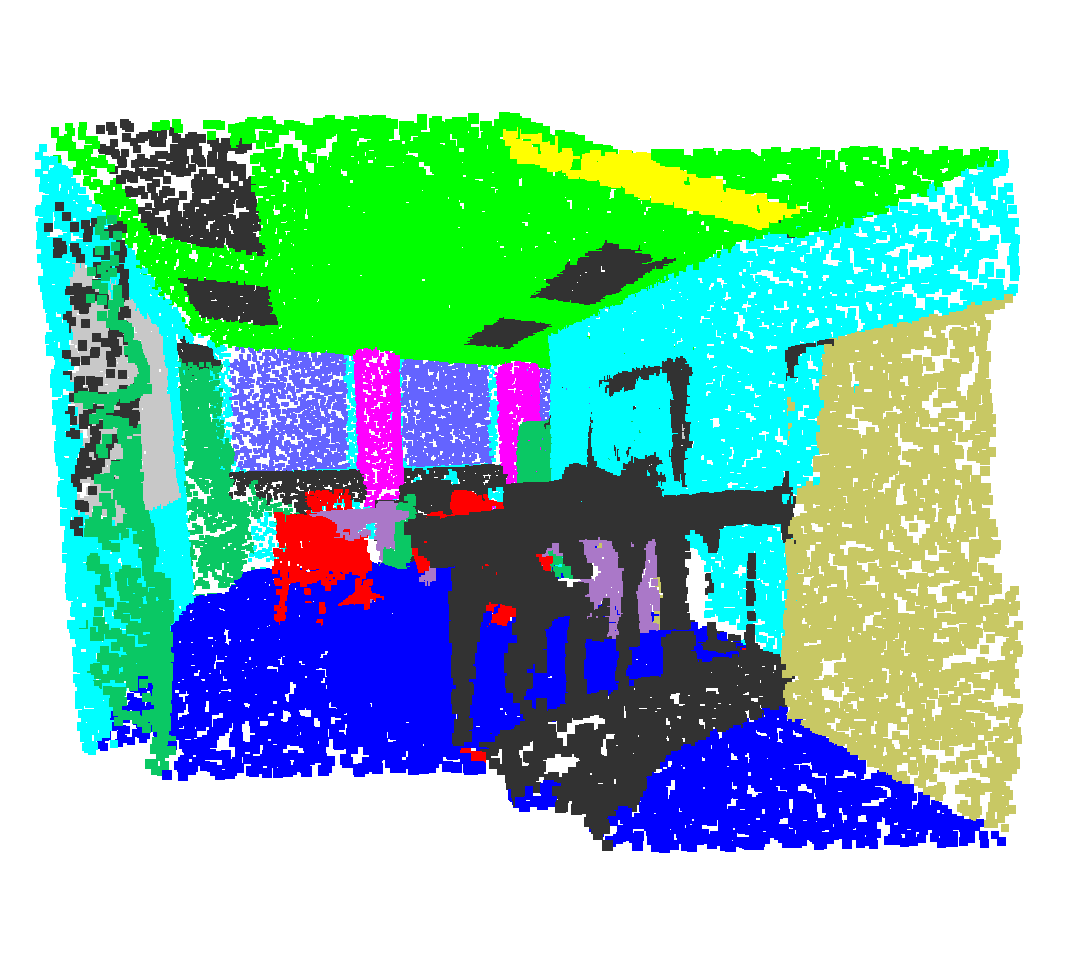} \phantomcaption}
\subfloat{\includegraphics[width=0.225\linewidth,trim={0 0 0 0}, clip]{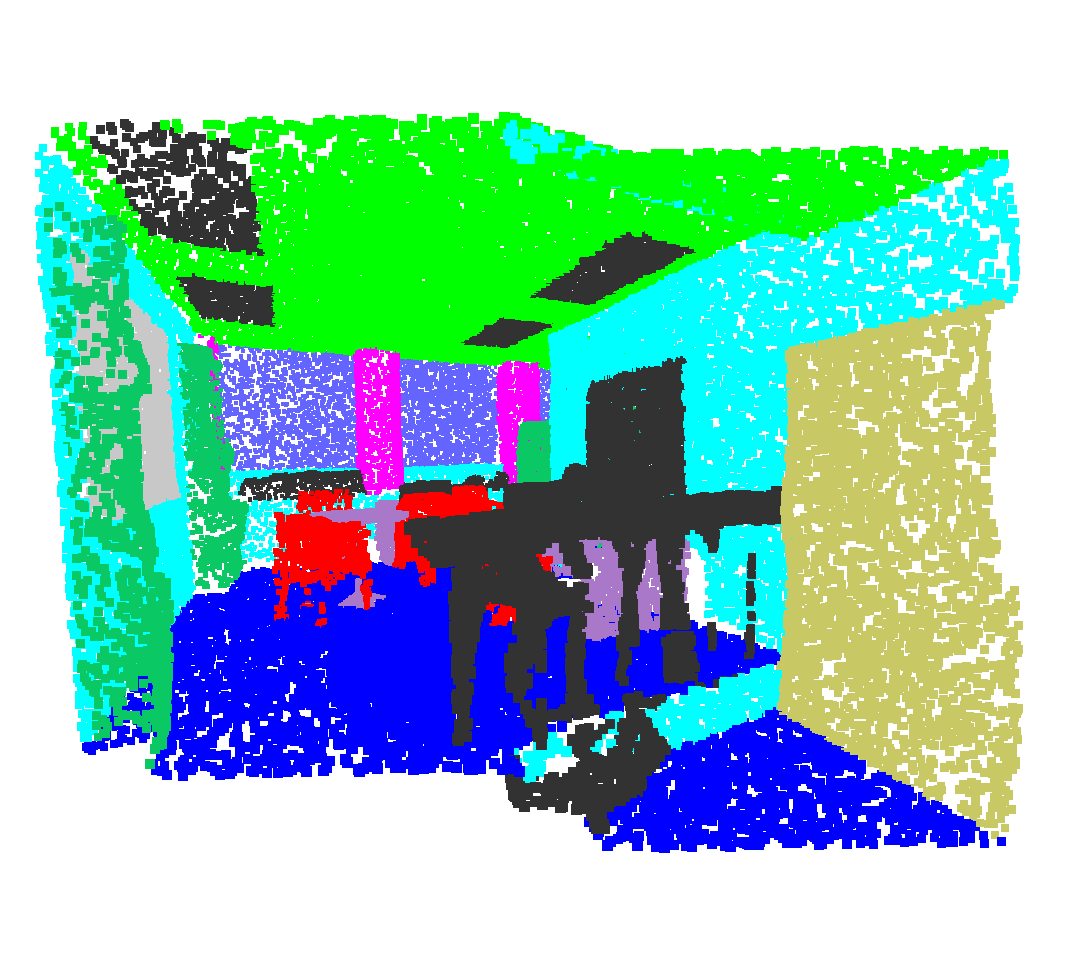} \phantomcaption}
\vspace{-3.5mm}
\subfloat{\includegraphics[width=0.225\linewidth,trim={0 2.0cm 0 4.0cm}, clip]{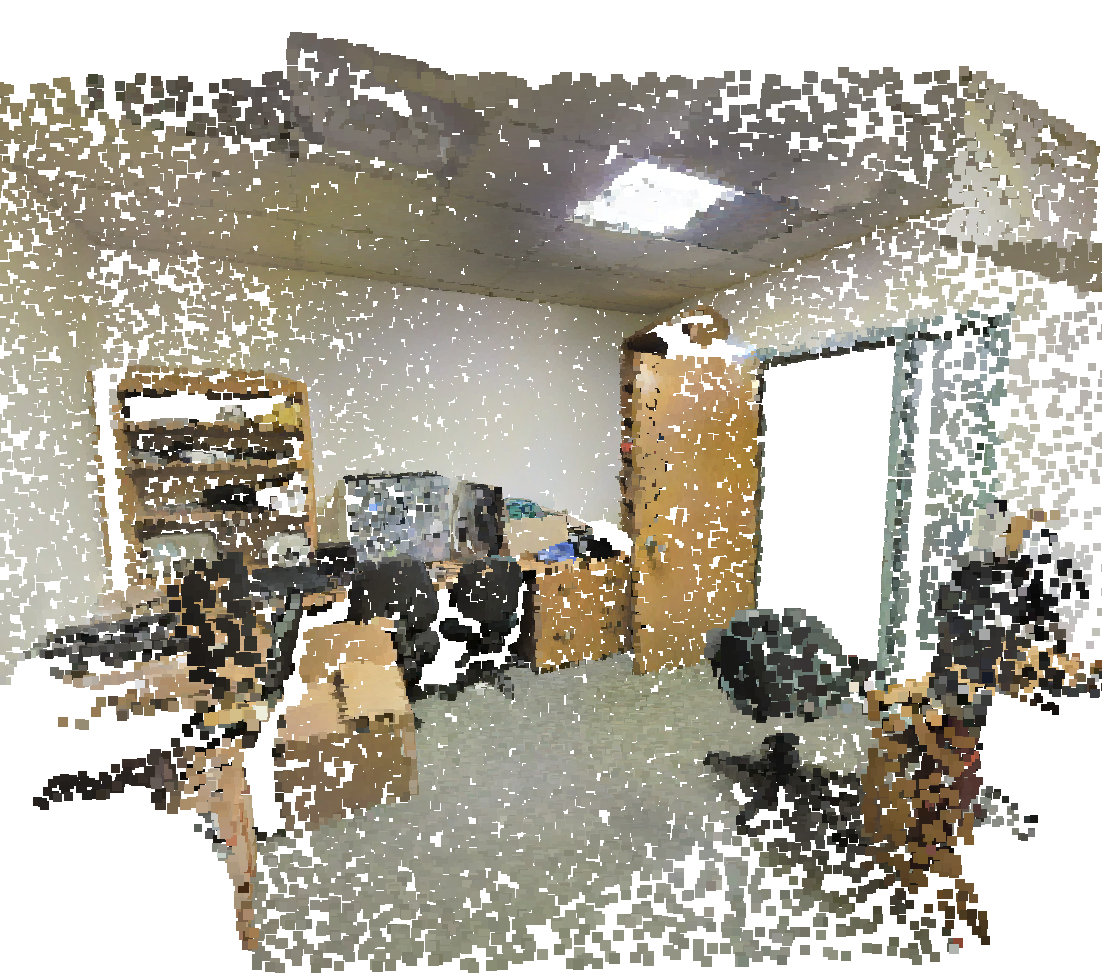} \phantomcaption}
\subfloat{\includegraphics[width=0.225\linewidth,trim={0 2.0cm 0 4.0cm}, clip]{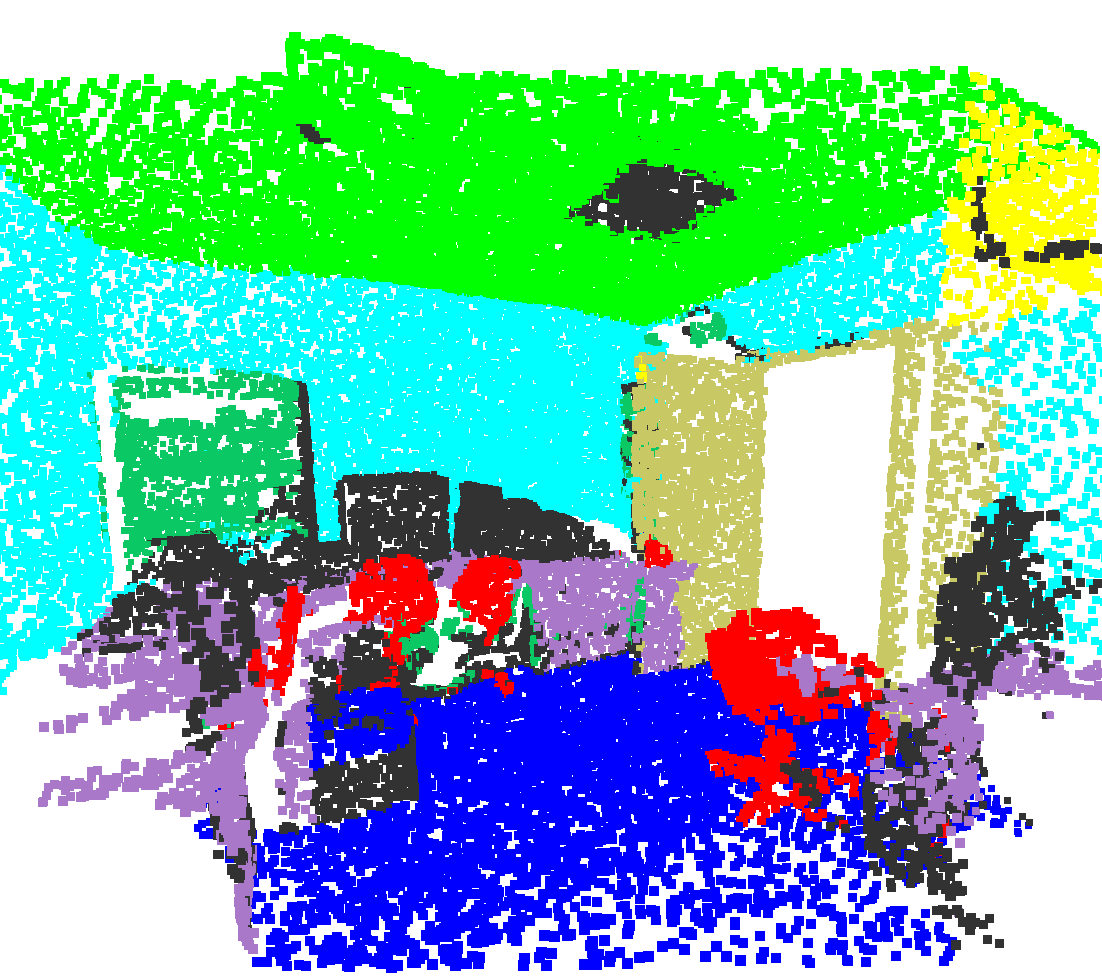} \phantomcaption}
\subfloat{\includegraphics[width=0.225\linewidth,trim={0 2.0cm 0 4.0cm}, clip]{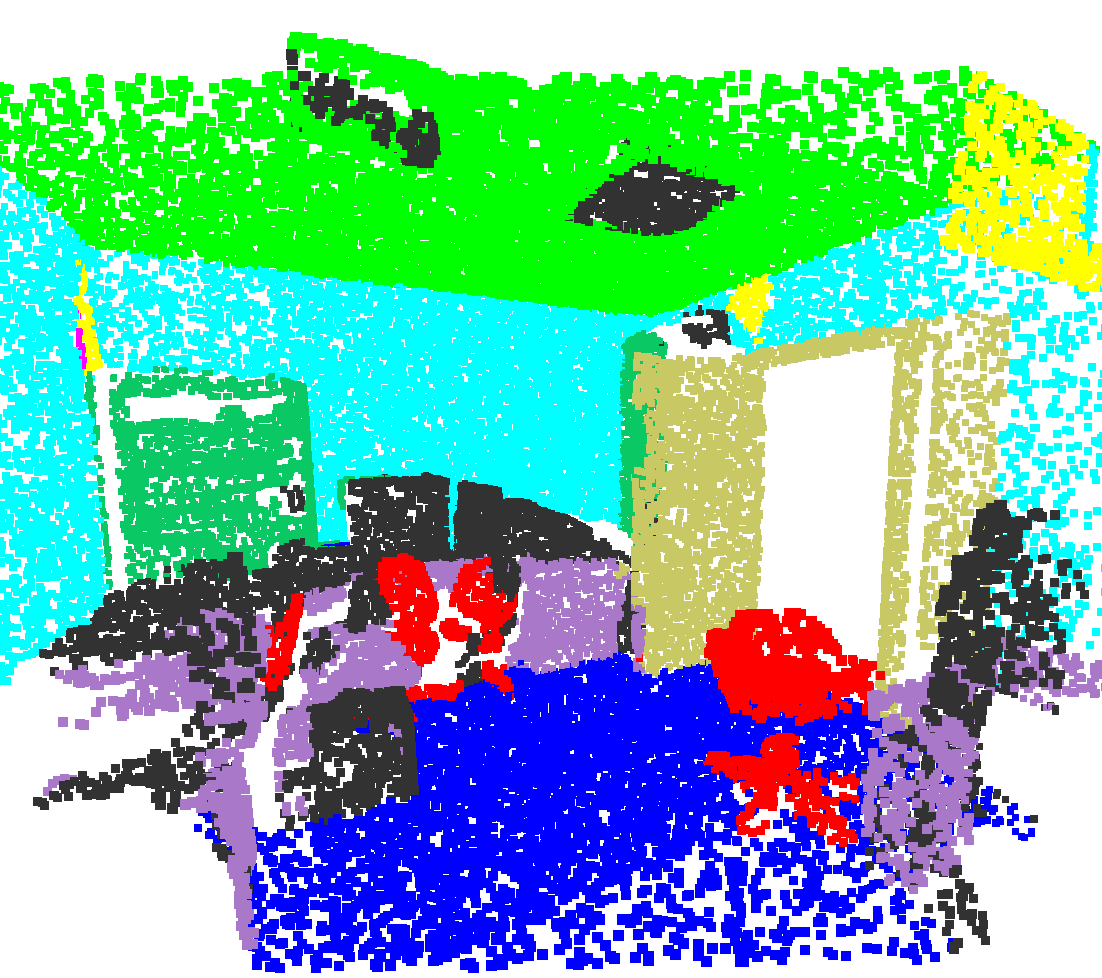} \phantomcaption}
\subfloat{\includegraphics[width=0.225\linewidth,trim={0 2.0cm 0 4.0cm}, clip]{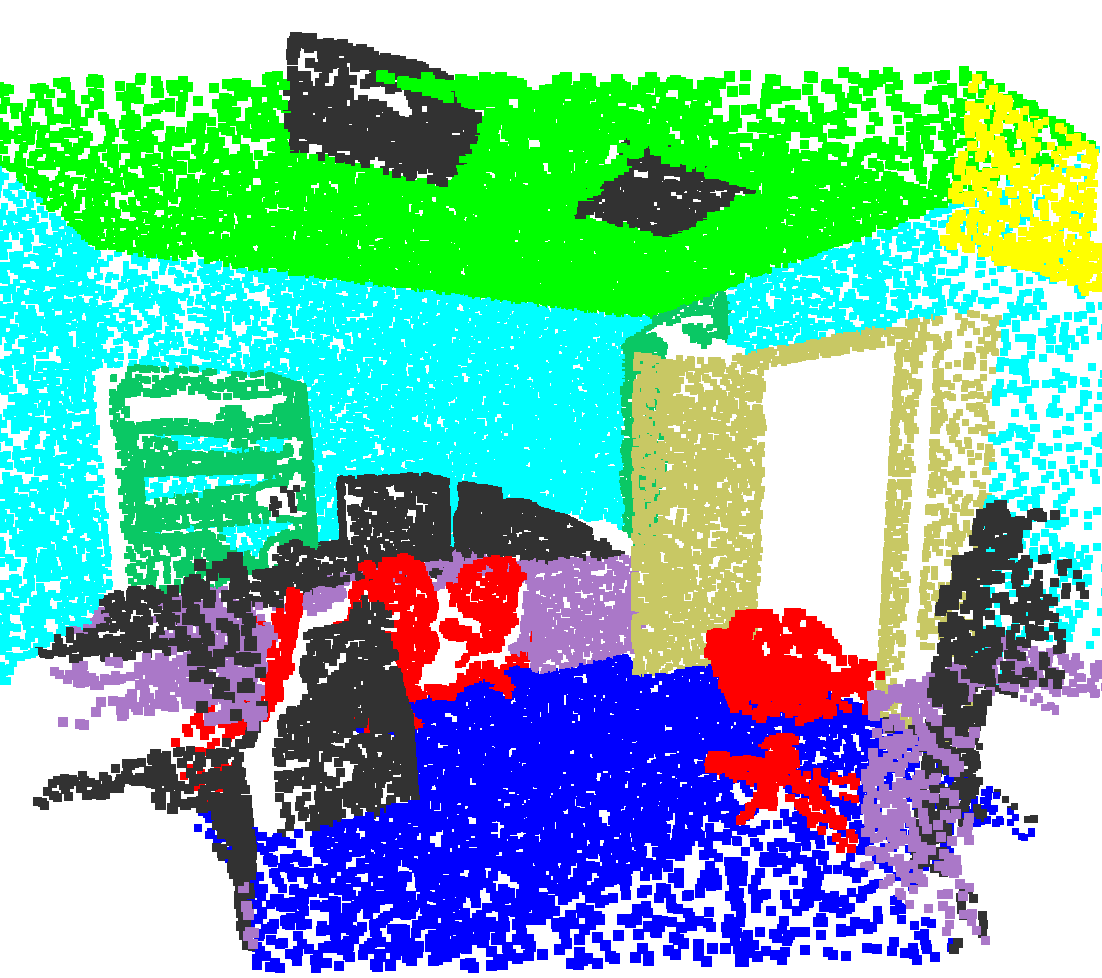} \phantomcaption} \vspace{-0.5mm}
\subfloat{\includegraphics[width=0.225\linewidth,trim={0 4.5cm 0 2.5cm}, clip]{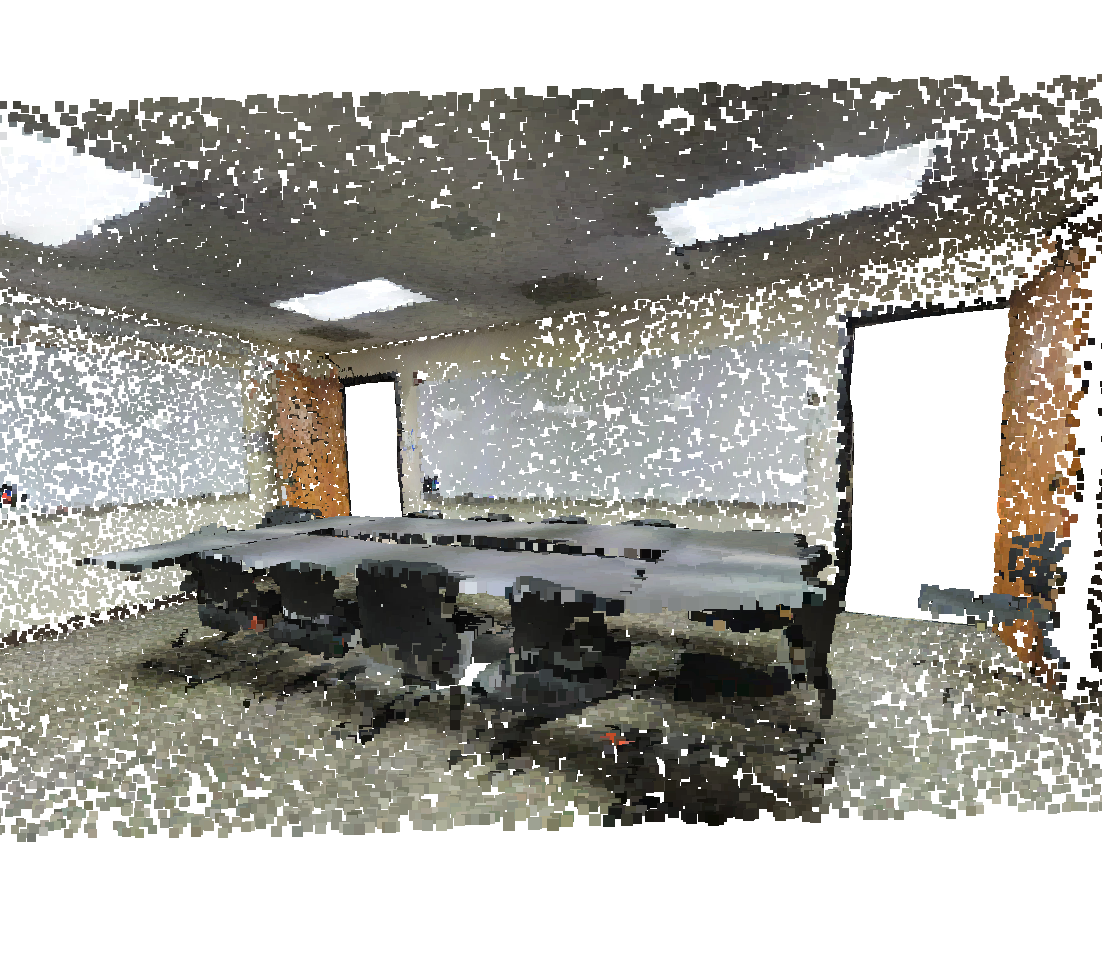} \phantomcaption}
\subfloat{\includegraphics[width=0.225\linewidth,trim={0 4.5cm 0 2.5cm}, clip]{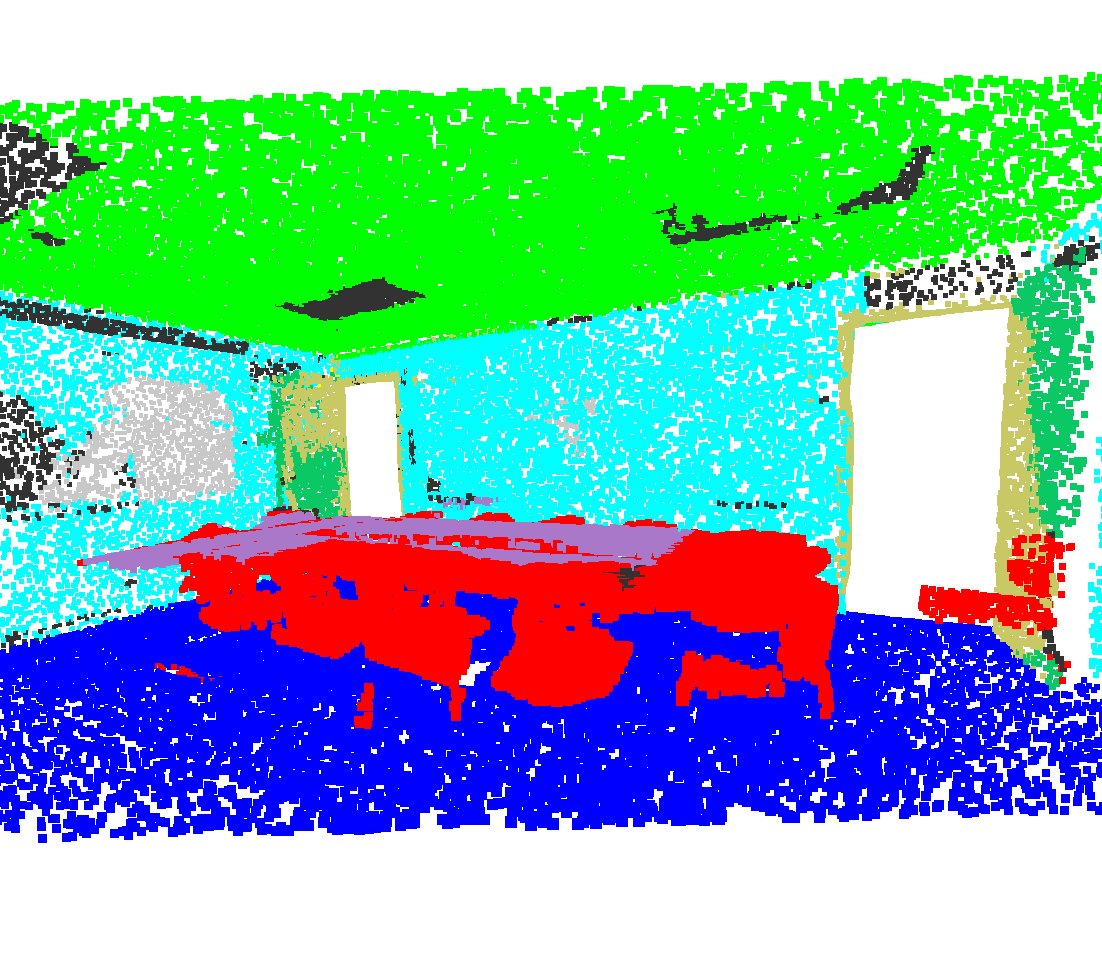} \phantomcaption}
\subfloat{\includegraphics[width=0.225\linewidth,trim={0 4.5cm 0 2.5cm}, clip]{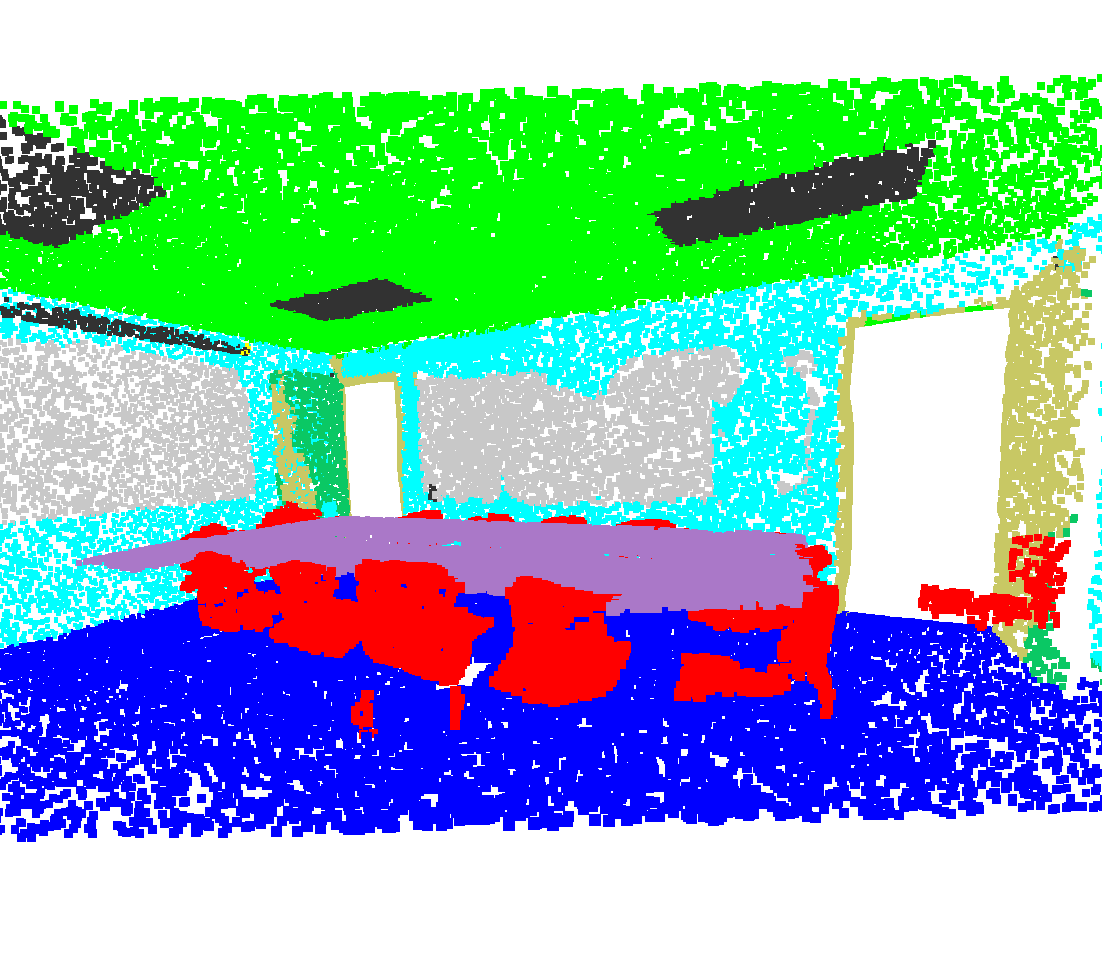} \phantomcaption}
\subfloat{\includegraphics[width=0.225\linewidth,trim={0 4.5cm 0 2.5cm}, clip]{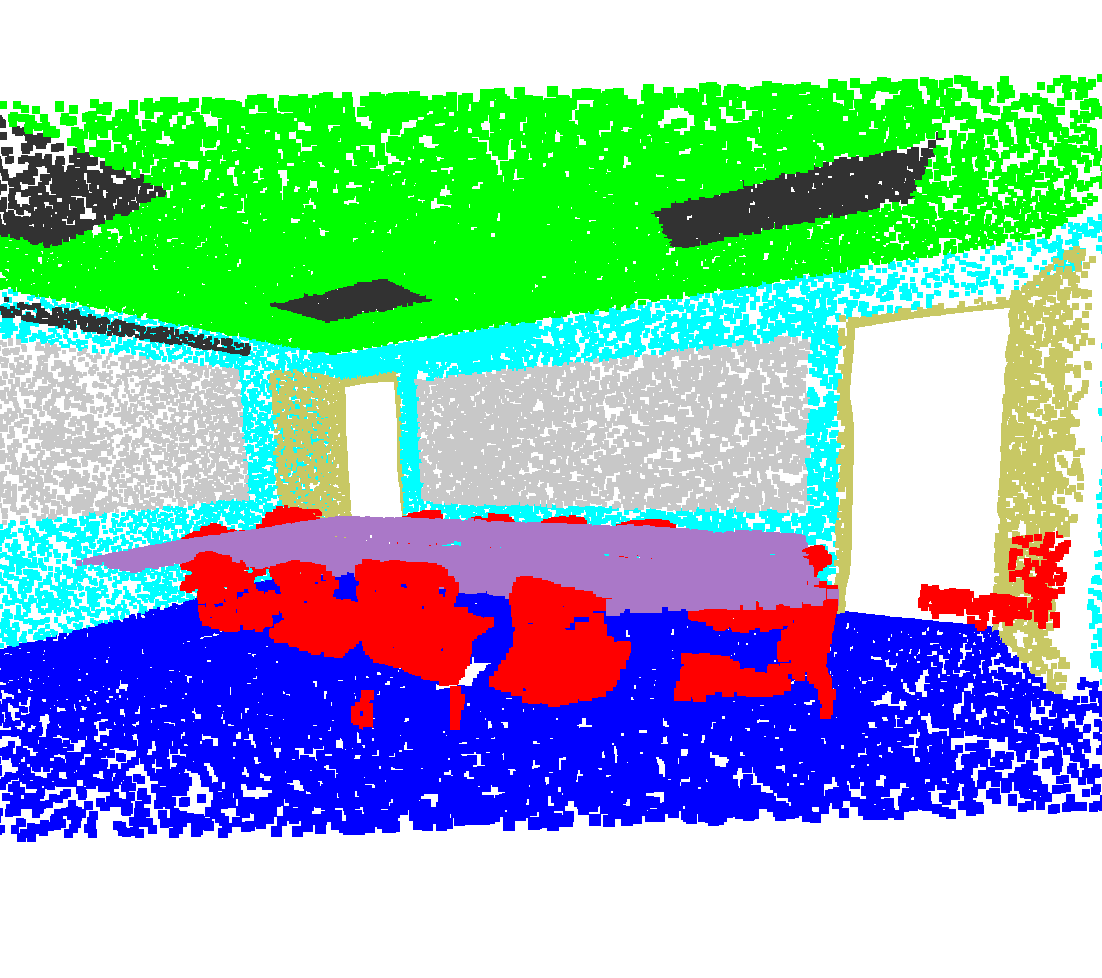} \phantomcaption} \vspace{-1.5mm}
\subfloat{\includegraphics[width=0.225\linewidth,trim={0 0 0 0}, clip]{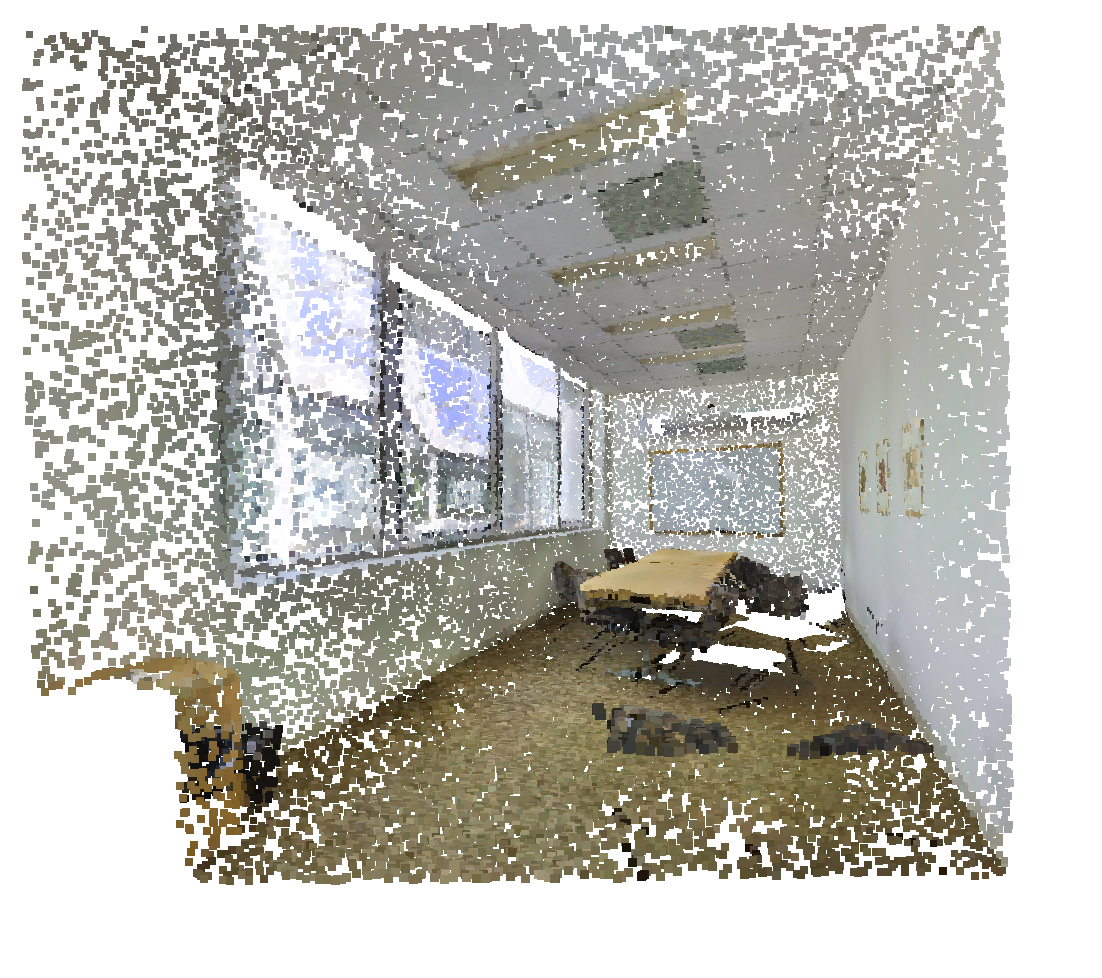} \phantomcaption}
\subfloat{\includegraphics[width=0.225\linewidth,trim={0 0 0 0}, clip]{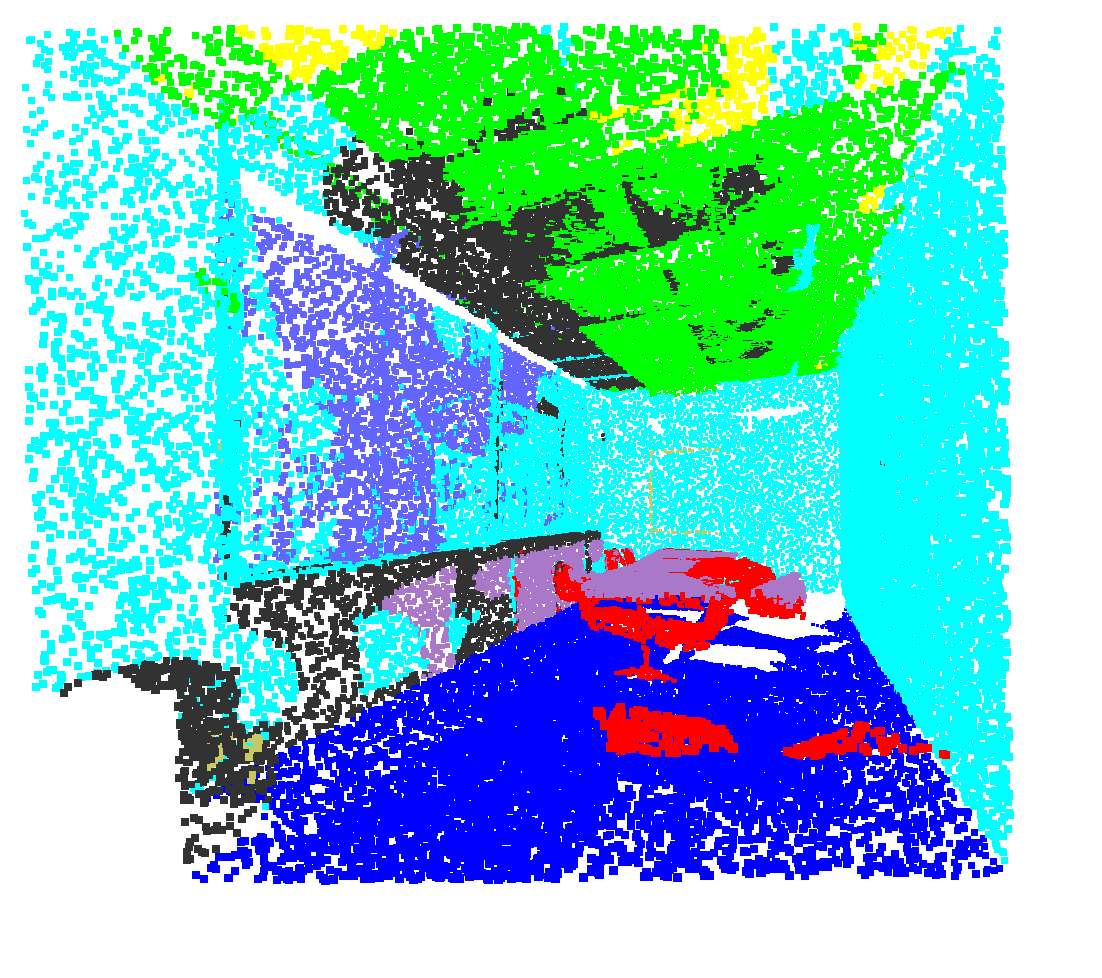} \phantomcaption}
\subfloat{\includegraphics[width=0.225\linewidth,trim={0 0 0 0}, clip]{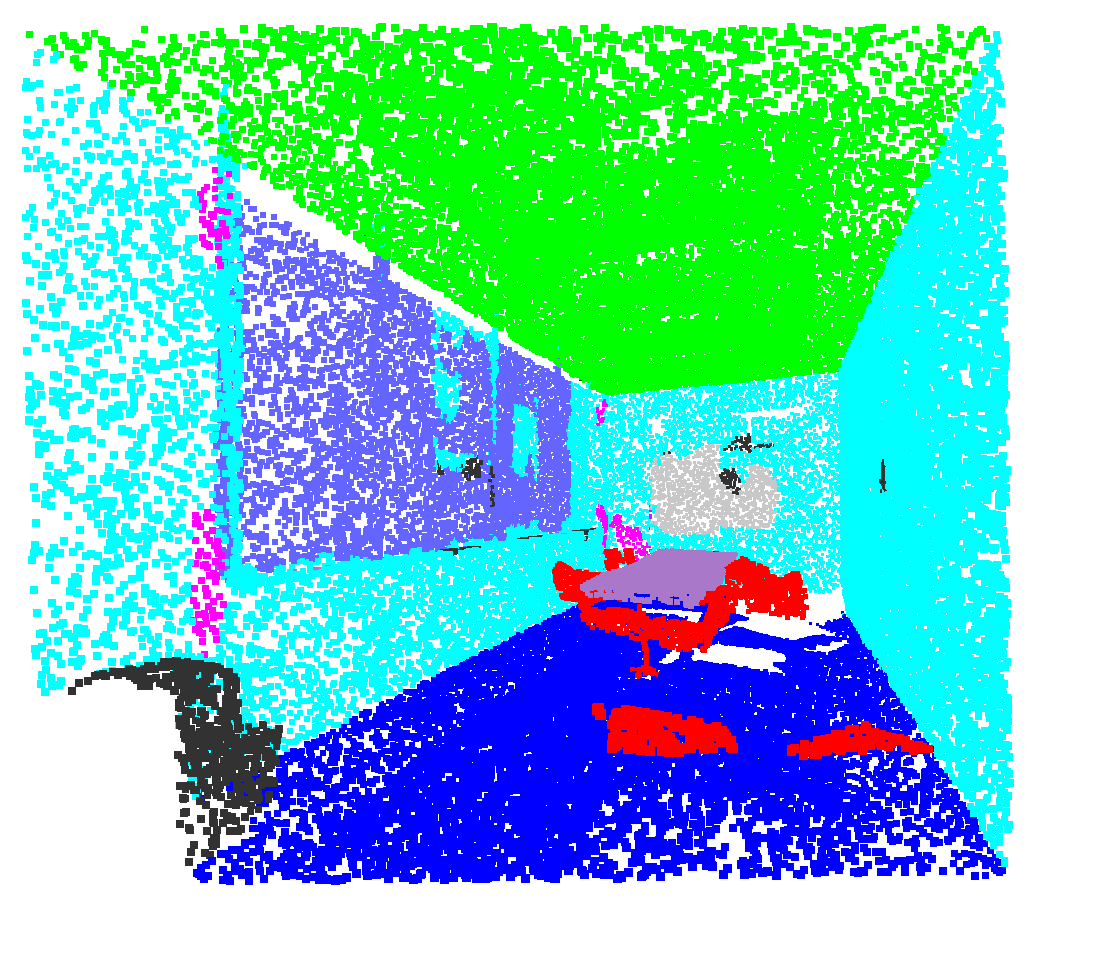} \phantomcaption}
\subfloat{\includegraphics[width=0.225\linewidth,trim={0 0 0 0}, clip]{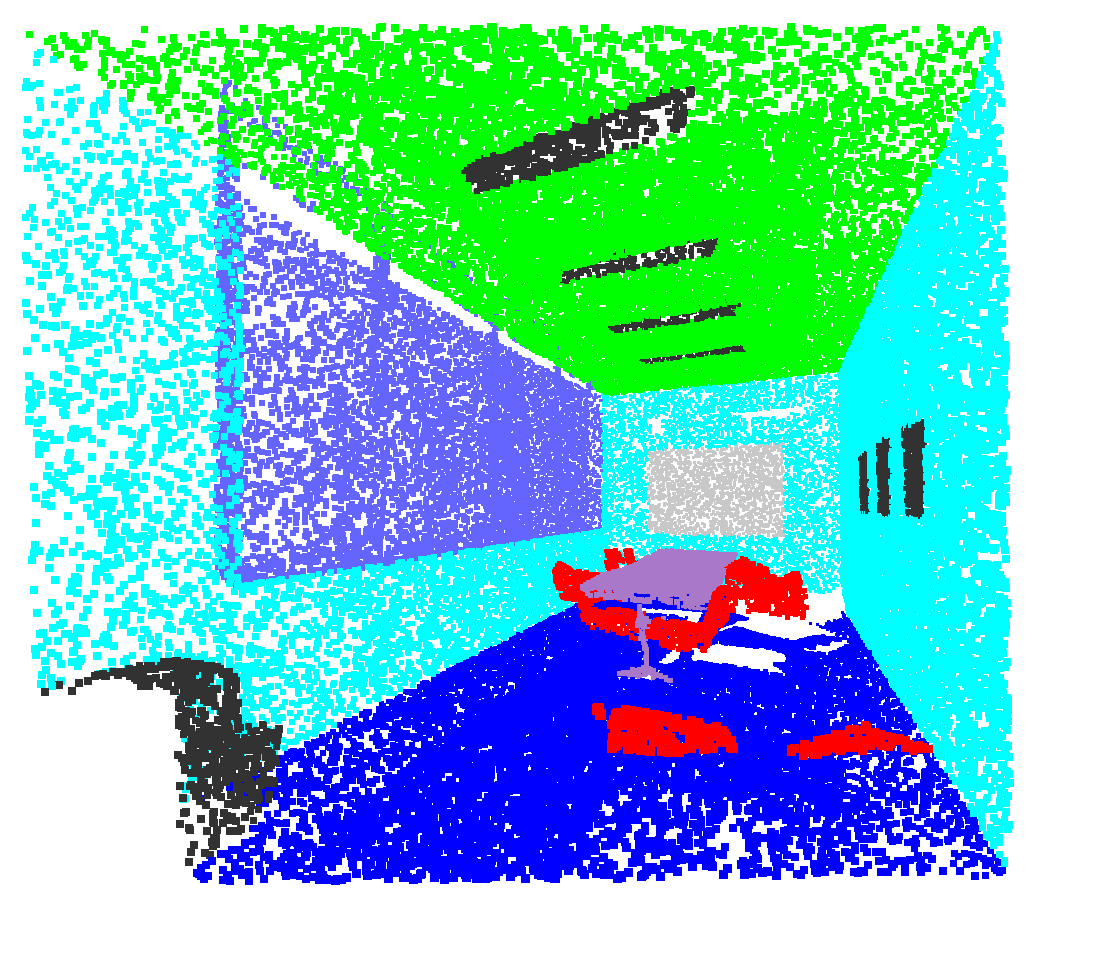} \phantomcaption} \vspace{-3.5mm}
\subfloat[\textbf{Input}]{\includegraphics[width=0.225\linewidth, trim={0 0 0 0}]{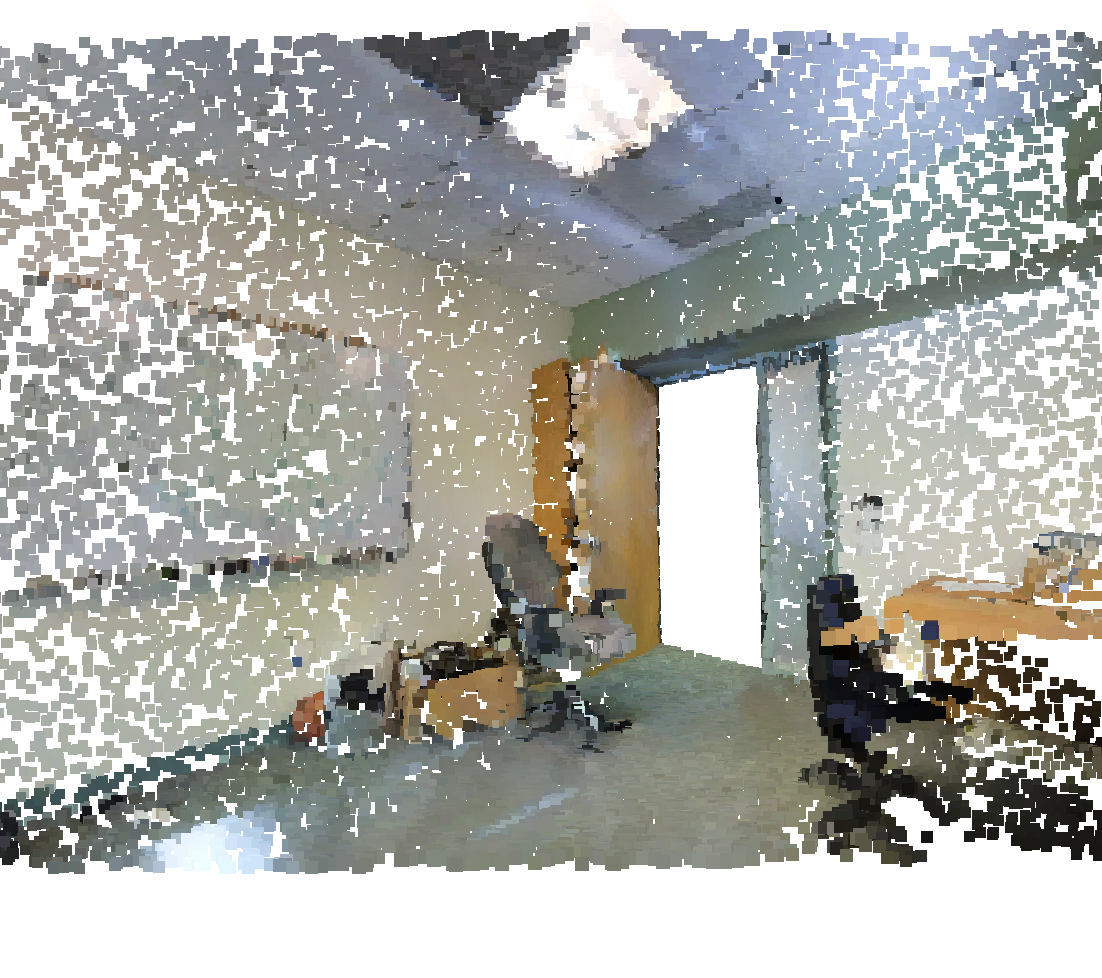}} \hspace{1.0mm}
\subfloat[\textbf{PointNet~\cite{qi2017pointnet}}]{\includegraphics[width=0.225\linewidth, trim={0 0 0 0}]{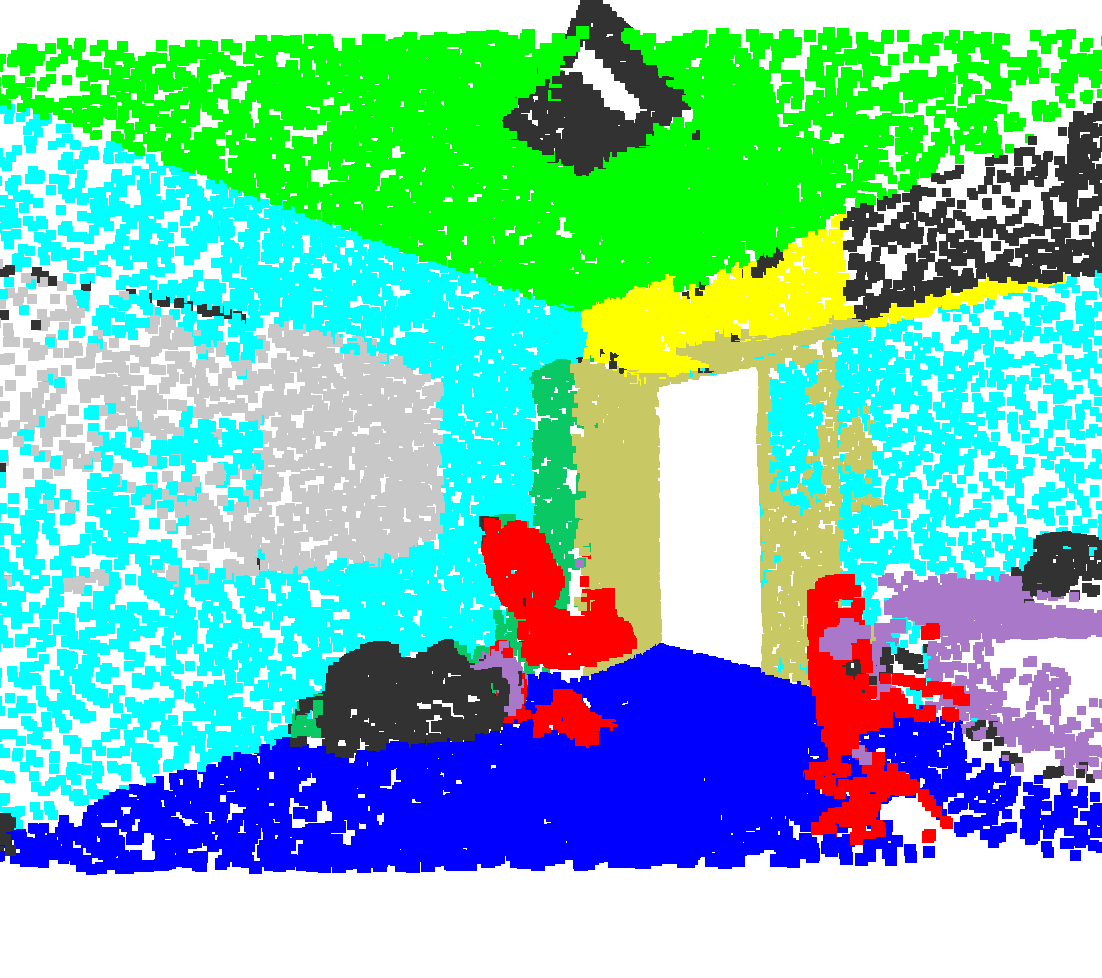}} \hspace{1.0mm}
\subfloat[\textbf{Our}]{\includegraphics[width=0.225\linewidth, trim={0 0 0 0}]{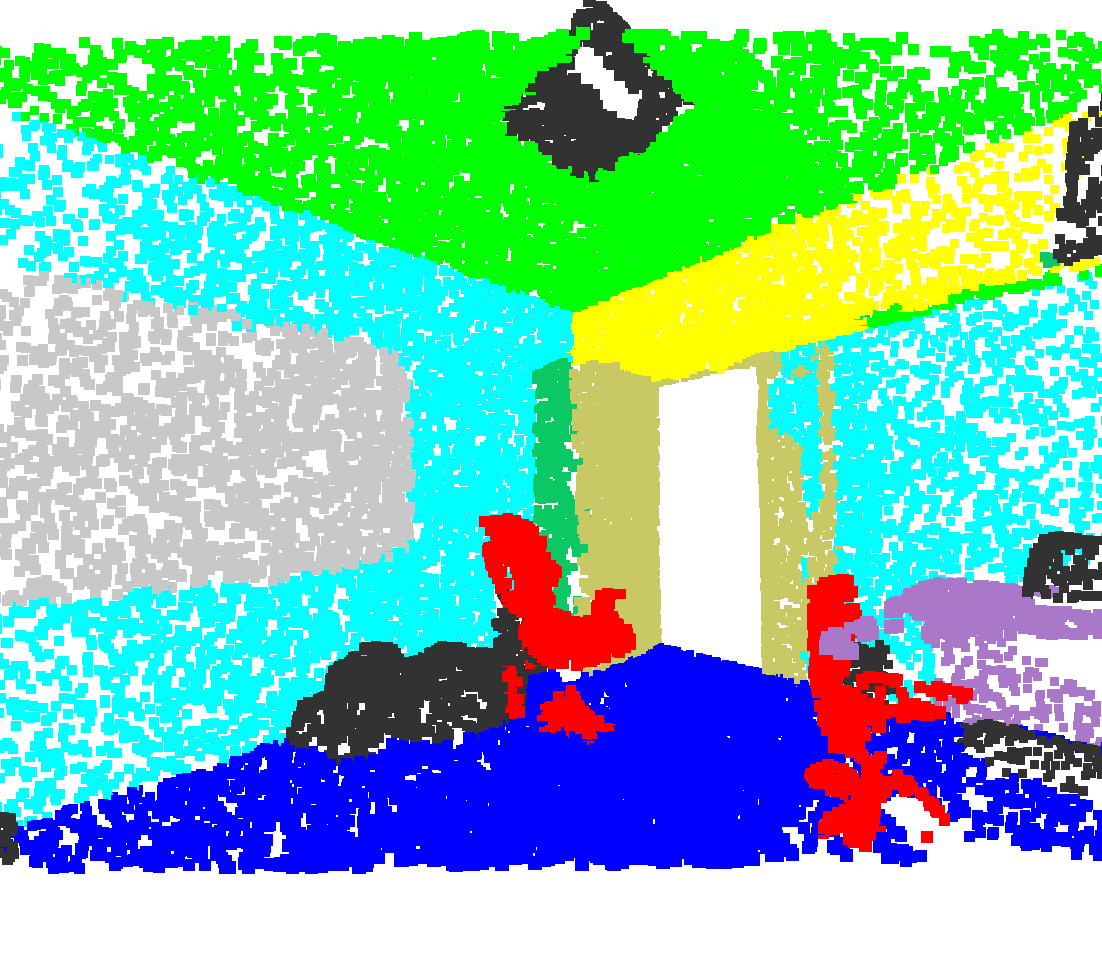}} \hspace{1.0mm}
\subfloat[\textbf{Ground Truth}]{\includegraphics[width=0.225\linewidth, trim={0 0 0 0}]{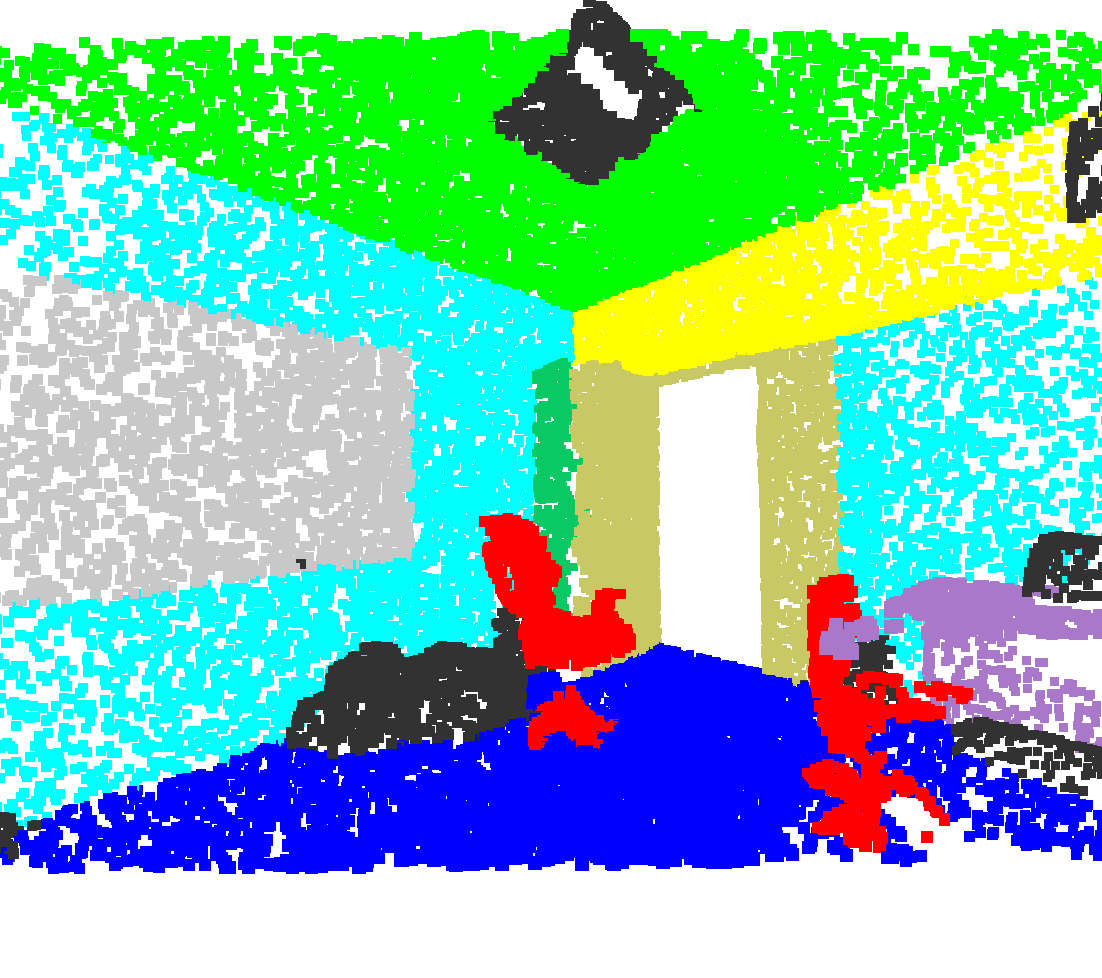}}
\caption{The visualization results on \emph{S3DIS} dataset. We compare our model with PointNet~\cite{qi2017pointnet} and the ground truth. The challenging sample rooms have been picked from the all six areas: area 1 (\textit{row 1}), area 2 (\textit{row 2}) area 3 (\textit{row 3}), area 4 (\textit{row 4}), area 5 (\textit{row 5}), and area 6 (\textit{row 6}).}
\vspace{-5.5mm}
\label{fig:suppl_s3dis_eval}
\end{figure*}

\setcounter{figure}{-7}
\begin{figure*}[t]
\centering
\subfloat{\includegraphics[width=0.9\linewidth, clip]{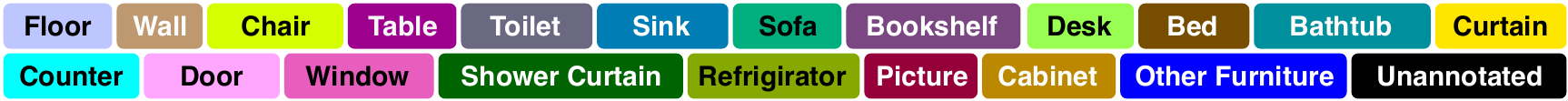}}
\vspace{-3.5mm}
\subfloat{\includegraphics[width=0.225\linewidth,trim={0 0 0 0}, clip]{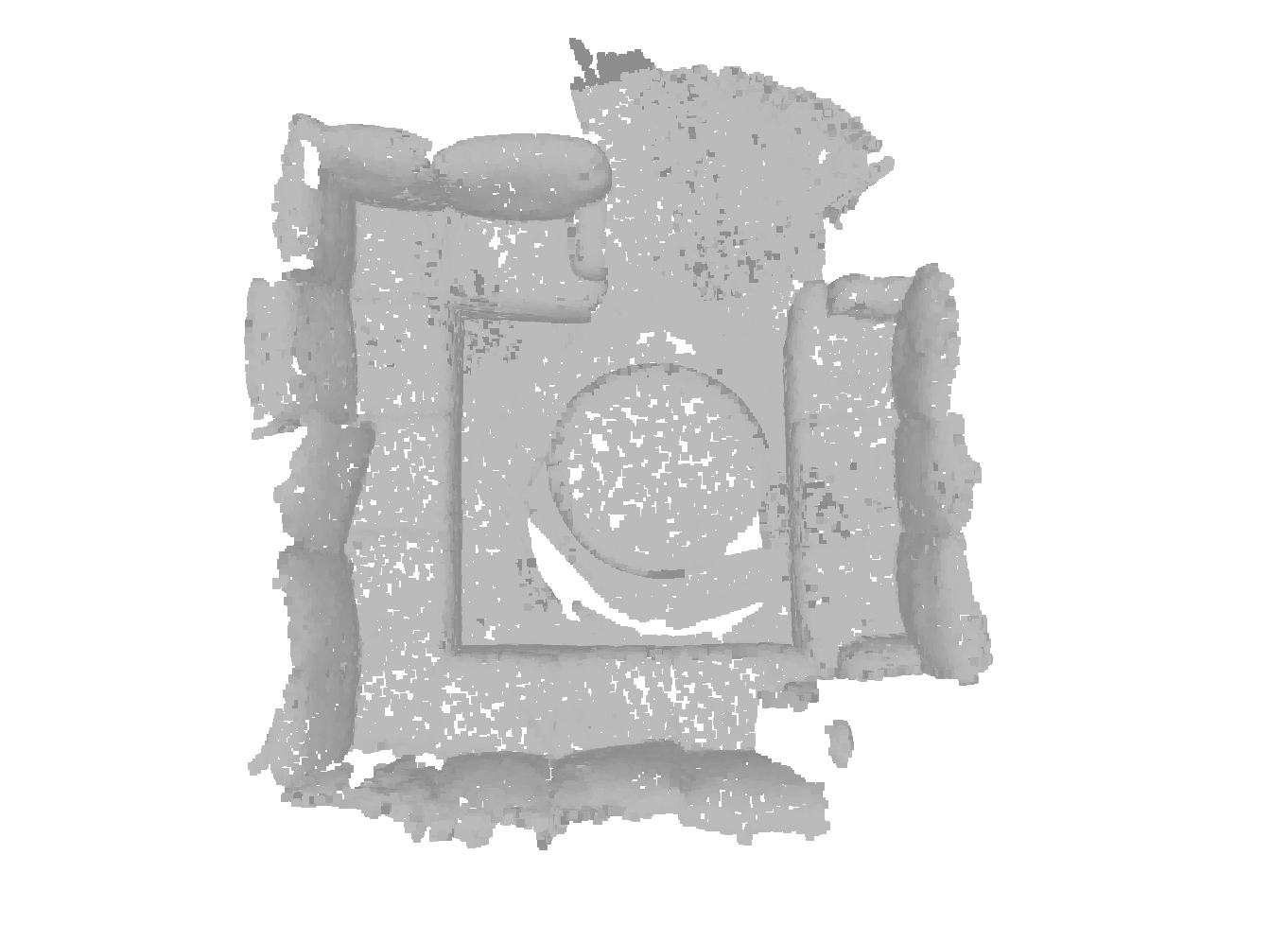} \phantomcaption}
\subfloat{\includegraphics[width=0.225\linewidth,trim={0 0 0 0}, clip]{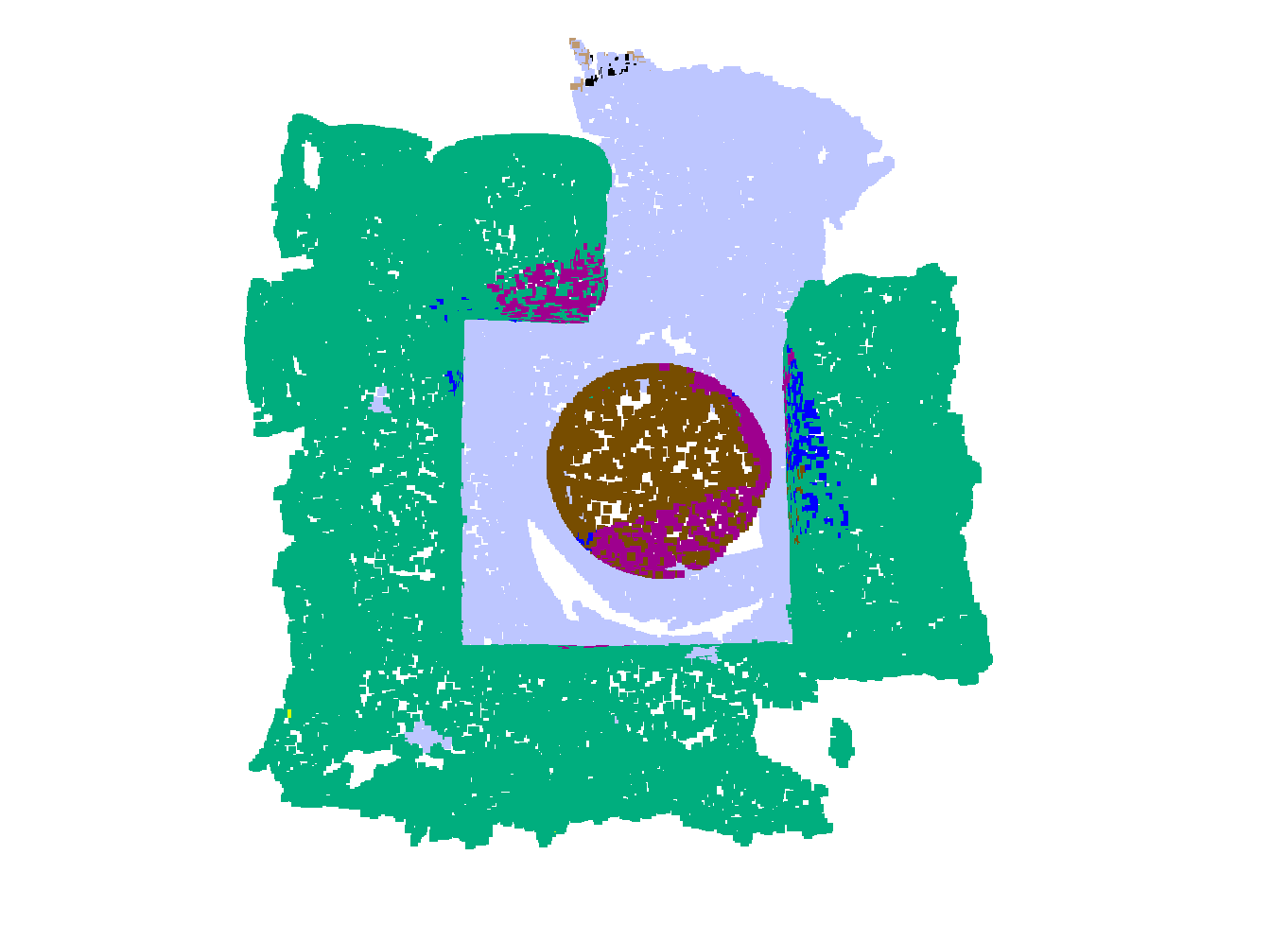} \phantomcaption}
\subfloat{\includegraphics[width=0.225\linewidth,trim={0 0 0 0}, clip]{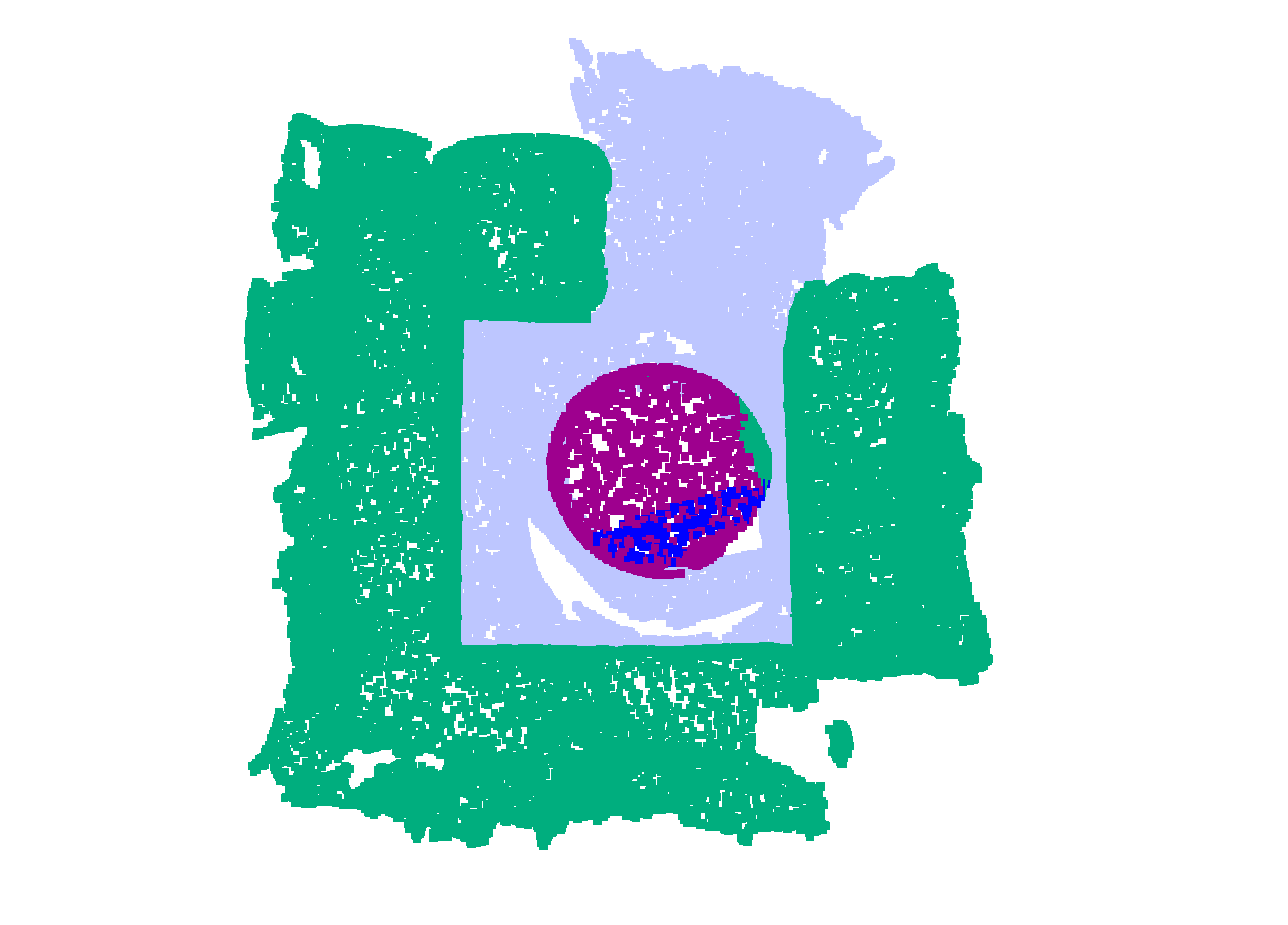} \phantomcaption}
\subfloat{\includegraphics[width=0.225\linewidth,trim={0 0 0 0}, clip]{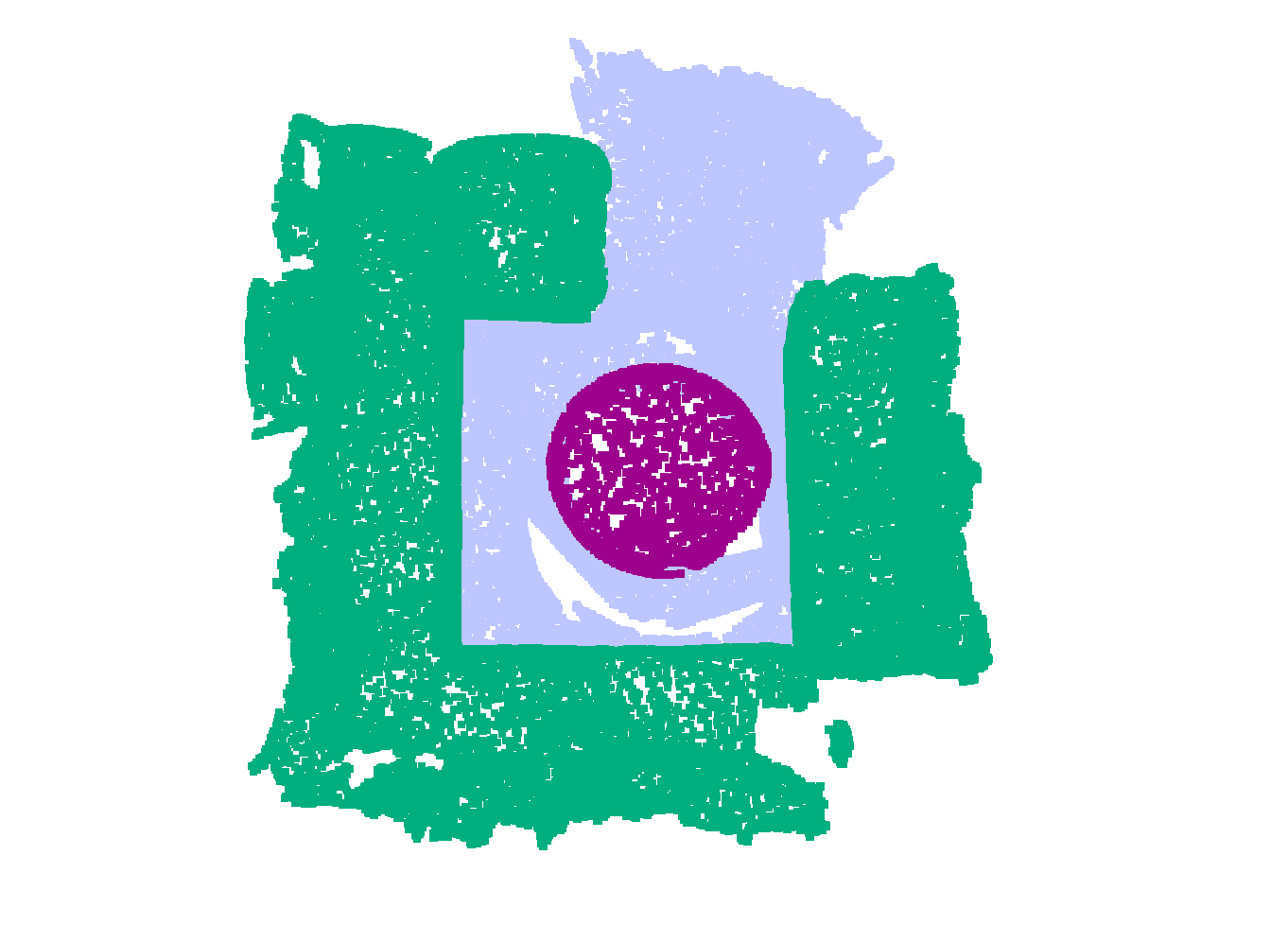} \phantomcaption}
\vspace{-3.5mm}
\subfloat{\includegraphics[width=0.225\linewidth,trim={0 0 0 0}, clip]{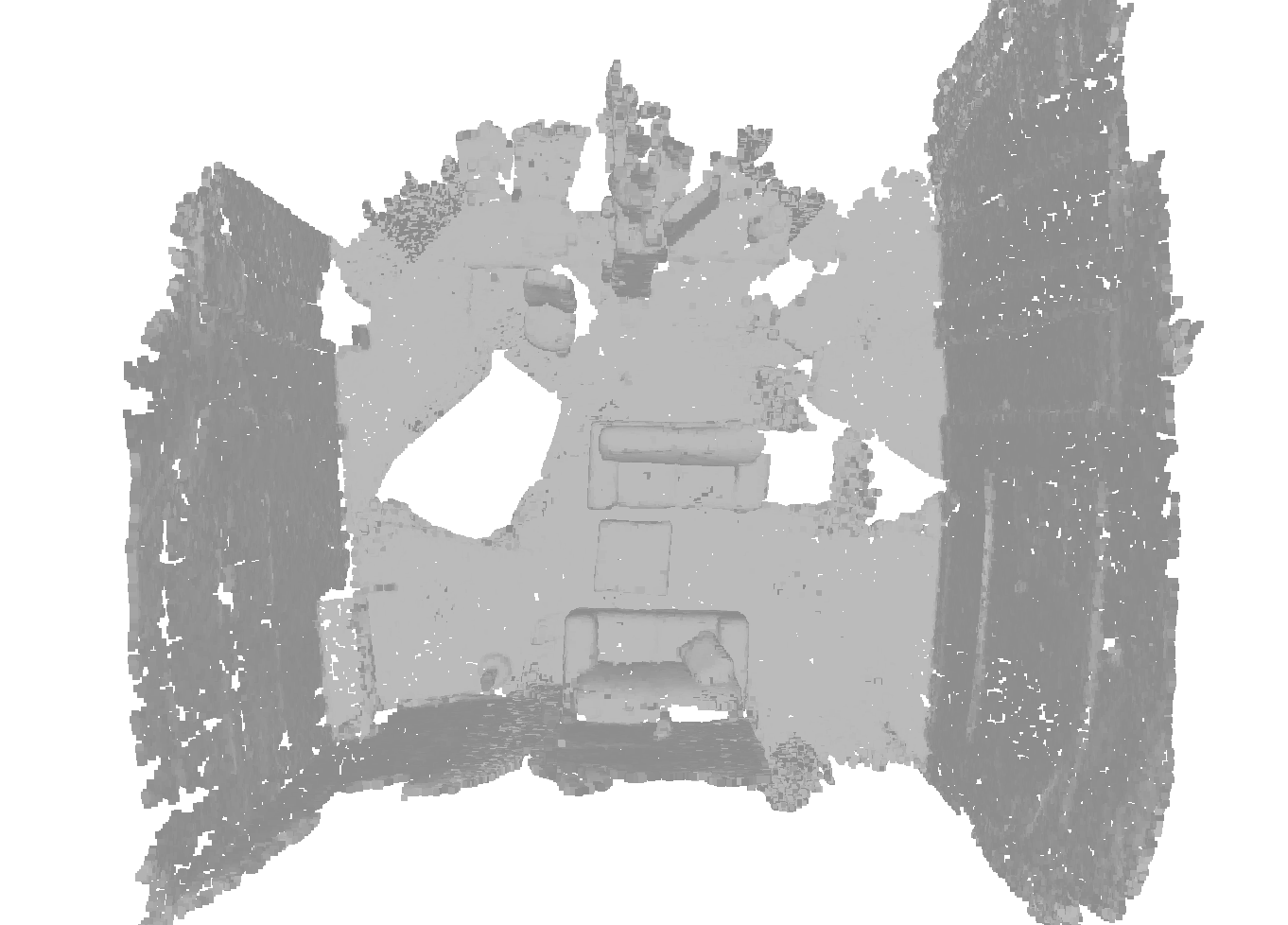} \phantomcaption}
\subfloat{\includegraphics[width=0.225\linewidth,trim={0 0 0 0}, clip]{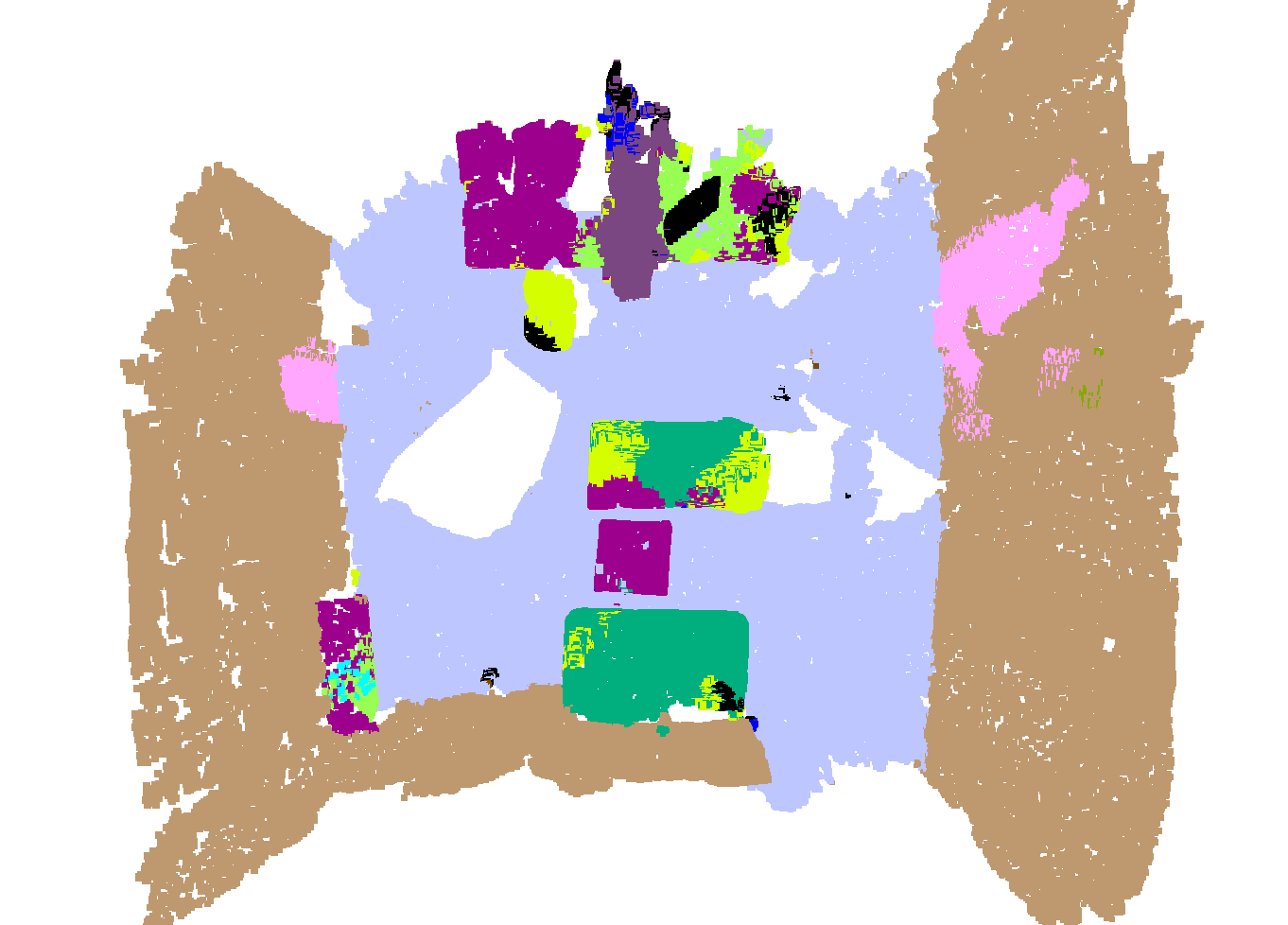} \phantomcaption}
\subfloat{\includegraphics[width=0.225\linewidth,trim={0 0 0 0}, clip]{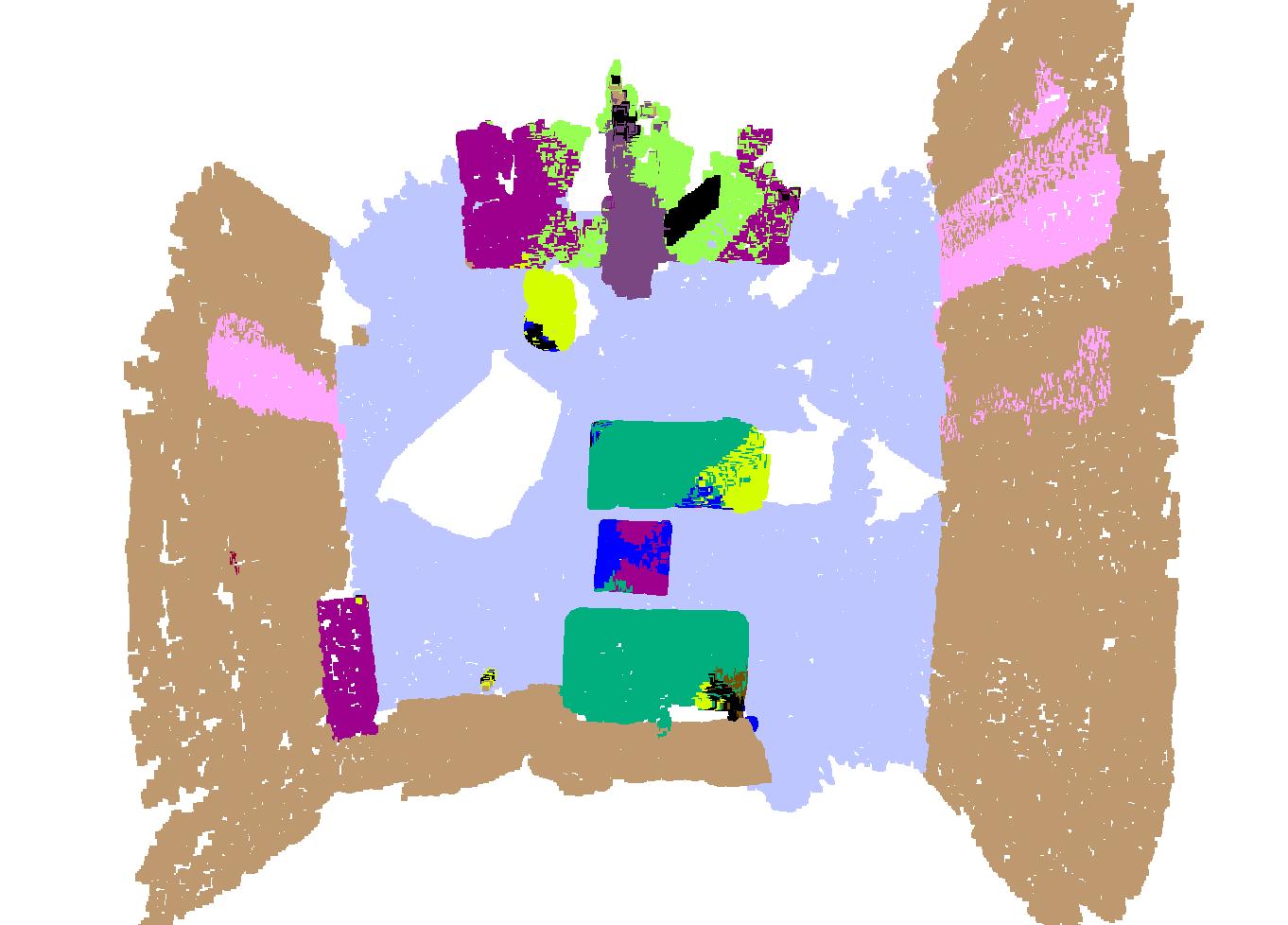} \phantomcaption}
\subfloat{\includegraphics[width=0.225\linewidth,trim={0 0 0 0}, clip]{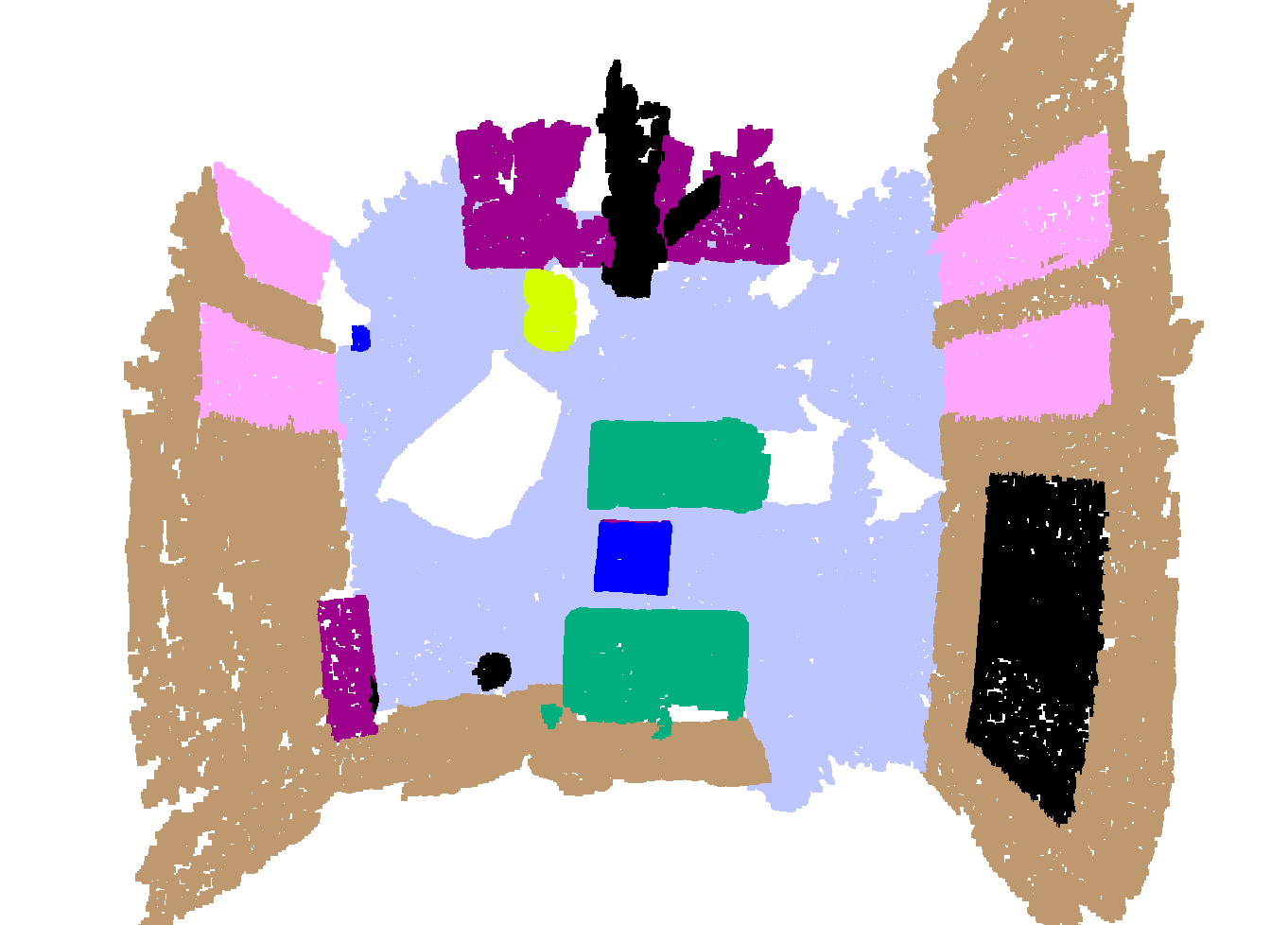} \phantomcaption}
\vspace{-3.5mm}
\subfloat{\includegraphics[width=0.225\linewidth,trim={0 0 0 0}, clip]{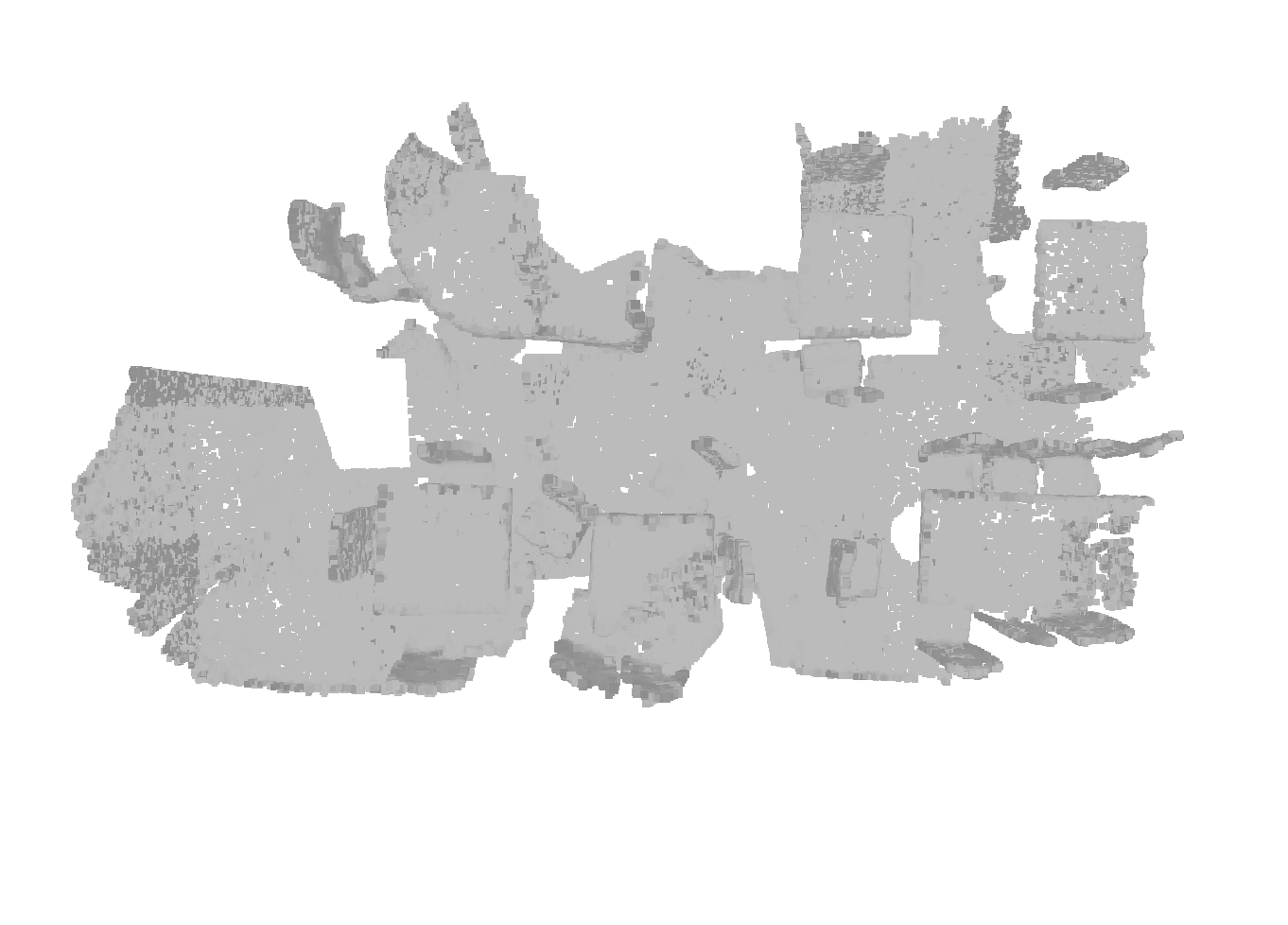} \phantomcaption}
\subfloat{\includegraphics[width=0.225\linewidth,trim={0 0 0 0}, clip]{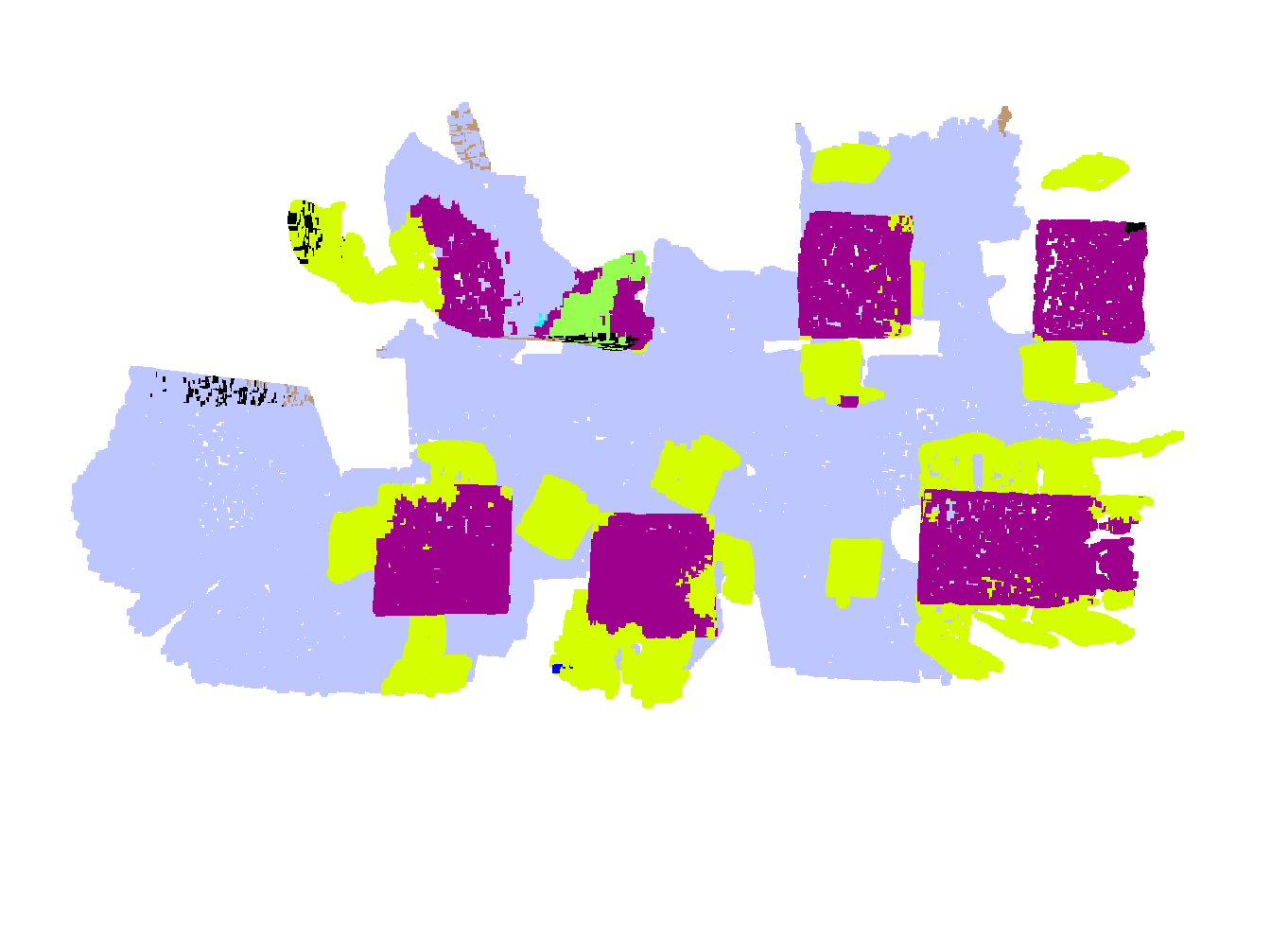} \phantomcaption}
\subfloat{\includegraphics[width=0.225\linewidth,trim={0 0 0 0}, clip]{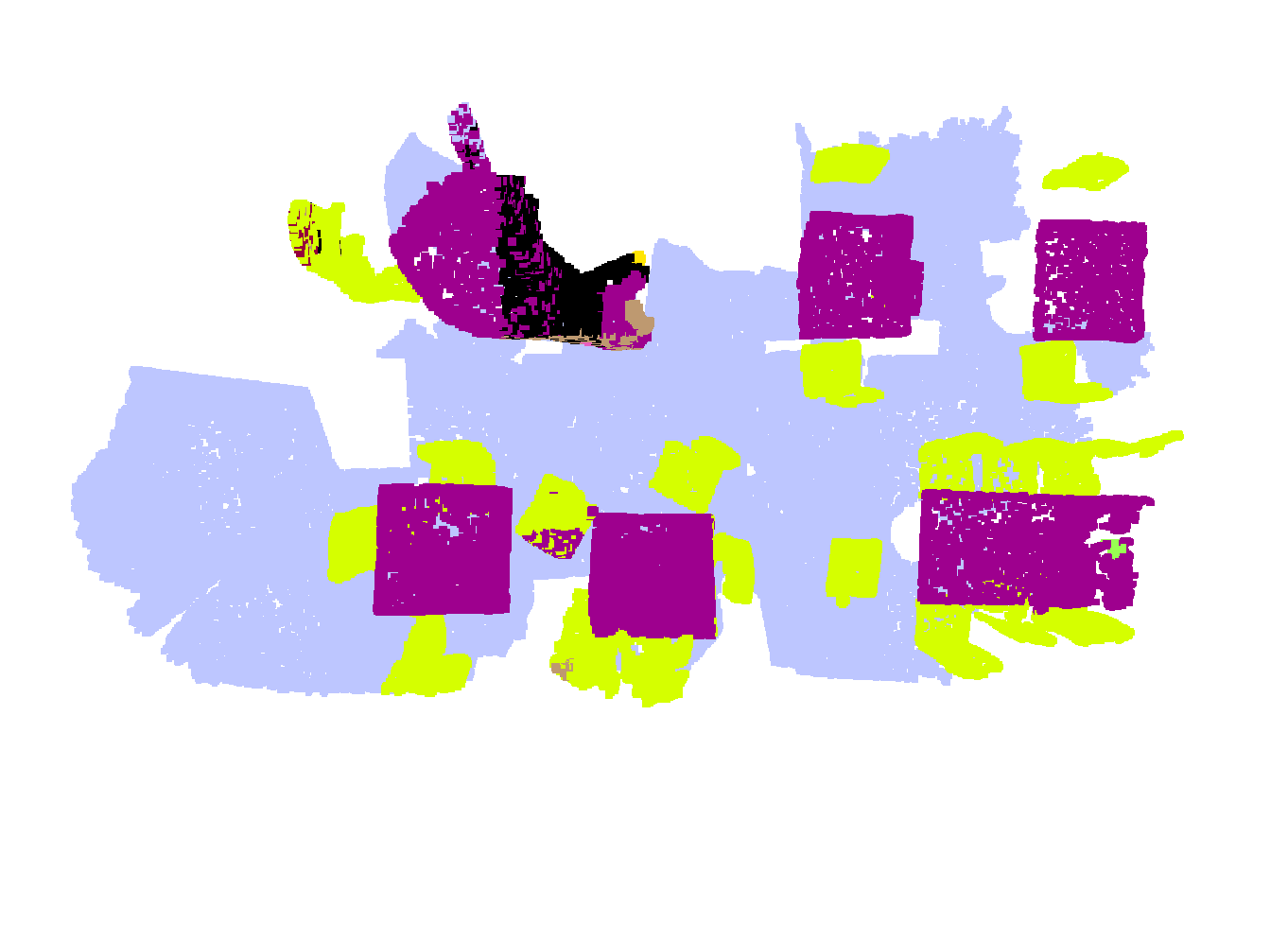} \phantomcaption}
\subfloat{\includegraphics[width=0.225\linewidth,trim={0 0 0 0}, clip]{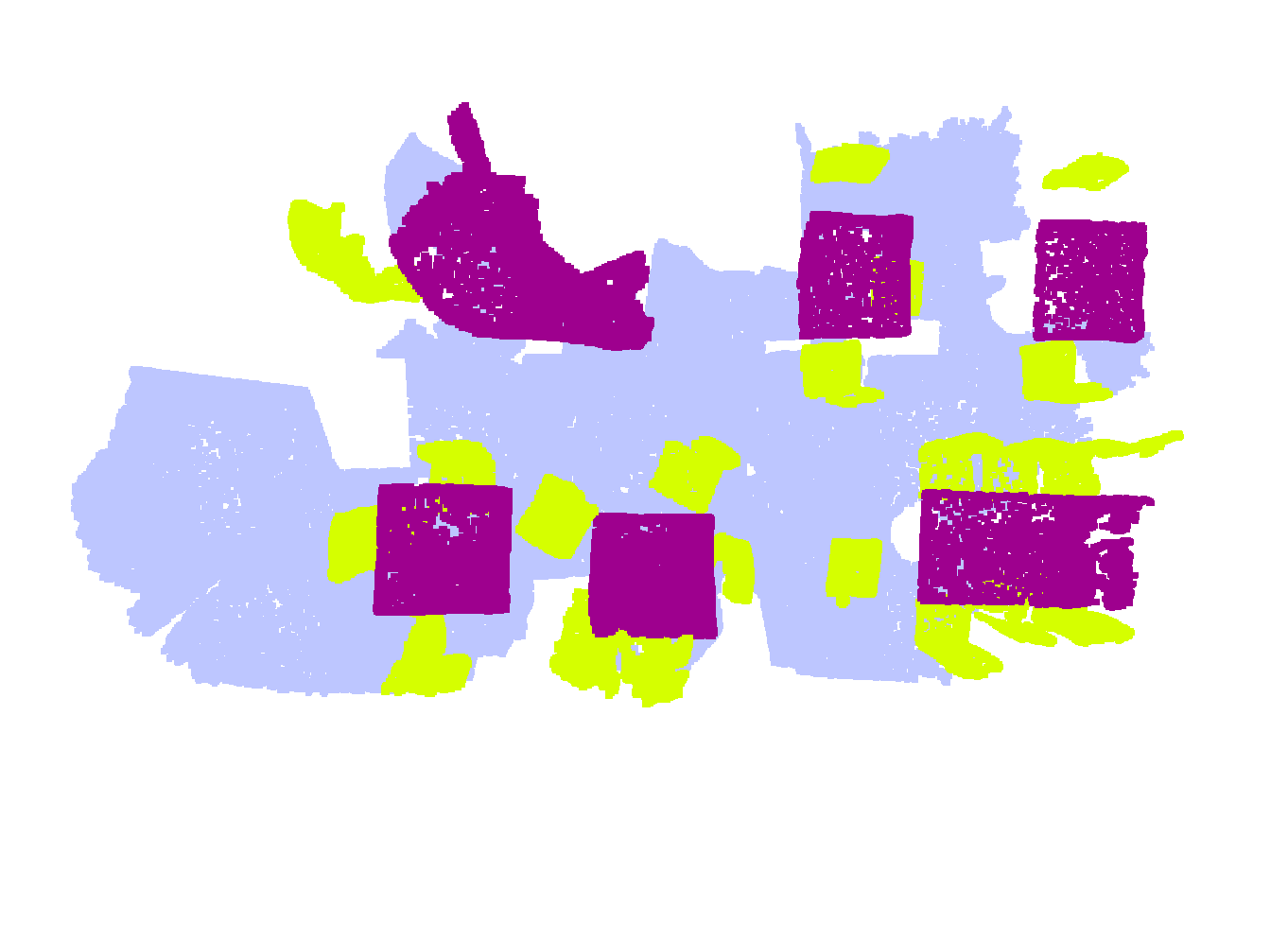} \phantomcaption}
\vspace{-3.5mm}
\subfloat{\includegraphics[width=0.225\linewidth,trim={0 0 0 0}, clip]{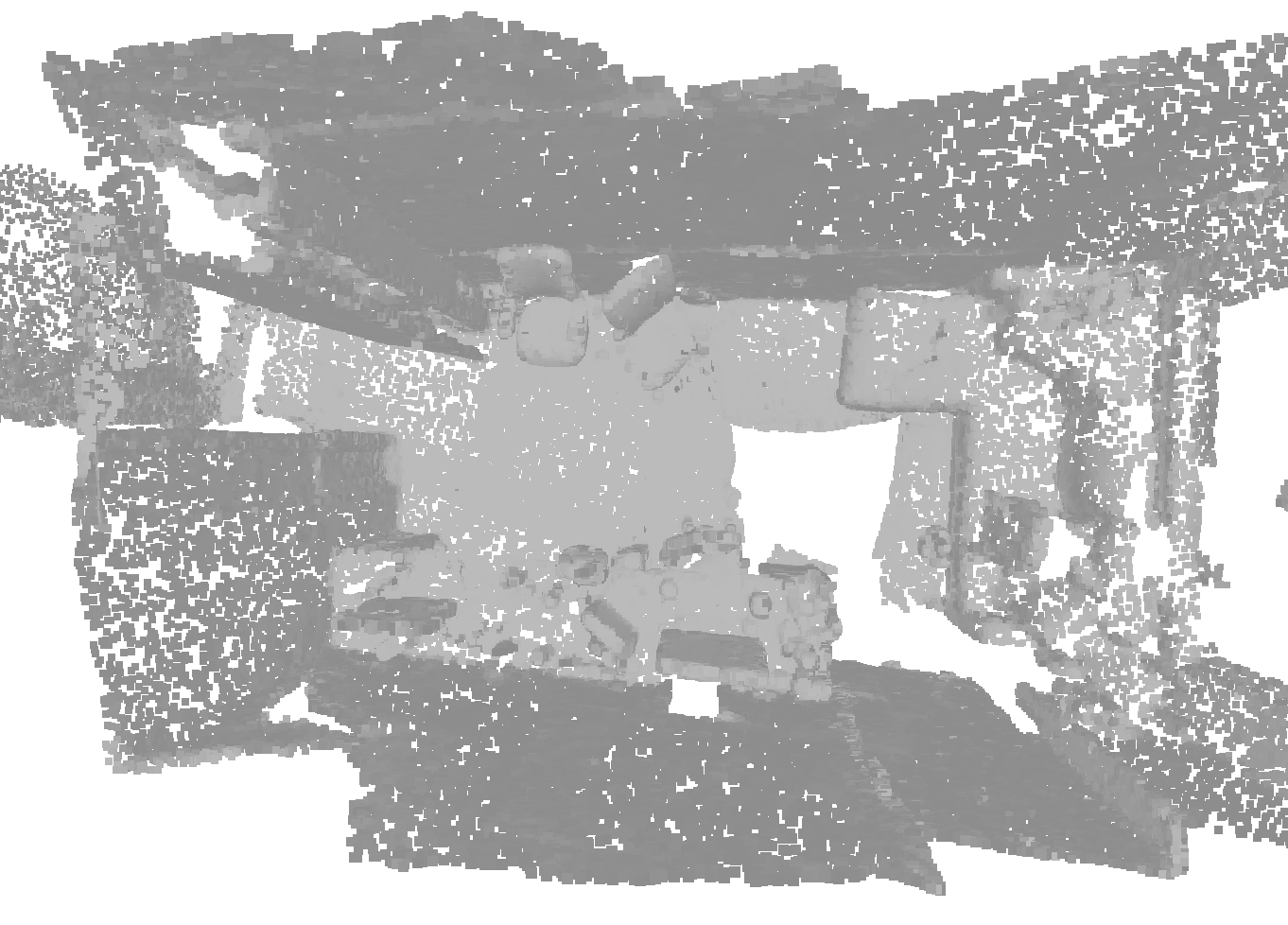} \phantomcaption}
\subfloat{\includegraphics[width=0.225\linewidth,trim={0 0 0 0}, clip]{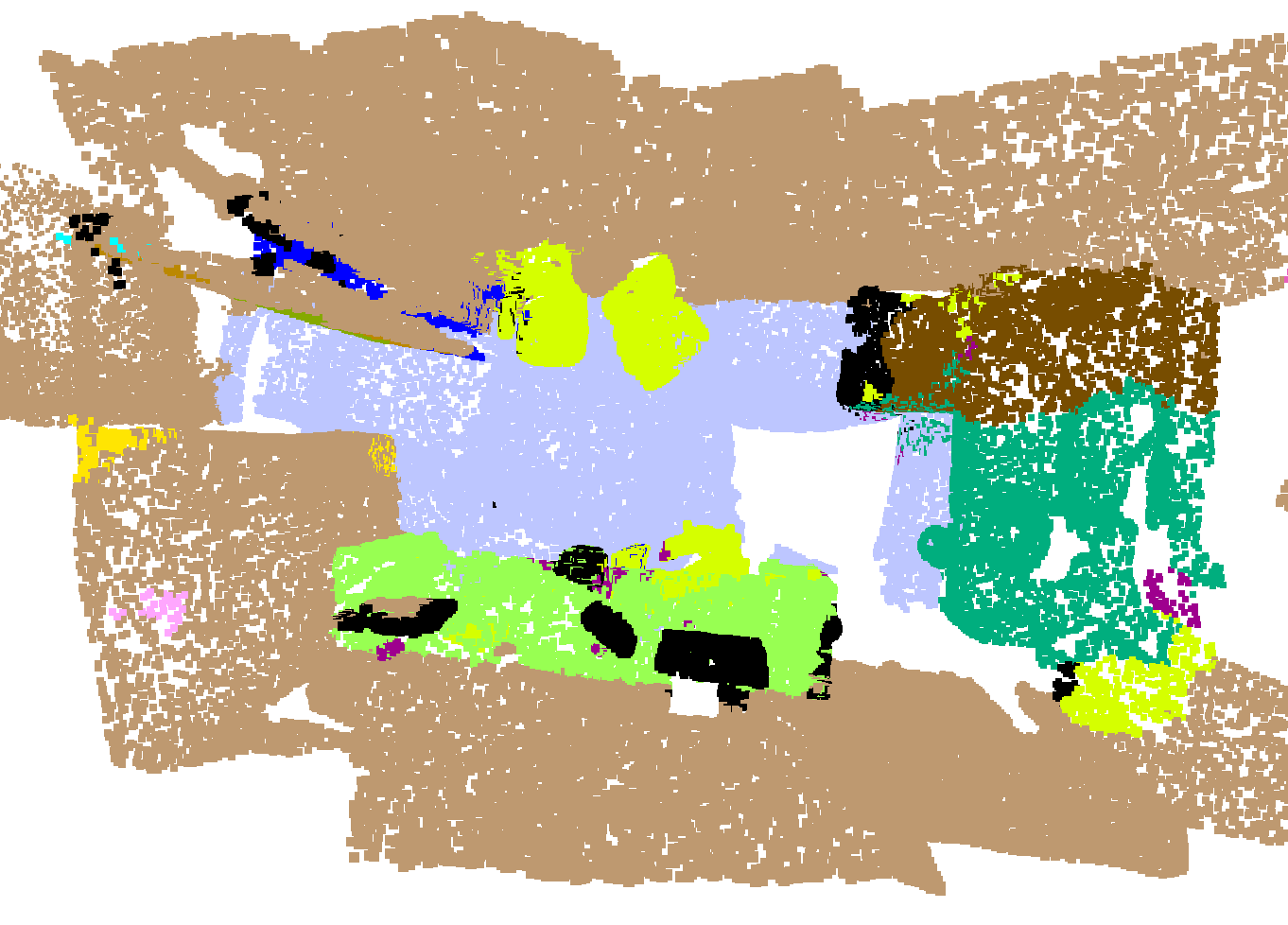} \phantomcaption}
\subfloat{\includegraphics[width=0.225\linewidth,trim={0 0 0 0}, clip]{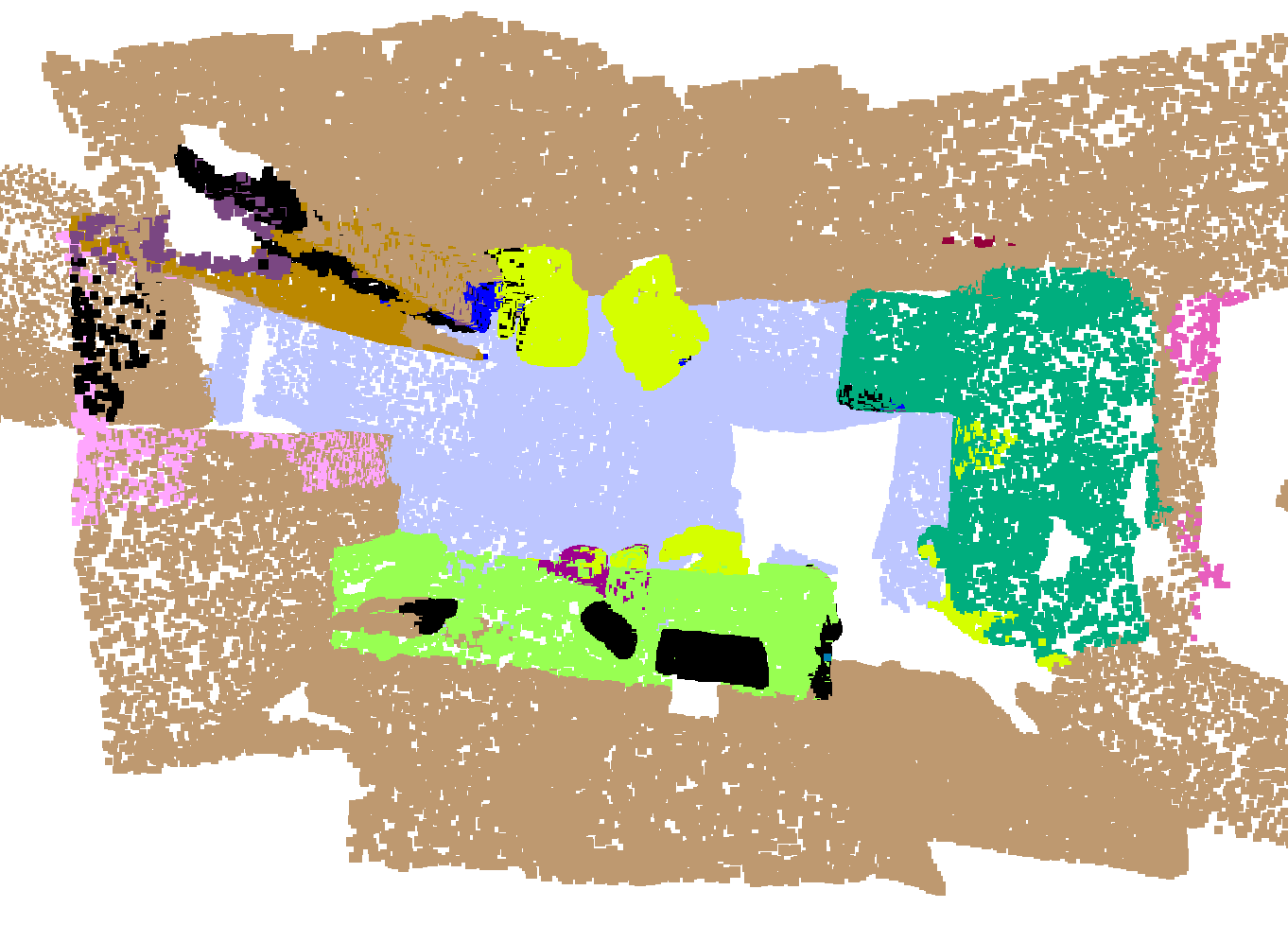} \phantomcaption}
\subfloat{\includegraphics[width=0.225\linewidth,trim={0 0 0 0}, clip]{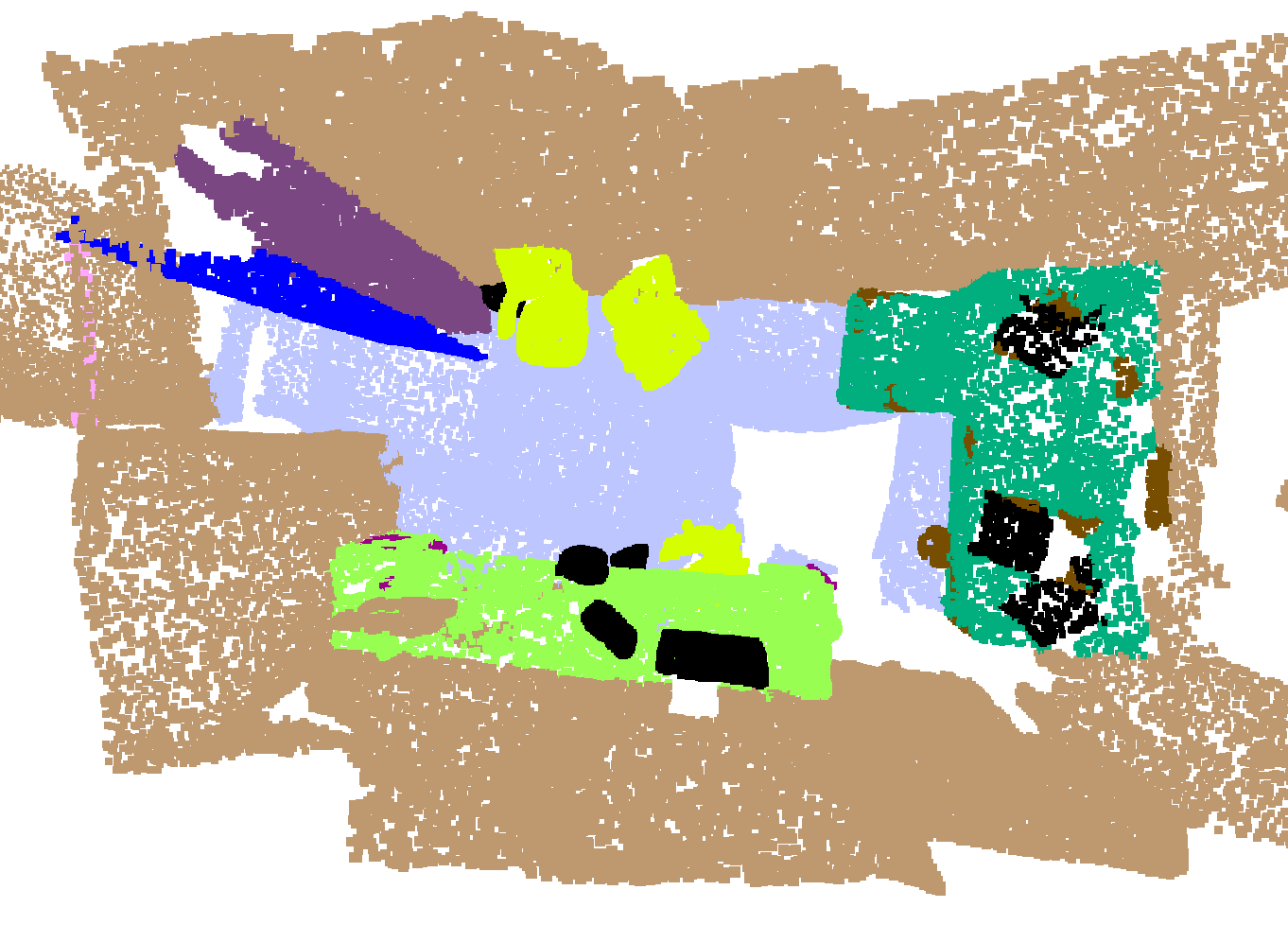} \phantomcaption}
\vspace{-3.5mm}
\subfloat{\includegraphics[width=0.225\linewidth,trim={0 0 0 0}, clip]{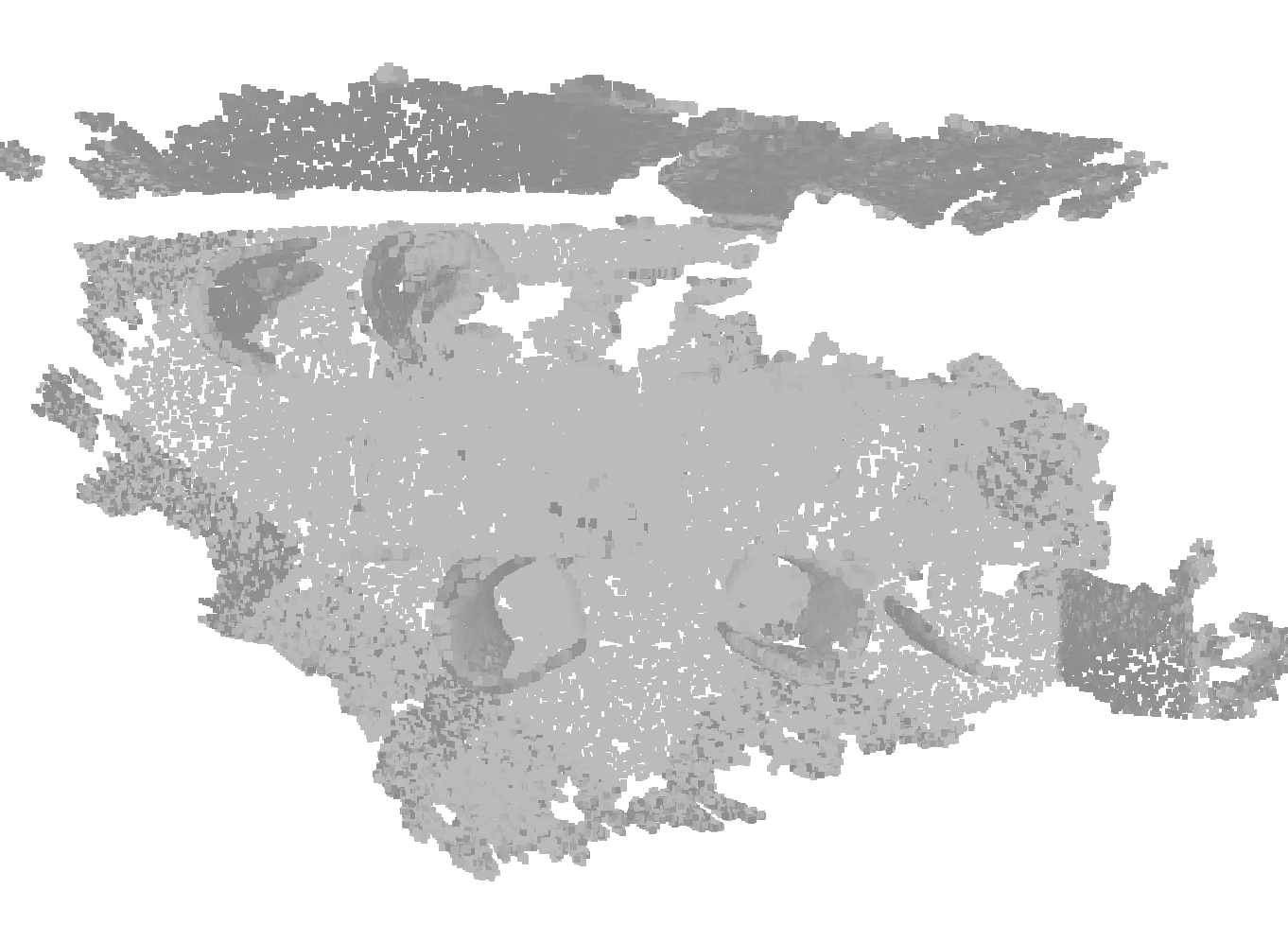} \phantomcaption}
\subfloat{\includegraphics[width=0.225\linewidth,trim={0 0 0 0}, clip]{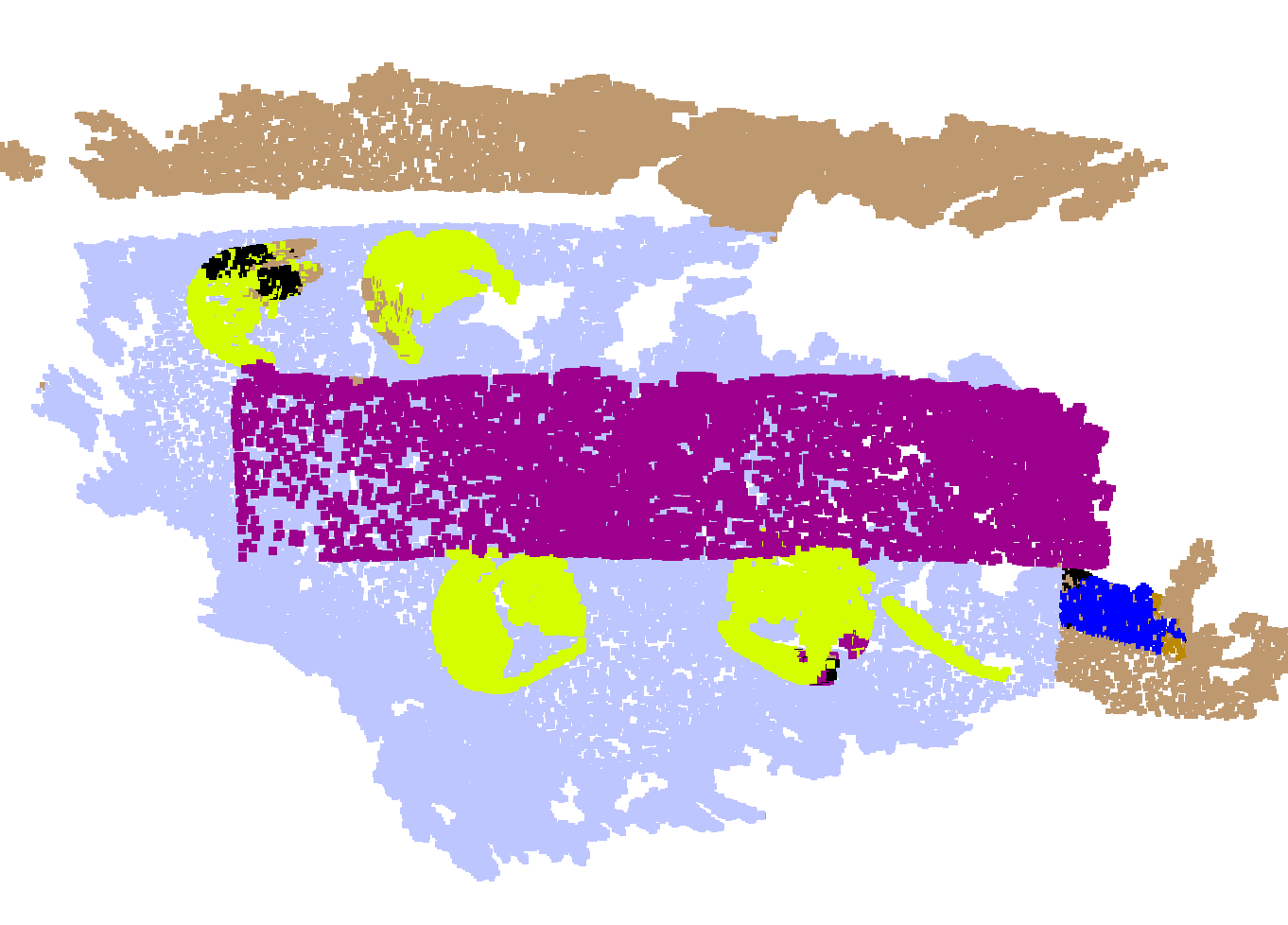} \phantomcaption}
\subfloat{\includegraphics[width=0.225\linewidth,trim={0 0 0 0}, clip]{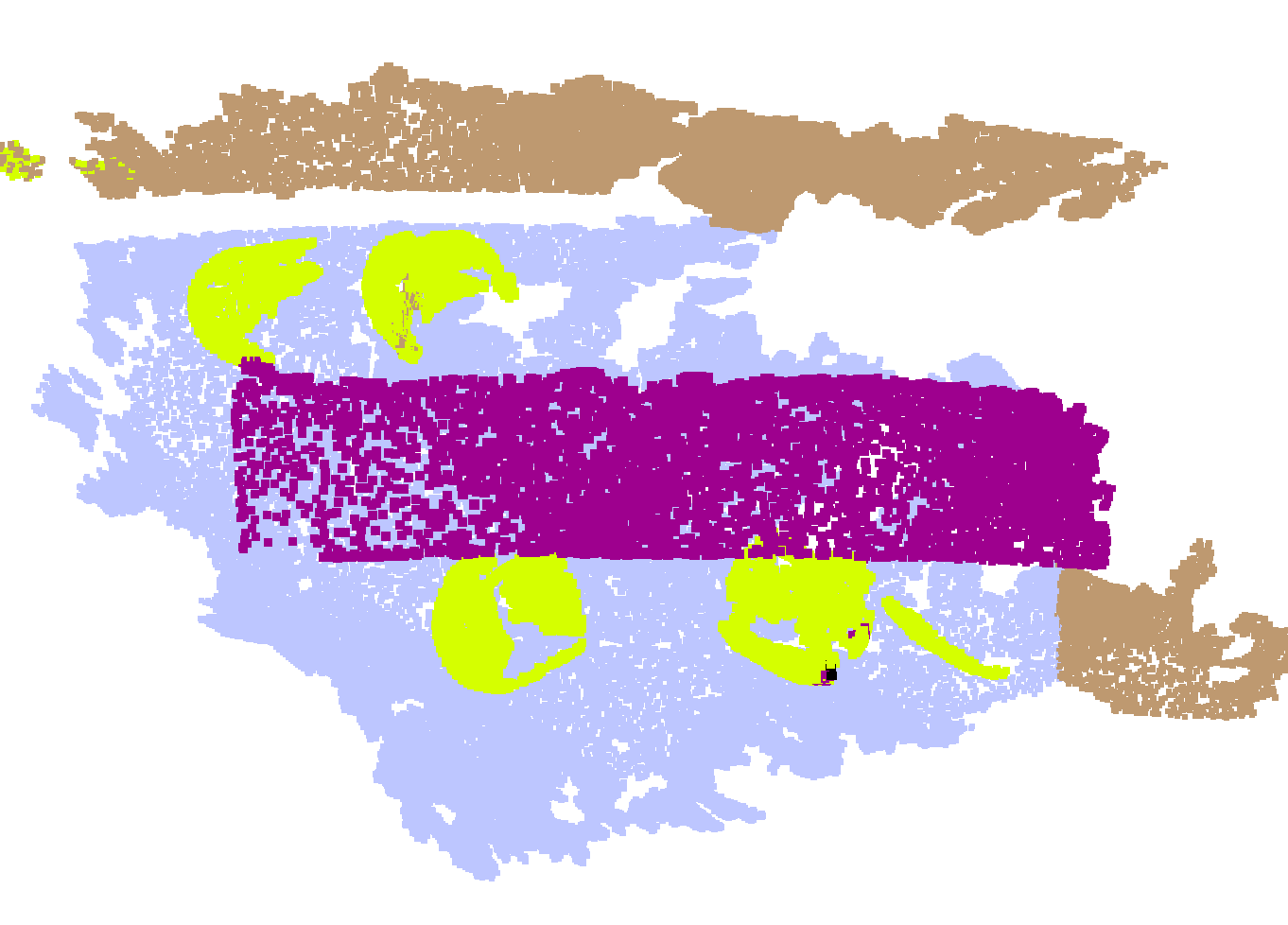} \phantomcaption}
\subfloat{\includegraphics[width=0.225\linewidth,trim={0 0 0 0}, clip]{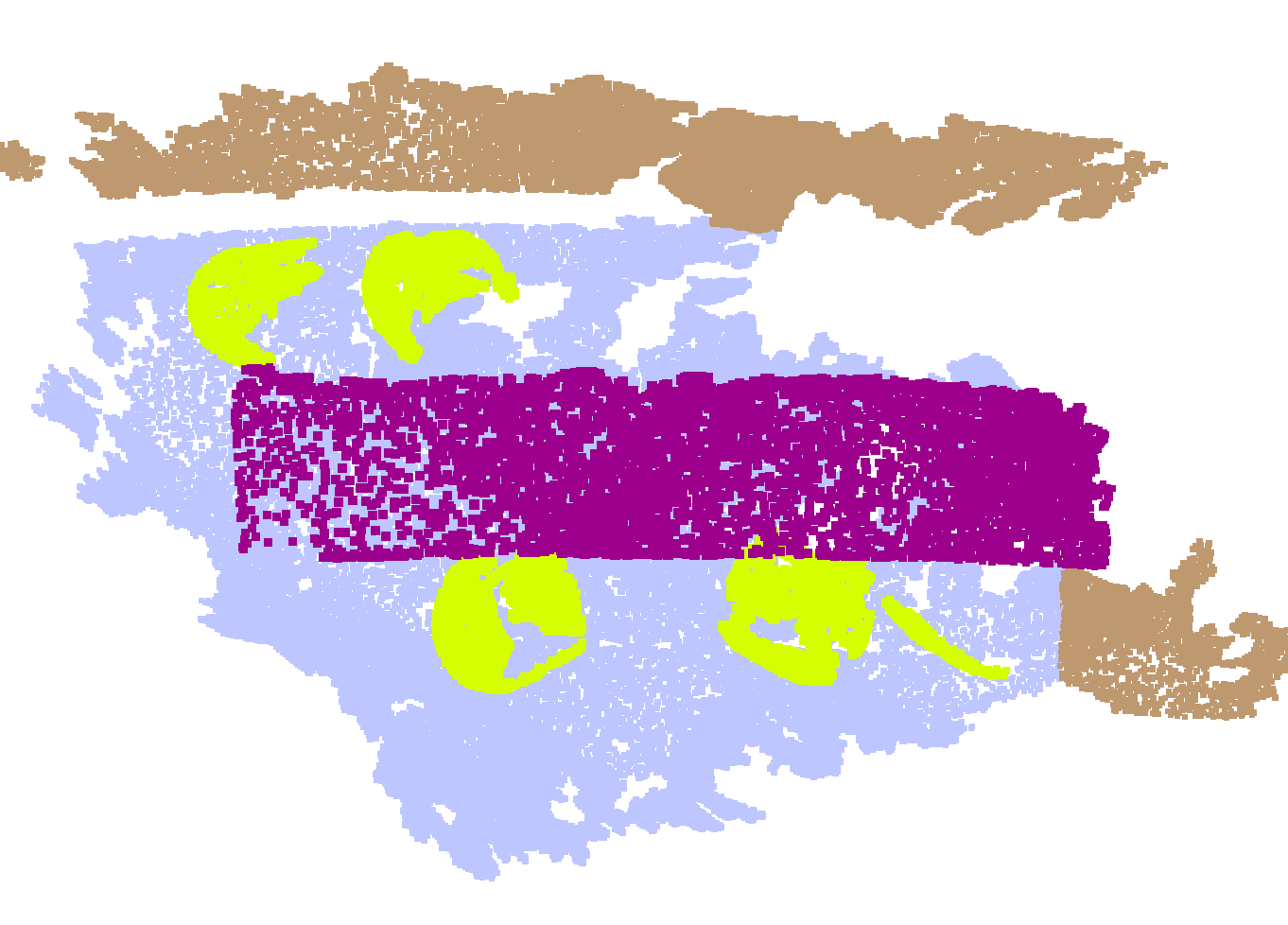} \phantomcaption}
\vspace{-3.5mm}
\subfloat[\textbf{Input}]{\includegraphics[width=0.225\linewidth, trim={0 0 0 0}]{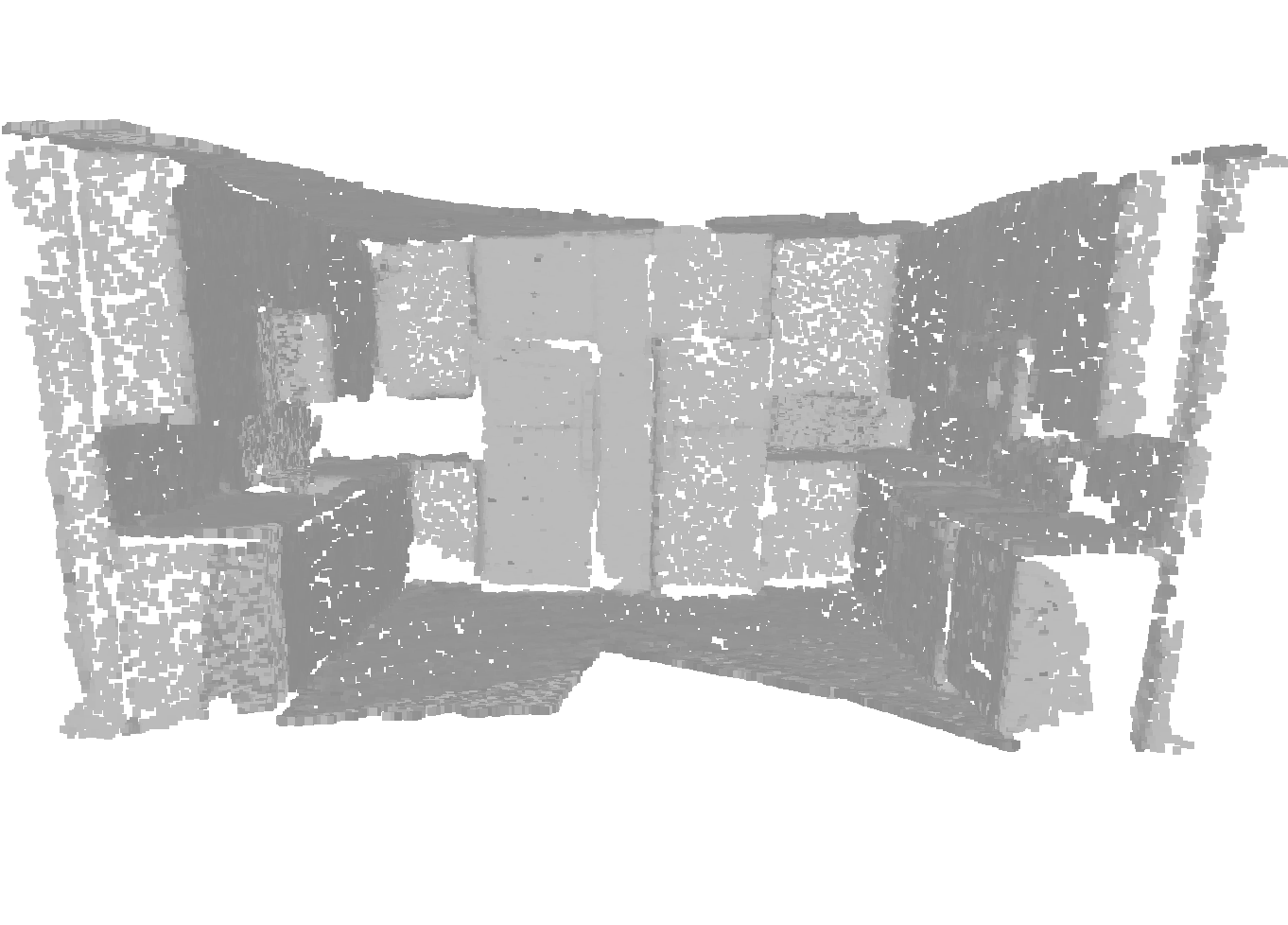}} \hspace{1.0mm}
\subfloat[\textbf{PointNet++~\cite{qi2017pointnet++}}]{\includegraphics[width=0.225\linewidth, trim={0 0 0 0}]{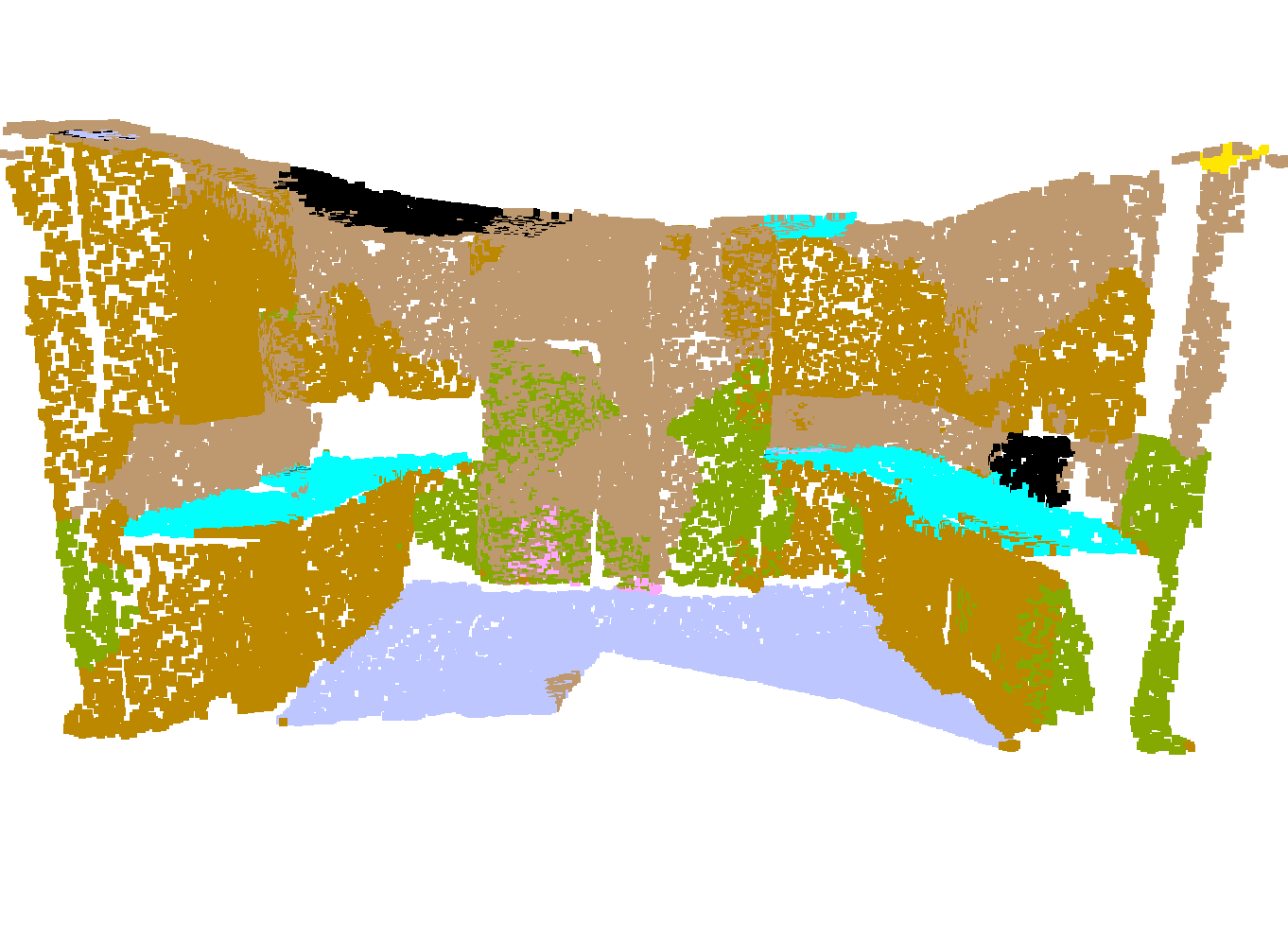}} \hspace{1.0mm}
\subfloat[\textbf{Our}]{\includegraphics[width=0.225\linewidth, trim={0 0 0 0}]{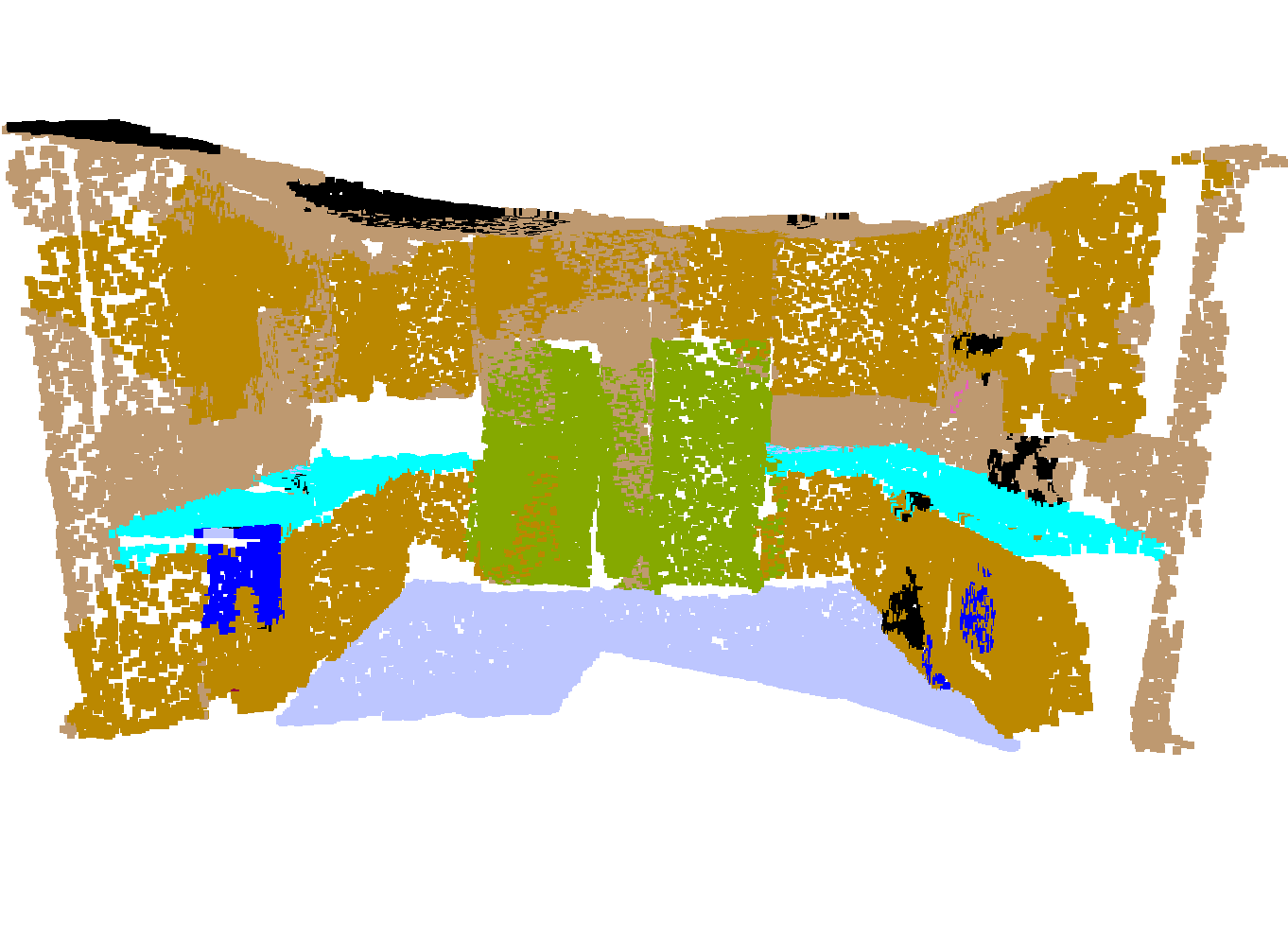}} \hspace{1.0mm}
\subfloat[\textbf{Ground Truth}]{\includegraphics[width=0.225\linewidth, trim={0 0 0 0}]{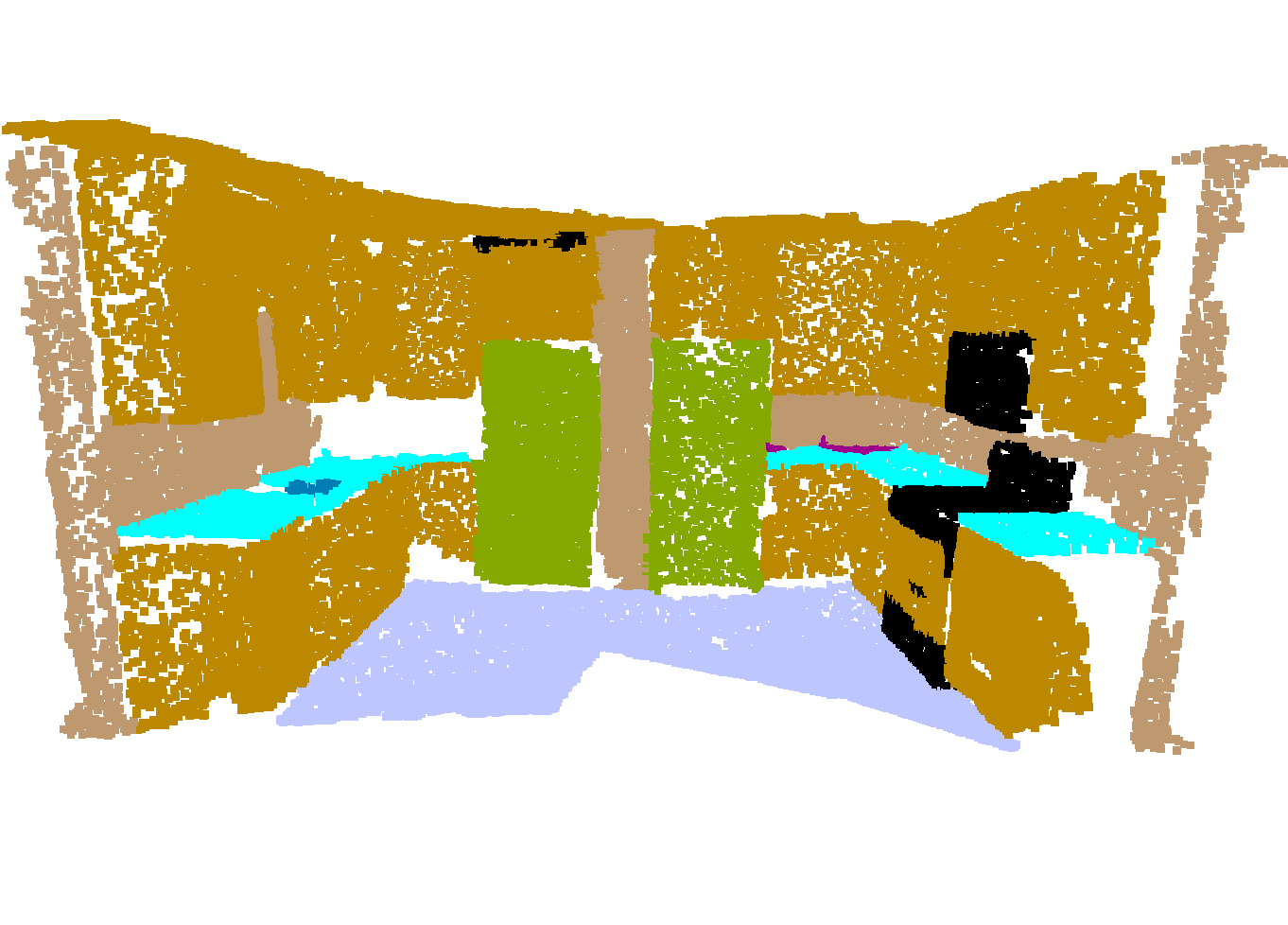}}
\caption{The visualization results on \emph{ScanNet} dataset. We compare our model with PointNet++~\cite{qi2017pointnet++} and the ground truth. The challenging sample rooms have been picked from the \emph{ScanNet} dataset.}
\vspace{-5.5mm}
\label{fig:suppl_scannet_eval}
\end{figure*}

\end{document}